%% file: main.tex
\documentclass{article}

\usepackage{arxiv}

\usepackage[utf8]{inputenc} 
\usepackage[T1]{fontenc}    
\usepackage{hyperref}       
\usepackage{url}            
\usepackage{booktabs}       
\usepackage{amsfonts, amsmath, amssymb, bm}       
\usepackage{nicefrac}       
\usepackage{microtype}      
\usepackage{lipsum}
\usepackage{graphicx}
\usepackage{xcolor, soul}
\graphicspath{ {./images/} }
\usepackage{algorithm}
\usepackage{algpseudocode}
\usepackage{mathtools}

\usepackage{caption}
\usepackage{subcaption}

\definecolor{edit}{rgb}{1.0, 0.0, 0.0}

\title{HypeRL: Hypernetwork-Based Reinforcement Learning for Control of Parametrized Dynamical Systems}

\author{
  Nicolò Botteghi \\
  MOX - Department of Mathematics\\
  Politecnico di Milano\\
  Milano, Italy \\
  \texttt{nicolo.botteghi@polimi.it} \\
  \And
  Stefania Fresca \\
  Department of Mechanical Engineering\\
  University of Washington\\
  Seattle, USA \\
  \AND
  Mengwu Guo \\
  Centre for Mathematical Sciences \\
  Lund University\\
  Lund, Sweden \\
  \And
  Andrea Manzoni \\
  MOX - Department of Mathematics\\
  Politecnico di Milano\\
  Milano, Italy \\
}

\begin{document}
\maketitle

\paragraph{Abstract.} In this work, we devise a new, general-purpose reinforcement learning strategy for the optimal control of parametric dynamical systems.
Such problems frequently arise in applied sciences and engineering and entail a significant complexity when control and/or state variables are distributed in high-dimensional space or depend on varying parameters. Traditional numerical methods, relying on either iterative minimization algorithms -- exploiting, e.g., the solution of the adjoint problem -- or dynamic programming -- also involving the solution of the Hamilton-Jacobi-Bellman (HJB) equation --  while reliable, often become computationally infeasible. Indeed, in either way, the optimal control problem has to be solved for each instance of the parameters, and this is out of reach when dealing with high-dimensional time-dependent and parametric dynamical systems. In this paper, we propose \textit{HypeRL}, a deep reinforcement learning (DRL) framework to overcome the limitations shown by traditional methods. HypeRL aims at approximating the optimal control policy directly, bypassing the need to numerically solve either the HJB equation explicitly for all possible states and parameters, or an adjoint problem within an iterative optimization loop for each parameter instance. Specifically, we employ an actor-critic DRL approach to learn an optimal feedback control strategy that can generalize across the range of variation of the parameters. 
To effectively learn such optimal control laws for different instances of the parameters, encoding the parameter information into the DRL policy and value function neural networks (NNs) is essential. 
To do so, HypeRL uses two additional NNs, often called \textit{hypernetworks}, to learn the weights and biases of the value function and the policy NNs. In this way, HypeRL effectively embeds the parametric information into the value function and policy NNs. We validate the proposed approach on two parametric control problems, namely \emph{(i)} a 1D parametric Kuramoto-Sivashinsky equation with in-domain control, and \emph{(ii)} a navigation problem of particle dynamics in a parametric 2D gyre flow. We show that the knowledge of physical and task-dependent information and the encoding of this information via a hypernetwork, are essential ingredients for learning parameter-dependent control policies that can generalize effectively to unseen scenarios, and for improving the sample efficiency of such policies.

\section{Introduction}
Many complex, distributed dynamical systems can be modeled through a set of parametrized partial differential equations (PDEs) and their optimal control (OC) represents a crucial challenge in many engineering and science applications, going way beyond the single, direct simulation of these systems. OC allows the integration of active control mechanisms into a control system and its most common application in addressing such problems involves determining optimal closed-loop controls that minimize a specified objective functional \cite{manzoni2021optimal}. To solve a PDE-constrained OC problem, one possibility is to rely on the Hamilton-Jacobi-Bellman (HJB) equation. However, the HJB is not easily tractable and is usually computationally expensive for high-dimensional and large-time horizon control problems. Another option is to locally solve the OC problem by exploiting the Pontryagin Maximum Principle (PMP). However, PMP involves the backward solution (in time) of the adjoint problem with the same dimension of the state equation. Hence, to find the OC law one should solve both the state and the adjoint equation repeatedly -- forward and backward in time, respectively -- in the whole space-time domain. For high-dimensional problems, storage and computational requirements make the PMP becoming quickly prohibitive. Traditional OC theory may present non-negligible shortcomings \cite{collis2004issues}, which are even more severe when the PDE parameters vary and the OC problem has to be solved for each new instance of the parameters. 

In this respect, reinforcement learning (RL) \cite{sutton2018reinforcement} is emerging as a new paradigm to address the solution of PDE-constrained OC problems and has been shown to outperform other OC strategies when the system's states are high-dimensional, noisy, or only partial measurements are available. RL avoids to solve the HJB or the adjoint equations explicitly, which would be untractable for extremely complex problems. Unlike the aforementioned OC approaches, RL aims to solve control problems by learning an OC law (the {\em policy}), while interacting with the dynamical system (the {\em environment}). RL assumes no prior knowledge of the system, thus yielding broadly applicable control approaches. Deep reinforcement learning (DRL) is the extension of RL using deep neural networks (NNs) to represent value functions and policies~\cite{arulkumaran2017deep, li2017deep, franccois2018introduction}. DRL has shown outstanding capabilities in complex control problems such as games \cite{mnih2015human, ye2020mastering, shao2019survey, lample2017playing, mnih2013playing, van2016deep}, simulated and real-world robotics \cite{lin1992reinforcement, kober2013reinforcement, polydoros2017survey, zhang2015towards, gu2017deep, zhao2020sim, botteghi2021low}, and recently PDEs, with particular emphasis on fluid dynamics \cite{bucci2019control, beintema2020controlling, buzzicotti2020optimal, fan2020reinforcement, rabault2019accelerating, xia2023active, peitz2023distributed, zolman2024sindy, botteghi2024parametric}. 

Despite its success, DRL still suffers from two major drawbacks, namely {\em (i)} the \textit{sample inefficiency}, making the DRL algorithms extremely data hungry, and {\em (ii)} the \textit{limited generalization} of the control strategies to changes in the environments. Tackling these two challenges is crucial for advancing DRL towards large-scale and real-world problems. 
These limitations are especially severe in the context of control of parametric PDEs, where obtaining (state) measurements is challenging due to the computational complexity of the (forward) PDE models. Moreover, little to no attempt has yet been made to devise DRL algorithms capable of handling changes in the systems’ dynamics resulting from variations in known PDE parameters. This means that for any new configuration of the system, the optimization problem must be solved from scratch. Additionally, while DRL generally decreases the computational complexity of traditional methods for the solution of OC problems, these algorithms still require huge training times and large amounts of data, i.e., we need the repeated evaluation of the solution to the system state equations. Consequently, applying DRL algorithms to address single problem scenarios would not be entirely justified.

Improving sample efficiency and generalization in DRL has been a key focus of recent research. Examples of such approaches are \textit{imitation learning} \cite{hussein2018, torabi2018behavioral}, where expert data are used to pre-train the control policies and to speed-up the (policy-)optimization process, \textit{transfer learning} \cite{torrey2010transfer, weiss2016survey}, where an optimal policy is transferred to a new environment with little or no retraining, and \textit{unsupervised representation learning} \cite{lesort2018, botteghi2022unsupervised}, where unsupervised learning techniques are exploited to learn compact representations of the data. Representation learning has been shown to improve the generalization of control policies to new environments and scenarios \cite{botteghi2022unsupervised}. Eventually, another prominent approach for enhancing the generalization capabilities of DRL agents is \textit{meta learning} \cite{hospedales2021meta, vilalta2002perspective}, where DRL policies are specifically built and optimized for adapting to new scenarios. However, these approaches have yet to be developed to tackle the challenging problem of controlling parametric PDEs, leaving a large gap for research and developments in the field.  

\begin{figure}[h!]
    \centering
    \includegraphics[width=0.8\textwidth]{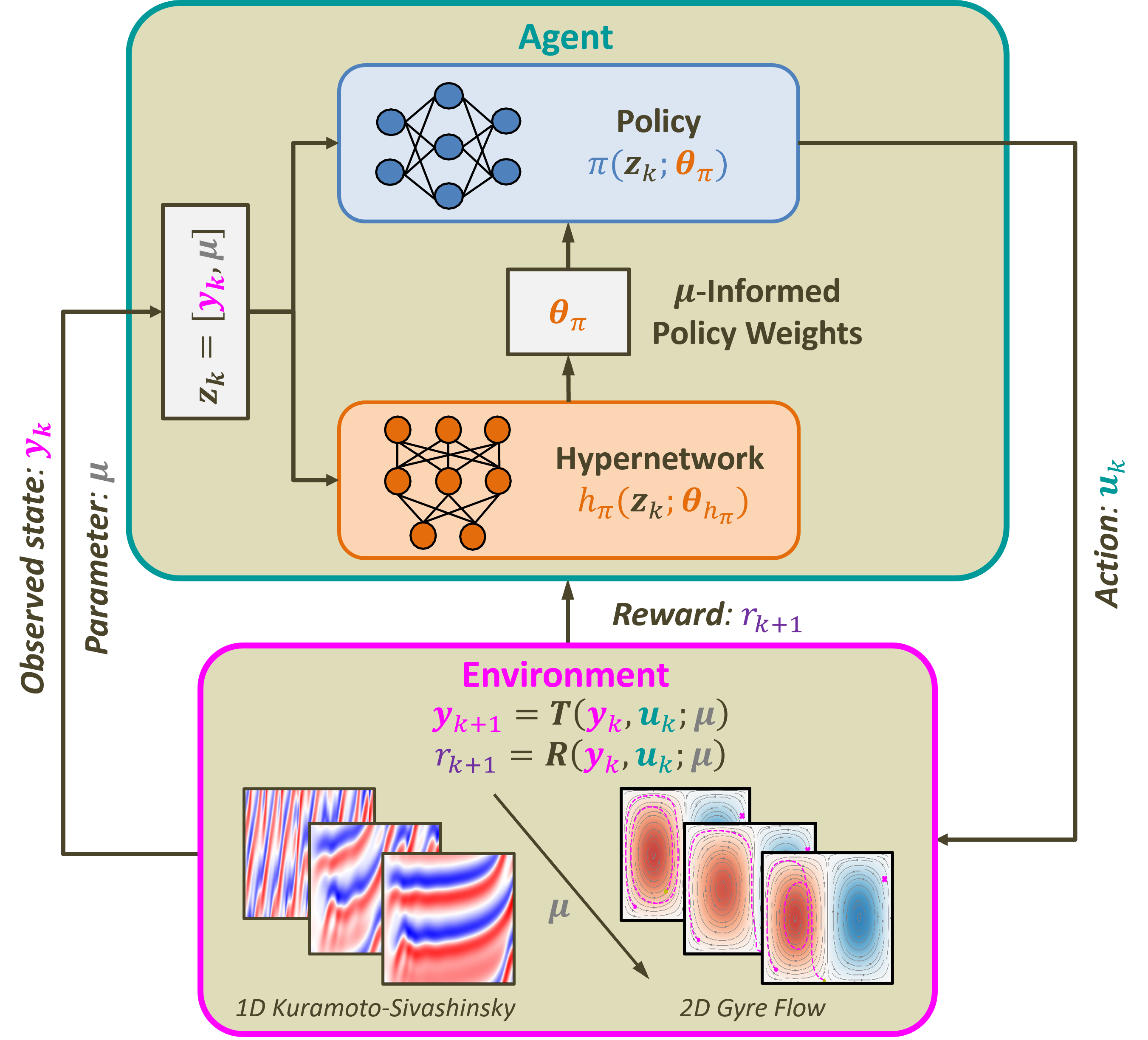}
    \caption{HypeRL for parametric PDE-constrained OC. We rely on a hypernetwork $h_{\pi}(\bm{z}_k; \boldsymbol{\theta}_{h_{\pi}})$ to learn the weights and biases of the policy neural network.}
    \label{fig:1}
\end{figure}

\textit{Hypernetworks} \cite{ha2016hypernetworks} are a class of NNs that provide the parameters, i.e., the weights and biases, of other NNs, often referred to as main or primary networks. 
Hypernetworks have shown promising results in a variety of deep learning problems, including continual learning, causal inference, transfer learning, weight pruning, uncertainty quantification, zero-shot learning, natural language processing, and recently DRL \cite{chauhan2023brief}. Indeed, hypernetworks are capable of enhancing the flexibility, expressivity, and performance of deep learning-based architecture, opening new doors for the development of novel and more advanced architectures. Hypernetworks in DRL were first used in \cite{sarafian2021recomposing} to learn the parameters of the value function or of the policy NNs. Enhancing DRL with hypernetworks has been done in the context of meta RL, zero-shot RL, and continual RL \cite{beck2023hypernetworks, rezaei2023hypernetworks,huang2021continual} for improving the performance of RL agents. However, to the best of our knowledge, no one has tackled the problem of controlling parametric dynamical systems with hypernetworks and DRL so far.

In this paper, we propose HypeRL, a novel DRL algorithm for control parameter-dependent PDE-constrained OC problems tailored to address the two aforementioned limitations of traditional DRL approaches, namely sample efficiency and generalization. In particular, with reference to Figure \ref{fig:1}, as per standard RL settings, we discretize in time and space the environment dynamics described by:
\begin{equation*}
\begin{split}
    \bm{y}_{k+1} = T(\bm{y}_{k}, \bm{u}_{k}; \boldsymbol{\mu})\, ,\\
    r_{k+1} = R(\bm{y}_{k}, \bm{u}_{k}; \boldsymbol{\mu})\, ,
\end{split}
\end{equation*}
where $T(\bm{y}_{k}, \bm{u}_{k}; \boldsymbol{\mu})$ and $R(\bm{y}_{k}, \bm{u}_{k}; \boldsymbol{\mu})$ are the transition and reward functions, $\bm{y}_{k+1}, \bm{y}_{k}$ indicate the state at timestep $k+1$ and $k$, respectively, $\bm{u}_k$ the control input, $r_{k+1}$ is the reward, and $\boldsymbol{\mu}$ the parametric dependency of the environment, which might include physical or task-dependent information. In addition, we define the agent's state at timestep $k$ as $\bm{z}_k=[\bm{y}_k,\boldsymbol{\mu}]$. We exploit the knowledge of parameters $\boldsymbol{\mu}$ to learn a parametrization of the control policy $\pi(\bm{z}_k;\boldsymbol{\theta}_\pi)$ and value function $Q(\bm{z}_k,\bm{u}_k;\boldsymbol{\theta}_Q)$ by means of two hypernetworks $h_{\pi}(\bm{z}_k; \boldsymbol{\theta}_{h_{\pi}})$ 
\begin{equation*}
    \begin{split}
        \boldsymbol{\theta}_{\pi} &= h_\pi(\bm{z}_k; \boldsymbol{\theta}_{h_\pi})\, , \\
        \bm{u}_k &= \pi(\bm{z}_k; \boldsymbol{\theta}_{\pi})\, ,
    \end{split}
\end{equation*}
and $h_{Q}(\bm{z}_k; \boldsymbol{\theta}_{h_{Q}})$
\begin{equation*}
    \begin{split}
        \boldsymbol{\theta}_{Q} &= h_Q(\bm{z}_k; \boldsymbol{\theta}_{h_Q})\, , \\
        q_k &= \pi(\bm{z}_k, \bm{u}_k; \boldsymbol{\theta}_{Q})\, ,
    \end{split}
\end{equation*}
where $\boldsymbol{\theta}_{h_\pi}$ and $\boldsymbol{\theta}_{h_Q}$ indicate the hypernetwork learnable parameters, $\boldsymbol{\theta}_{\pi}$ the policy learnable parameters, and $\boldsymbol{\theta}_{Q}$ the value function ones. We test our hypernetwork-based method on two challenging parametric control problems, namely \emph{(i)} the stabilization of a parametric Kuramoto-Sivashinsky equation to arbitrary references and \emph{(ii)} the control of a particle to arbitrary target locations in a parametric gyre flow. Throughout our experiments,  we show that the encoding via hypernetworks of the physical and task-dependent information enables effective coordination and enhances generalization capabilities compared to traditional and state-of-the-art DRL algorithms.

The paper is organized as follows: in Section \ref{sec:preliminaries}, we introduce the building blocks of our framework, namely OC, RL, and hypernetworks. In Section \ref{sec:methodology}, we describe our parameter-informed HypeRL framework combining DRL with hypernetworks, and in Section \ref{sec:numerical_experiments} we show the results and discuss the findings. The conclusions are finally reported in Section \ref{sec:conclusion}.

\section{Preliminaries}\label{sec:preliminaries}

In this section, we frame the class of optimal control (OC) problems we aim at solving, and we introduce the building blocks -- namely reinforcement learning (RL) and hypernetworks -- that will be combined, at a later stage, to obtain the new HypeRL framework for OC of parametric partial differential equations (PDEs).

\subsection{Optimal Control Problems}

Along the paper we consider time-continuous OC problems under the following form  \cite{manzoni2021optimal}:
\begin{equation}
\begin{split}
    \min_{\bm{y}\in \mathcal{Y},~\bm{u}\in \mathcal{U}_\text{ad}}\ \ \ &{J}(\bm{y}, \bm{u}) \\
    \text{subject to ~~~~} & \frac{\partial \bm{y}}{\partial t} (t) = \bm{F}(\bm{y}(t),\bm{u}(t);\boldsymbol{\mu})\quad \text{for}\; t\in [t_0, t_f] \,, \;\text{with}\; \bm{y}(t_0)=\bm{y}_0\,,\\
\end{split}
\label{eq:OCP}
\end{equation}
in which $\bm{y}: [t_0,t_f]\to \mathbb{R}^{|\bm{y}|}$ denotes the state solution in a function space $\mathcal{Y} = \mathbb{R}^{|\bm{y}|} \otimes C^1[t_0, t_f]$, and $\bm{u}:[t_0,t_f]\to U_\text{ad} \subset \mathbb{R}^{|\bm{u}|}$ denotes the control input functions in an admissible space $\mathcal{U}_\text{ad} = U_\text{ad} \otimes C[t_0, t_f]$, $U_\text{ad}$ being the admissible set of input control vectors at each time instance. The state equation represents a semi-discretized form of a nonlinear time-dependent PDE characterized by a parameter vector $\boldsymbol{\mu} \in \mathcal{M}\subset \mathbb{R}^{|\boldsymbol{\mu}|}$, $\mathcal{M}$ being the compact parameter space, $\bm{F}$ is continuous in $\bm{u}$ and Lipschitz continuous in $\bm{y}$ (with a Lipschitz constant independent of $\bm{u}$)\footnote{We refer, e.g., to the Picard–Lindelöf theorem \cite{lindelof1894sur}.}, ${J}:\mathcal{Y}\times \mathcal{U}_\text{ad}\to \mathbb{R}$ is a cost functional, and $\bm{y_0}$ gives a prescribed initial condition. 

In this work, we consider cost functionals of the following form:
\begin{equation}
{J}(\bm{y}, \bm{u}) = \int_{t_0}^{t_f}  L(\bm{y}(\tau), \bm{u}(\tau),\tau)  ~\mathrm{d}\tau  \,,
    \label{eq:J}
\end{equation}
in which
\begin{equation}
\begin{split}
 L: \mathbb{R}^{|\bm{y}|}\times U_\text{ad} \times [t_0,t_f] \rightarrow \mathbb{R}, \quad (\bm{x},\bm{z},\tau) \mapsto \frac{1}{2} ||\bm{x} - \bm{y}_{\text{ref}}(\tau)||^2 + \frac{\alpha}{2} ||\bm{z} - \bm{u}_{\text{ref}}(\tau)||^2  \,, \\
\end{split}
\label{eq:L}
\end{equation}
where $\alpha$ is a scalar coefficient balancing the contribution of the two terms, $\bm{y}_{\text{ref}}$ and $\bm{u}_{\text{ref}}$ are the reference values of the state and of the control, respectively, and $\|\cdot\|_2$ represents the Euclidean norm.

The OC problem in \eqref{eq:OCP} can be addressed by introducing the value function $V: \mathbb{R}^{|\bm{y}|}\times [t_0, t_f] \to \mathbb{R}$ as follows:
\begin{equation}
    V(\bm{y}(t), t) := \min_{\bm{u}\in U_\text{ad}\otimes C[t, t_f]}\int_{t}^{t_f}  L(\bm{y}(\tau), \bm{u}(\tau),\tau) ~\mathrm{d}\tau \,.
\end{equation}
Note that the state equation is satisfied over the time interval $[t,t_f]$ with the state starting with $\bm{y}(t)$.
To note, such a value function is the solution to the Hamilton-Jacobi-Bellman (HJB) \cite{kirk2004optimal, fleming2006controlled} equation:
\begin{equation}
    -\frac{\partial V(\bm{y},t)}{\partial t} = \min_{\bm{u}(t)\in U_\text{ad}}\left\{  L(\bm{y}, \bm{u}(t),t) + \bm{F}(\bm{y},\bm{u}(t);\boldsymbol{\mu})^T\frac{\partial V(\bm{y},t)}{\partial \bm{y}} \right\}, \quad (\bm{y},t) \in  \mathbb{R}^{|\bm{y}|}\times [t_0, t_f]\,,
\label{eq;HJB_continuoustime}
\end{equation}
which provides, at least in principle, the solution of the OC problem \eqref{eq:OCP}, that is, $\min J = V(\bm{y}_0, t_0)$. 

The HJB equation is typically a high-dimensional PDE as $|\bm{y}|$ is often large. Moreover, due to the dependency of $\bm{F}(\cdot)$ on the (many) parameters $\boldsymbol{\mu}$, the HJB equation has to be solved for each sampling location of the parameters, because each sample defines an individual optimal control problem for the specific $\boldsymbol{\mu}$. Therefore,  \eqref{eq;HJB_continuoustime} is not easily tractable and may be computationally prohibitive for high-dimensional OC problems constrained by PDEs. This trait is commonly known as the \textit{curse of dimensionality} \cite{sutton2018reinforcement}. We also note that an alternative approach to solving the constrained OC problem \eqref{eq:OCP} uses the Karush-Kuhn-Tucker optimality conditions \cite{karush1939minima, kuhn1951nonlinear} via the introduction of a Lagrange multiplier. However, also in this case the iterative nature of the optimization methods (like, e.g., gradient-based, Newton, quasi-Newton, sequential quadratic programming) makes the numerical solution of each PDE-constrained optimization problem usually very hard. In the case of multiple OC problems, for different parameter values, the overall computational cost would be therefore prohibitive.

\subsection{From Dynamic Programming to Reinforcement Learning}\label{subsec:reinforcement_learning}
Solving the HJB equation in \eqref{eq;HJB_continuoustime} requires reformulating the problem in a time-discrete framework. Algorithms solving the fully-discretized HJB are typically referred to as dynamic programming (DP) approaches \cite{sutton2018reinforcement, bertsekas2012dynamic, bertsekas2024course}. DP methods can handle all kinds of hybrid systems, even with non-differentiable dynamics, and stochastic OC problems \cite{diehl2011numerical}. Examples of DP algorithms are policy iteration and value iteration \cite{sutton2018reinforcement}.
DP utilizes the Markov Decision Process (MDP) as underlying mathematical framework in order to account for stochastic systems' dynamics and tackle a broader class of stochastic OC problems \cite{bertsekas2012dynamic}. 

A MDP is a tuple $\langle Y, U_{\text{ad}}, T, R \rangle$ where $ Y \subset \mathbb{R}^{|\bm{y}|}$ is the set of observable states, $U_{\text{ad}} \subset \mathbb{R}^{|\bm{u}|}$ is the set of admissible actions, $T: Y \times Y \times U_{\text{ad}} \longrightarrow [0,1]^{|\bm{y}|}$ such that $(\bm{y}_{k+1},\bm{y}_k,\bm{u}_k) \longmapsto T(\bm{y}_{k+1},\bm{y}_k,\bm{u}_k)$ is the transition function, and $R: Y \times U_{\text{ad}}\longrightarrow \mathbb{R}$ such that $ (\bm{y}_k,\bm{u}_k)\longmapsto R(\bm{y}_k,\bm{u}_k)$ denotes the reward function. The transition function $(\bm{y}_{k+1},\bm{y}_k,\bm{u}_k) \longmapsto T(\bm{y}_{k+1},\bm{y}_k,\bm{u}_k)$ 
describes the probability of reaching state $\bm{y}_{k+1}$ from state $\bm{y}_k$ while taking action $\bm{u}_k$,
\begin{equation*}
    p(\text{Y}_{k+1} = \bm{y}_{k+1}| \text{Y}_k=\bm{y}_k,\text{U}_k=\bm{u}_k)\, ,
\end{equation*}
fulfilling
\begin{equation*}
    \sum_{\bm{y}_{k+1} \in Y} p(\text{Y}_{k+1} = \bm{y}_{k+1}| \text{Y}_k=\bm{y}_k,\text{U}_k=\bm{u}_k) = 1, \qquad   \forall \ \  \bm{y}_k \in Y, \bm{u}_k \in U_{\text{ad}}.
\end{equation*}
It is worth mentioning that a deterministic transition function $\bm{y}_{k+1}=T(\bm{y}_k, \bm{u}_k)$ is a special case arising when $p(\bm{y}_{k+1}|\bm{y}_k, \bm{u}_k)=1$ and that the reward function is the fully-discretized counterpart of the running cost $L(\cdot)$, but with opposite sign. Therefore, the value function in DP problems is typically written as a maximization problem over the possible controls rather than a minimization one:
\begin{equation}
    V(\bm{y}_k) = \max_{\bm{u} \in U_{\text{ad}}}  \mathbb{E}\big[ R(\text{Y}_k, \text{U}_k) + V(\text{Y}_{k+1}) | \text{Y}_k=\bm{y}_k, \text{U}_k=\bm{u}_k \big ]\, , 
\label{eq:value_function_DP}
\end{equation}
where we indicate with $\mathbb{E}[\cdot]$ the expected value of a random variable, and $\text{Y}_{k+1} \sim T(\bm{y}_{k+1}, \bm{y}_k, \bm{u}_k)$. Equation \eqref{eq:value_function_DP} can be seen as the fully-discretized HJB in stochastic settings.

The key idea of DP is to estimate the value function $V(\cdot)$ using the perfectly-known environment dynamics $R(\cdot)$ and $T(\cdot)$ and then use it to structure the search for good control strategies. Despite their success, DP algorithms {\em (i)} require perfect knowledge of $R(\cdot)$ and $T(\cdot)$  -- despite they are, in many scenarios, often not known exactly or extremely expensive to compute, especially when the dimensionality of the state is very high -- and {\em (ii)} are unpractical to use in problems with large number of states due to the need of solving the HJB for all possible states. These two drawbacks drastically limit the application of DP algorithms to complex and large-scale problems such as those arising in the OC of parametric PDEs.

Reinforcement learning (RL) \cite{sutton2018reinforcement} is a promising machine learning approach to solve sequential decision-making problems through a trial-and-error process.  Unlike DP, RL does not try to solve the HJB for all possible states; rather, it aims at deriving OC laws from {\em (i)} measurements of the system -- often referred to as observations, and {\em (ii)} reward samples, without direct knowledge of $T(\cdot)$ and $R(\cdot)$ \cite{sutton2018reinforcement, lewis2009reinforcement}. In RL, we can identify two main entities: the \textit{agent} and the \textit{environment} (see Figure \ref{fig:1}). The agent aims to find the best strategy to solve a given task by interacting with an unknown environment. Similarly to DP, the optimality of the strategy learned by the agent is defined by a task-dependent reward function. In particular, we can use RL to solve OC problems, such as the one in Equation $\eqref{eq:OCP}$, by learning an OC law from data without the need to explicitly solve the HJB equation (either in continuous or in discrete settings) for all possible states. Similarly to DP, we can rely on MDPs to mathematically formulate the RL problem.
The goal of any RL algorithm is to find the optimal policy (control law) maximizing the expected cumulative return~$G_k$:
\begin{equation}
G_{k+1} = r_{k+1} + \gamma r_{k+2} + \gamma^2 r_{k+3} + \cdots = \sum_{j=1}^{H} \gamma^j r_{k+j}\,,
\label{eq:cumulative_rew}
\end{equation}
where the control horizon\footnote{The control horizon can be either finite or infinite.} $H$ is defined as $H=(t_f - t_0)/\Delta t$, the subscript $k$ denotes the time-step,  $r_k=R(\bm{y}_k, \bm{u}_k)$ is the instantaneous reward received by the agent at time-step $k$, and $\gamma$ is a discount factor balancing the contribution of present and future rewards, where $0 \leq \gamma \leq 1$. 
It is worth mentioning that the expected return $G_k$ is the analogous of the cost functional ${J}(\cdot)$ in Equation \eqref{eq:J} in a fully-discretized and discounted setting.

Almost all RL algorithms revolve around estimating the value function without any knowledge of the true transition and reward functions, i.e., without explicitly solving the HJB for all possible states. Thus, the value function is learned from state-action-reward trajectories, i.e., from data. 
Starting from an initial estimate of the value function and of the optimal policy, RL algorithms iteratively improve these estimates until the value function and the optimal policy are found.
We can rewrite the value function $V(\cdot)$ using the expected return $G_k$:
\begin{equation}
    V(\bm{y}_k) = \mathbb{E}_{\pi^*}[G_{k+1}|\text{Y}_k=\bm{y}_k],
    \label{eq:value_function_RL}
\end{equation}
where $\mathbb{E}_{\pi^*}[\cdot | \text{Y}_k=\bm{y}_k ]$ denotes the conditional expectation  of a random variable if the agent follows the optimal policy $\pi^*$ {on a time-step of length $H$}, given the starting value $\text{Y}_k = \bm{y}_k$. In general, a control policy $\pi$ can be either stochastic, i.e.,  $\pi: Y \times U_{\text{ad}} \rightarrow [0, 1]^{|\bm{u}|}$, or deterministic, i.e.,  $\pi: Y \rightarrow U_{\text{ad}}$.

Similarly, we can define the action value function $Q:Y \times U_{\text{ad}} \to \mathbb{R}$, as the value of taking action $\bm{u}_k$ at a certain state $\bm{y}_k$:
\begin{equation}
    Q(\bm{y}_k, \bm{u}_k) = \mathbb{E}_{\pi^*}[G_{k+1}|\text{Y}_k=\bm{y}_k, \text{U}_k=\bm{u}_k]. 
\label{eq:actiovaluefunction}
\end{equation}
It is worth mentioning that there exists a direct relation between the value function $V(\bm{y}_k)$ and the action-value function $Q(\bm{y}_k, \bm{u}_k)$. In particular, we can write:
\begin{equation*}
    V(\bm{y}_k) = \max_{\bm{u}_k \in U_{\text{ad}}} Q(\bm{y}_k, \bm{u}_k).
\label{eq:valuevsactionvalue}
\end{equation*}

RL algorithms are usually classified as \textit{model-based} or \textit{model-free} methods \cite{banerjee2023survey}. In this context, the keyword \textit{model} indicates whether the agent relies (model-based) or not (model-free) on an environment model, often built from the interaction data, to learn the value function and the policy.  Another important distinction can be found between \textit{online} and \textit{offline} approaches. Online RL aims at learning the optimal policy while interacting with the environment. Conversely, offline RL aims to learn the policy offline given a fixed dataset of trajectories. While online methods better embody the interactive nature of RL, in many (safety-critical) applications it is not possible to apply random actions to explore the environment and offline approaches are preferred. Eventually, we can distinguish among \textit{value-based}, \textit{policy-based}, and \textit{actor-critic} algorithms \cite{sutton2018reinforcement}. Value-based algorithms rely only on the estimation of the (action) value function and derive the optimal policy by greedily selecting the action with the highest value at each time-step. Examples of value-based algorithms are Q-learning \cite{watkins1992q} and its extensions relying on deep neural networks \cite{mnih2013playing, mnih2015human, van2016deep, wang2016dueling}. Second, policy-based algorithms directly optimize the parameters of the policies with the aim of maximizing the return $G_k$ via the policy gradient \cite{sutton2018reinforcement}. One of the first and most famous policy-based algorithms is REINFORCE \cite{williams1992simple}. Third, actor-critic algorithms learn value function and policy at the same time. The keyword \textit{actor} refers to the policy acting on the environment, while \textit{critic} refers to the value function assessing the quality of the policy. Examples are deep deterministic policy gradient (DDPG) \cite{lillicrap2015continuous}, proximal policy optimization (PPO) \cite{schulman2017proximal}, and soft actor-critic (SAC) \cite{haarnoja2018soft}. Eventually, we can identify \textit{on-policy} and \textit{off-policy} methods. On-policy approaches utilize the same policy for exploration and exploitation. Therefore, they often optimize a stochastic policy that can either explore the environment, but also exploit good rewards. An example of on-policy approach is PPO. On the other side, off-policy algorithms maintain two distinct policies for exploration and exploitation, making it possible to reuse the interaction data to update the models. Examples of off-policy algorithms are DDPG and SAC. 

\subsubsection{Twin-Delayed Deep Deterministic Policy Gradient}\label{subsec:TD3}

In our numerical experiments, we utilize a model-free, online, off-policy, and actor-critic approach, namely Twin-Delayed Deep Deterministic Policy Gradient (TD3) \cite{fujimoto2018addressing}. However, our method can be directly used by any other RL algorithm. TD3 learns a deterministic policy $\pi(\cdot)$, i.e., the actor, and the action-value function $Q(\cdot)$, i.e., the critic. The actor and the critic are parametrized by means of two DNNs of parameters $\boldsymbol{\theta}_Q$ and $\boldsymbol{\theta}_{\pi}$, respectively.  We indicate the parametrized policy with $\pi(\bm{y}_k; \boldsymbol{\theta}_{\pi})$ and the action-value function with $Q(\bm{y}_k, \bm{u}_k;\boldsymbol{\theta}_Q)$. TD3 can handle continuous\footnote{Space and time are discretized but each variable can assume any continuous value in the admissible ranges.} state and action spaces, making it a suitable candidate for controlling parametric PDEs using smooth control strategies.

To learn the optimal action-value function\footnote{Because we do not directly solve the HJB equation to obtain the value function, its initial estimate may be far from the true one. Therefore, we highlight the keyword "optimal" to distinguish the true value function from the estimated one.}, TD3 relies on temporal-difference (TD) learning \cite{sutton2018reinforcement}. In particular, starting from the definition of the action-value function in Equation \eqref{eq:actiovaluefunction}, we can write:
\begin{equation}
\begin{split}
    Q(\bm{y}_k, \bm{u}_k) &= \mathbb{E}_{\pi}[\Sigma_{j=1}^{H} \gamma^j r_{k+j}|\text{Y}_k=\bm{y}_k, \text{U}_k=\bm{u}_k]\, , \\
    &= \mathbb{E}_{\pi}[r_{k+1} + \Sigma_{j=2}^{H} \gamma^j r_{k+j}|\text{Y}_k=\bm{y}_k, \text{U}_k=\bm{u}_k]\, , \\
    &= \mathbb{E}_{\pi}[r_{k+1} + \gamma \max_{\bm{u}_k\in U_{\text{ad}}}Q(\text{Y}_{k+1}, \cdot)|\text{Y}_k=\bm{y}_k, \text{U}_k=\bm{u}_k]\, , \\
\end{split}
\end{equation}
where we use \textit{bootstrapping} to express the value of a state-action pair $Q(\bm{y}_k,\bm{u}_k)$ by using in the update rule \eqref{eq:Q-learning-update rule} the estimate of the action-value function $Q(\bm{y}_{k+1},\cdot)$  instead of observed returns from complete trajectories.  The expectation is now with respect to a generic policy $\pi$, which may be far from the optimal policy $\pi^*$. Using TD learning, we can iteratively update the estimate of the action-value function as:
\begin{equation}
    Q(\bm{y}_k, \bm{u}_k) \leftarrow Q(\bm{y}_k, \bm{u}_k) + \alpha \Big(r_{k+1} + \gamma \max_{\bm{u}_k \in U_{\text{ad}}}Q(\bm{y}_{k+1}, \cdot) - Q(\bm{y}_k, \bm{u}_k)\Big)\, ,
\label{eq:Q-learning-update rule}
\end{equation}
where $\alpha$ is the learning rate.
However, in the case of a continuous action space the update rule in Equation \eqref{eq:Q-learning-update rule} cannot be used directly. Indeed, while for discrete action spaces evaluating the maximum of the Q-value for all the possible actions is straightforward, for continuous actions the bootstrap of the target Q-value would require solving an (expensive) optimization problem over the entire action space. Therefore, for continuous actions the following update rule is commonly used:
\begin{equation}
    Q(\bm{y}_k, \bm{u}_k) \leftarrow Q(\bm{y}_k, \bm{u}_k) + \alpha \Big( r_{k+1} + \gamma Q(\bm{y}_{k+1}, \bm{u}_{k+1}) - Q(\bm{y}_k, \bm{u}_k)\Big)\, ,
\label{eq:TD-learning-update rule continuous}
\end{equation}
where $\bm{u}_{k+1}=\pi(\bm{y}_{k+1})$ is selected accordingly to the current estimate of the "optimal" policy.

TD3 relies on a memory buffer $\mathcal{M}$ to store the interaction data $(\bm{y}_k, \bm{u}_k, r_{k+1}, \bm{y}_{k+1})$ for all time-steps $k$.
Given a randomly-sampled batch of interaction tuples, we can employ Equation \eqref{eq:TD-learning-update rule continuous} as a loss function for updating the parameters $\boldsymbol{\theta}_Q$ of the action-value function $Q(\bm{y}_k, \bm{u}_k;\boldsymbol{\theta}_Q)$ as:
\begin{equation}
\begin{split}
        \mathcal{L}(\boldsymbol{\theta}_Q) &= \mathbb{E}_{\bm{y}_k, \bm{u}_k, \bm{y}_{k+1}, r_{k+1} \sim \mathcal{M}}[(r_{k+1} + \gamma \bar{Q}(\bm{y}_{k+1}, \bm{u}_{k+1}; \boldsymbol{\theta}_{\bar{Q}}) - Q(\bm{y}_k, \bm{u}_k; \boldsymbol{\theta}_Q))^2]\, , \\
        &= \mathbb{E}_{\bm{y}_k, \bm{u}_k, \bm{y}_{k+1}, r_{k+1} \sim \mathcal{M}}[(\underbrace{r_{k+1} + \gamma \bar{Q}(\bm{y}_{k+1}, \bar{\pi}(\bm{y}_{k+1};\boldsymbol{\theta}_{\bar{\pi}})+\boldsymbol{\epsilon}; \boldsymbol{\theta}_{\bar{Q}})}_{\text{target value}} - Q(\bm{y}_k, \bm{u}_k; \boldsymbol{\theta}_Q))^2]\,,
\end{split}
\label{eq:valuefunctionloss}
\end{equation}
where the so-called target networks $\bar{Q}(\bm{y}_k, \bm{u}_k; \boldsymbol{\theta}_{\bar{Q}})$ and $\bar{\pi}(\bm{y}_k;\boldsymbol{\theta}_{\bar{\pi}})$ are copies of $Q(\bm{y}_k, \bm{u}_k;\boldsymbol{\theta}_Q)$ and $\pi(\bm{y}_k; \boldsymbol{\theta}_{\pi})$, respectively, with frozen parameters, i.e., they are not updated in the backpropagation step to improve the stability of the training. We indicate with $\boldsymbol{\epsilon} \sim \text{clip}(\mathcal{N}(\bm{0}, \bar{\boldsymbol{\sigma}}), -c, c)$ the noise added to estimate the action value in the interval $[-c, c]$ around the target action.
To reduce the problem of overestimation of the target Q-values, TD3 estimates two independent action-value functions, namely $Q_1(\bm{y}_k,\bm{u}_k;\boldsymbol{\theta}_{Q_1})$ and  $Q_2(\bm{y}_k,\bm{u}_k;\boldsymbol{\theta}_{Q_2})$, and two target action-value functions $\bar{Q}_1(\bm{y}_k,\bm{u}_k;\boldsymbol{\theta}_{\bar{Q}_1})$ and  $\bar{Q}_2(\bm{y}_k,\bm{u}_k;\boldsymbol{\theta}_{\bar{Q}_2})$, and computes the target value for regression (see \eqref{eq:valuefunctionloss}) as:
\begin{equation*}
    \underbrace{    r_{k+1} + \gamma  \min_{i=1, 2}\bar{Q}_i(\bm{y}_{k+1},  \bm{u}_{k+1};\boldsymbol{\theta}_{\bar{Q}_i}).}_{\text{target value}}
\end{equation*}

The action-value function $Q_1(\bm{y}_k, \bm{u}_k;\boldsymbol{\theta}_{Q_1})$ is used to update the parameters of the deterministic policy $\pi(\bm{y}_k;\boldsymbol{\theta}_{\pi})$ according to the deterministic policy gradient theorem \cite{silver2014deterministic}. In particular, the gradient of the critic guides the improvements of the actor and the policy parameters are updated to ascend the action-value function:
\begin{equation*}
    \mathcal{L}(\boldsymbol{\theta}_{\pi}) = \mathbb{E}_{\bm{y}_k \sim \mathcal{M}}[-\nabla_{\bm{u}_k} Q_1(\bm{y}_k, \pi(\bm{y}_k; \boldsymbol{\theta}_{\pi}); \boldsymbol{\theta}_{Q_1})].
\label{eq:policytd3}
\end{equation*}

The target networks, parametrized by $\boldsymbol{\theta}_{\bar{Q}_1}$, $\boldsymbol{\theta}_{\bar{Q}_2}$, and $\boldsymbol{\theta}_{\bar{\pi}}$, respectively, are updated with a slower frequency than the actor and the critic according to:
\begin{equation*}
\begin{split}
\boldsymbol{\theta}_{\bar{Q}_1} &= \rho \boldsymbol{\theta}_{Q_1} + (1-\rho)\boldsymbol{\theta}_{\bar{Q}_1}\, , \\
\boldsymbol{\theta}_{\bar{Q}_2} &= \rho \boldsymbol{\theta}_{Q_2} + (1-\rho)\boldsymbol{\theta}_{\bar{Q}_1}\, , \\
\boldsymbol{\theta}_{\bar{\pi}} &= \rho \boldsymbol{\theta}_{\pi} + (1-\rho)\boldsymbol{\theta}_{\bar{\pi}}\, , \\
\end{split}
\label{target_nns_update}
\end{equation*}
where $\rho$ is a constant factor determining the speed of the updates of the target parameters.

\subsection{Hypernetworks}\label{subsec:hypernetwork}

A hypernetwork \cite{ha2016hypernetworks} is a neural network (NN) that generates the weights and biases of another NN, often referred to as main or primary network. 
Formally, a hypernetwork $h: Z \subset 
\mathbb{R}^{|\bm{z}|}\to \mathbb{R}^{|\boldsymbol{\theta}_f|}$ learns the parameters $\boldsymbol{\theta}_f$ of the main network $f:X \subset \mathbb{R}^{|\bm{x}|}\to W \subset \mathbb{R}^{|\bm{w}|}$. If we consider a standard supervised learning regression task, and we assume the availability of a dataset of $N$ input-output pairs $[(\bm{x}^{(1)}, \bm{w}^{(1)}), \cdots, (\bm{x}^{(N)}, \bm{w}^{(N)})]$ and hypernetwork inputs $[\bm{z}^{(1)}, \ldots,\bm{z}^{(N)}]$, we can write:
\begin{equation*}
\begin{split}
    \boldsymbol{\theta}_{f} &= h(\bm{z}^{(i)}; \boldsymbol{\theta}_h)\, , \\
    \hat{\bm{w}}^{(i)} &= f(\bm{x}^{(i)};\boldsymbol{\theta}_{f}) \, ,
\end{split}
\end{equation*}
where $\bm{z}$, often called context vector, can be a task-conditioned, data-conditioned, or noise-conditioned input \cite{chauhan2023brief}, and $\boldsymbol{\theta}_h$ denotes the set of parameters of the hypernetwork $h(\cdot)$.
The parameters of the two networks $\boldsymbol{\theta}_h$ and $\boldsymbol{\theta}_f$ can be updated jointly by minimizing a prescribed loss function. In the specific case of a regression task, the loss function is usually the mean-squared error between the target and predicted values:
\begin{equation*}
    \mathcal{L}(\boldsymbol{\theta}_f, \boldsymbol{\theta}_h) = \sum_{i=1}^N ||\bm{w}^{(i)}-\hat{\bm{w}}^{(i)}||_2^2.
\end{equation*}

\section{Methodology for HypeRL}\label{sec:methodology}

Given all the elements introduced in the previous section, we are now ready to set our proposed strategy to address OC of parametric dynamical systems through RL. 

\subsection{Problem Settings}

We cast the OC problem of parametric dynamical systems, introduced in Equation $\eqref{eq:OCP}$, as RL problem, where an agent, i.e., the controller, aims to learn the OC strategy by interacting with an unknown environment, governed by the dynamical system. Differently from the majority of the literature on RL for OC of dynamical systems, we focus on devising a control strategy that is adaptable to changes of the systems' dynamics deriving from variations of known parameters $\boldsymbol{\mu}$.
The RL environment is defined by a transition and reward functions:
\begin{equation*}
\begin{split}
    \bm{y}_{k+1} &= T(\bm{y}_k, \bm{u}_k; \boldsymbol{\mu})\, , \\
   r_{k+1} &= R(\bm{y}_k, \bm{u}_k;; \boldsymbol{\mu}) \, , \\
\end{split}
\end{equation*}
where the transition and reward function are assumed to be deterministic and dependent on the parameter vector $\boldsymbol{\mu}$. The parameter vector $\boldsymbol{\mu}$ might describe a physical parameter of the system, a reference value, or a navigation target.
The transition function corresponds to the fully-discretized form of the state equation (see Equation \eqref{eq:OCP}) obtained by introducing a suitable time-integration scheme over a partition of $[t_0, t_f]$ made by $N_t$ time-steps $\{t_k\}_{k=0}^{N_t}$ such that the step size is $\Delta t = (t_f - t_0)/N_t$. For example, by using an explicit Runge-Kutta scheme, we obtain the following full-order model:  
\begin{equation*}
\begin{split}
    \bm{y}_{k+1}= \underbrace{\bm{y}_{k} + \Delta t \Phi(t_{k}, \bm{y}_{k}, \bm{u}_{k}; \Delta t, \bm{F}, \boldsymbol{\mu})} _{:=  \, T(\bm{y}_{k}, \bm{u}_{k};\boldsymbol{\mu})}, 
    \qquad 
     k = 0,\ldots,N_t-1,
     \label{eq:state_update}
\end{split}
\end{equation*}
where $\bm{y}_{k} \approx \bm{y}(t_k)$ and $\bm{u}_{k} \approx \bm{u}(t_k)$, and $\Phi$ denotes the integration method’s increment function related to the state equation. The agent may observe the full state $\bm{y}_k$ or only a part of it. In addition, we assume accurate knowledge of the parameter $\boldsymbol{\mu}$. While the knowledge of physical parameters might seem a strong hypothesis, the advances of machine learning have opened new doors toward estimating such parameters from data with high accuracy, e.g., \cite{tomasetto2025reduced}. We indicate the agent's state at timestep $k$ with $\bm{z}_k=[\bm{y}_k, \boldsymbol{\mu}]$. 

With reference to Algorithm \ref{alg:RL_pseudo_code}, the agent-environment interaction scheme is organized in two loops: the outer loop indicates the training episode; instead, the inner loop refers to the number of time-steps for each episode. At the beginning of each new episode we sample a random initial condition and a PDE parameters value $\boldsymbol{\mu}$ to obtain the initial state. The agent utilizes the state at the current time-step and the PDE parameters to select a control input. The control input is fed to the transition and reward models to obtain the state at the next time-step and the reward. The tuples $(\bm{z}_k, \bm{u}_k, r_k, \bm{z}_{k+1})$ collected at each time-step $k$ of the interaction are used to train the policy (and value function) deep NN to maximize the expected cumulative reward at each episode.
\begin{algorithm}[ht!]
\caption{Episodic RL for control of parametric dynamical system}\label{alg:RL_pseudo_code}
\small
\begin{algorithmic}
\State Initialize policy and value function parameters $\boldsymbol{\theta}_{\pi}$ and $\boldsymbol{\theta}_{Q}$
\For{$e=1:E_{\max}$}
\State Sample an initial condition and input parameter $\boldsymbol{\mu}$ \State Get initial measurement $\bm{y}_k$
\For{$k=1:K_{\max}$}
\State Sample action from a policy $\bm{u}_k \sim \pi(\bm{z}_k; \boldsymbol{\theta}_{\pi}) + \boldsymbol{\epsilon}$, where $ \boldsymbol{\epsilon} \sim \mathcal{N}(\bm{0}, \boldsymbol{\sigma})$
\State Observe reward $r_{k+1}=R(\bm{y}_k, \bm{u}_k;\boldsymbol{\mu})$ and new state $\bm{y}_{k+1}=T(\bm{y}_k, \bm{u}_k;\boldsymbol{\mu})$
\If{train models}
\State Update policy and value function using the tuple $(\bm{z}_k, \bm{u}_k, r_{k+1}, \bm{z}_{k+1})$
\EndIf
\EndFor
\EndFor
\end{algorithmic}
\end{algorithm}

In this setting, we present a novel RL framework that can efficiently learn control policies for parametric PDEs from limited samples and that can generalize to new, unseen instances of the systems' parameters $\boldsymbol{\mu}$ (see Figure \ref{fig:1}). In contrast with the widely-used concatenation of information in the agent's state, to enhance the sample-efficiency and generalization capabilities of RL agents, we propose a parameter-informed HypeRL architecture relying on hypernetworks (see Section \ref{subsec:hypernetwork}). Hypernetworks allow us to express the weights and biases of the value function and policy as functions of $\bm{y}_k$ and $\boldsymbol{\mu}$. This new paradigm for encoding the information of physical and task-dependent parameters drastically improves the performance of the RL agent in terms of total cumulative reward, sample efficiency, and generalization with respect to state-of-the-art RL approaches. 

\subsection{HypeRL-TD3}

In this work, we enhance the TD3 algorithm (see Section \ref{subsec:TD3}) with hypernetworks. However, our method can be easily and directly applied to other RL algorithms, such as PPO and SAC. Analogously to the TD3 algorithm, we rely on the estimation of two action-value functions  $Q_1(\bm{z}_k, \bm{u}_k;\boldsymbol{\theta}_{Q_1})$ and $Q_2(\bm{z}_k, \bm{u}_k;\boldsymbol{\theta}_{Q_2})$ and a policy  $\pi(\bm{z}_k;\boldsymbol{\theta}_{\pi})$ by means of deep NNs. However, the parameters of the main networks are now learned using three hypernetworks $h_{Q_1}(\bm{z}_k; \boldsymbol{\theta}_{h_{Q_1}})$, $h_{Q_2}(\bm{z}_k; \boldsymbol{\theta}_{h_{Q_2}})$, and $h_{\pi}(\bm{z}_k; \boldsymbol{\theta}_{h_{\pi}})$. In this way, we are able to learn a representation of state $\bm{y}_k$ and input parameter vector $\boldsymbol{\mu}$ and encode this information in the weights and biases of the main networks. 
The \emph{hyper}-policy is defined as:
\begin{equation*}
\begin{split}
    \boldsymbol{\theta}_{\pi} &= h_{\pi}(\bm{z}_k;\boldsymbol{\theta}_{h_{\pi}})\, , \\
    \bm{u}_k &= \pi(\bm{z}_k; \boldsymbol{\theta}_{\pi})\, , \\   
\end{split}
\end{equation*}
and analogously, the \emph{hyper}-value functions as:
\begin{equation*}
\begin{split}
    \boldsymbol{\theta}_{Q_i} &= h_{Q_i}(\bm{z}_k;\boldsymbol{\theta}_{h_{Q_i}})\, , \\
    q_{i,k} &= Q_i(\bm{z}_k, \bm{u}_k; \boldsymbol{\theta}_{Q_i})\, , \qquad 
     i = 1,2 \, ,\\   
\end{split}
\end{equation*}
where $q_{i,k}$ is the predicted Q-value by the action-value function $Q_i(\cdot, \cdot)$.
The hypernetwork parameters are jointly optimized with the main network parameters by simply allowing the gradient of the loss functions to flow through the hypernetworks. Policy and value functions are updated using the TD3 training objectives:
\begin{equation*}
    \begin{split}
             \mathcal{L}(\boldsymbol{\theta}_{\pi}, \boldsymbol{\theta}_{h_{\pi}}) &= \mathbb{E}_{\bm{z}_k \sim \mathcal{M}}[-\nabla_{\bm{u}_k} Q_1(\bm{z}_k, \pi(\bm{z}_k; \boldsymbol{\theta}_{\pi}); \boldsymbol{\theta}_{Q_1})]\, , \\
              \mathcal{L}(\boldsymbol{\theta}_{Q_i}, \boldsymbol{\theta}_{h_{Q_i}}) &= \mathbb{E}_{\bm{z}_k, \bm{u}_k, r_{k+1},  \bm{z}_{k+1} \sim \mathcal{M}}[\texttt{Huber}(r_{k+1} + \gamma \min_{i=1, 2}\bar{Q}_i(\bm{z}_{k+1}, \bm{u}_{k+1}; \boldsymbol{\theta}_{\bar{Q}_i}) - Q_i(\bm{z}_k, \bm{u}_k; \boldsymbol{\theta}_{Q_i}))]\, , \\
    \end{split}
\end{equation*}
where, similarly to \cite{fujimoto2023sale}, we found beneficial to the agent's performance replacing the mean-squared error with the \texttt{Huber} loss to update the hyper-value function.

In Algorithm \ref{alg:HypeRL TD3}, we show the HypeRL-TD3 pseudo code, where we highlight in blue the changes to the original TD3 algorithm. 
\begin{algorithm}[h!]
\caption{HypeRL-TD3}\label{alg:HypeRL TD3}
\small
\begin{algorithmic}
\State
\State {\color{blue}Initialize hypernetworks $h_{Q_1}(\cdot; \boldsymbol{\theta}_{h_{Q_1}})$, $h_{Q_2}(\cdot; \boldsymbol{\theta}_{h_{Q_2}})$, and $h_{\pi}(\bm{z}; \boldsymbol{\theta}_{h_{\pi}})$
with random parameters $\boldsymbol{\theta}_{h_{Q_1}}, \boldsymbol{\theta}_{h_{Q_2}}, \boldsymbol{\theta}_{h_{\pi}}$
\State Initialize target hypernetworks $\boldsymbol{\theta}_{h_{\bar{Q}_1}} \leftarrow \boldsymbol{\theta}_{h_{Q_1}}$, $\boldsymbol{\theta}_{h_{\bar{Q}_2}} \leftarrow \boldsymbol{\theta}_{h_{Q_2}}$, $\boldsymbol{\theta}_{h_{\bar{\pi}}} \leftarrow \boldsymbol{\theta}_{h_{\pi}}$}
\State Initialize main networks  $Q_1(\cdot,\cdot;\boldsymbol{\theta}_{Q_1})$, $Q_2(\cdot,\cdot;\boldsymbol{\theta}_{Q_2})$, and $\pi(\cdot; \boldsymbol{\theta}_{\pi})$
\State Initialize target networks  $\bar{Q}_1(\cdot,\cdot;\boldsymbol{\theta}_{\bar{Q}_1})$, $\bar{Q}_2(\cdot,\cdot;\boldsymbol{\theta}_{\bar{Q}_2})$, and $\bar{\pi}(\cdot; \boldsymbol{\theta}_{\bar{\pi}})$
\State Initialize memory buffer $\mathcal{M}$
\State
\For{$e=1:E_{\max}$}
\State Initialize the system and get initial measurement $\bm{y}_k,\boldsymbol{\mu}$
\State
\For{$t=1:T_{\max}$}
\State {\color{blue} Set $\bm{z}_k=[\bm{y}_k, \boldsymbol{\mu}]$}
\State {\color{blue}Sample policy parameters $\boldsymbol{\theta}_{\pi}=h_{\pi}(\bm{z}_k; \boldsymbol{\theta}_{h_{\pi}})$}
\State Sample action $\bm{u}_k \sim \pi(\bm{z}_k; \boldsymbol{\theta}_{\pi}) + \boldsymbol{\epsilon}$, where $ \boldsymbol{\epsilon} \sim \mathcal{N}(\bm{0}, \boldsymbol{\sigma})$
\State Observe reward $r_{k+1}$ and new state $\bm{y}_{k+1}$
\State Store tuple $(\bm{z}_k, \bm{u}_k, r_{k+1}, \bm{z}_{k+1})$ in $\mathcal{M}$
\State
\If{train models}
\State Sample mini-batch $(\bm{z}_k, \bm{u}, r_{k+1}, \bm{z}_{k+1})$ from $\mathcal{M}$
\State {\color{blue}Sample target policy parameters $\boldsymbol{\theta}_{\bar{\pi}}=h_{\bar{\pi}}(\bm{z}_{k+1}; \boldsymbol{\theta}_{h_{\bar{\pi}}})$}
\State $\bm{u}_{k+1} \leftarrow \bar{\pi}(\bm{z}_{k+1}; \boldsymbol{\theta}_{\bar{\pi}}) +  \boldsymbol{\epsilon}$, where $\boldsymbol{\epsilon} \sim \text{clip}(\mathcal{N}(\bm{0}, \bar{\boldsymbol{\sigma}}), -c, c)$
\State  {\color{blue}Sample target value functions parameters $\boldsymbol{\theta}_{\bar{Q}_1}=h_{\bar{Q}_1}(\bm{z}_k; \boldsymbol{\theta}_{h_{\bar{Q}_1}})$ and $\boldsymbol{\theta}_{\bar{Q}_2}=h_{\bar{Q}_2}(\bm{z}_k; \boldsymbol{\theta}_{h_{\bar{Q}_2}})$} 
\State $q^k \leftarrow r_{k+1} + \gamma  \min_{i=1, 2}\bar{Q}(\bm{z}_{k+1},  \bm{u}_{k+1}; \boldsymbol{\theta}_{\bar{Q}_i})$
\State  {\color{blue}Sample value functions parameters $\boldsymbol{\theta}_{Q_1}=h_{\bar{Q}_1}(\bm{z}_k; \boldsymbol{\theta}_{h_{Q_1}})$ and $\boldsymbol{\theta}_{Q_2}=h_{Q_2}(\bm{z}_k; \boldsymbol{\theta}_{h_{Q_2}})$} 
\State Update critic parameters {\color{blue} and hypernetworks parameters} according:
\State $\mathcal{L}(\boldsymbol{\theta}_{Q_i}, {\color{blue}\boldsymbol{\theta}_{h_{Q_i}}}) = \mathbb{E}_{\bm{z}_k, \bm{u}_k, r_{k+1}, \bm{z}_{k+1} \sim \mathcal{M}}[ {\color{blue}\texttt{Huber}}(q_k - Q_i(\bm{z}_k, \bm{u}_k; \boldsymbol{\theta}_{Q_i}))]$ \ \ \   with  \ \ \ $i \in \{1, 2\}$
\State
\If{train actor}
\State {\color{blue}Sample policy parameters $\boldsymbol{\theta}_{\pi}=h_{\pi}(\bm{z}_k; \boldsymbol{\theta}_{h_{\pi}})$}
\State Update policy parameters {\color{blue} and hypernetworks parameters} according to:
\State $    \mathcal{L}(\boldsymbol{\theta}_{\pi}, {\color{blue}\boldsymbol{\theta}_{h_{\pi}}}) = \mathbb{E}_{\bm{z}_k \sim \mathcal{M}}[-\nabla_{\bm{u}_k} Q_1(\bm{z}_k, \pi(\bm{z}_k; \boldsymbol{\theta}_{\pi}); \boldsymbol{\theta}_{Q_1})]$
\State Update target networks {\color{blue} by updating the hypernetworks parameters:
\State $\boldsymbol{\theta}_{h_{\bar{Q}_1}} = \rho \boldsymbol{\theta}_{h_{Q_1}} + (1-\rho)\boldsymbol{\theta}_{h_{\bar{Q}_1}}$
\State $\boldsymbol{\theta}_{h_{\bar{Q}_2}} = \rho \boldsymbol{\theta}_{h_{Q_2}} + (1-\rho)\boldsymbol{\theta}_{h_{\bar{Q}_2}}$
\State $\boldsymbol{\theta}_{h_{\bar{\pi}}} = \rho \boldsymbol{\theta}_{h_{\pi}} + (1-\rho)\boldsymbol{\theta}_{h_{\bar{\pi}}}$}
\EndIf
\EndIf
\EndFor
\EndFor
\end{algorithmic}
\end{algorithm}

\subsubsection{Neural Network Architectures}\label{subsec:HypeRL_architecture}
\begin{figure}[h!]
    \centering
    \includegraphics[width=0.8\linewidth]{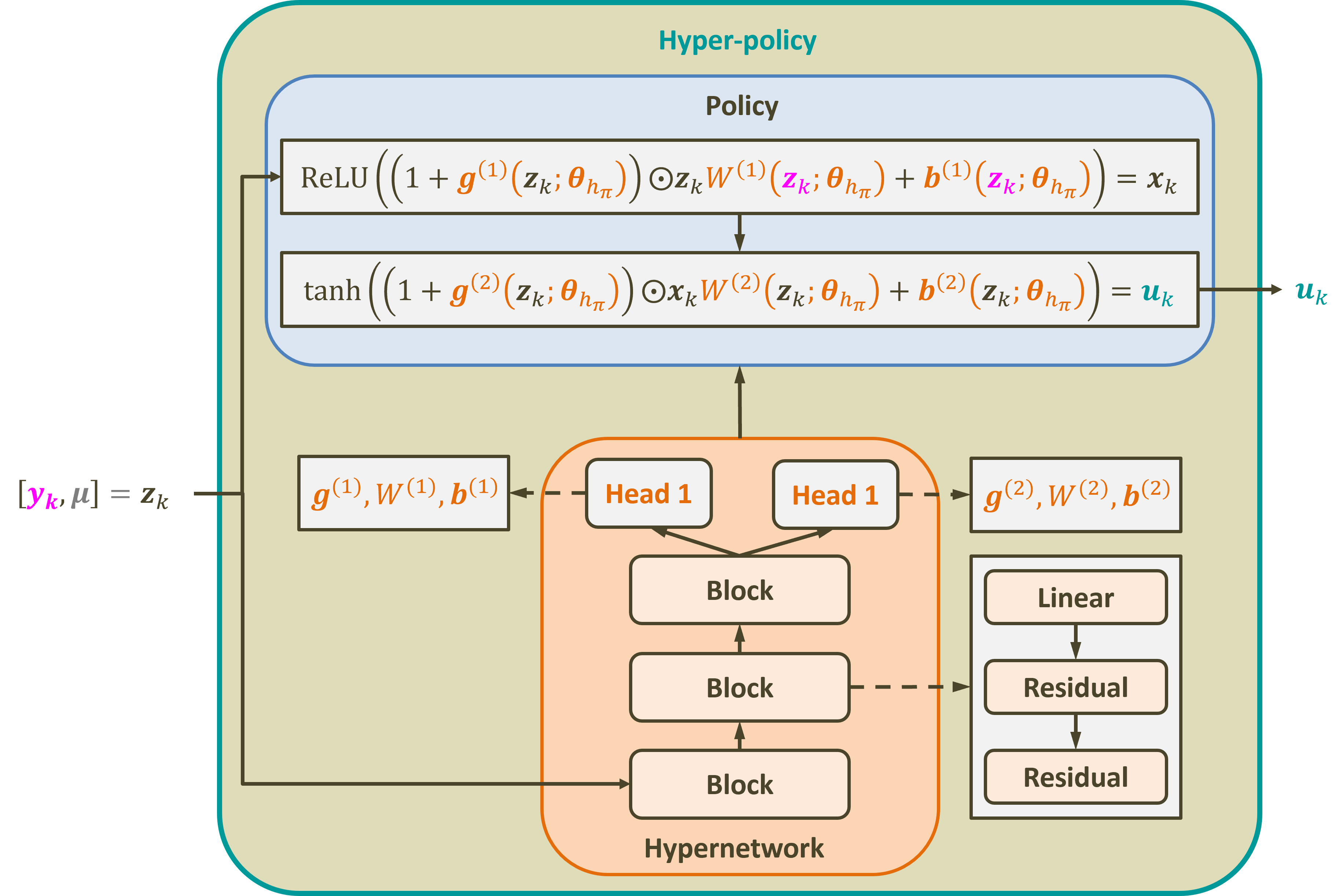}
    \caption{Hyper-policy architecture. }
    \label{fig:hyperpolicy}
\end{figure}

The main policy network $\pi(\bm{z}_k;\boldsymbol{\theta}_{\pi})$ is composed of one input layer of dimension equal to the agent's state dimension $\bm{z}_k \in \mathbb{R}^{|\bm{y}|+|
\boldsymbol{\mu}|}$, one hidden layer with $256$ neurons, and an output layer of dimension equal to the control action dimension $|\bm{u}_k| \in \mathbb{R}^{|\bm{u}|}$.  In particular, we can write the main policy network as:
\begin{equation*}
    \begin{split}
        \bm{x}_k &= \texttt{ReLU}\big((1+\bm{g}^{(1)}(\bm{z}_k;\boldsymbol{\theta}_{h_\pi}))\odot \bm{z}_k W^{(1)}(\bm{z}_k;\boldsymbol{\theta}_{h_\pi})+\bm{b}^{(1)}(\bm{z}_k;\boldsymbol{\theta}_{h_\pi}) \big)\, , \\
        \bm{u}_k &= \texttt{tanh}\big((1+\bm{g}^{(2)}(\bm{z}_k;\boldsymbol{\theta}_{h_\pi})\odot \bm{x}_k W^{(2)}(\bm{z}_k;\boldsymbol{\theta}_{h_\pi})+\bm{b}^{(2)}(\bm{z}_k;\boldsymbol{\theta}_{h_\pi}) \big)\, , \\
    \end{split}
\end{equation*}
where $\bm{g}^{(1)}, W^{(1)}, \bm{b}^{(1)}$ and $\bm{g}^{(2)}, W^{(2)}, \bm{b}^{(2)}$ are respectively the gain vectors, the weights matrices, and the biases of the two layers of the main networks that are learned via the hypernetwork $h(\bm{z}_k;\boldsymbol{\theta}_{h_\pi})$, i.e., the hypernetworks outputs, $\boldsymbol{\theta}_{\pi}=[\bm{g}^{(1)}, W^{(1)}, \bm{b}^{(1)},\bm{g}^{(2)}, W^{(2)}, \bm{b}^{(2)}]$, and $\odot$ indicates the element-wise product. This type of architecture is motivated by the work in \cite{littwin2019deep} and it is depicted in Figure \ref{fig:hyperpolicy}.
Similarly to the policy, the main value function networks $Q_1(\bm{z}_k, \bm{u}_k;\boldsymbol{\theta}_{Q_1})$ and $Q_2(\bm{z}_k, \bm{u}_k;\boldsymbol{\theta}_{Q_2})$ are composed of one input layer of dimension $|\bm{z}_k| + |\bm{u}_k|$, one hidden layer with $256$ neurons, and an output layer of  dimension 1. By indicating with $\boldsymbol{\upsilon}_k=[\bm{z_k,\bm{u}_k}]$, we can write: 
\begin{equation*}
    \begin{split}
        \boldsymbol{\chi}_k &= \texttt{ReLU}\big((1+\bm{g}^{(1)}(\bm{z}_k;\boldsymbol{\theta}_{h_{Q_i}}))\odot \boldsymbol{\upsilon}_k W^{(1)}(\bm{z}_k;\boldsymbol{\theta}_{h_{Q_i}})+\bm{b}^{(1)}(\bm{z}_k;\boldsymbol{\theta}_{h_{Q_i}}) \big)\, , \\
        q_{i,k} &= (1+\bm{g}^{(2)}(\bm{z}_k;\boldsymbol{\theta}_{h_{Q_i}})\odot \boldsymbol{\chi}_k W^{(2)}(\bm{z}_k;\boldsymbol{\theta}_{h_{Q_i}})+\bm{b}^{(2)}(\bm{z}_k;\boldsymbol{\theta}_{h_{Q_i}})\, . \\
    \end{split}
\end{equation*}
The hypernetworks use the same architecture and weight initialization proposed in \cite{sarafian2021recomposing}. Each hypernetwork is composed of three blocks that map $\bm{z}_k$ to $\bm{g}^{(1)}, W^{(1)}, \bm{b}^{(1)}$ and $\bm{g}^{(2)}, W^{(2)}, \bm{b}^{(2)}$. Each block is composed of a linear layer and two residual blocks and each residual block is composed of two linear layers with \texttt{ReLU} activation and a skip connection.

\section{Numerical Experiments}\label{sec:numerical_experiments}

We validate our proposed approach on two control baselines, namely  (\emph{i}) the stabilization of a parametric Kuramoto-Sivashinsky equation with distributed in-domain actuators and sensors, and (\emph{ii}) a 2D navigation problem in a gyre flow, where the goal is to control the velocity of particle to reach arbitrary targets.

We compare the HypeRL-TD3 agent with:
\begin{enumerate}
    \item TD3. The algorithm uses the default and widely-used architecture for policy and value functions \cite{fujimoto2018addressing} composed of an input layer, two hidden layers with $256$ neurons each and an output layer. Analogously to HypeRL-TD3, the input of the policy networks is $\bm{z}_k$ and the input of the value function networks is $[\bm{z}_k, \bm{u}_k]$.
    \item TD3 (\texttt{Huber}). This is a variant of TD3 where the \texttt{Huber} loss replaces the mean-squared error loss to train the critic networks to match the objective used by HypeRL-TD3.
    \item TD3 (2 layers). This a variant of TD3 where a single hidden layer is used to match the HypeRL-TD3 main policy and value function networks.
    \item TD3 (no $\boldsymbol{\mu}$). We remove the knowledge of the parameter vector $\boldsymbol{\mu}$ to TD3 to show the need for making the agents parameter-informed in parametric control problems. 
\end{enumerate}
In all our numerical experiments, each agent is trained for $2000$ episodes. The evaluation occurs every $200$ training episodes and we apply a warmup of $100$ episodes in which the agents' actions are sampled from a uniform distribution $\in \mathcal{U}(-\bm{u}_{\min}, \bm{u}_{\max})$. The evaluation metric is the average cumulative reward of $10$ evaluation episodes in which the optimal deterministic policy is deployed.
 
\subsection{Stabilization of a 1D Kuramoto-Sivashinsky Equation}
\begin{figure*}
    \begin{minipage}{0.49\linewidth}
    \centering \subfloat{\includegraphics[height=0.55\textwidth]{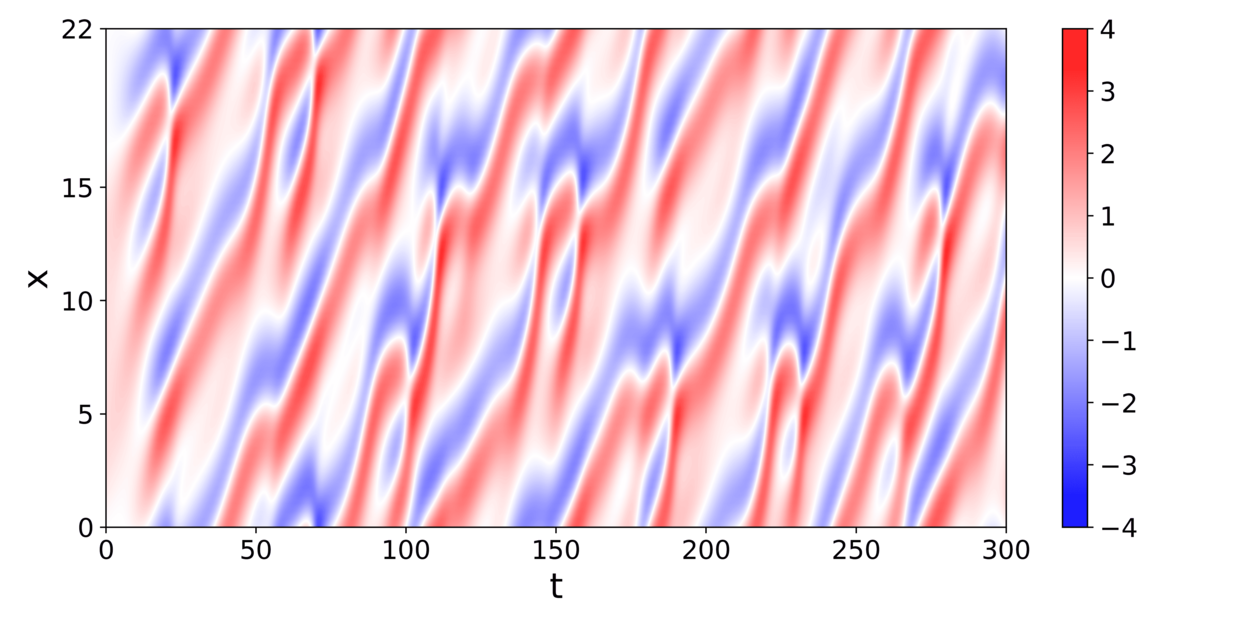}}
    \vspace{-0.2cm}
    \subfloat{\includegraphics[height=0.55\textwidth]{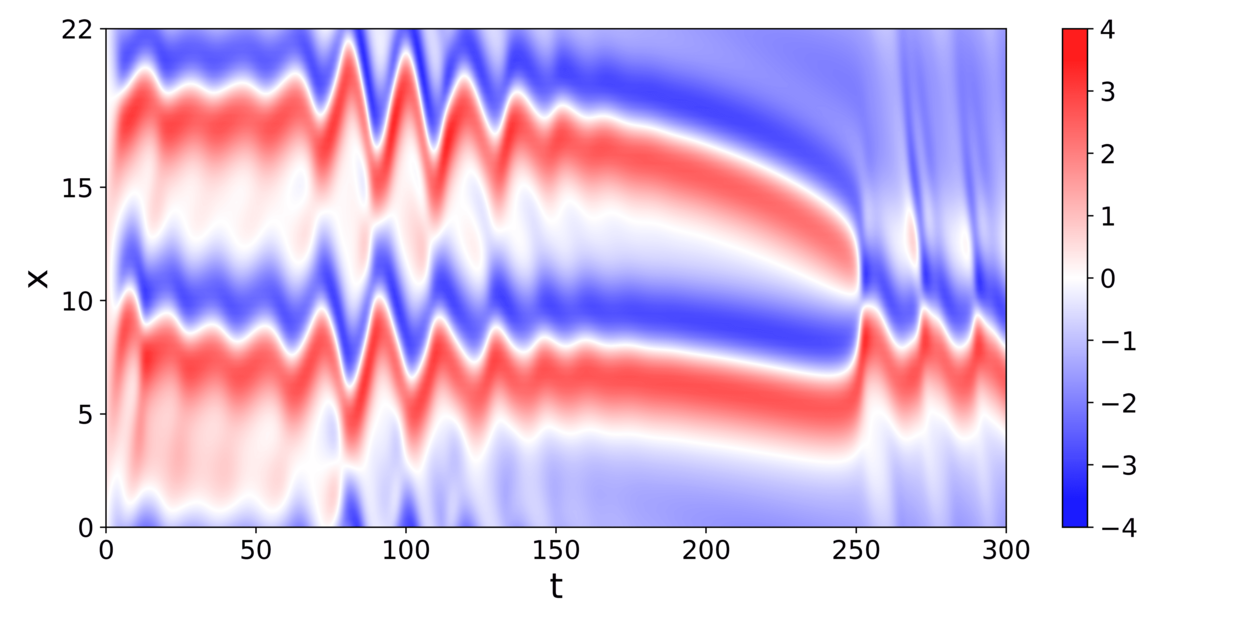}}
    \end{minipage}
    \begin{minipage}{0.49\linewidth}
    \centering \subfloat{\includegraphics[height=0.55\textwidth]{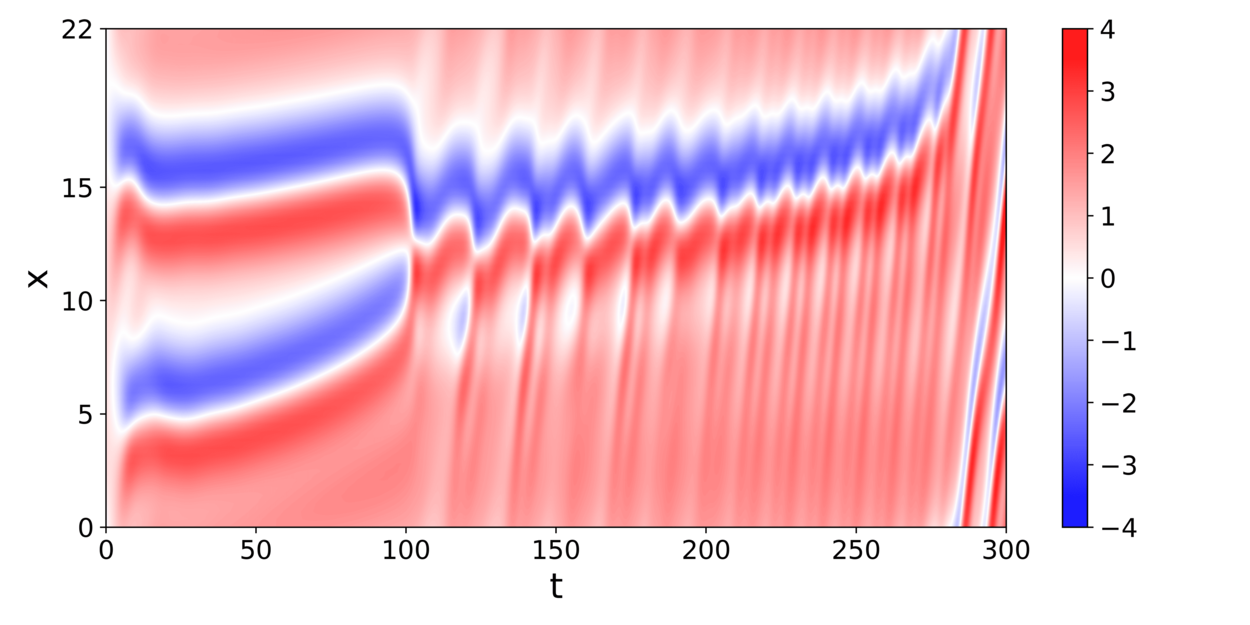}}
    \vspace{-0.2cm}
    \subfloat{\includegraphics[height=0.55\textwidth]{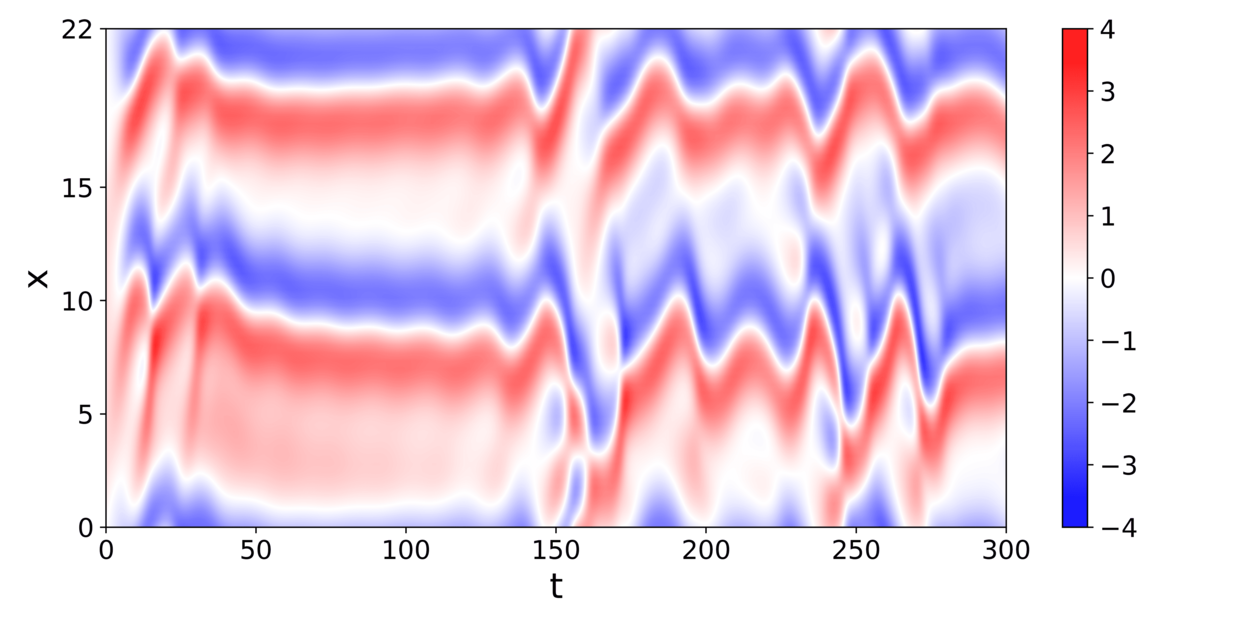}}
    \end{minipage}
    \caption{Solutions of the uncontrolled KS equation for different values of $\nu$, namely, $\nu=0$ (top left), $\nu=0.2$ (top right), $\nu=-0.25$ (bottom left), and $\nu=-0.1$ (bottom right).}
    \label{fig:KS_nu}
\end{figure*}

The Kuramoto-Sivashinsky (KS) equation is a nonlinear 1D PDE describing pattern and instability in fluid dynamics, plasma physics, and combustion, e.g., the diffusive-thermal instabilities in a laminar flame front \cite{kudryashov1990exact}. Similarly to \cite{peitz2023distributed, botteghi2024parametric}, we write the KS PDE with state $s(x,t)$ and forcing term $u(x,t)$ with the addition of a parametric cosine term, breaking the spatial symmetries of the equation and making the search for the optimal control policy more challenging:
\begin{equation}
\begin{split}
     \frac{\partial s(x,t)}{\partial t} + s\frac{\partial s(x,t)}{\partial x} + \frac{\partial^2 s(x,t)}{\partial x^2} + \frac{\partial^4 s(x,t)}{\partial x^4} + \nu \cos{\Big(\frac{4\pi x}{L}\Big)} &=  u(x,t)\, , \\
     u(x,t) &= \sum_{i=1}^{N_a} a_i(t) \psi(x, m_i) \, , \\
     \psi(x, m_i) &=\frac{1}{2}\exp\Big(-{\Big(\frac{x-m_i}{\sigma}\Big)}^2\Big)\, , \\
\end{split}
\label{eq:ks}
\end{equation}
where $u$ is the time-dependent control input function with $a_i(t) \in [-1, 1]$, $\psi(x, c_i)$ is a Gaussian kernel of mean $c_i$ and standard deviation $\sigma=0.8$, $\nu \in [-0.25, 0.25]$ is the parameter of interest of the system, and $\mathcal{D}=[0, 22]$ is the spatial domain with periodic boundary conditions. We numerically solve the KS equation for $T=300$s with a $dt=0.1$ and we discretize the spatial domain with $N_{x}=64$, leading to $s(x,t_k)\mapsto \bm{y}_k\in\mathbb{R}^{N_x}$ for each time instance $t_k$. In Figure \ref{fig:KS_nu} we show examples of solution for different values of the parameter $\nu$. It is worth highlighting that small variations of $\nu$ lead to vastly different solutions.
We assume to have $N_a=8$ equally spaced actuators.  We evolve the KS equation for $100$ timesteps before activating the controller.  Similar settings were used in \cite{peitz2023distributed, botteghi2024parametric}. However, instead of looking for a  controller that solely stabilizes the KS state to the zero reference, i.e., $\bm{y}_{\text{ref}}=\bm{0}$, we consider two more challenging control problems:
\begin{enumerate}
    \item stabilization of the state of a KS equation to an arbitrary reference $\in \mathcal{U}(-3.0, 3.0)$ with $\nu=0.0$, and
    \item stabilization of the state of a parametric KS equation to an arbitrary reference $\in \mathcal{U}(-3.0, 3.0)$ with $\nu \in \mathcal{U}(-0.25, 0.25)$\, ,
\end{enumerate}
where $\mathcal{U}$ indicates the uniform distribution. In both cases, the agent's state $\bm{z}_k=[\bm{y}_k, \boldsymbol{\mu}]$, where we indicate with $\boldsymbol{\mu}=[\bm{y}_{\text{ref}}, \nu]$.

The reward function corresponds to the fully-discretized (in space and time) counterpart of the running cost in Equation \eqref{eq:L} with opposite sign. Our goal is to steer the measured state $\bm{y}_k$ to a reference value $\bm{y}_{\text{ref}}$ with minimal control effort. Therefore, we utilize the following reward function:
\begin{equation}
    R(\bm{y}_k, \bm{u}_k; \boldsymbol{\mu}) = -
\underbrace{||\bm{y}_k - \bm{y}_{\text{ref}}||_2^2}_{\text{state cost}} -  \underbrace{\alpha ||\bm{u}_k||_2^2}_{\text{action cost}} \, ,
\label{eq:reward_function_pde}
\end{equation}
where $\bm{y}_{\text{ref}}$ indicates the reference values for the state, and $\alpha=0.1$ is a scalar positive coefficient balancing the contribution of the two terms.

\subsubsection{Stabilization of the State of a KS Equation to an Arbitrary Reference}

In this first example, the goal is to steer the state of a KS equation to an arbitrary reference that is randomly sampled from a uniform distribution in $\in  \mathcal{U}(-3.0, 3.0)$ at the beginning of each training and evaluation episode.

\begin{figure}[h!]
     \centering
     \begin{subfigure}[b]{0.49\textwidth}
         \centering
         \includegraphics[width=\textwidth]{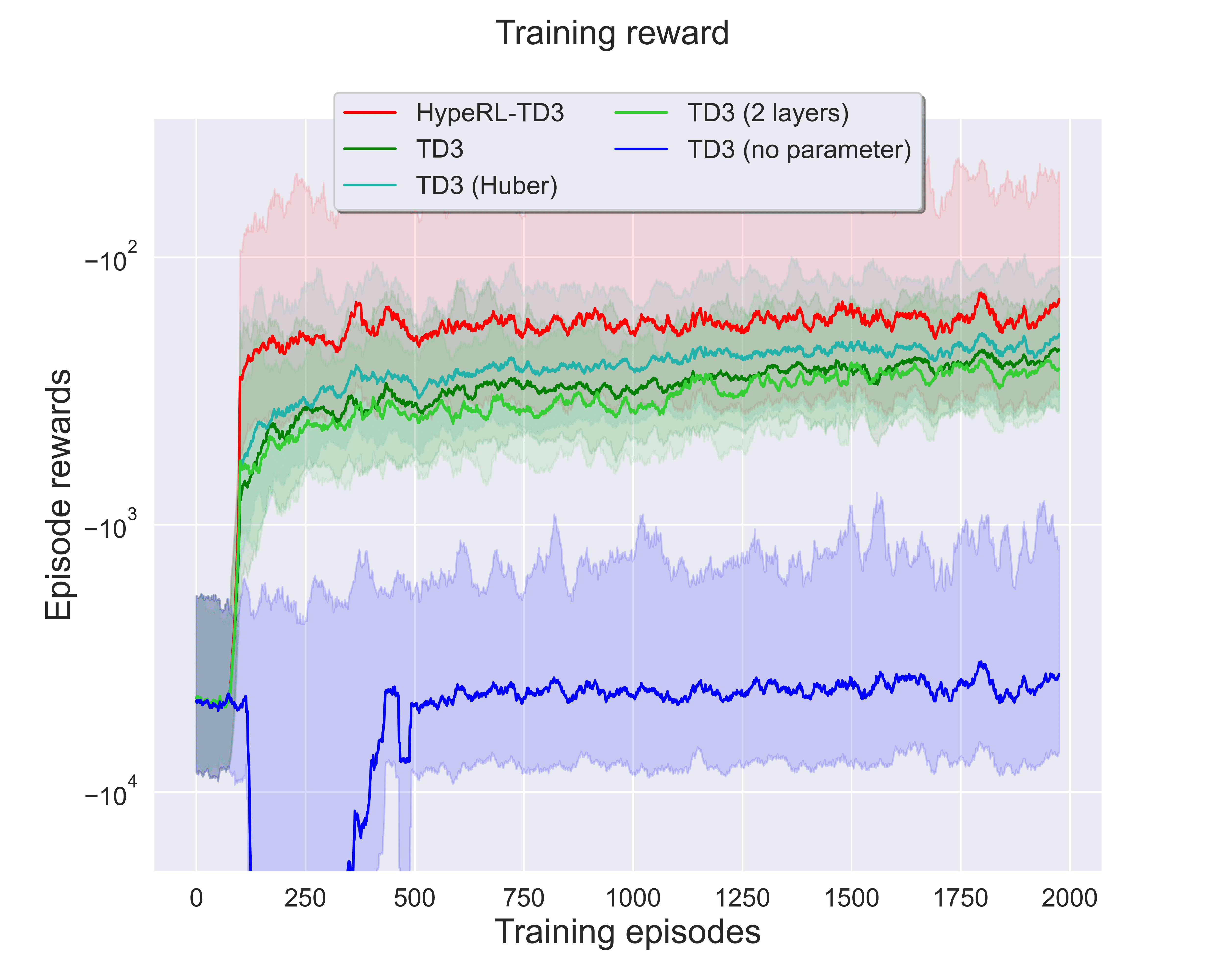}
         \caption{}
         \label{fig:ks_train_rewards_ks}
     \end{subfigure}
          \begin{subfigure}[b]{0.49\textwidth}
         \centering
         \includegraphics[width=\textwidth]{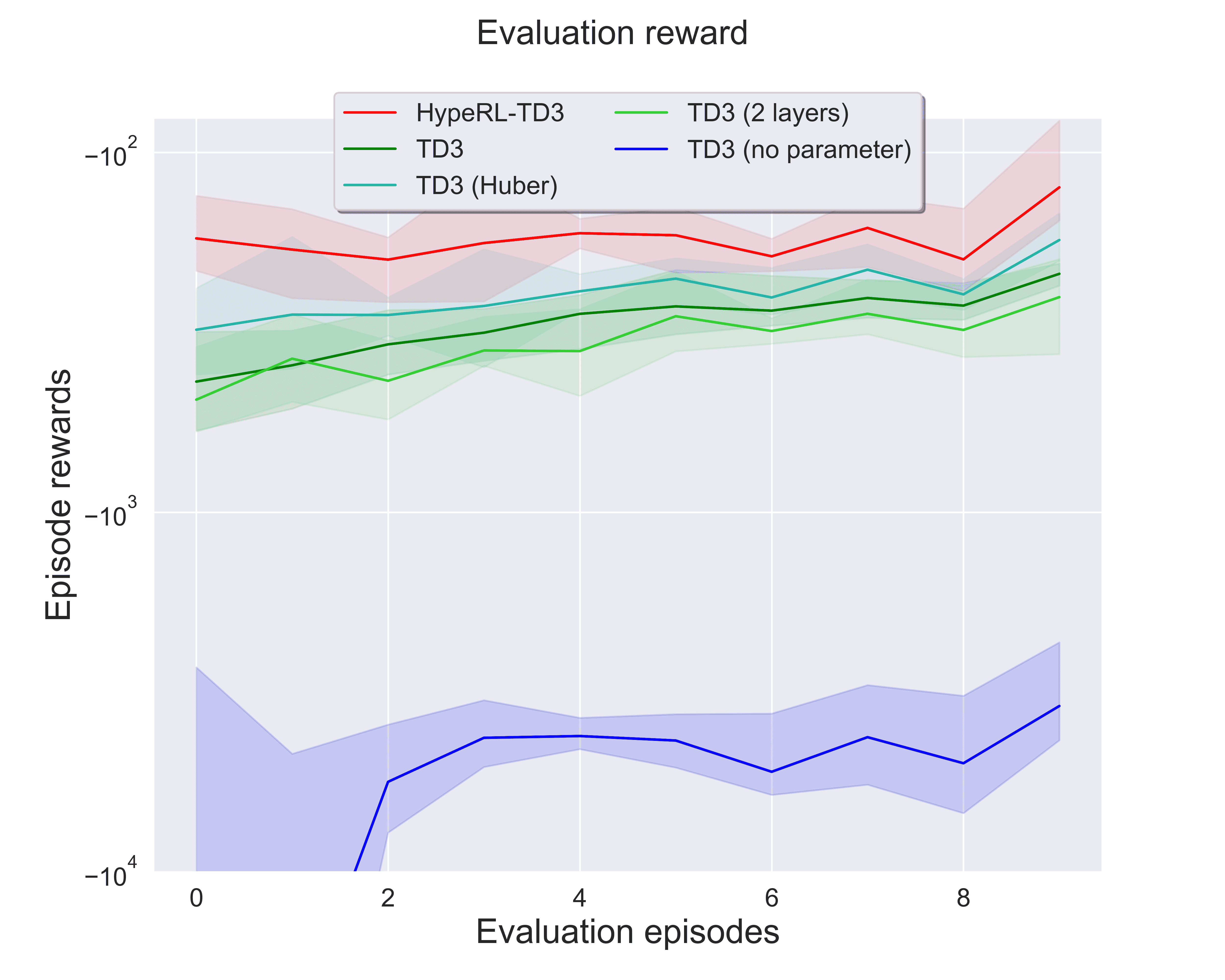}
         \caption{}
         \label{fig:ks_eval_rewards_ks}
     \end{subfigure}
        \caption{Stabilization of the state of a KS equation to an arbitrary reference -- training and evaluation results. The solid line represents the mean and the shaded area the minimum and maximum values observed over 5 different random seeds. }
        \label{fig:2}
\end{figure}

In Figure \ref{fig:2}, we show the mean and the standard deviation of the cumulative reward during training and evaluation, respectively. It is possible to notice the superior performance of the hypernetwork-based agent in early and later stages of the training and during evaluation.

In Table \ref{tab:KS_res}, we show the training and evaluation rewards collected by the different agents over 5 different random seeds, where we highlight in bold the highest scores of the training and evaluation. In the calculation of these scores, we remove for every agent the rewards collected during $100$ warmup episodes as common to each agent.
\begin{table}[ht!]
    \caption{Mean and standard deviation of the cumulative reward over training (left) and evaluation (right) collected by the different algorithms. The results report the average performance over 5 different random seeds.}
    \label{tab:KS_res}
    \begin{minipage}{.5\linewidth}
      \centering
\begin{tabular}{|| c | c ||} 
 \hline
Average cumulative reward & mean $\pm$ std \\ 
 \hline\hline
 HypeRL-TD3 & $\bm{-180.41 \pm 62.97}$ \\ 
 \hline
TD3 & $-313.17 \pm 116.74$  \\
 \hline
 TD3 (Huber) & $-263.78 \pm 93.49$  \\  
 \hline
  TD3 (2 layers) & $-344.13 \pm  119.91$  \\  
 \hline
  TD3 (no $\boldsymbol{\mu}$) & $-12522.60 \pm 25874.76$   \\ 
 \hline
\end{tabular}
    \end{minipage}%
    \begin{minipage}{.5\linewidth}
      \centering
\begin{tabular}{|| c | c ||} 
 \hline
Average cumulative reward & mean $\pm$ std \\ 
 \hline\hline
 HypeRL-TD3 & $\bm{-175.33 \pm 20.85}$ \\ 
 \hline
TD3 & $-304.27 \pm 63.11$  \\
 \hline
 TD3 (Huber) & $-249.76 \pm 37.31$  \\  
 \hline
  TD3 (2 layers) & $-344.52 \pm  68.47$  \\  
 \hline
  TD3 (no $\boldsymbol{\mu}$) & $-17473.19 \pm 31416.09$   \\ 
 \hline
\end{tabular}
    \end{minipage} 
\end{table}
The results show that the HypeRL-TD3 outperforms the TD3 agents, showing that the way state and parametric information is encoded are crucial ingredients for sample efficiency and generalization of the RL algorithms.

\begin{figure*}[h!]
    \begin{minipage}{0.49\linewidth}
    \centering \subfloat{\includegraphics[height=0.85\textwidth]{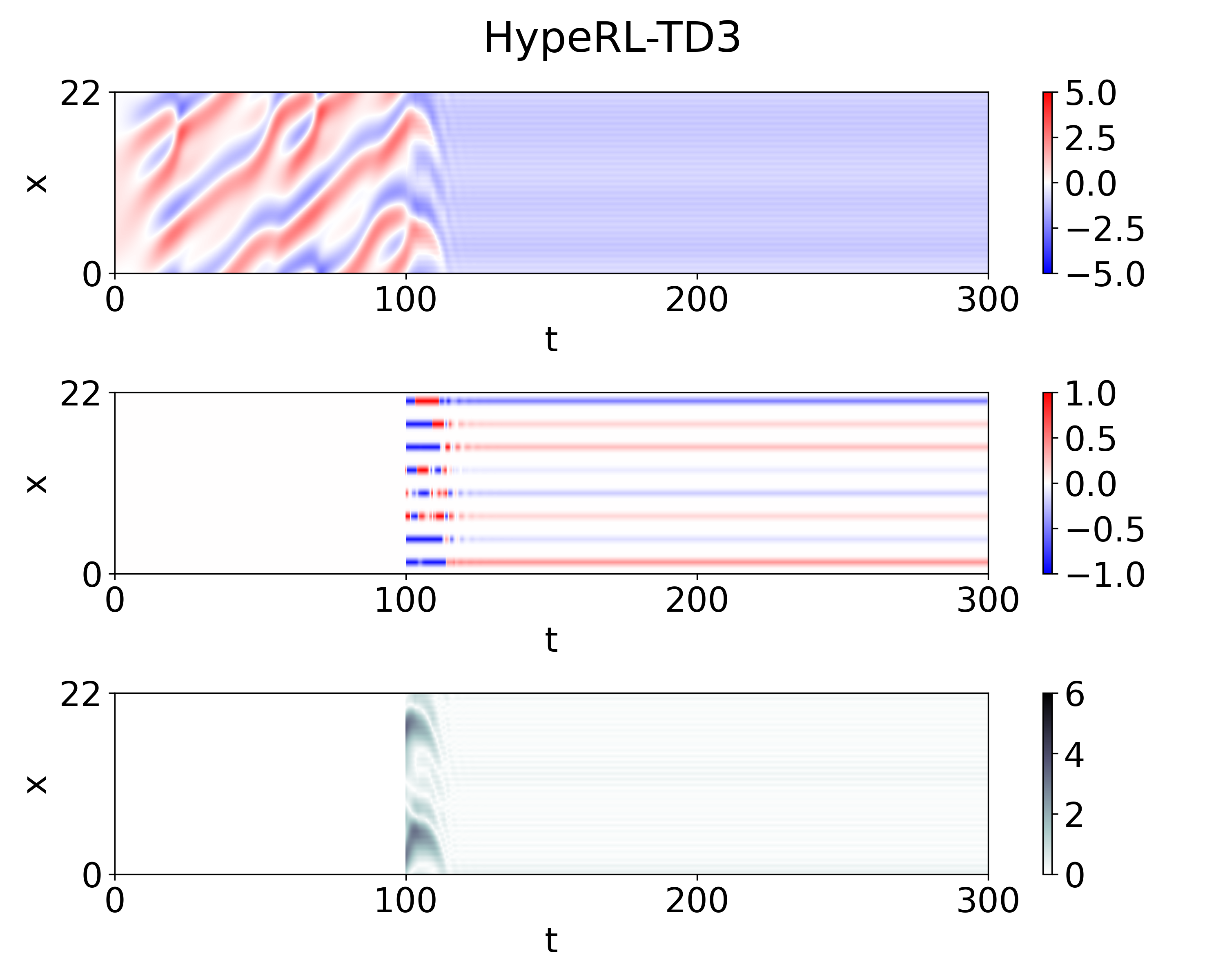}}
    \vspace{-0.1cm}
    \subfloat{\includegraphics[height=0.85\textwidth]{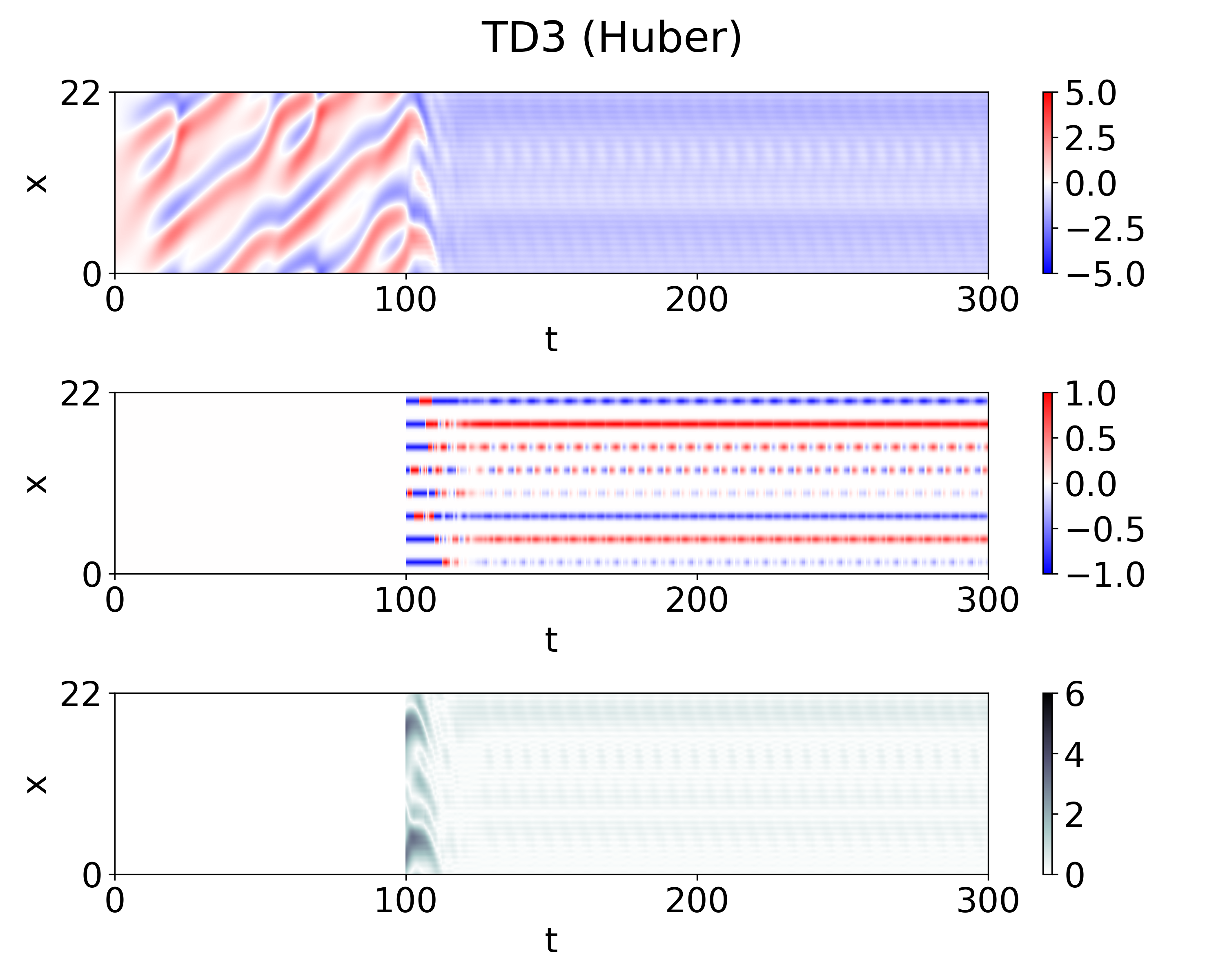}}
    \end{minipage}
    \begin{minipage}{0.49\linewidth}
    \centering \subfloat{\includegraphics[height=0.85\textwidth]{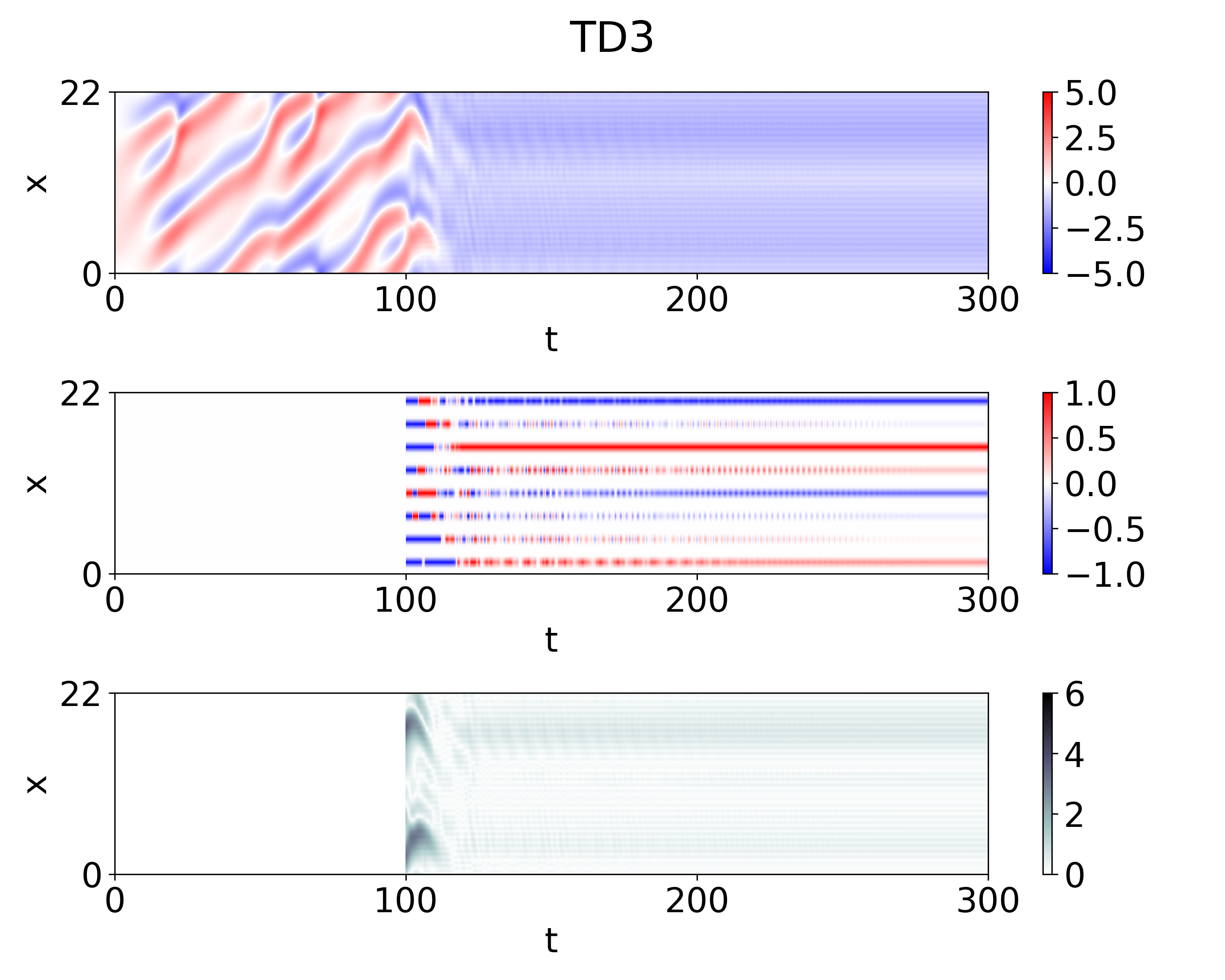}}
    \vspace{-0.1cm}
    \subfloat{\includegraphics[height=0.85\textwidth]{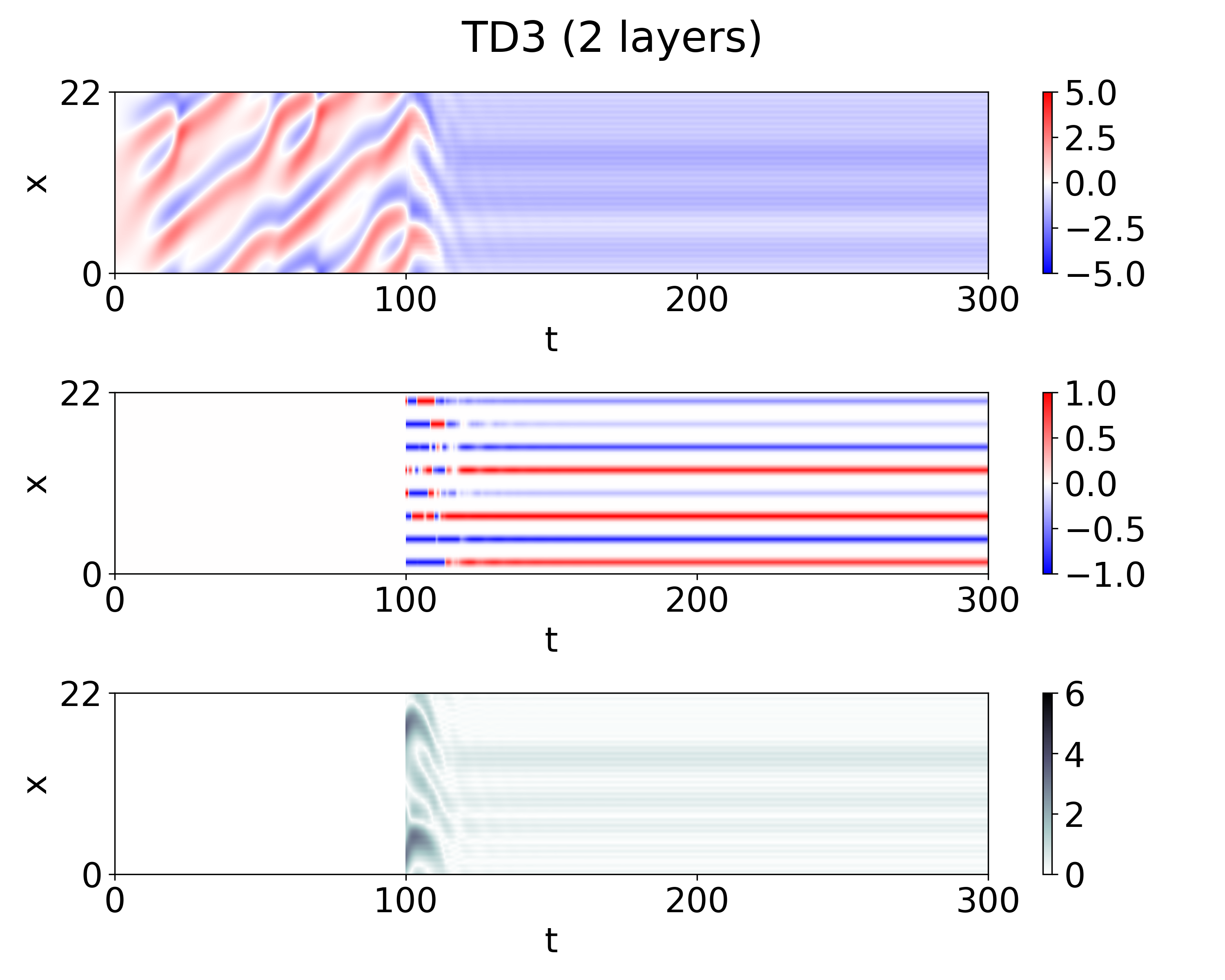}}
    \end{minipage}
    \caption{Controlled solutions of the KS equation when using the different agents. The first row shows the state, the second the control, and the third one the error $|\bm{y}_k - \bm{y}_{\text{ref}}|$.}
    \label{fig:KS_controlled}
\end{figure*}

In Figure \ref{fig:KS_controlled}, we show an example of controlled solution for $\bm{y}_{\text{ref}}=-\bm{1}$, where the first row depicts the evolution of the state, the second one of the control, and the third one of the state-tracking error $|\bm{y}_k - \bm{y}_{\text{ref}}|$. Additional results, including the performance of the TD3 agent without access to $\boldsymbol{\mu}$, and state and action costs over training and evaluation, as defined in Equation \eqref{eq:reward_function_pde}, are in Appendix \ref{app:KS_extra_results}.

\subsubsection{Stabilization of the State of a Parametric KS Equation to an Arbitrary Reference}
In this second example, the goal is to steer the state of a \emph{parametric} KS equation to an arbitrary reference that is randomly sampled from a uniform distribution in $\in  \mathcal{U}(-3.0, 3.0)$ at the beginning of each training and evaluation episode together with the parameter $\nu \in \mathcal{U}(-0.25, 0.25)$.

\begin{figure}[h!]
     \centering
     \begin{subfigure}[b]{0.49\textwidth}
         \centering
         \includegraphics[width=\textwidth]{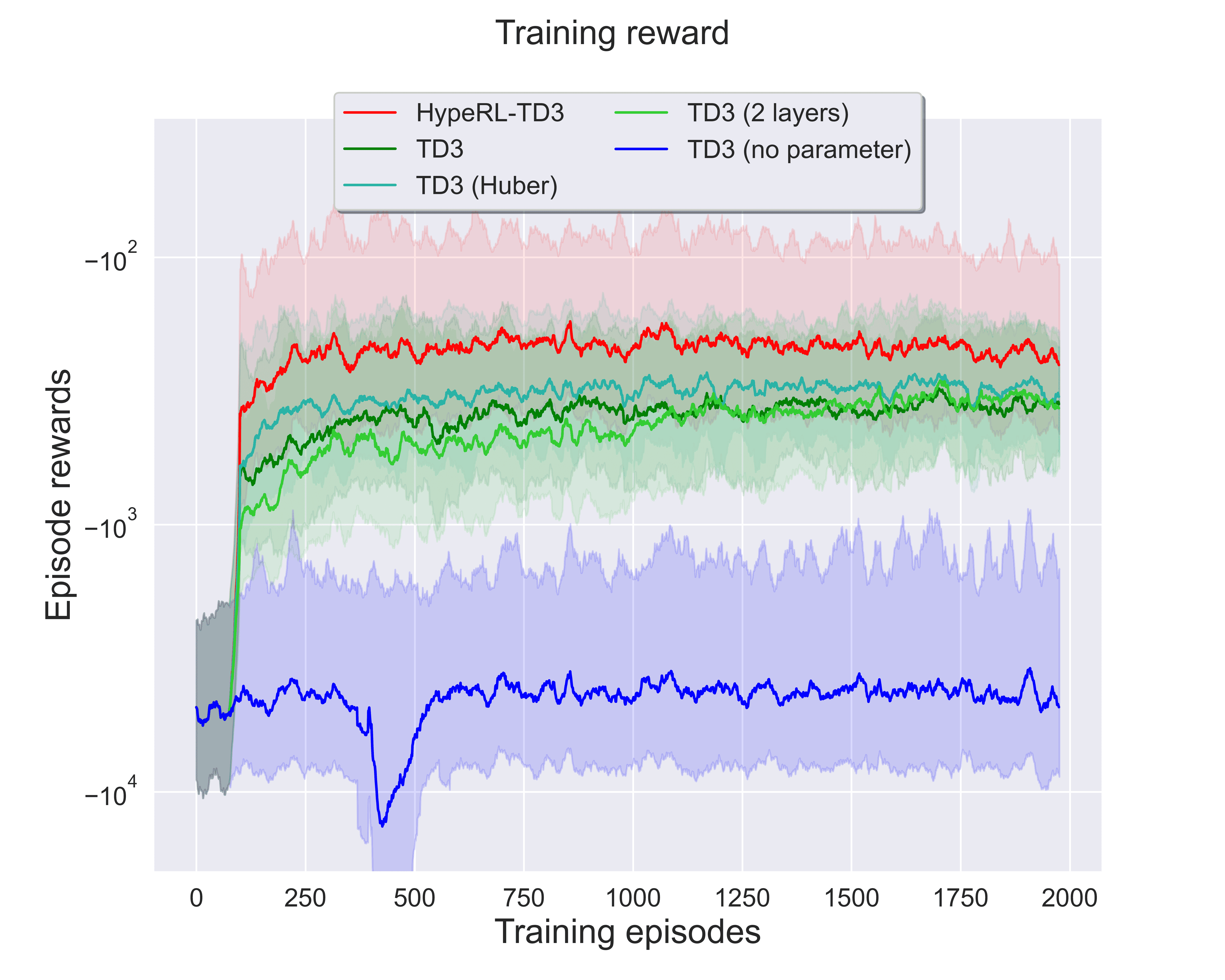}
         \caption{}
         \label{fig:param_ks_train_rewards_ks}
     \end{subfigure}
          \begin{subfigure}[b]{0.49\textwidth}
         \centering
         \includegraphics[width=\textwidth]{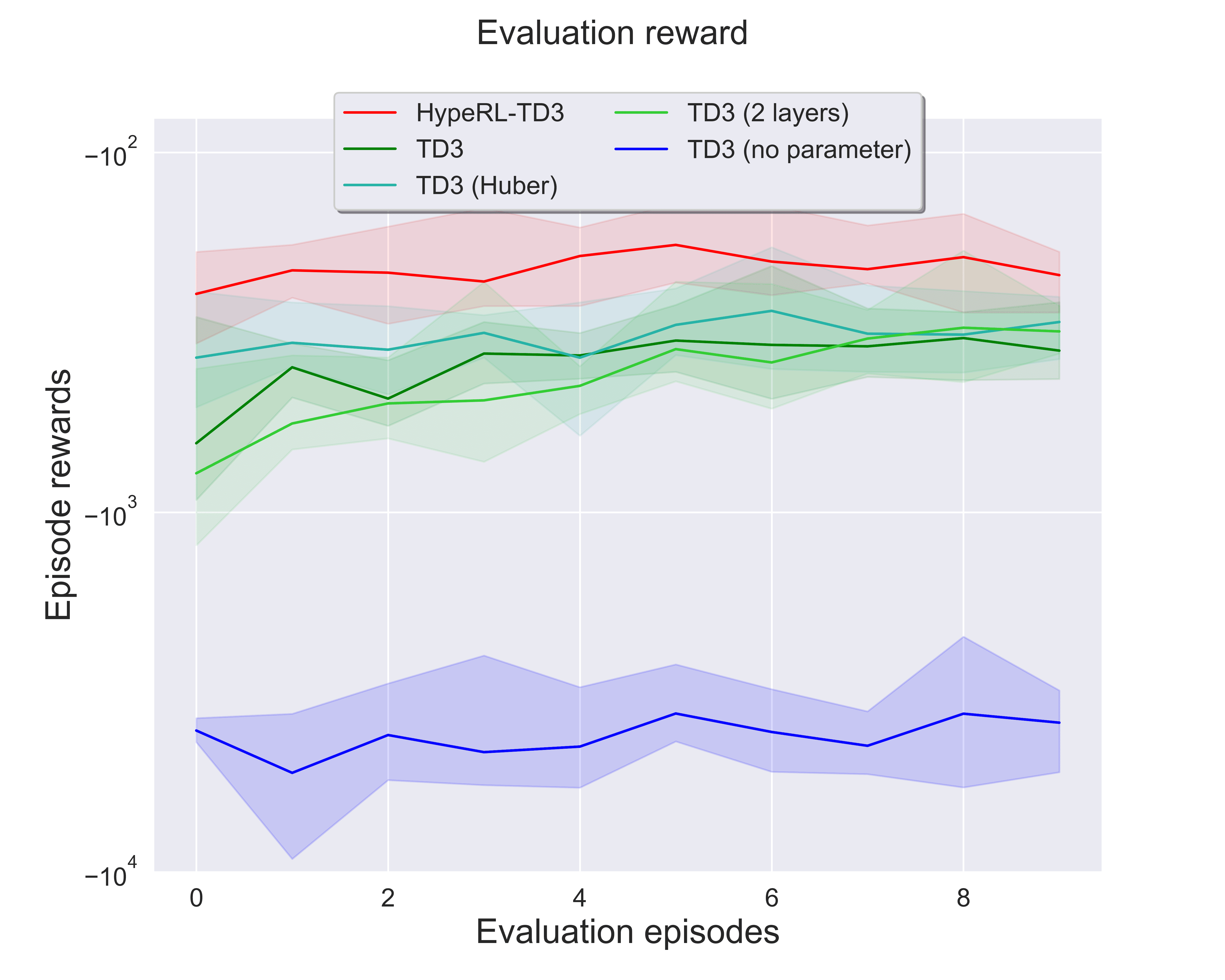}
         \caption{}
         \label{fig:param_ks_eval_rewards_ks}
     \end{subfigure}
        \caption{Stabilization of the state of a parametric KS equation to an arbitrary reference -- training and evaluation results. The solid line represents the mean and the shaded area the minimum and maximum values observed over 5 different random seeds. }
        \label{fig:ks_param_rewards}
\end{figure}

In Figure \ref{fig:ks_param_rewards}, we show the mean and the standard deviation of the cumulative reward during training and evaluation, respectively. It is possible to notice the superior performance of the hypernetwork-based agent in early and later stages of the training and during evaluation. 

In Table \ref{tab:param_KS_res}, we show the training and evaluation rewards collected by the different agents over 5 different random seeds, where we highlight in bold the highest scores of the training and evaluation. In the calculation of the scores, we remove for every agent the $100$ warmup episodes. 
\begin{table}[h!]
    \caption{Mean and standard deviation of the cumulative reward over training (left) and evaluation (right) collected by the different algorithms. The results report the average performance over 5 different random seeds.}
    \label{tab:param_KS_res}
    \begin{minipage}{.5\linewidth}
      \centering
\begin{tabular}{|| c | c ||} 
 \hline
Average cumulative reward & mean $\pm$ std \\ 
 \hline\hline
 HypeRL-TD3 & $\bm{-223.70 \pm 74.20}$ \\ 
 \hline
TD3 & $-396.56 \pm 133.33$  \\
 \hline
 TD3 (Huber) & $-329.48 \pm 94.98$  \\  
 \hline
  TD3 (2 layers) & $-450.86 \pm  188.58$  \\  
 \hline
  TD3 (no $\boldsymbol{\mu}$) & $-4598.64 \pm 2294.87$   \\ 
 \hline
\end{tabular}
    \end{minipage}%
    \begin{minipage}{.5\linewidth}
      \centering
\begin{tabular}{|| c | c ||} 
 \hline
Average cumulative reward & mean $\pm$ std \\ 
 \hline\hline
 HypeRL-TD3 & $\bm{-210.50 \pm 18.04}$ \\ 
 \hline
TD3 & $-395.40 \pm 92.77$  \\
 \hline
 TD3 (Huber) & $-326.36 \pm 30.47$  \\  
 \hline
  TD3 (2 layers) & $-446.20 \pm  138.94$  \\  
 \hline
  TD3 (no $\boldsymbol{\mu}$) & $-4222.67 \pm 485.95$   \\ 
 \hline
\end{tabular}
    \end{minipage} 
\end{table}
Again, the results show that the hypernetwork-based algorithm outperforms the TD3 agents, showing that encoding parametric information via a hypernetwork is crucial for sample efficiency and generalization of the RL algorithms.

\begin{figure*}[h!]
    \begin{minipage}{0.49\linewidth}
    \centering \subfloat{\includegraphics[height=0.85\textwidth]{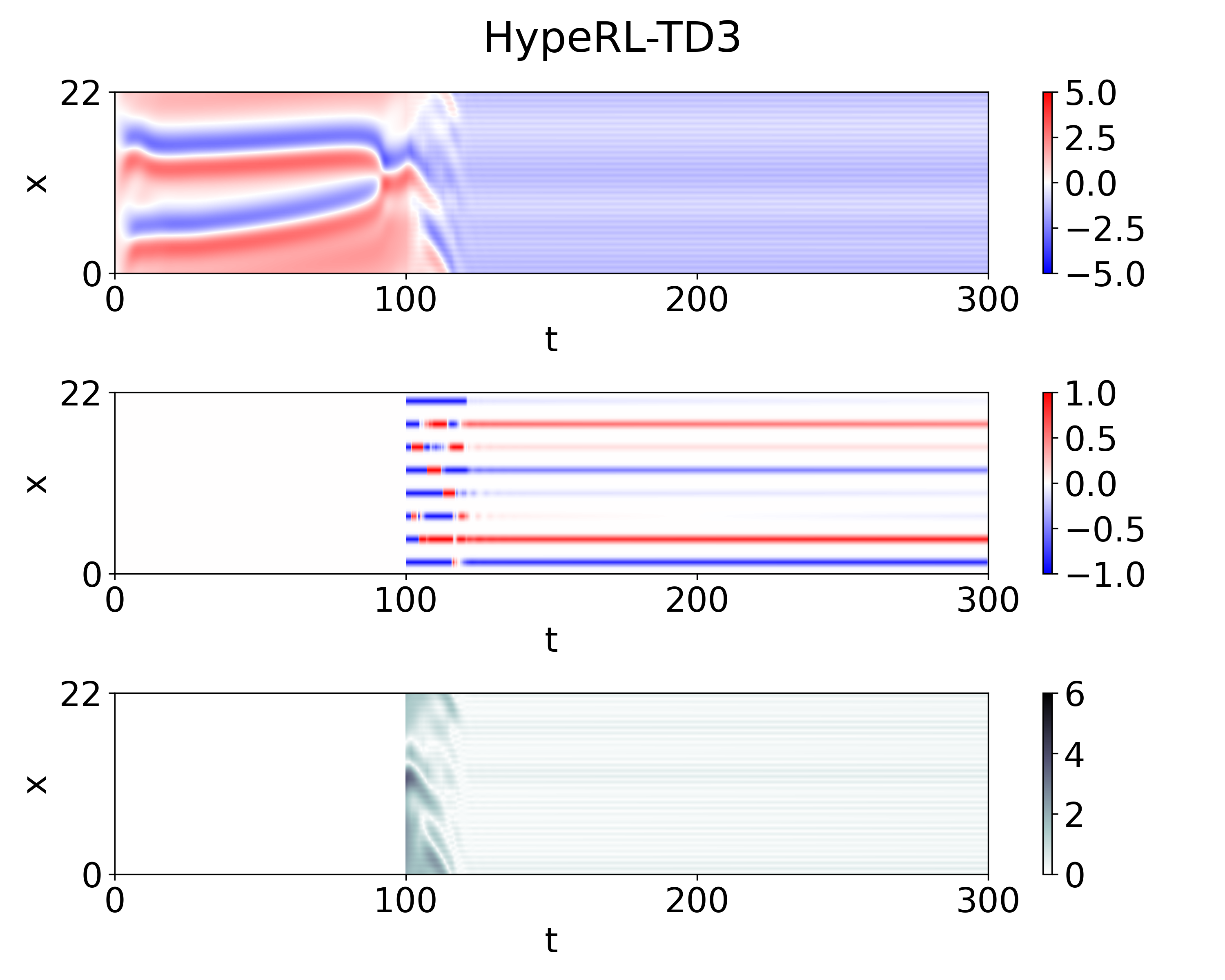}}
    \vspace{-0.1cm}
    \subfloat{\includegraphics[height=0.85\textwidth]{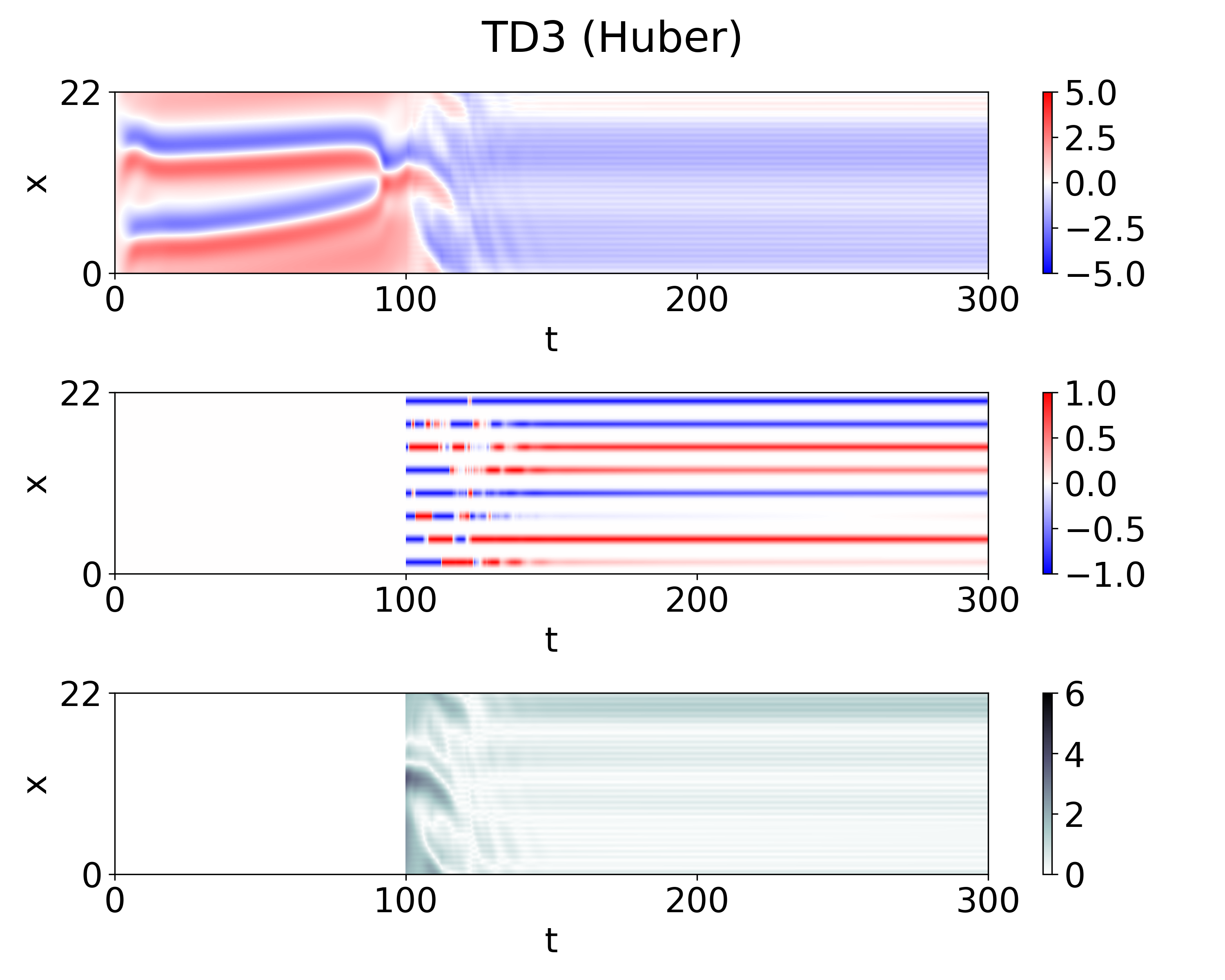}}
    \end{minipage}
    \begin{minipage}{0.49\linewidth}
    \centering \subfloat{\includegraphics[height=0.85\textwidth]{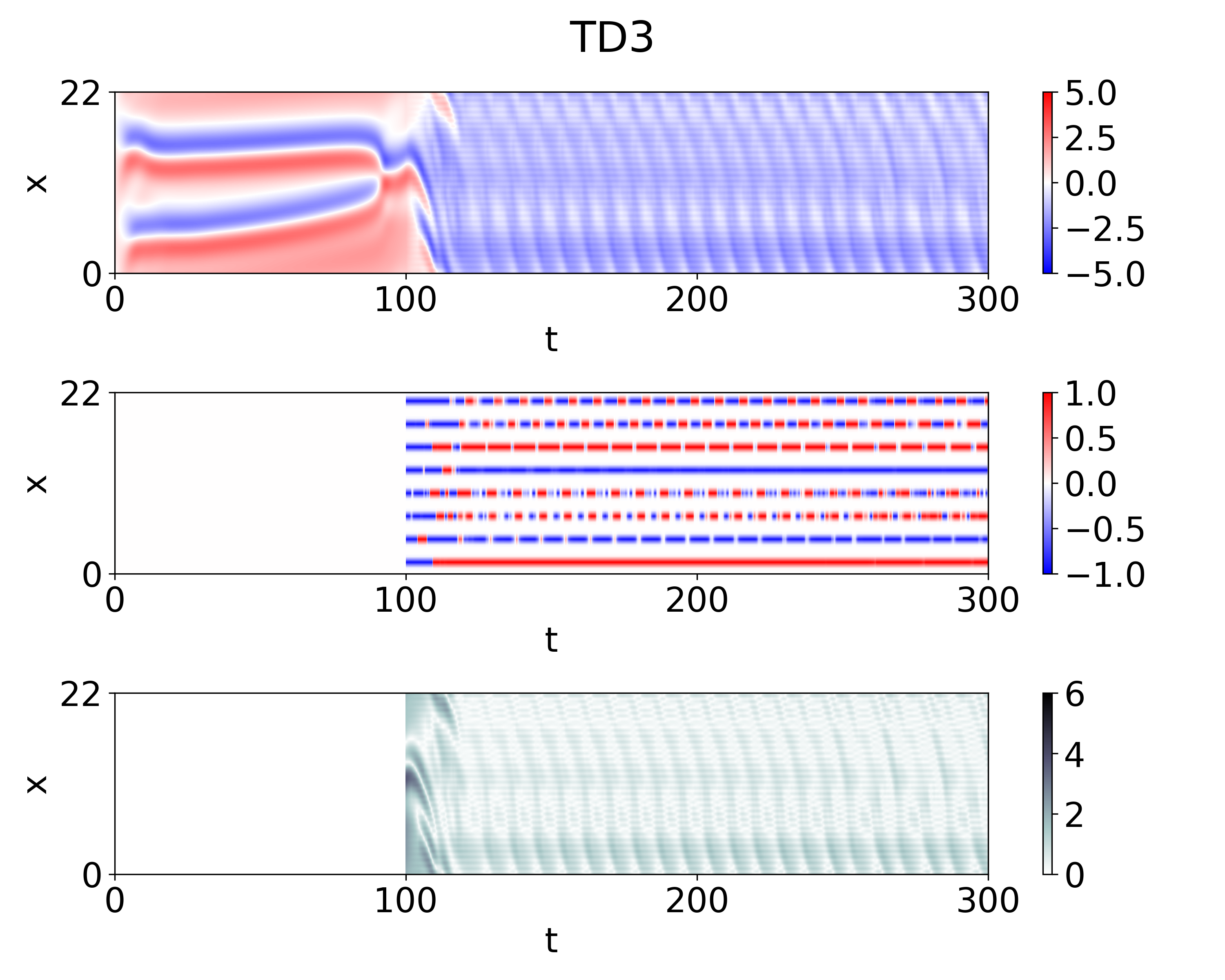}}
    \vspace{-0.1cm}
    \subfloat{\includegraphics[height=0.85\textwidth]{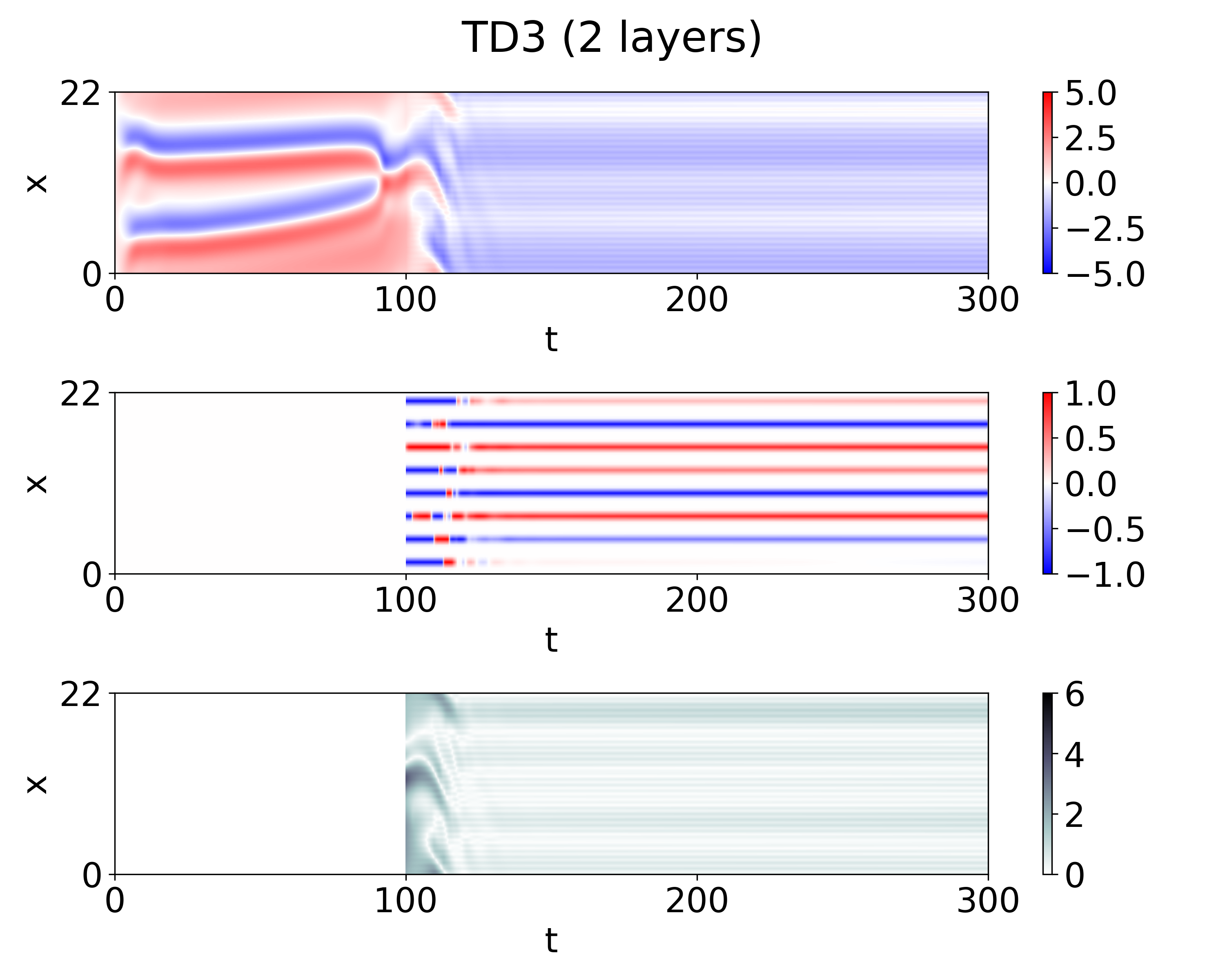}}
    \end{minipage}
    \caption{Controlled solutions of the parametric KS equation when using the different agents. The first row shows the state, the second the control, and the third one the error $|\bm{y}_k - \bm{y}_{\text{ref}}|$.}
    \label{fig:parametric_KS_controlled}
\end{figure*}

In Figure \ref{fig:parametric_KS_controlled}, we show an example of controlled solution for $\bm{y}_{\text{ref}}=-\bm{1}$ and $\nu=0.25$, where the first row depicts the evolution of the state, the second one the control, and the third one the state-tracking error $|\bm{y}_k - \bm{y}_{\text{ref}}|$.
Additional results are reported in Appendix \ref{app:param_KS_extra_results}.

\subsection{Navigation of a Particle in a Gyre Flow}

\begin{figure*}[h!]
    \begin{minipage}{0.49\linewidth}
    \centering \subfloat{\includegraphics[height=0.7\textwidth]{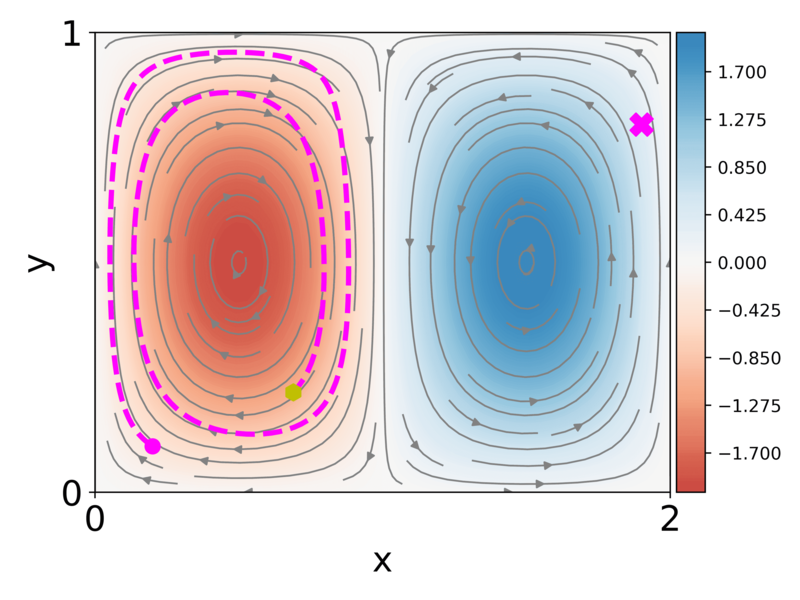}}
    \vspace{-0.2cm}
    \subfloat{\includegraphics[height=0.7\textwidth]{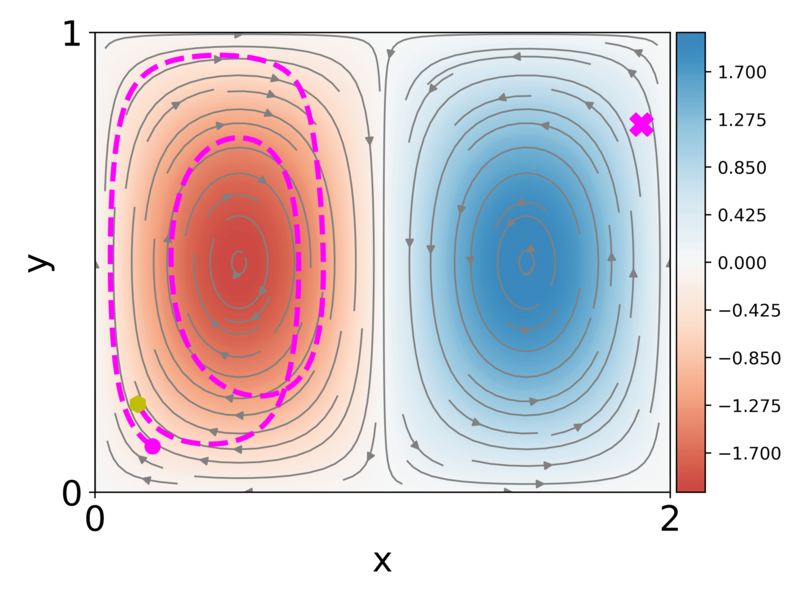}}
    \end{minipage}
    \begin{minipage}{0.49\linewidth}
    \centering \subfloat{\includegraphics[height=0.7\textwidth]{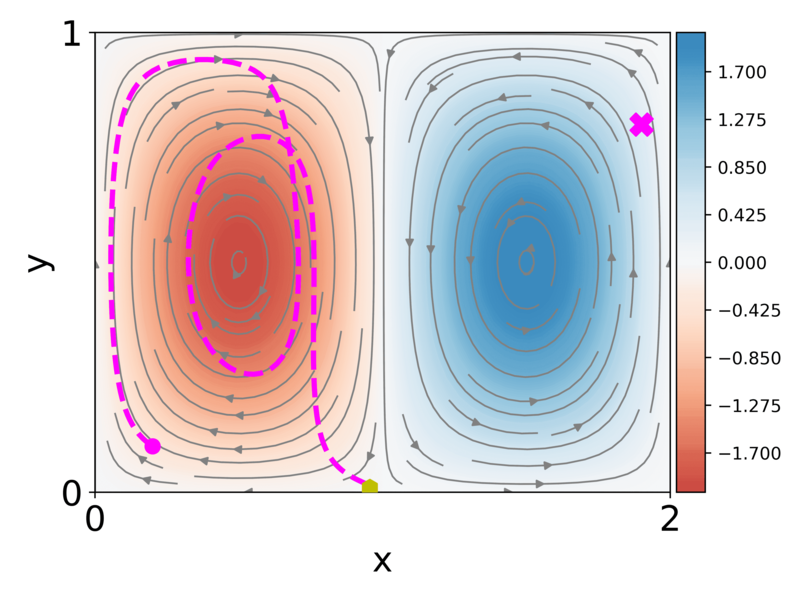}}
    \vspace{-0.2cm}
    \subfloat{\includegraphics[height=0.7\textwidth]{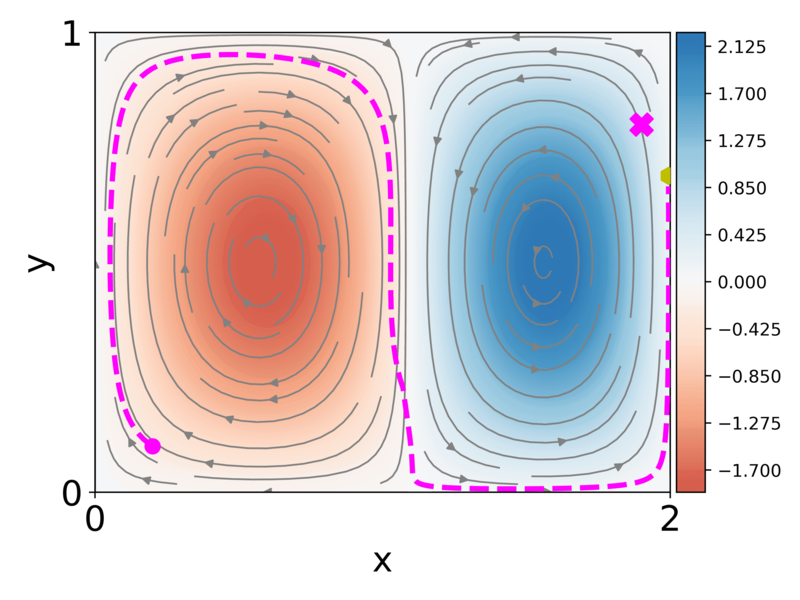}}
    \end{minipage}
    \caption{Trajectory (magenta dashed line) of the particle (yellow hexagon) in a gyre flow at different timesteps for  different values of $A$ and $\omega$ -- $A=0.1, \omega=2\pi/10$ (top left), $A=0.2, \omega=2\pi/10$ (bottom left), $A=0.3, \omega=2\pi/10$ (top right), and $A=0.1,\omega=2\pi/6$ (bottom right). The magenta circle indicates the starting location and the magenta cross the target one.}
    \label{fig:gyro_param}
\end{figure*}

As a second test case, we consider the navigation problem of a mobile particle in a double gyre flow field \cite{gunnarson2021learning, krishna2023finite}. The flow field is defined as:
\begin{equation*}
    \phi(x, y, t) = A\sin{\big(\pi f(x,t)\big)}\sin{\big(\pi y\big)} \, , 
\end{equation*}
where $x,y$ are the spatial coordinates, $t$ the time, $\pi$ (with a slightly abuse of notation) represents pi, and
\begin{equation*}
    f(x,t) = \big[ \epsilon \sin(\omega t)\big] x^2 + \big[1 - 2\epsilon \sin(\omega t) \big] x \, , 
\end{equation*}
Hence, an ODE system governing the particle dynamics in this gyro flow field is given by:
\begin{equation}
\begin{split}
    \frac{dx}{dt}(t) &= -\frac{\partial \phi}{\partial y}(x(t),y(t),t) + u_x(t) =  -\pi A \sin{\big(\pi f(x(t),t)\big) \cos{(\pi y(t))} + u_x(t)}\, , \\
    \frac{dy}{dt}(t) &= \frac{\partial \phi}{\partial x} (x(t),y(t),t) + u_y (t)= \pi A \cos{\big(\pi f(x(t),t)) \sin{(\pi y(t)\big)}\frac{\partial f}{\partial x}(x(t),t) + u_y(t)}\, , \\
\end{split}
\label{eq:gyre_flow}
\end{equation}
in which $u_x, u_y$ are the control inputs.
The flow is restrained within a domain $(x,y)^T \in \mathcal{D}=[0,2]\times[0, 1]$ and $A, \epsilon, \omega$ are model parameters controlling the flow pattern. In particular, $A$ affects the velocity magnitude, $\epsilon$ the magnitude of the oscillation in the $x-$direction, and $\omega$ the angular oscillation frequency. Similarly to \cite{krishna2023finite}, we solve Equation $\eqref{eq:gyre_flow}$ for $T=80$s and $\Delta t=0.1$s. However, differently from \cite{gunnarson2021learning, krishna2023finite} where the agent is trained to reach a single target, we consider two more challenging control problems:
\begin{enumerate}
    \item navigation of a particle from a random initial position to a random target in a  gyre flow with fixed parameters, namely $A=0.1, \omega=2\pi/10$, and $\epsilon=0.25$,
    \item navigation of a particle from a random initial position to a random target in a parametric gyre flow with $A \in \mathcal{U}(0.1 ,0.4), \omega 
    \in \mathcal{U}(2\pi/10, 2\pi/2)$, and $\epsilon=0.25$.
\end{enumerate}
To force the controller to exploit the dynamics of the gyre flow to navigate as the largest flow velocity is $\approx A\pi$, we limit $u_x, u_y \in [-0.2, 0.2]$.
In both cases, the agent's state $\bm{z}_k=[x_k, y_k, k, \boldsymbol{\mu}]$ in which the collection $[x_k, y_k, k]$ serves as $\bm{y}_k$ from Section 3, with  $\boldsymbol{\mu}=[x_{\text{target}}, y_{\text{target}}, A, \omega]$, and $x_{\text{target}}, y_{\text{target}}$ are the coordinates of the target location.
In Figure \ref{fig:gyro_param}, we show examples of particle trajectories for different values of the amplitude parameter.

Our goal is to steer a particle in a gyre flow, defined as in Equation \eqref{eq:gyre_flow}, from a randomly sampled initial condition $\in \mathcal{U}(0.1, 1.9) \times \mathcal{U}(0.1, 0.9)$ to a randomly sampled target position $(x_{\text{target}}, y_{\text{target}})\in \mathcal{U}(0.1, 1.9) \times \mathcal{U}(0.1, 0.9)$ with minimal control effort. Therefore, we utilize the following reward function:
\begin{equation}
    R(x_k, y_k, \bm{u}_k; \boldsymbol{\mu}) = -
\underbrace{(x_k - x_{\text{target}})^2 + (y_k - y_{\text{target}})^2}_{\text{state cost}} -  \underbrace{\alpha ||\bm{u}_k||_2^2}_{\text{action cost}} \, ,
\label{eq:reward_function_gyro}
\end{equation}
where $\alpha=0.1$ is a scalar positive coefficient balancing the contribution of the action cost.

\begin{figure*}[h!]
    \begin{minipage}{0.49\linewidth}
    \centering \subfloat{\includegraphics[height=0.7\textwidth]{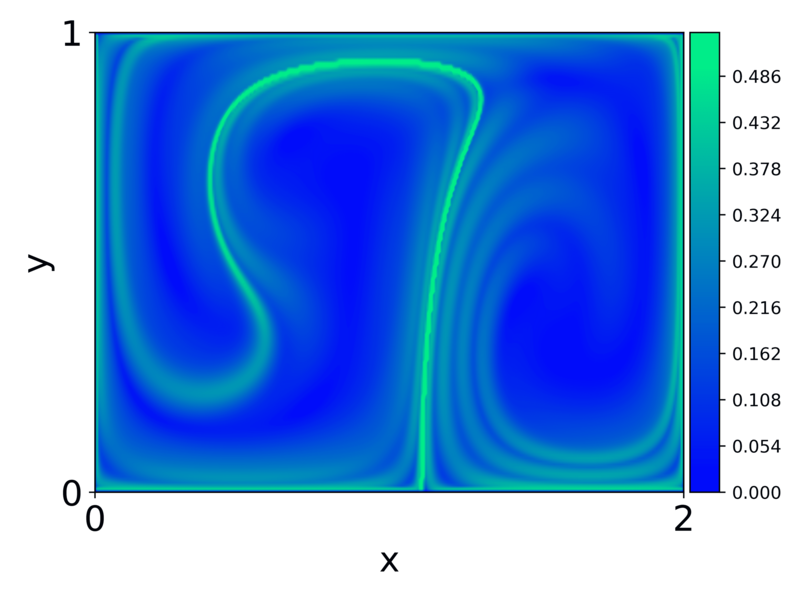}}
    \vspace{-0.2cm}
    \subfloat{\includegraphics[height=0.7\textwidth]{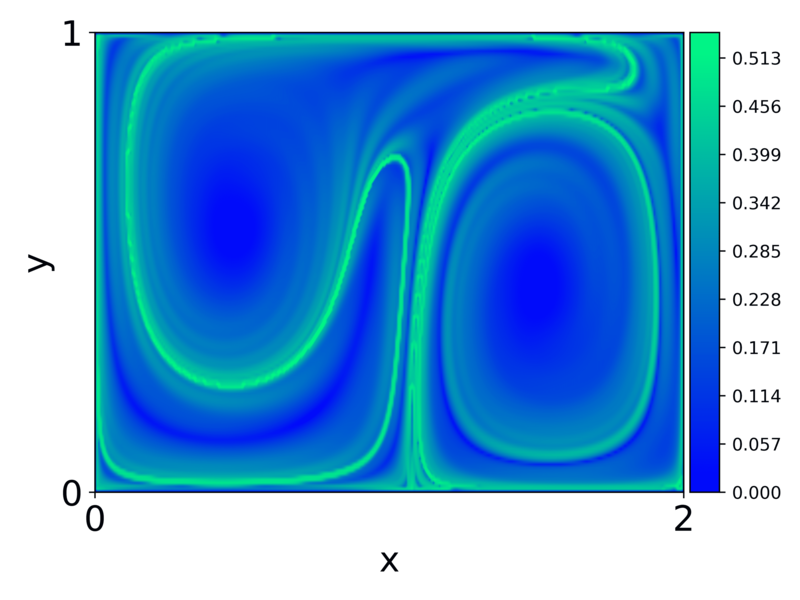}}
    \end{minipage}
    \begin{minipage}{0.49\linewidth}
    \centering \subfloat{\includegraphics[height=0.7\textwidth]{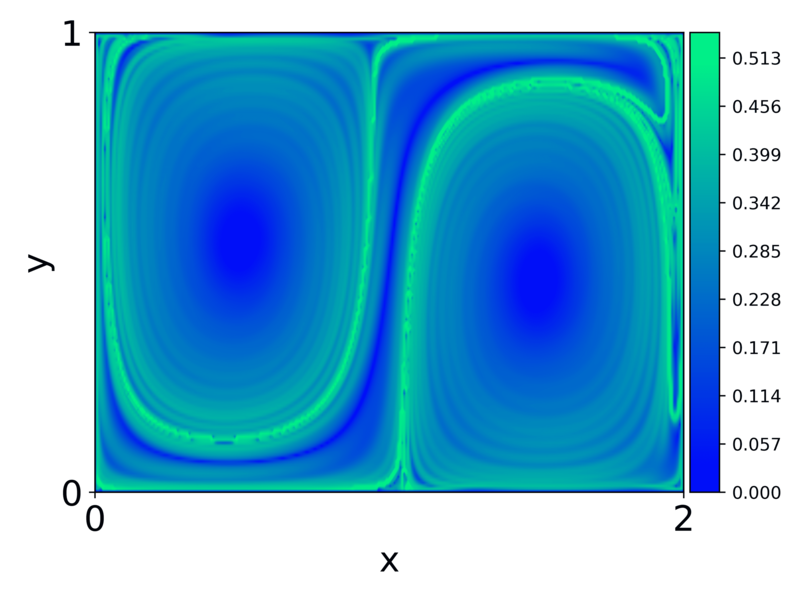}}
    \vspace{-0.2cm}
    \subfloat{\includegraphics[height=0.7\textwidth]{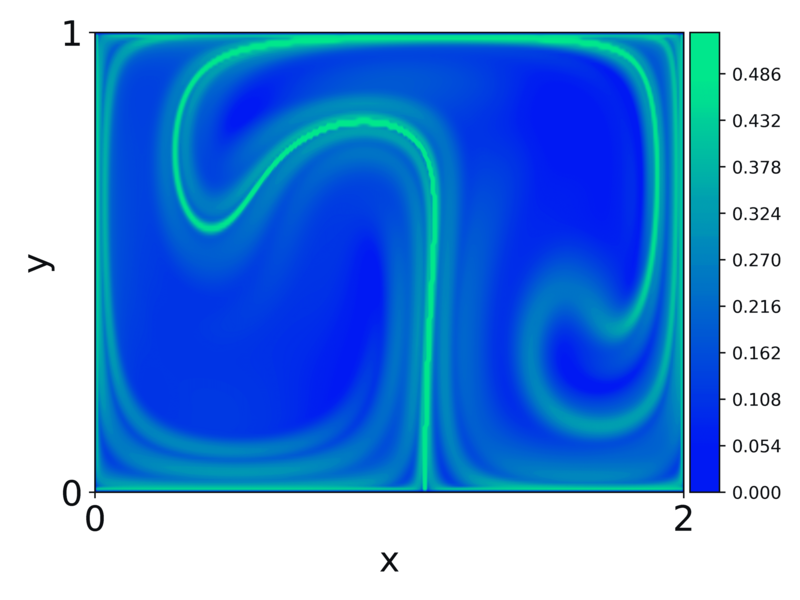}}
    \end{minipage}
    \caption{Finite-time Lyapunov exponent field calculated for $T=10$s and  different values of the amplitude $A$ and frequency $\omega$ parameters - $A=0.1, \omega=2\pi/10$ (top left), $A=0.2, \omega=2\pi/10$ (bottom left), $A=0.3, \omega=2\pi/10$ (top right), and $A=0.1,\omega=2\pi/6$ (bottom right).}
    \label{fig:FTLE_gyro_param}
\end{figure*}

To assess the effectiveness of our control strategy in exploiting the flow, we introduce the Lyapunov exponents \cite{wolf1985determining, shadden2005definition, sandri1996numerical}. The Lyapunov exponents (LE) play a crucial role in the investigation of the behaviour of chaotic dynamical systems. LEs measures the rate of convergence or divergence of a trajectory when the initial condition is perturbed. The finite-time Lyapunov exponent (FTLE) generalizes the LE to finite times, i.e., it measures the sensitivity of a trajectory to its initial condition over a finite-time horizon. Through the FTLE of the flow field, we are capable of identifying the areas of the domain where passive particles get trapped in and their ridges, separating the different stable regions. We leave the numerical computations of the FTLE to Appendix \ref{app:computations_FTLE}. In general, regions with high FTLE requires higher control effort than regions with lower FTLE, making the latter more favorable for energy-efficient control. Examples of FTLE fields for different values of the parameters $A$ and $\omega$ are shown in Figure \ref{fig:FTLE_gyro_param}. The FTLE fields are computed for $T=10$s, $dt=0.1$s, $N_x=300$, and $N_y=150$. It is worth highlighting that small variations of the gyre flow parameters produce large variations in the FTLE fields, making the learning of optimal control policies extremely challenging.

\subsubsection{Navigation of a Particle to Arbitrary Targets in a Gyre Flow}

\begin{figure}[h!]
     \centering
     \begin{subfigure}[b]{0.49\textwidth}
         \centering
         \includegraphics[width=\textwidth]{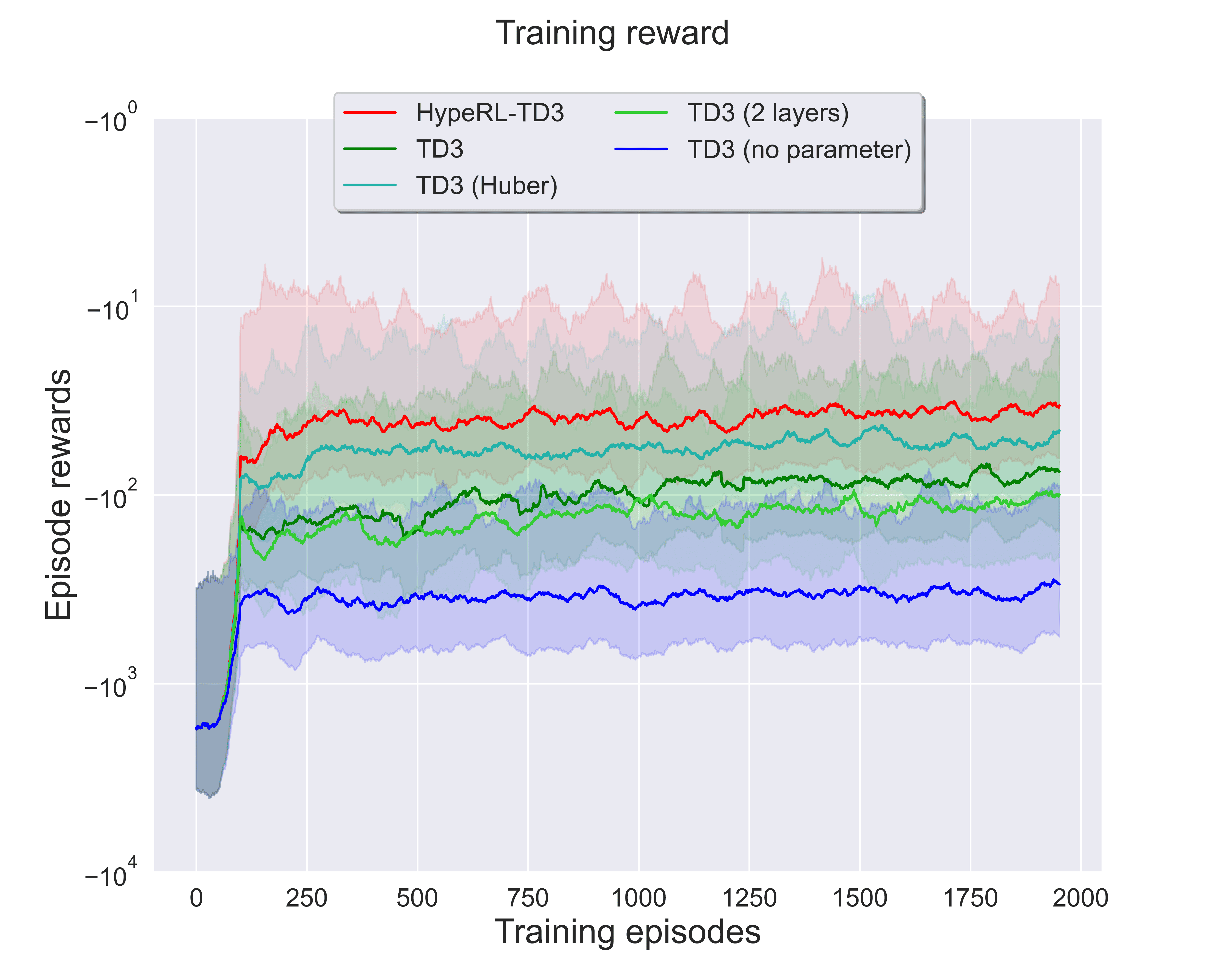}
         \caption{}
         \label{fig:gyro_train_rewards_ks}
     \end{subfigure}
          \begin{subfigure}[b]{0.49\textwidth}
         \centering
         \includegraphics[width=\textwidth]{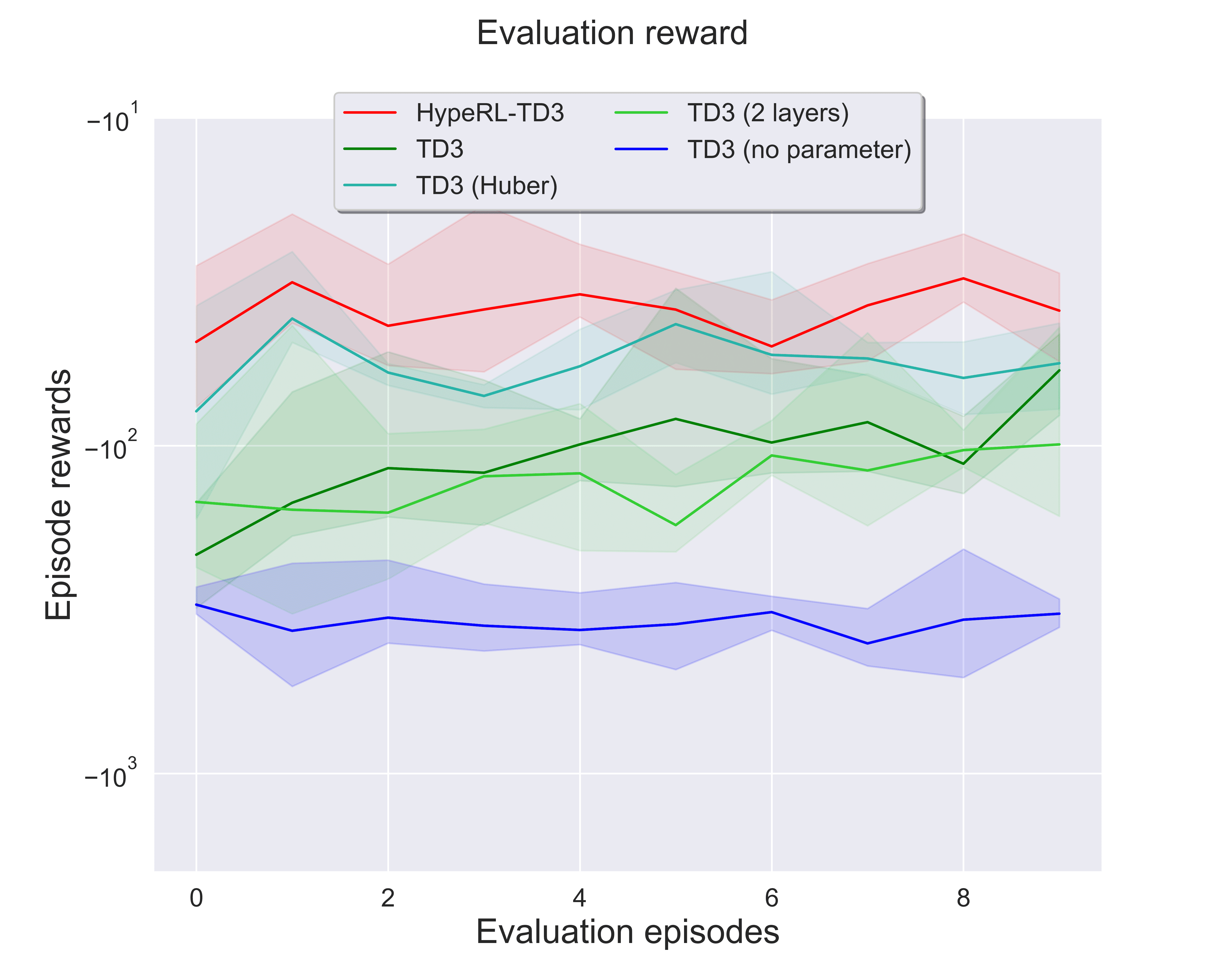}
         \caption{}
         \label{fig:gyro_eval_rewards_ks}
     \end{subfigure}
        \caption{Training and evaluation results. The solid line represents the mean and the shaded area the minimum and maximum values observed over 5 different random seeds.}
        \label{fig:10}
\end{figure}

In this first example, the goal is to control a particle in a gyre flow to a randomly sampled target location inside the domain. 
In Figure \ref{fig:10}, we show the mean and the standard deviation of the cumulative reward during training and evaluation, respectively. It is possible to notice the superior performance of the hypernetwork-based agent in early and later stages of the training and during evaluation.

In Table \ref{tab:gyro_res} we report the cumulative rewards over training and evaluation of the different agents, averages over 5 different random seed. The HypeRL-TD3 agent outperforms all the other agent by achieving the lowest average reward upon training and evaluation.
\begin{table}[h!]
    \caption{Mean and standard deviation of the cumulative reward over training (left) and evaluation (right) collected by the different algorithms. The results report the average performance over 5 different random seeds.}
    \label{tab:gyro_res}
    \begin{minipage}{.5\linewidth}
      \centering
\begin{tabular}{|| c | c ||} 
 \hline
Average cumulative reward & mean $\pm$ std \\ 
 \hline\hline
 HypeRL-TD3 & $\bm{ -40.08 \pm  18.45}$ \\ 
 \hline
TD3 & $-102.5 \pm  60.46$  \\
 \hline
 TD3 (Huber) & $-57.77 \pm 26.708$  \\  
 \hline
  TD3 (2 layers) & $-131.81 \pm 71.74$  \\  
 \hline
  TD3 (no $\boldsymbol{\mu}$) & $-344.34 \pm  131.50$   \\ 
 \hline
\end{tabular}
    \end{minipage}%
    \begin{minipage}{.5\linewidth}
      \centering
\begin{tabular}{|| c | c ||} 
 \hline
Average cumulative reward & mean $\pm$ std \\ 
 \hline\hline
 HypeRL-TD3 & $\bm{-39.16 \pm  6.00}$ \\ 
 \hline
TD3 & $-114.04 \pm 41.10$  \\
 \hline
 TD3 (Huber) & $-57.55 \pm  10.82$  \\  
 \hline
  TD3 (2 layers) & $-131.48 \pm  25.24$  \\  
 \hline
  TD3 (no $\boldsymbol{\mu}$) & $-346.58 \pm  25.93$   \\ 
 \hline
\end{tabular}
    \end{minipage} 
\end{table}

\begin{figure*}[h!]
    \begin{minipage}{0.49\linewidth}
    \centering \subfloat{\includegraphics[height=0.7\textwidth]{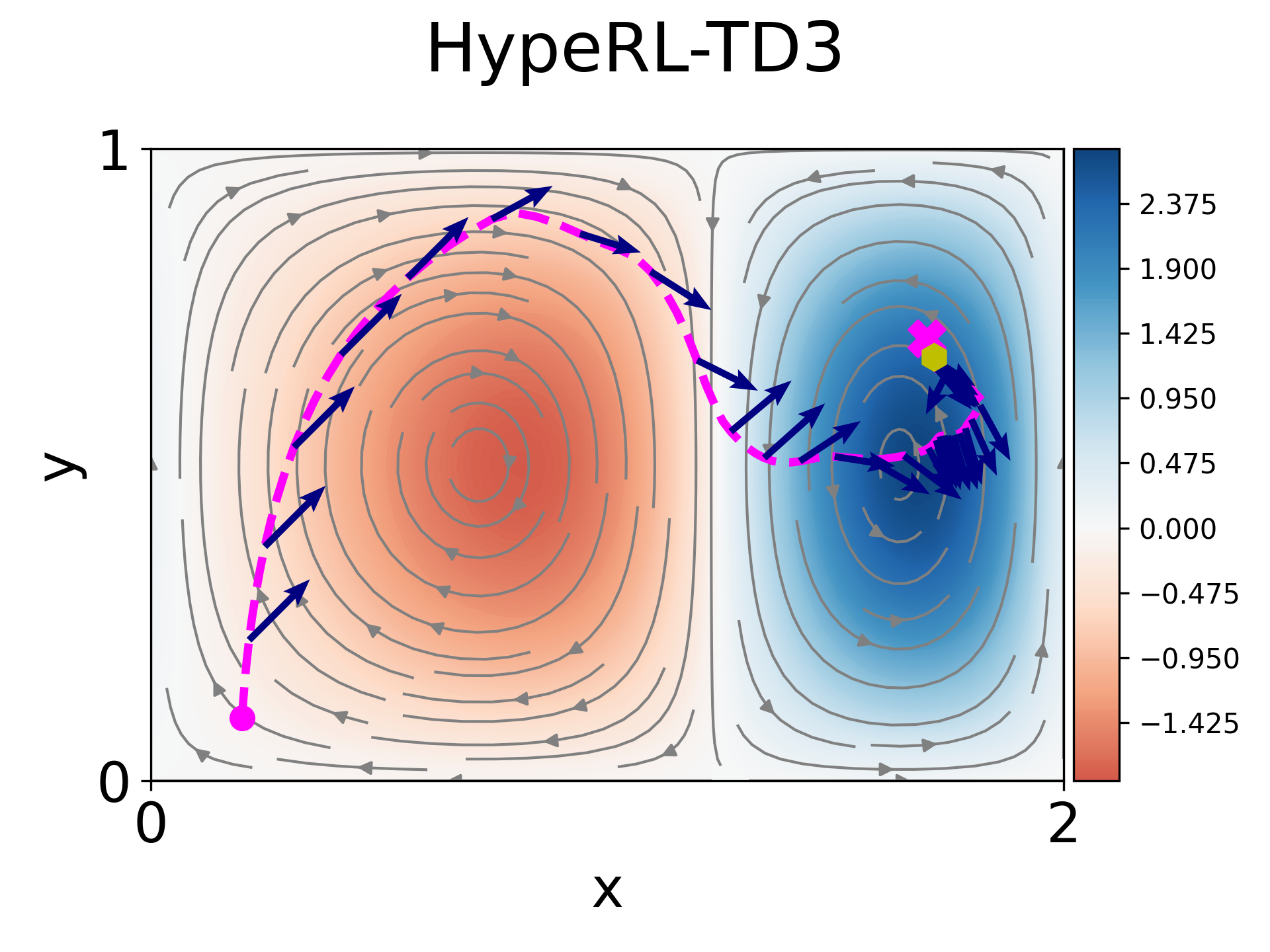}}
    \vspace{-0.1cm}
    \subfloat{\includegraphics[height=0.7\textwidth]{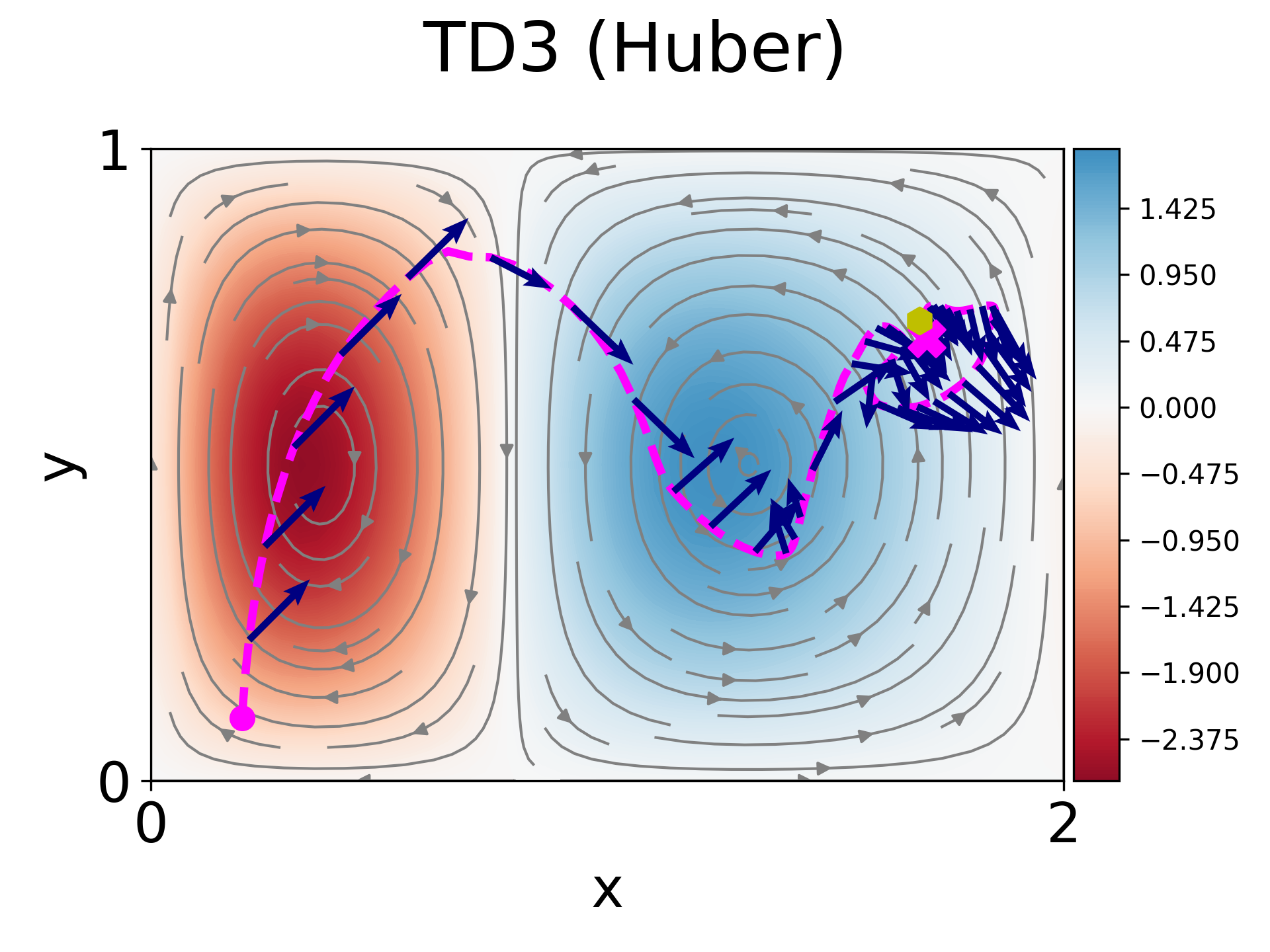}}
    \end{minipage}
    \begin{minipage}{0.49\linewidth}
    \centering \subfloat{\includegraphics[height=0.7\textwidth]{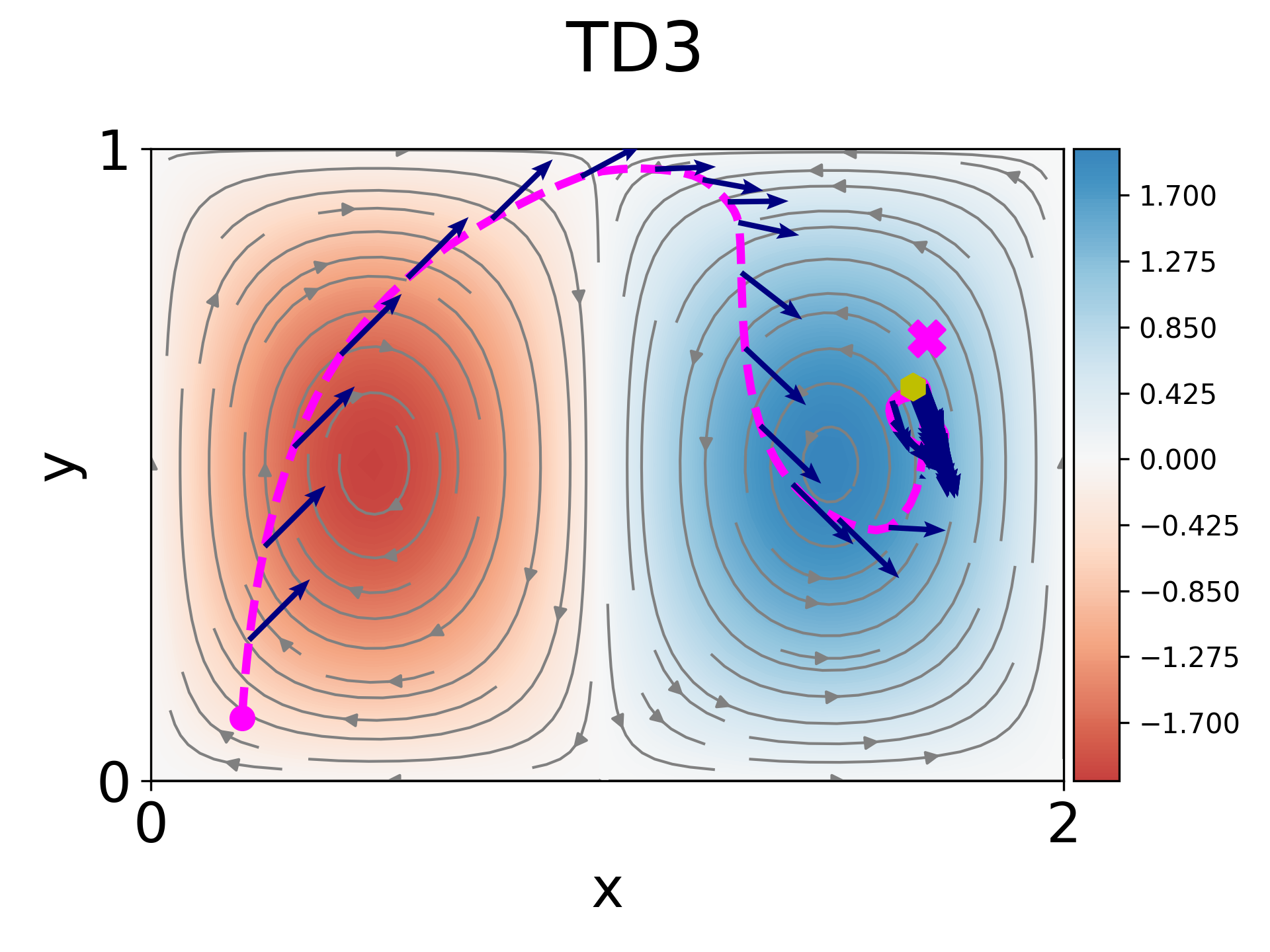}}
    \vspace{-0.1cm}
    \subfloat{\includegraphics[height=0.7\textwidth]{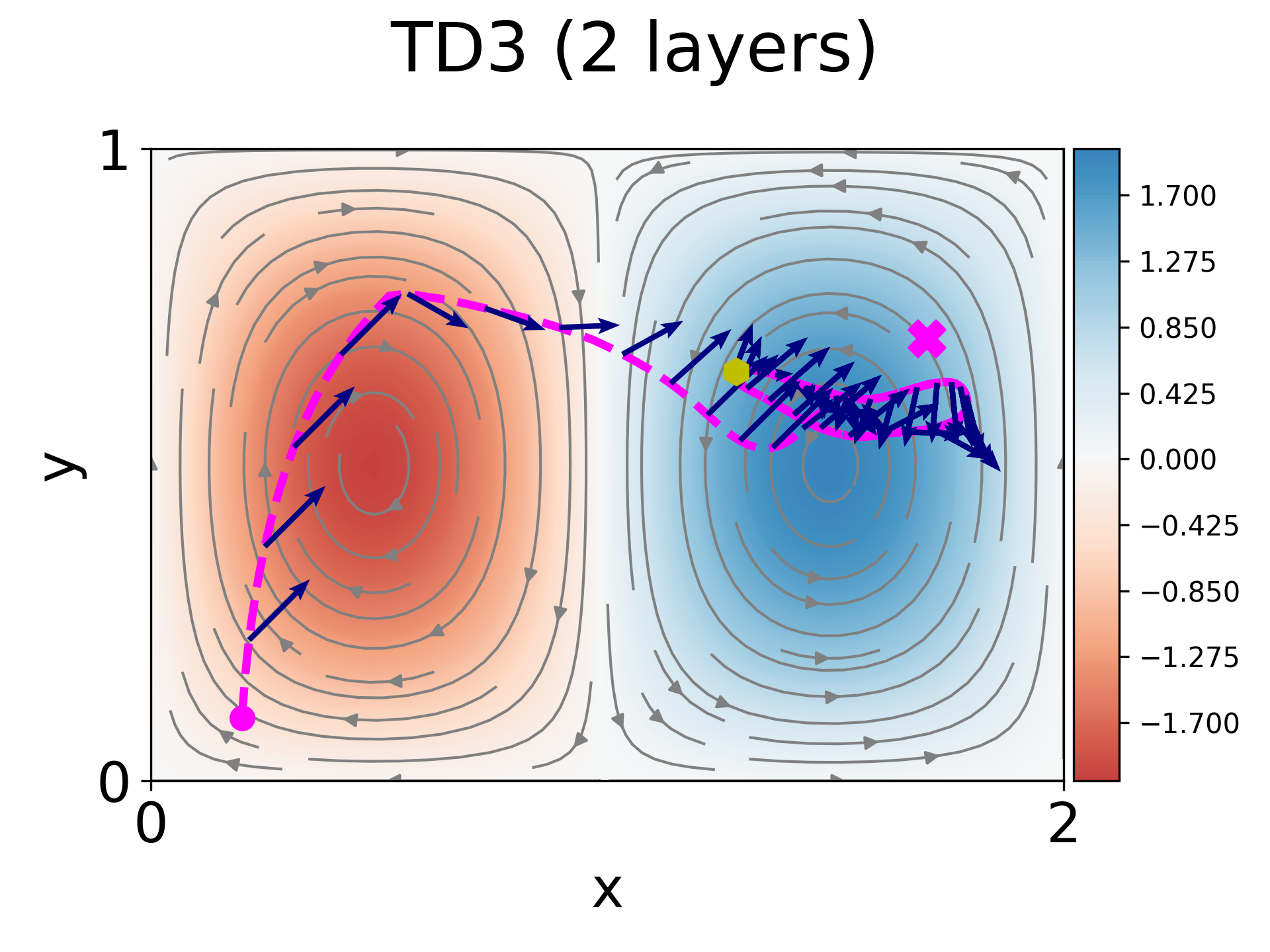}}
    \end{minipage}
    \caption{Trajectories (magenta dash line) of a controlled particle (yellow hexagon) in a gyre flow when using the different agents. The magenta circle indicates the starting location and the magenta cross the target one and the blue arrows represent the control inputs.}
    \label{fig:gyro_controlled}
\end{figure*}

\begin{figure*}[h!]
    \begin{minipage}{0.49\linewidth}
    \centering \subfloat{\includegraphics[height=0.7\textwidth]{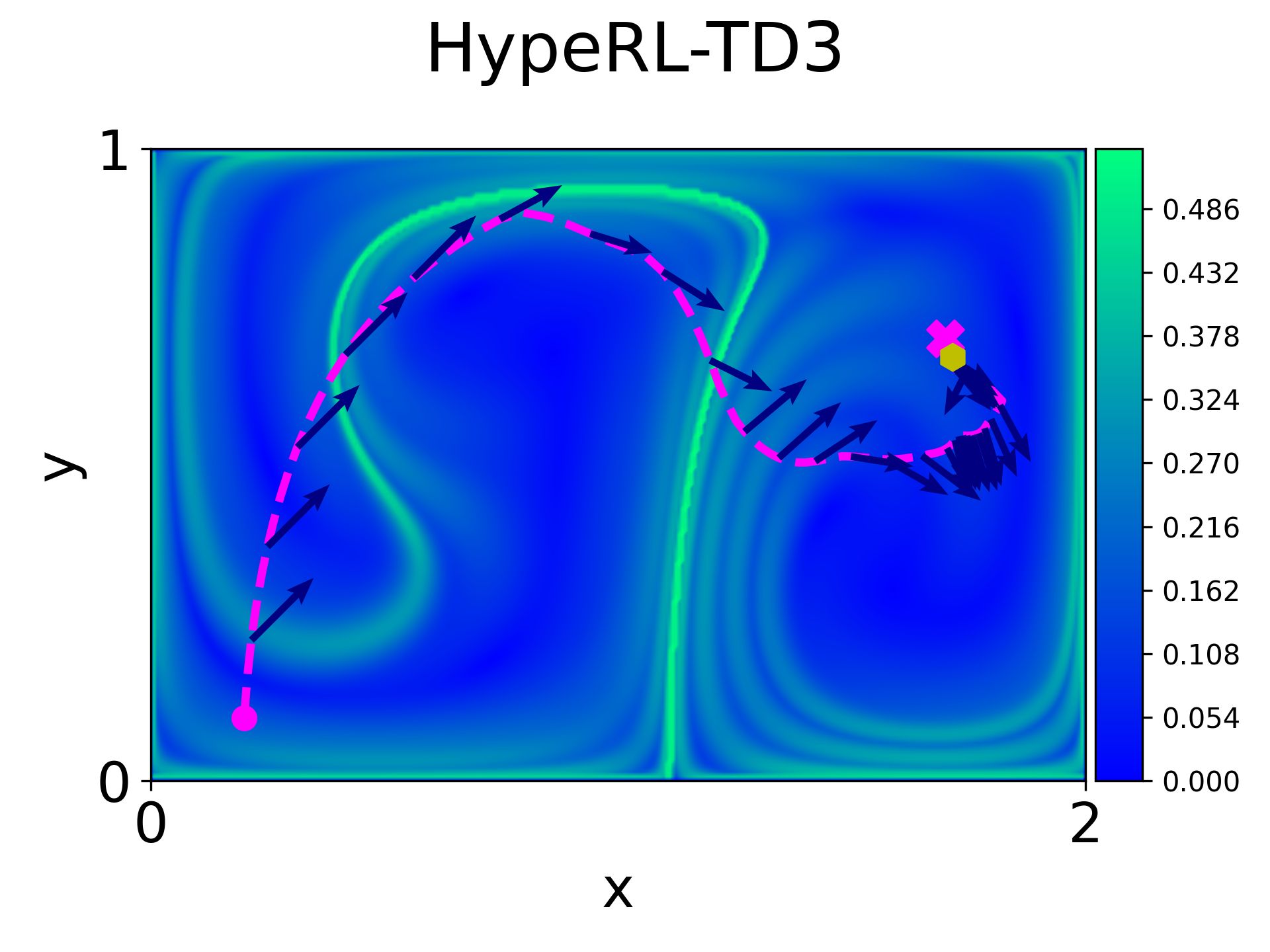}}
    \vspace{-0.1cm}
    \subfloat{\includegraphics[height=0.7\textwidth]{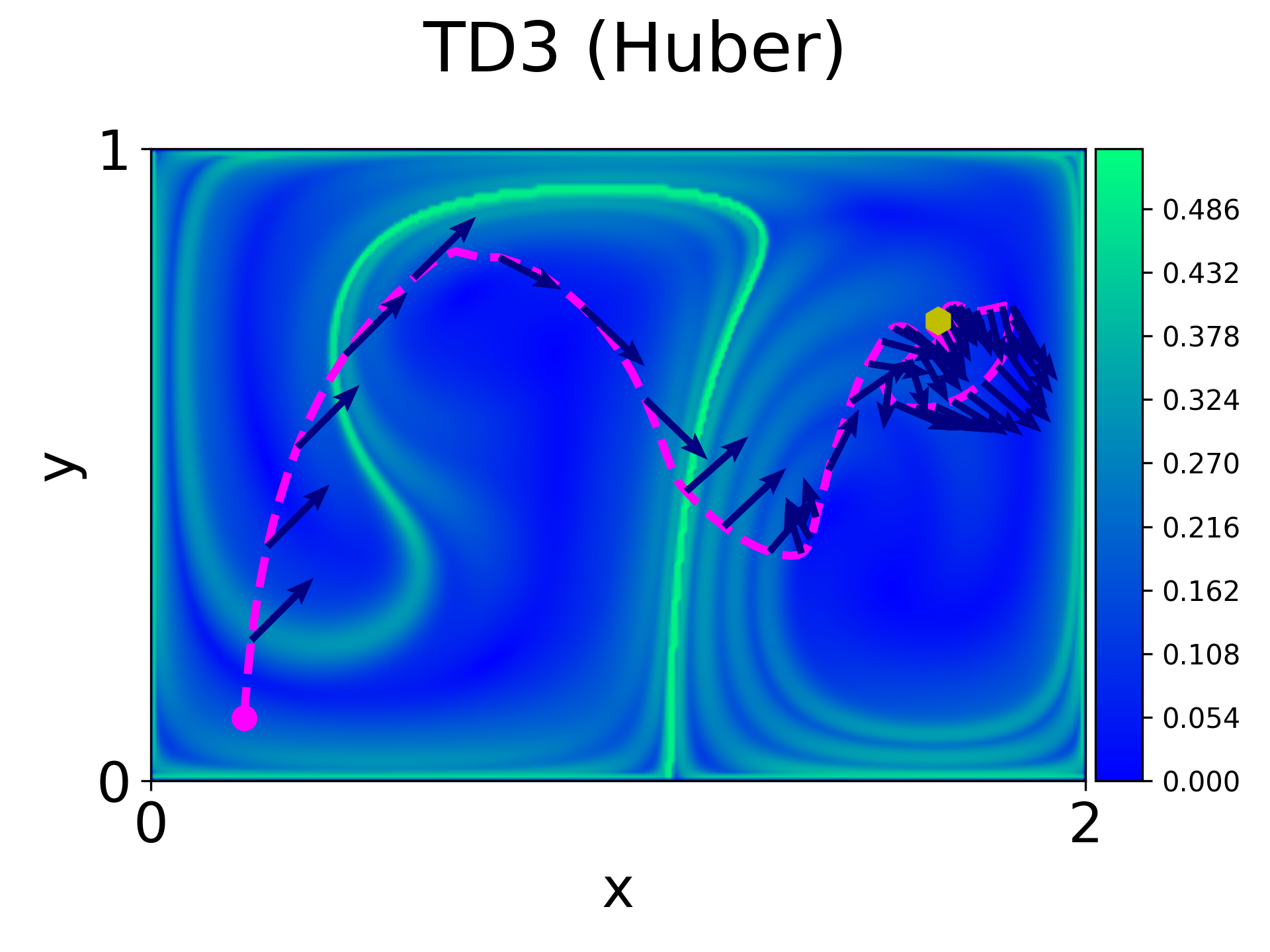}}
    \end{minipage}
    \begin{minipage}{0.49\linewidth}
    \centering \subfloat{\includegraphics[height=0.7\textwidth]{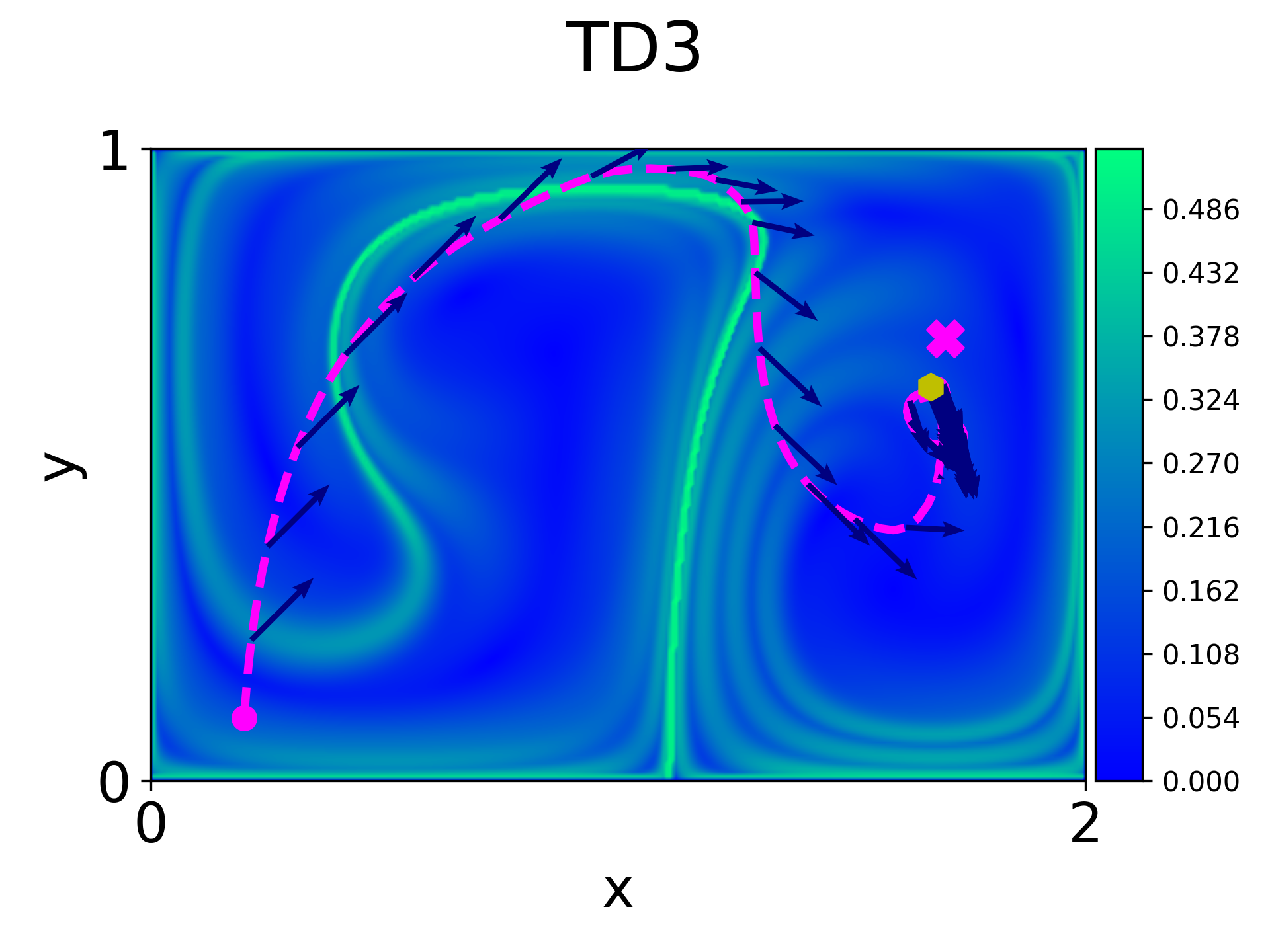}}
    \vspace{-0.1cm}
    \subfloat{\includegraphics[height=0.7\textwidth]{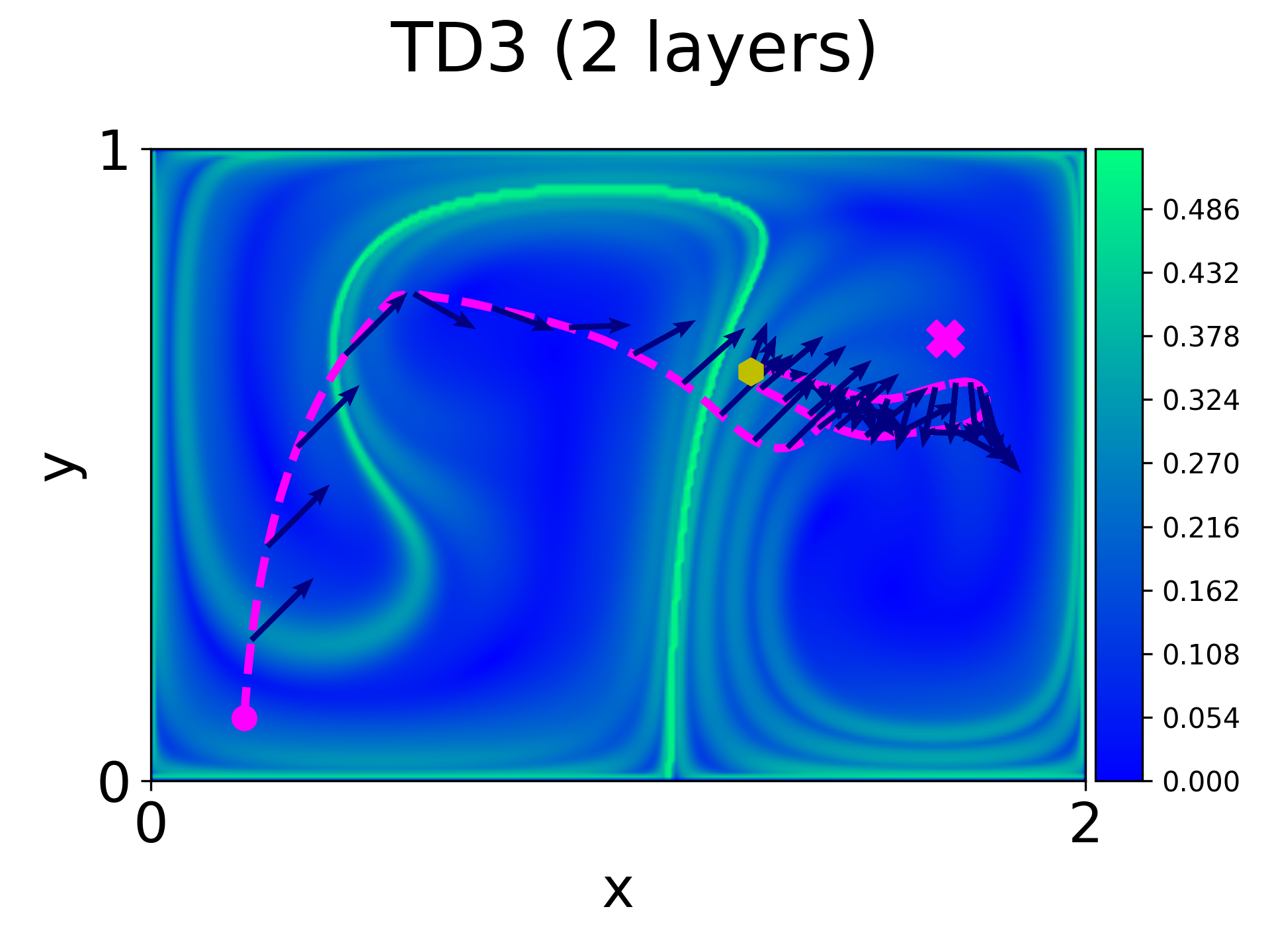}}
    \end{minipage}
    \caption{Trajectories (magenta dash line) of a controlled particle (yellow hexagon) in a gyre flow when using the different agents overlaid with the FTLE. The magenta circle indicates the starting location and the magenta cross the target one and the blue arrows represent the control inputs.}
    \label{fig:gyro_controlled_FTLE}
\end{figure*}

Eventually, in Figure \ref{fig:gyro_controlled} and \ref{fig:gyro_controlled_FTLE}, we show the trajectory of the particle when controlled by the different agents where the background represents the flow field at the last timestep of the trajectory and the FTLE computed over an interval of $T=10$s with a $dt=0.1$, respectively. While the different variants of TD3 get influenced by the ridges of the FTLE (see Figure \ref{fig:gyro_controlled_FTLE}) and achieve trajectories that oscillate around the target that require more time and control effort, HypeRL-TD3 smoothly and efficiently reaches the target by exploiting the stable regions of the gyre flow field.

Additional results, including the performance of the TD3 agent without access to the knowledge of $\boldsymbol{\mu}$, are shown in Appendix  \ref{app:gyro_extra}.

\subsubsection{Navigation of a Particle to Arbitrary Targets in a Parametric Gyre Flow}

In the last test case, the goal is to control a particle in a parametric gyre flow. We use the same settings as the previous case with the addition of randomly sampling the parameter $A \in \mathcal{U}(0.1,0.4)$ and the parameter $\omega \in \mathcal{U}(2\pi/10,2\pi/2)$ at the beginning of each training and evaluation episode.

\begin{figure}[h!]
     \centering
     \begin{subfigure}[b]{0.49\textwidth}
         \centering
         \includegraphics[width=\textwidth]{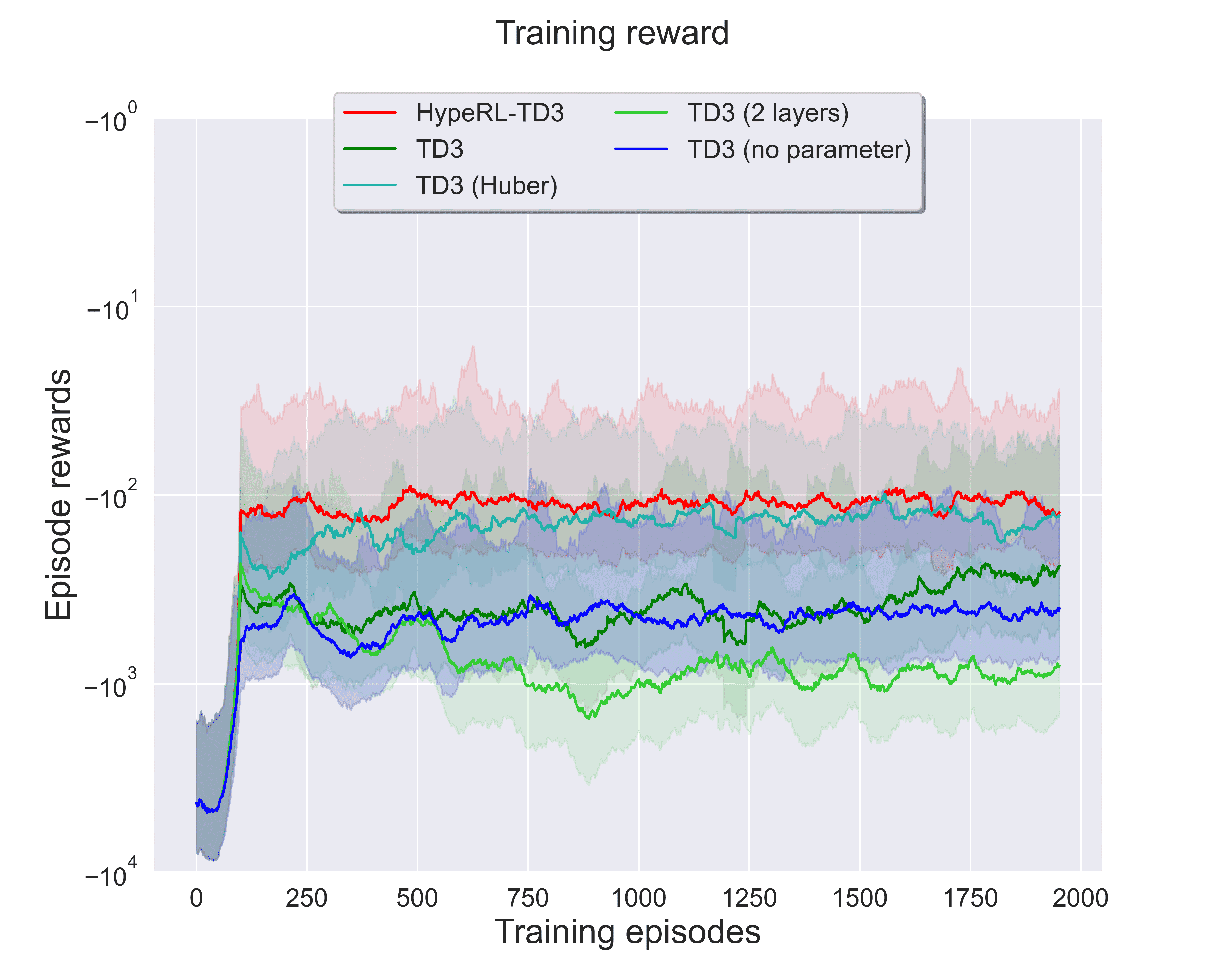}
         \caption{}
         \label{fig:param_gyro_train_rewards}
     \end{subfigure}
          \begin{subfigure}[b]{0.49\textwidth}
         \centering
         \includegraphics[width=\textwidth]{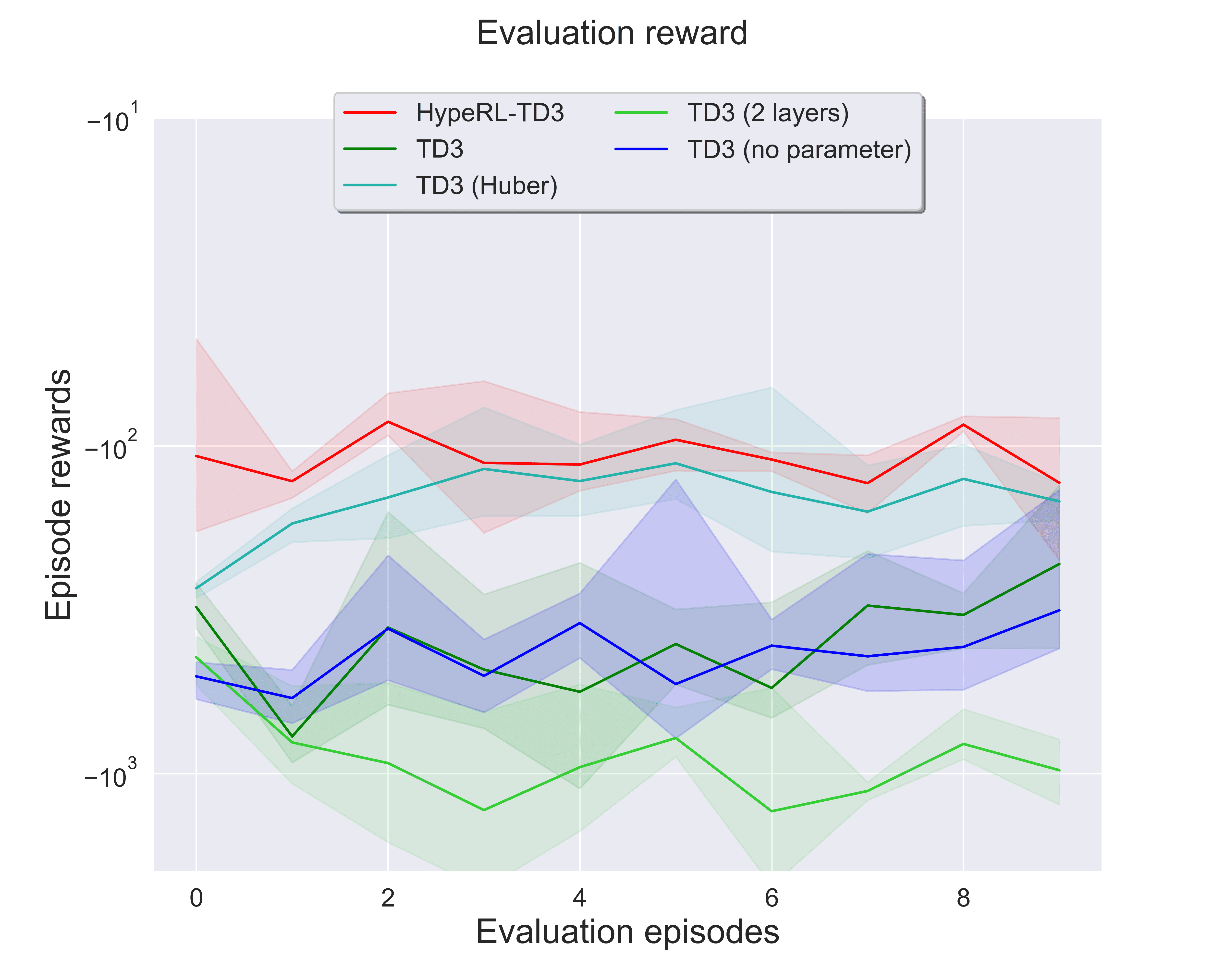}
         \caption{}
         \label{fig:param_gyro_eval_rewards}
     \end{subfigure}
        \caption{Training and evaluation results. The solid line represents the mean and the shaded area the minimum and maximum values observed over 5 different random seeds.}
        \label{fig:param_gyro_rewards}
\end{figure}

In Figure \ref{fig:param_gyro_rewards}, we show the mean and the standard deviation of the cumulative reward during training and evaluation, respectively. It is possible to notice the superior performance of the hypernetwork-based agent in early and later stages of the training and during evaluation. 

\begin{table}[ht!]
    \caption{Mean and standard deviation of the cumulative reward over training (left) and evaluation (right) collected by the different algorithms. The results report the average performance over 5 different random seeds.}
    \label{tab:param_gyro_res}
    \begin{minipage}{.5\linewidth}
      \centering
\begin{tabular}{|| c | c ||} 
 \hline
Average cumulative reward & mean $\pm$ std \\ 
 \hline\hline
 HypeRL-TD3 & $\bm{-112.38 \pm 70.55}$ \\ 
 \hline
TD3 & $-398.12 \pm 332.63$  \\
 \hline
 TD3 (Huber) & $-147.08 \pm 109.23$  \\  
 \hline
  TD3 (2 layers) & $-801.62 \pm  576.11$  \\  
 \hline
  TD3 (no $\boldsymbol{\mu}$) & $-457.47 \pm  266.10$   \\ 
 \hline
\end{tabular}
    \end{minipage}%
    \begin{minipage}{.5\linewidth}
      \centering
\begin{tabular}{|| c | c ||} 
 \hline
Average cumulative reward & mean $\pm$ std \\ 
 \hline\hline
 HypeRL-TD3 & $\bm{-110.06 \pm 16.05}$ \\ 
 \hline
TD3 & $-430.34 \pm  153.73$  \\
 \hline
 TD3 (Huber) & $ -152.05 \pm  43.72$  \\  
 \hline
  TD3 (2 layers) & $ -942.89 \pm  245.60$  \\  
 \hline
  TD3 (no $\boldsymbol{\mu}$) & $-441.75 \pm  84.12$   \\ 
 \hline
\end{tabular}
    \end{minipage} 
\end{table}

\begin{figure*}[h!]
    \begin{minipage}{0.49\linewidth}
    \centering \subfloat{\includegraphics[height=0.7\textwidth]{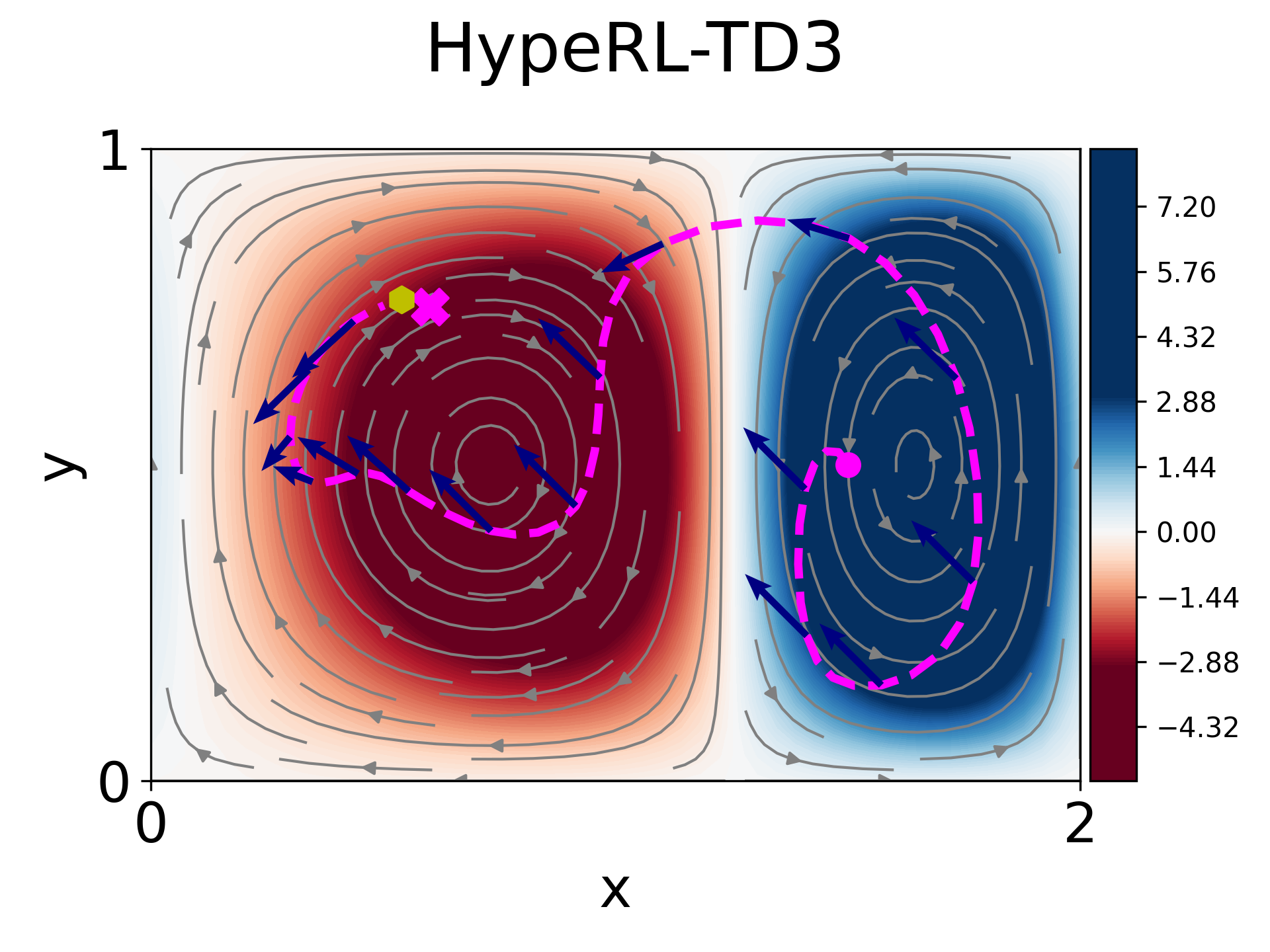}}
    \vspace{-0.1cm}
    \subfloat{\includegraphics[height=0.7\textwidth]{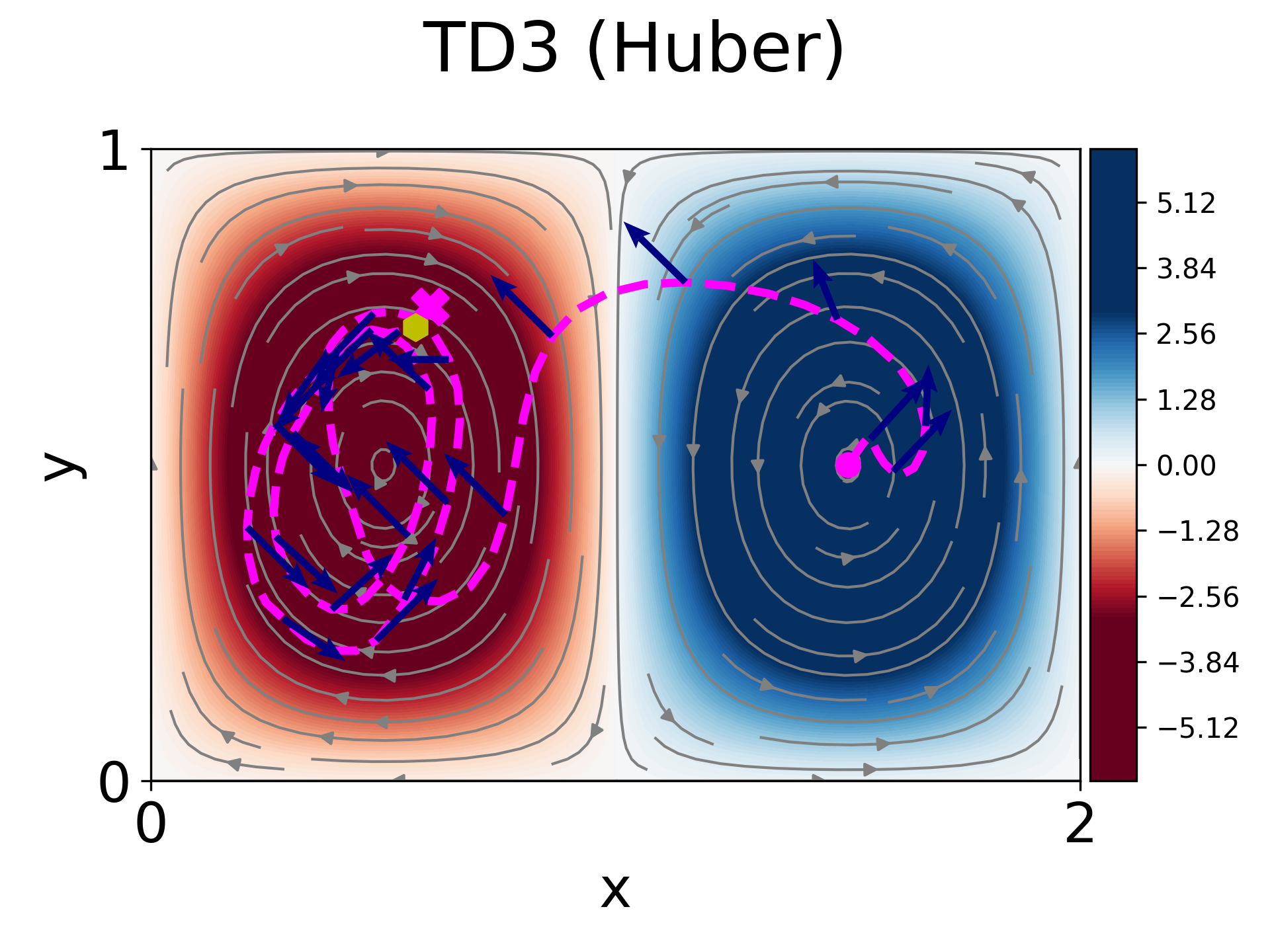}}
    \end{minipage}
    \begin{minipage}{0.49\linewidth}
    \centering \subfloat{\includegraphics[height=0.7\textwidth]{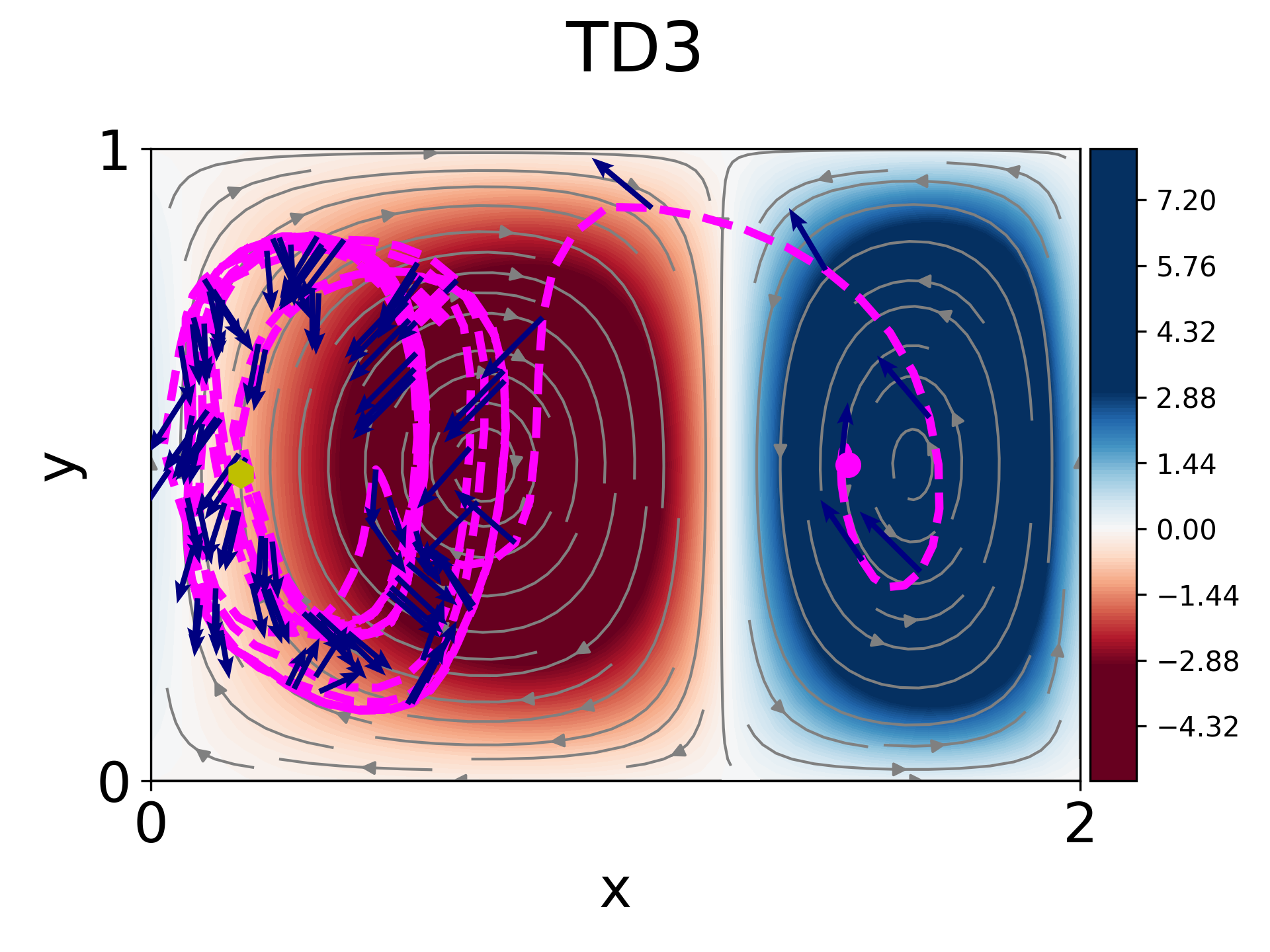}}
    \vspace{-0.1cm}
    \subfloat{\includegraphics[height=0.7\textwidth]{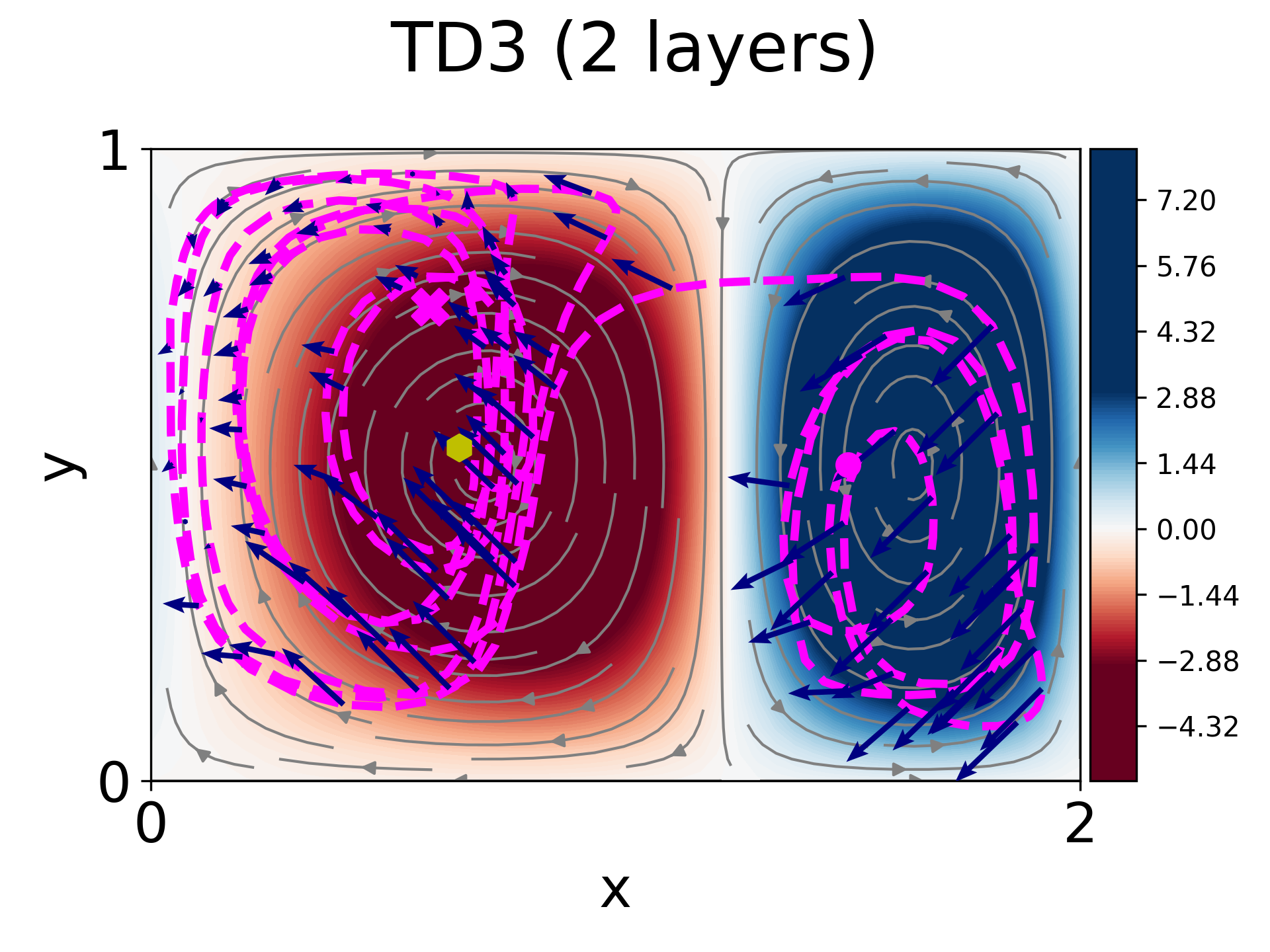}}
    \end{minipage}
    \caption{Trajectories (magenta dash line) of a controlled particle (yellow hexagon) in a gyre flow when using the different agents. The magenta circle indicates the starting location and the magenta cross the target one and the blue arrows represent the control inputs.}
    \label{fig:param_gyro_controlled}
\end{figure*}

\begin{figure*}[h!]
    \begin{minipage}{0.49\linewidth}
    \centering \subfloat{\includegraphics[height=0.7\textwidth]{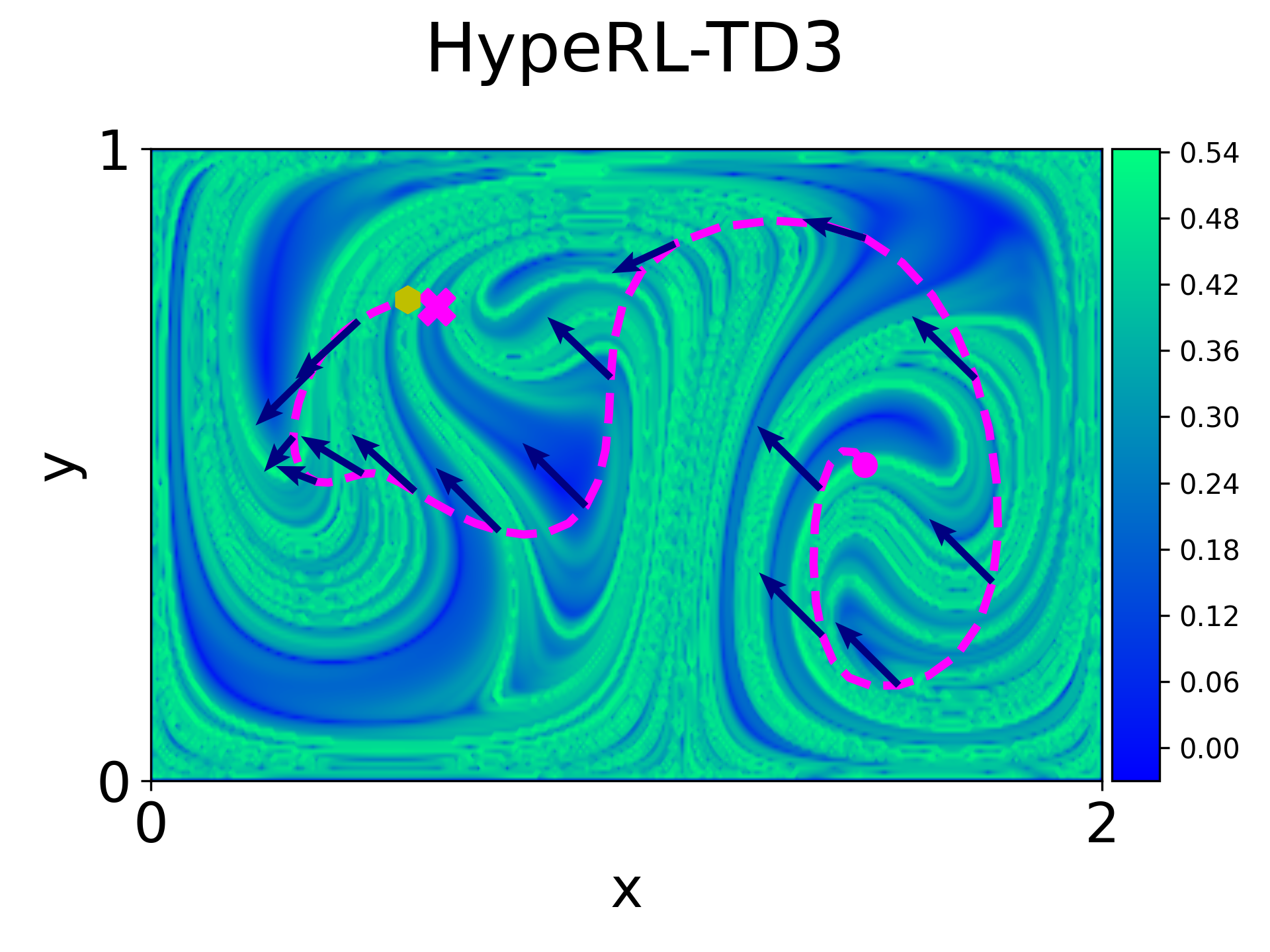}}
    \vspace{-0.1cm}
    \subfloat{\includegraphics[height=0.7\textwidth]{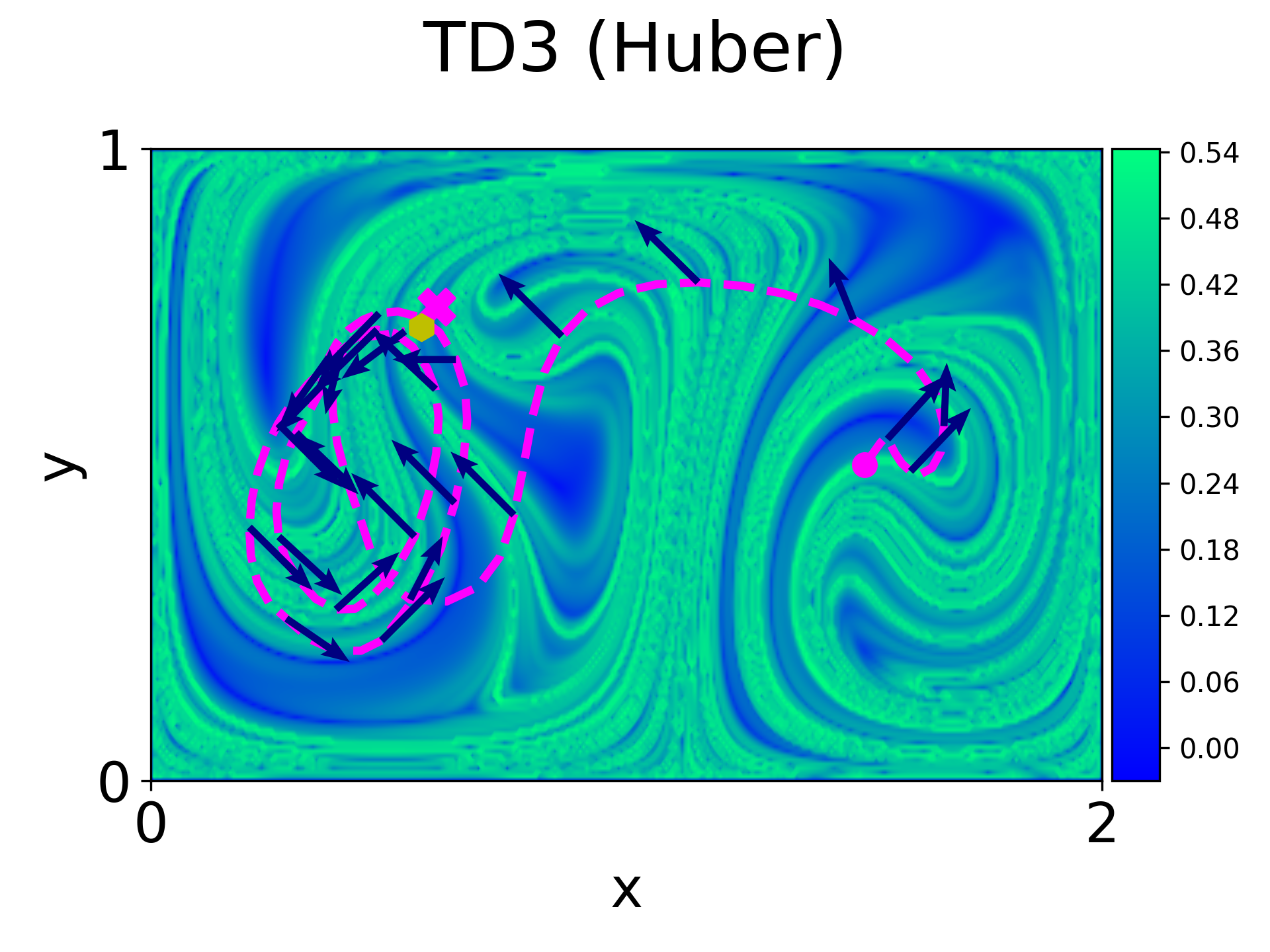}}
    \end{minipage}
    \begin{minipage}{0.49\linewidth}
    \centering \subfloat{\includegraphics[height=0.7\textwidth]{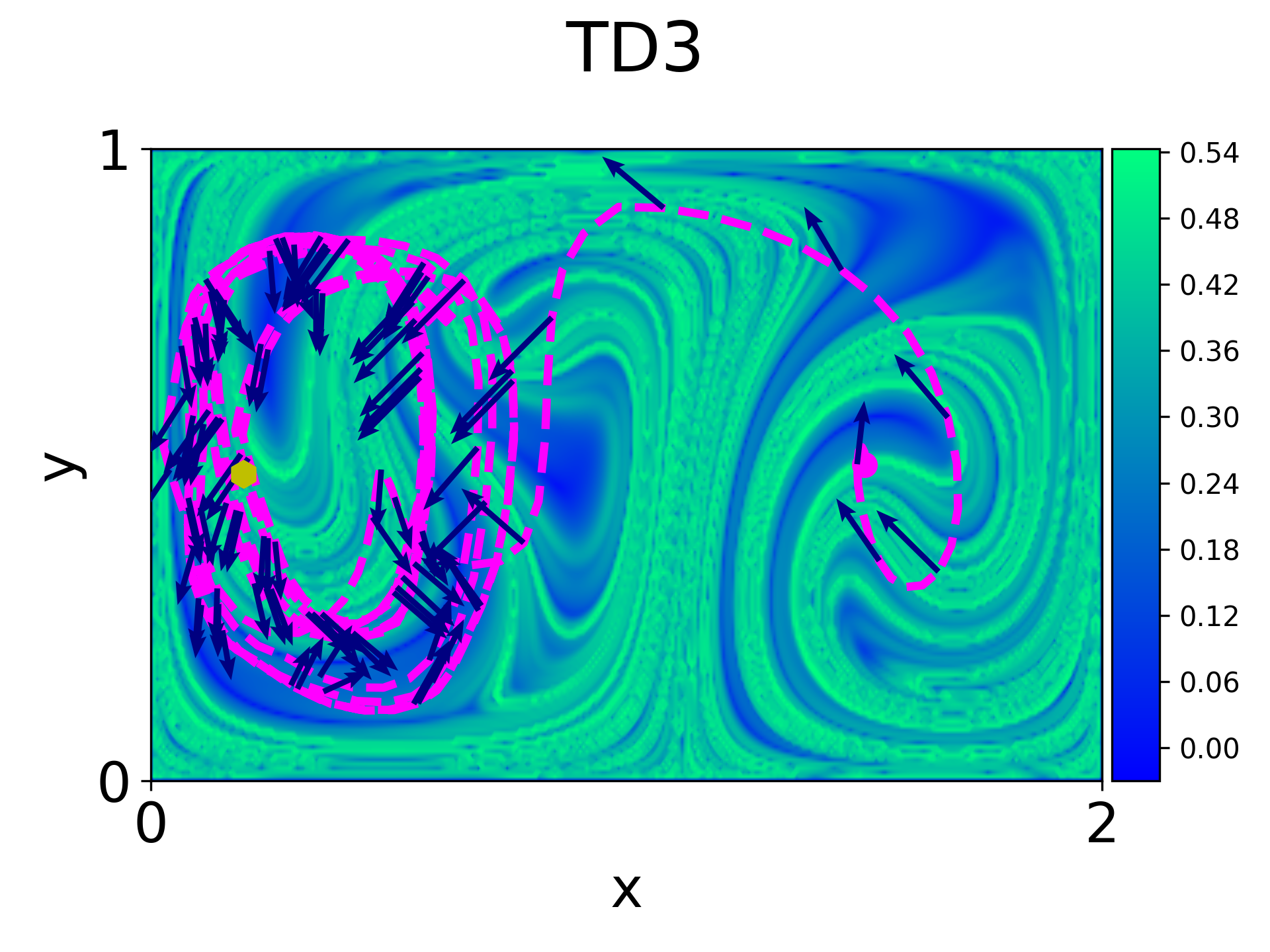}}
    \vspace{-0.1cm}
    \subfloat{\includegraphics[height=0.7\textwidth]{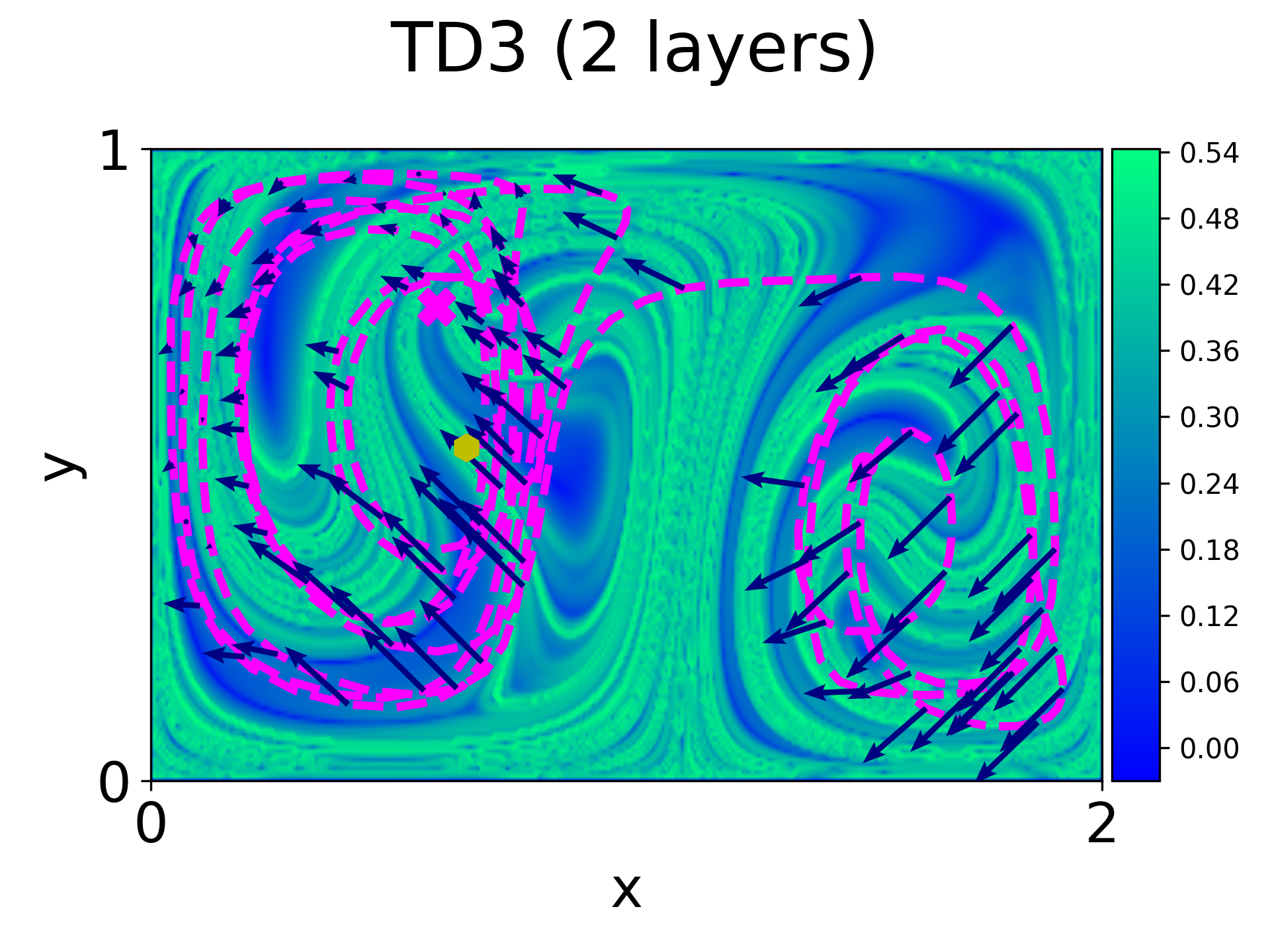}}
    \end{minipage}
    \caption{Trajectories (magenta dash line) of a controlled particle (yellow hexagon) in a gyre flow when using the different agents overlaid with the FTLE. The magenta circle indicates the starting location and the magenta cross the target one and the blue arrows represent the control inputs.}
    \label{fig:param_gyro_controlled_FTLE}
\end{figure*}

Eventually, in Figure \ref{fig:param_gyro_controlled} and \ref{fig:param_gyro_controlled_FTLE}, we show the trajectory of the particle when controlled by the different agents where the background represents the flow field at the last timestep of the trajectory and the FTLE computed over an interval of $T=10$s with a $dt=0.1$, respectively. While the different variants of TD3 get influenced by the ridges of the FTLE (see Figure \ref{fig:param_gyro_controlled_FTLE}) and achieve trajectories that oscillate around the target that require more time and control effort, HypeRL-TD3 smoothly and efficiently reaches the target by exploiting the stable regions of the gyre flow field.

\section{Discussion and Conclusion}\label{sec:conclusion}

In this paper, we proposed a novel approach for optimal control of parametric dynamical systems using RL and hypernetworks that we named HypeRL. In particular, we focused on developing a framework for enhancing the performance of RL algorithms in terms of sample efficiency and generalization to new instances of the dynamical system parameters. We tested the capabilities of HypeRL on two challenging control problems of chaotic and parametric dynamical, namely a 1D Kuramoto Sivashinsky and a navigation problem in a 2D gyre flow. We showed that knowledge of the dynamical systems parameters and how this information is encoded, i.e., via hypernetworks, is an essential ingredient for sample-efficiency and for learning of control policies that can generalize effectively.

In this work, we assumed perfect knowledge of the dynamical system and task-dependent parameters. While the knowledge of the physical parameters might seem a strong assumption, the advances of machine learned has opened the possibility of estimating the physical parameters from raw data, e.g., in \cite{tomasetto2025reduced}. As future work, we plan to incorporate a parameter-estimation step in our control pipeline and develop a two-stage procedure to avoid the need for exact knowledge of the physical parameters.

In our numerical approaches, we showed that the parametrization of policy and value function matters for improving the performance of the agent. Starting from the TD3 agent with three fully connected layers, we tested a two-layer architecture to match the HypeRL-TD3. However, removing one layer and reducing the learnable parameters of the TD3 agent drastically reduces the cumulative reward achieved by the agent. In addition, we replace the mean-squared error loss with the Huber loss to update the value function to match the HypeRL-TD3 update, but while it was beneficial for the improving mean reward and reducing the standard deviation across the different seed compared to TD3, it is still not enough to achieve comparable performance to HypeRL-TD3. These results strengthen our claim that hypernetwork-based parametrizations better encode state, task-dependent, and physical parameters into the learnable parameters of the main policy and value function networks.  Although the total learnable parameters are higher due to the presence of hypernetworks, the optimization of these learnable parameters becomes simpler, requires less data, and allows the learning of better controllers.

\section*{Acknowledgments}
NB and AM acknowledge the Project “Reduced Order Modeling and Deep Learning for the real-time approximation of PDEs (DREAM)” (Starting Grant No. FIS00003154), funded by the Italian Science Fund (FIS) - Ministero dell'Università e della Ricerca. AM also acknowledges the project “Dipartimento di Eccellenza” 2023-2027 funded by MUR and the project FAIR (Future Artificial Intelligence Research), funded by the NextGenerationEU program within the PNRR-PE-AI scheme (M4C2, Investment 1.3, Line on Artificial Intelligence). NB and AM are members of the National Group of Scientific Computing (GNCS).

\bibliographystyle{unsrt}  
\bibliography{references}

\appendix
\input{appendix}

\end{document}

%% file: appendix.tex
\clearpage
\section{Numerical Computation of the Finite Time Lyapunov Exponent}\label{app:computations_FTLE}
To compute the FTLE field for the dynamical systems $\bm{F}(\bm{y}(t))$ we follow the same procedure as in \cite{krishna2023finite}. We initialize the grid of passive particles at time $t_0=0.0$ and integrate them using $\bm{F}(\bm{y}(t))$ for a fixed amount of time $T=10.0$. This results in a flow map:
\begin{equation}
    \boldsymbol{\Phi}_{t_0}^{t_0+T}:\bm{y}(t_0) \mapsto \bm{y}(t_0) + \int_{t_0}^{t_0+T}\bm{F}(\bm{y}(\tau)) d\tau.
\end{equation}
The flow map operator $\boldsymbol{\Phi}_{t_0}^{t_0+T}$ maps an initial state at $t=t_0$ to a final state at $t=t_0 + T$. After computing the flow map, its Jacobian can be computed using finite differences as:
\begin{equation}
    (\bm{D}\boldsymbol{\Phi}_{t_0}^{t_0+T})_{i,j} = \Bigg[
    \begin{matrix}
        d_{0,0} & d_{0,1} \\
        d_{1,0} & d_{1,1} \\
    \end{matrix}\Bigg]
\end{equation}
where $i, j$ correspond to the indices of the particles in the grid and
\begin{equation}
    \begin{split}
        d_{0,0} &= \frac{x_{i+1, j}(t_0+T)-x_{i-1, j}(t_0+T)}{x_{i+1, j}(t_0)-x_{i-1, j}(t_0)} \, , \\
        d_{0,1} &= \frac{x_{i, j+1}(t_0+T)-x_{i, j-1}(t_0+T)}{y_{i, j-1}(t_0)-y_{i, j-1}(t_0)}\, , \\
        d_{1,0} &= \frac{y_{i+1, j}(t_0+T)-y_{i-1, j}(t_0+T)}{x_{i+1, j}(t_0)-x_{i-1, j}(t_0)}\, , \\
        d_{1,1} &= \frac{y_{i, j+1}(t_0+T)-y_{i, j-1}(t_0+T)}{y_{i, j+1}(t_0)-y_{i, j-1}(t_0)}\, , \\
    \end{split}
\end{equation}
where $\bm{y} = [x, y]$. Using the Jacobian of the flow map, we can compute the Cauchy-Green deformation tensor
\begin{equation}
    \boldsymbol{\Delta}_{i,j} = (\bm{D}\boldsymbol{\Phi}_{t_0}^{t_0+T})_{i,j}^{\top}(\bm{D}\boldsymbol{\Phi}_{t_0}^{t_0+T})_{i,j}\, ,
\end{equation}
where $(\cdot)^\top$ represents the matrix transpose. Eventually, we can compute compute the FTLE field from $\boldsymbol{\Delta}_{i,j}$ for each particle as:
\begin{equation}
    \sigma_{i, j} = \frac{1}{T}\ln\sqrt{(\lambda_{\max})_{i,j}}\, ,
\end{equation}
where $\lambda_{\max}$ indicates the largest eigenvalue. 

\clearpage
\section{Additional Results: Stabilization of a Kuramoto-Sivashinsky Equation}\label{app:additional_results}

\subsection{Stabilization of the State of a KS Equation to an Arbitrary Reference}\label{app:KS_extra_results}
\begin{figure}[h!]
     \centering
     \begin{subfigure}[b]{0.49\textwidth}
         \centering
         \includegraphics[width=0.85\textwidth]{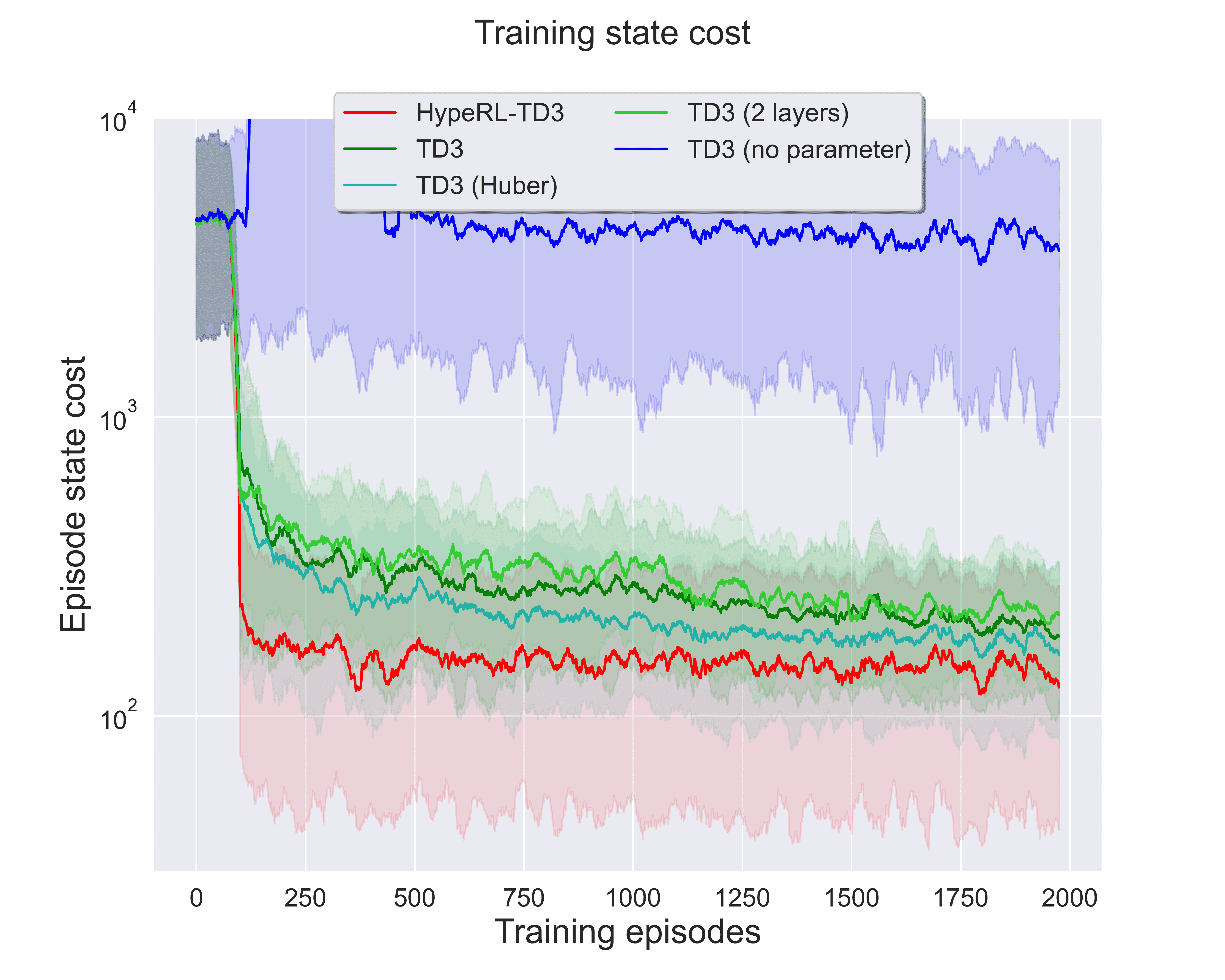}
         \caption{}
    \end{subfigure}
    \begin{subfigure}[b]{0.49\textwidth}
         \centering
         \includegraphics[width=0.85\textwidth]{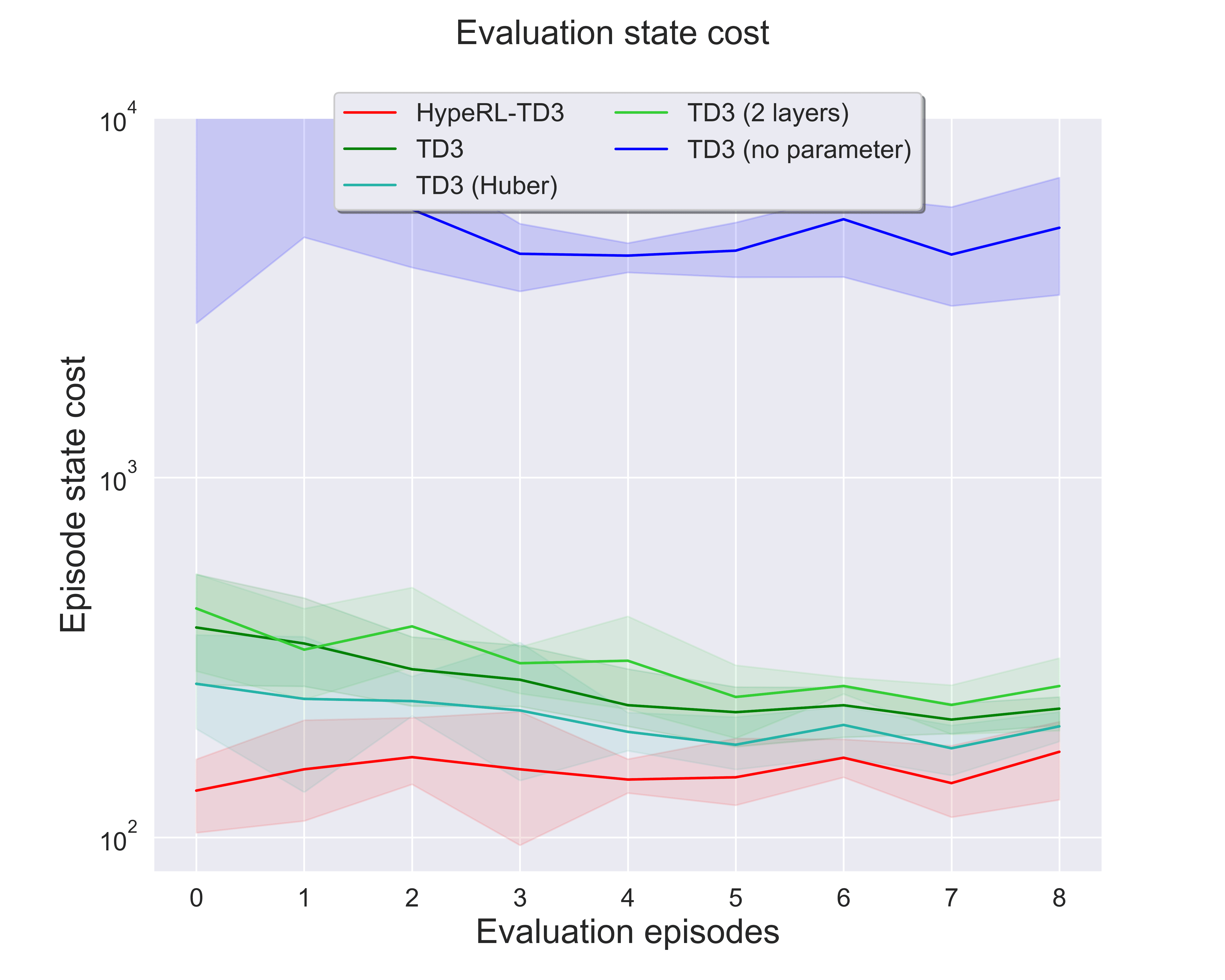}
         \caption{}
     \end{subfigure}
    \begin{subfigure}[b]{0.49\textwidth}
         \centering
         \includegraphics[width=0.85\textwidth]{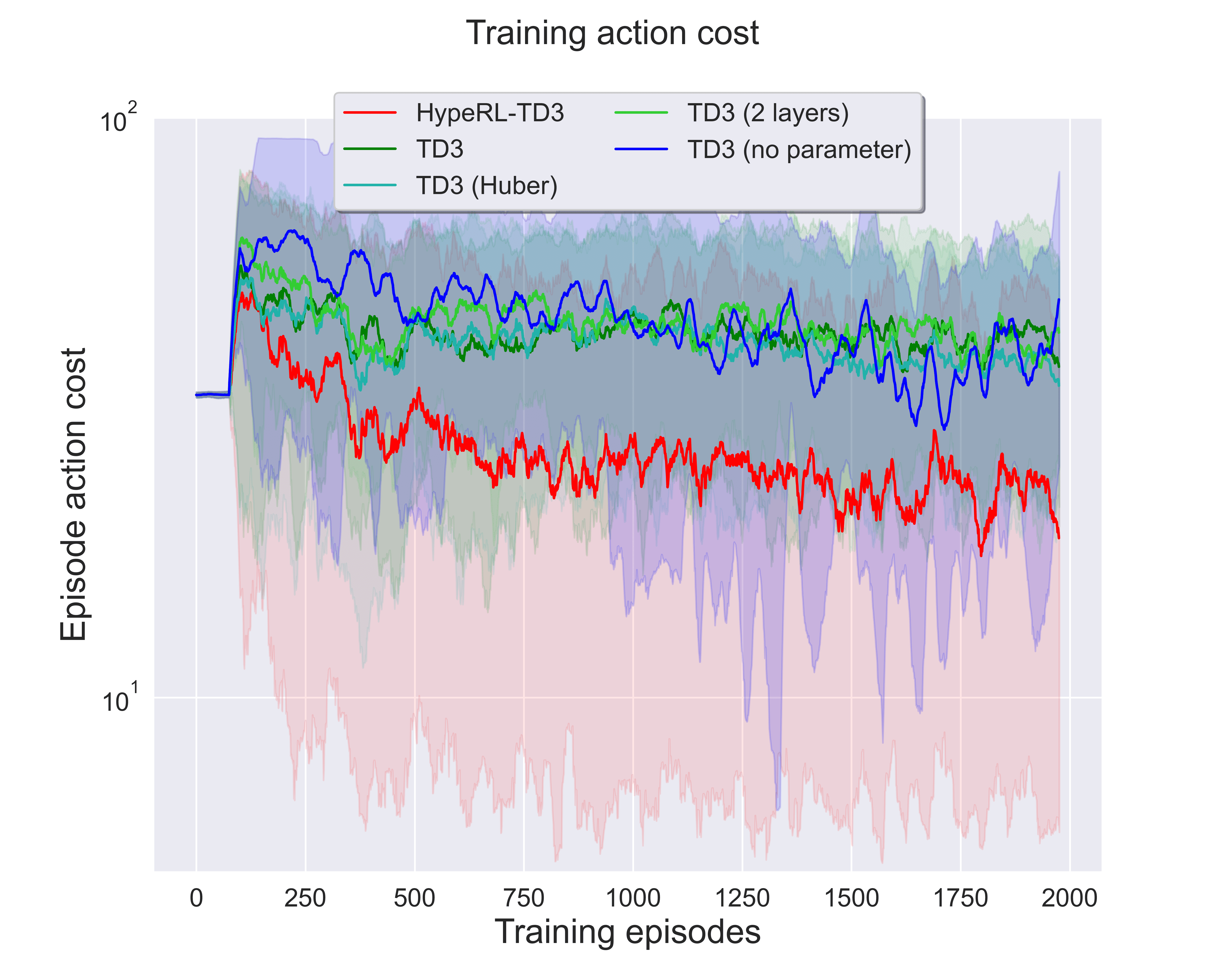}
         \caption{}
     \end{subfigure}
          \centering
    \begin{subfigure}[b]{0.49\textwidth}
         \centering
         \includegraphics[width=0.85\textwidth]{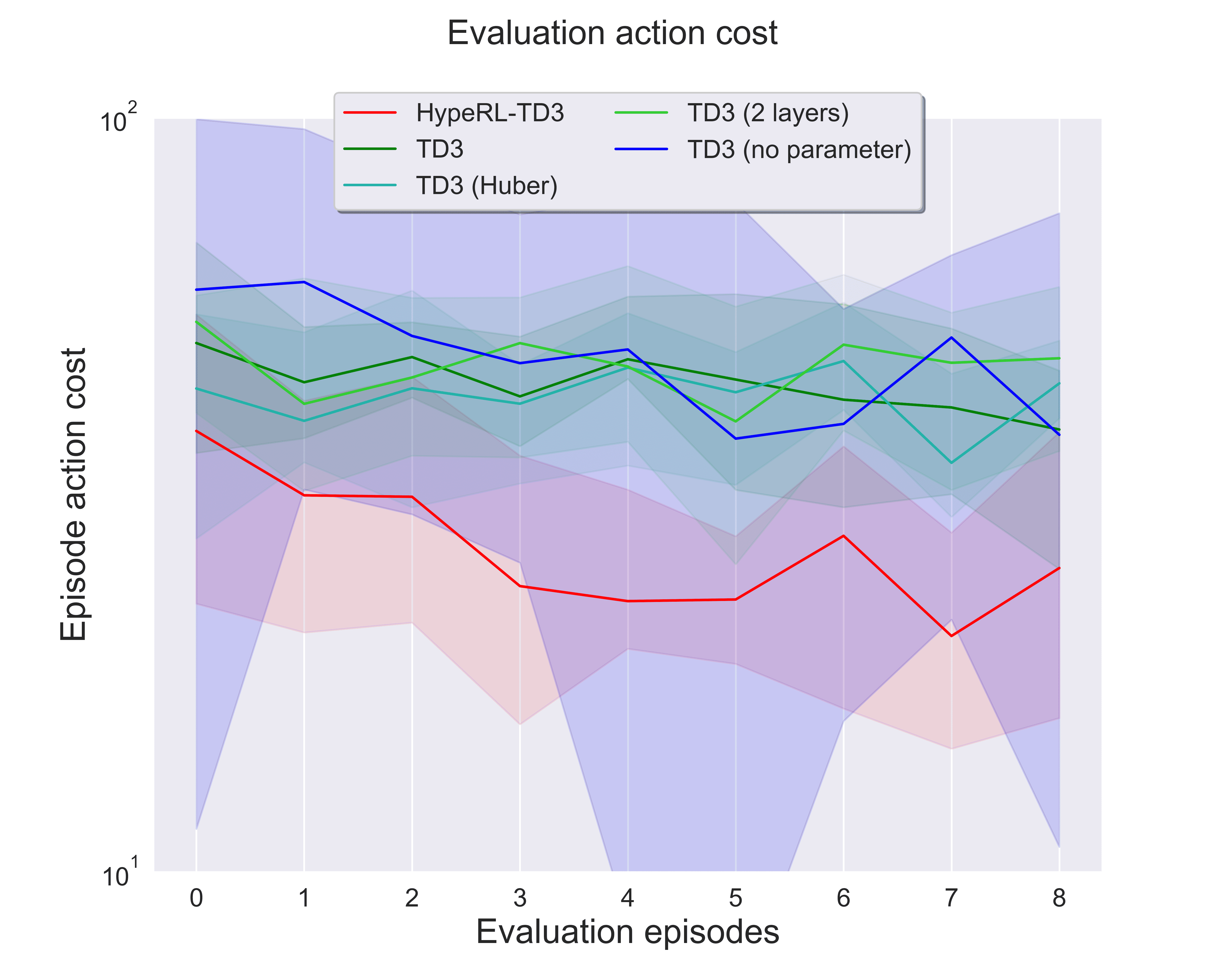}
         \caption{}
     \end{subfigure}
        \caption{Training and evaluation results. The solid line represents the mean and the shaded area the minimum and maximum values observed over 5 different random seeds.}
        \label{fig:2app}
\end{figure}

In Figure \ref{fig:2app}, we show the state and action costs over training and evaluation for the KS equation.

\begin{figure*}[h!]
    \begin{minipage}{0.49\linewidth}
    \centering \subfloat{\includegraphics[height=0.85\textwidth]{Pics/controlled_solutions_KS/hypeRL_td3_v2_1_1.png}}
    \vspace{-0.1cm}
    \subfloat{\includegraphics[height=0.85\textwidth]{Pics/controlled_solutions_KS/td3_huber_1_1.png}}
    \end{minipage}
    \begin{minipage}{0.49\linewidth}
    \centering \subfloat{\includegraphics[height=0.85\textwidth]{Pics/controlled_solutions_KS/td3_1_1.png}}
    \vspace{-0.1cm}
    \subfloat{\includegraphics[height=0.85\textwidth]{Pics/controlled_solutions_KS/td3_twolayers_1_1.png}}
    \end{minipage}
    \centering
    \begin{minipage}{0.49\linewidth}
    \centering \subfloat{\includegraphics[height=0.85\textwidth]{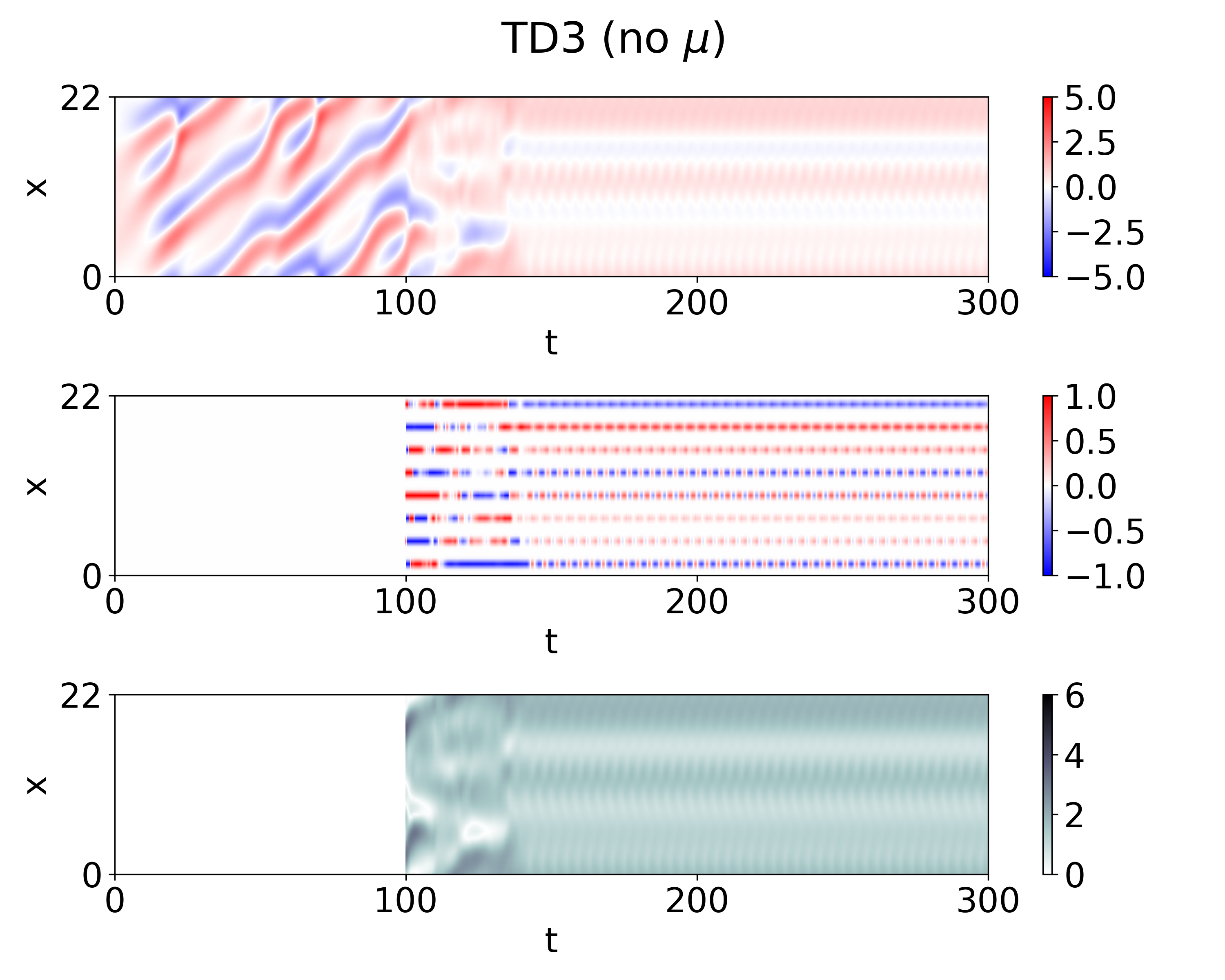}}
    \end{minipage}
    \caption{Controlled solutions of the KS equation when using the different agents. The first row shows the state, the second the control, and the third one the error $|\bm{y}_k - \bm{y}_{\text{ref}}|$.}
    \label{fig:KS_controlled_full}
\end{figure*}

\begin{figure*}
    \begin{minipage}{0.49\linewidth}
    \centering \subfloat{\includegraphics[height=0.85\textwidth]{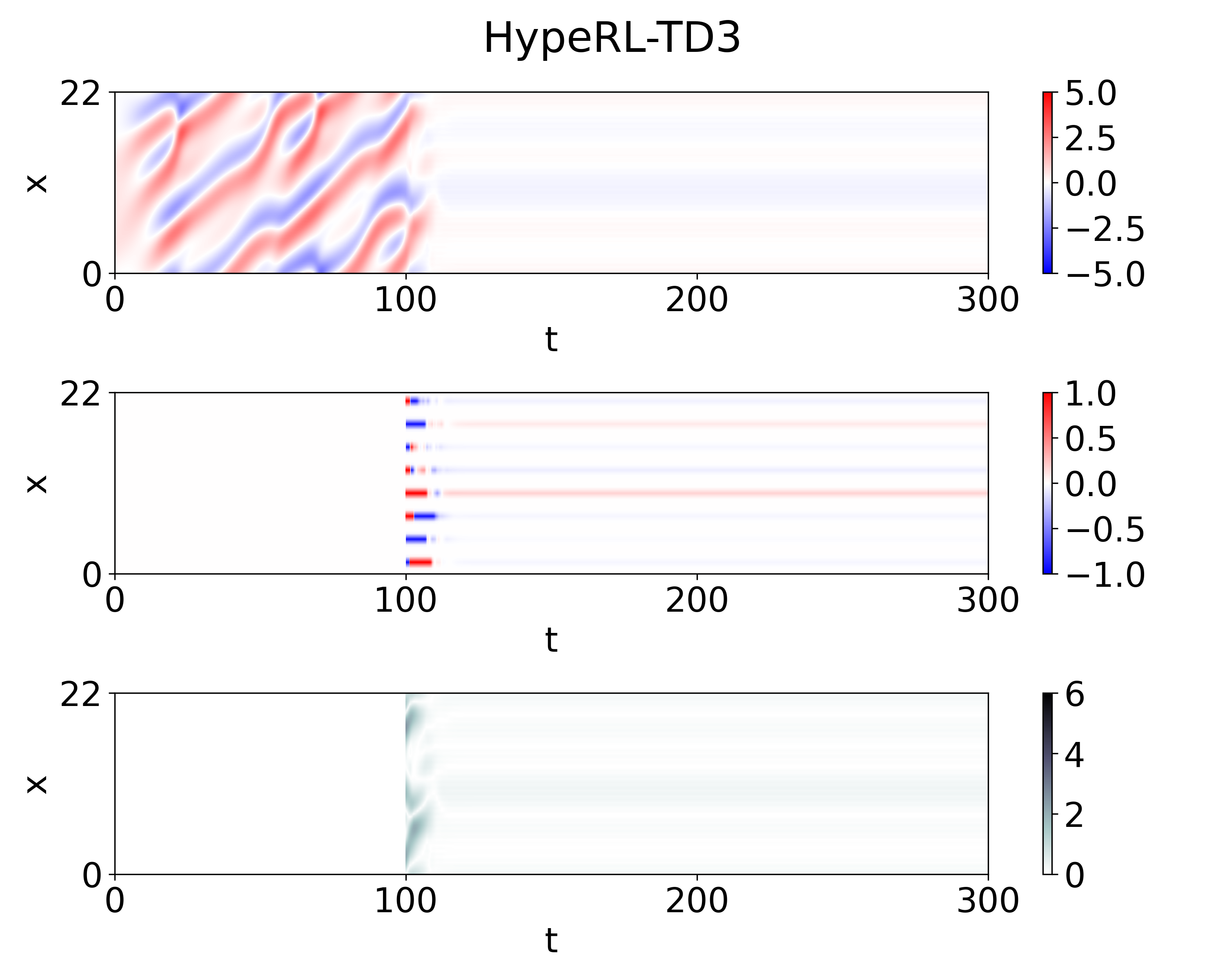}}
    \vspace{-0.1cm}
    \subfloat{\includegraphics[height=0.85\textwidth]{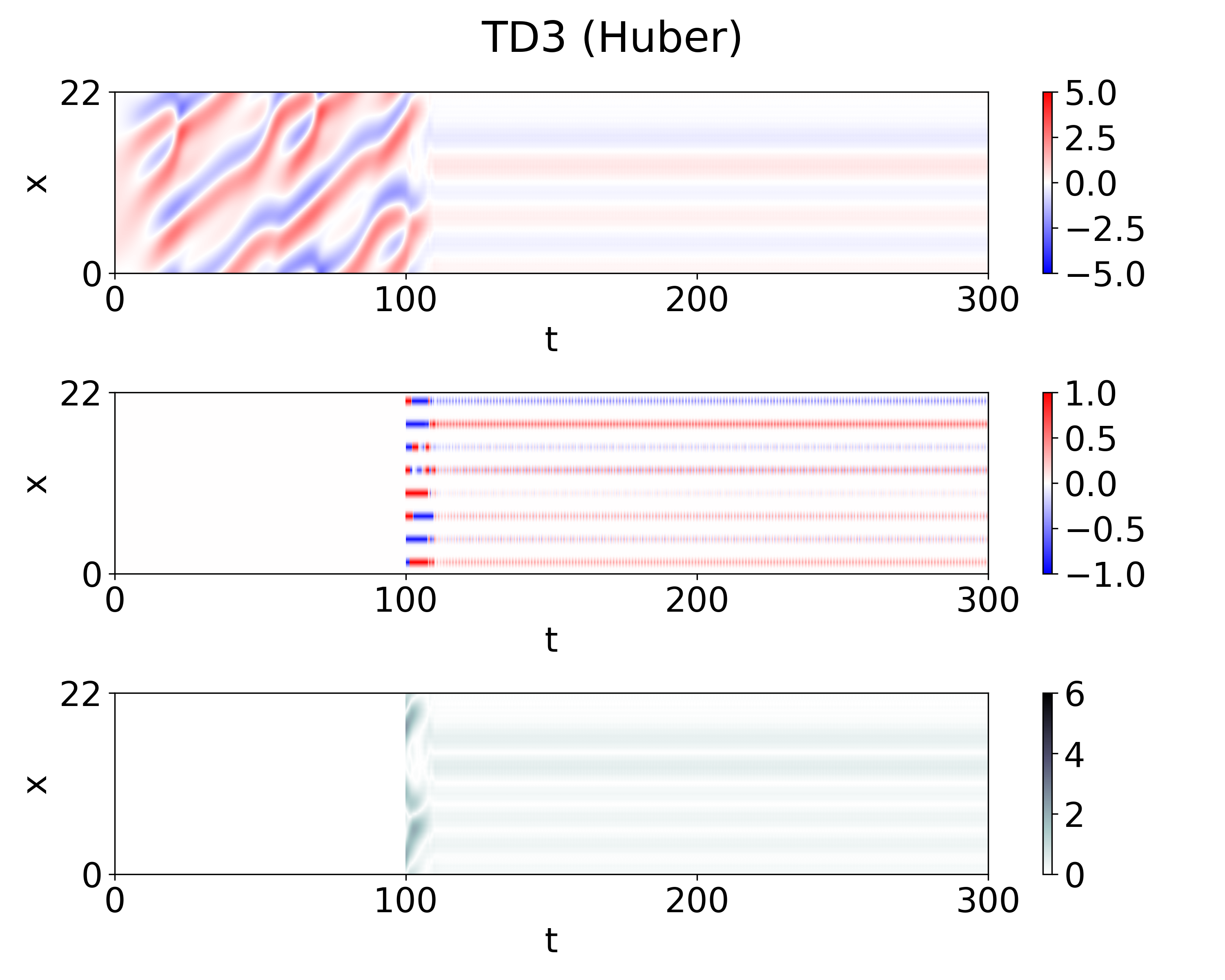}}
    \end{minipage}
    \begin{minipage}{0.49\linewidth}
    \centering \subfloat{\includegraphics[height=0.85\textwidth]{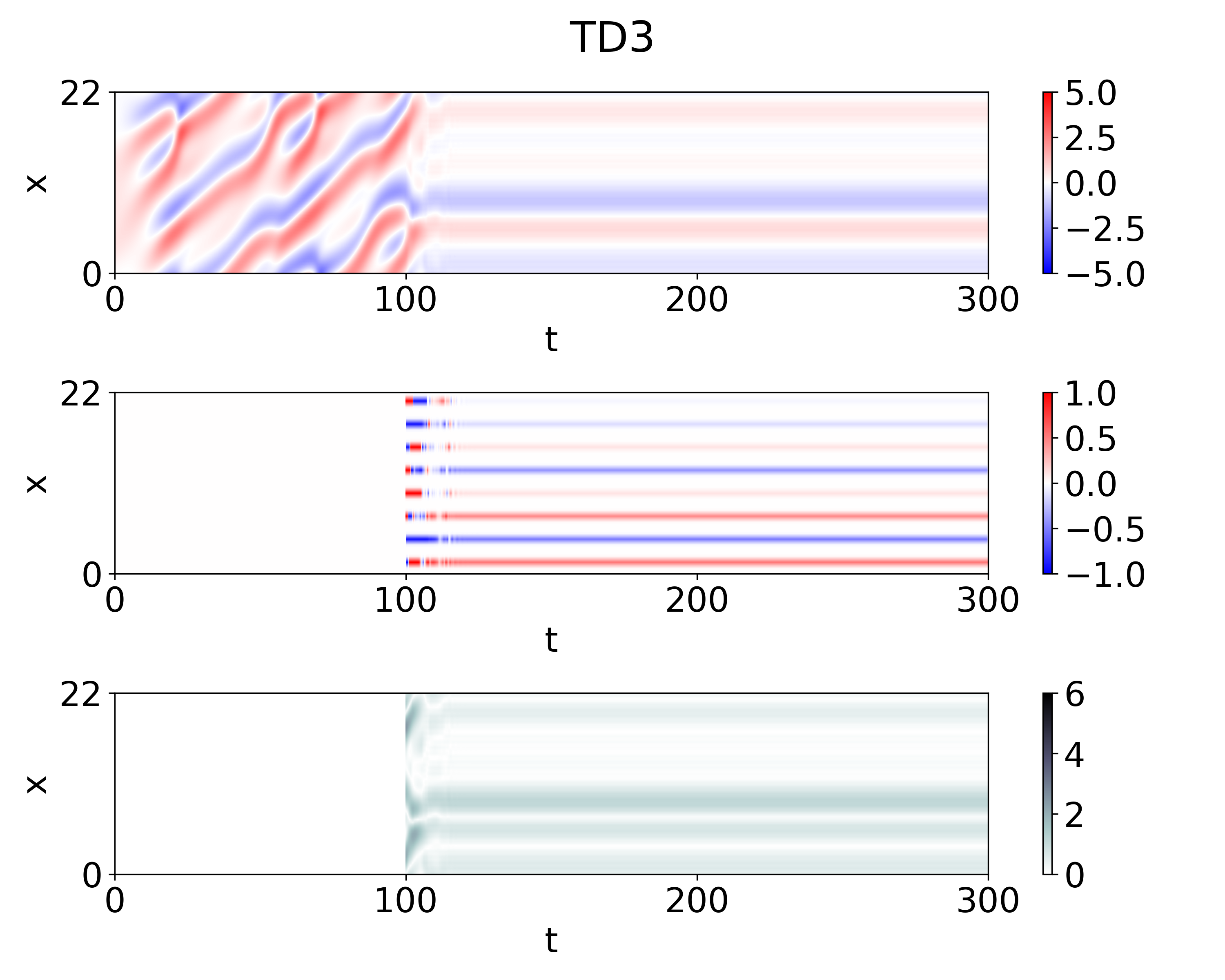}}
    \vspace{-0.1cm}
    \subfloat{\includegraphics[height=0.85\textwidth]{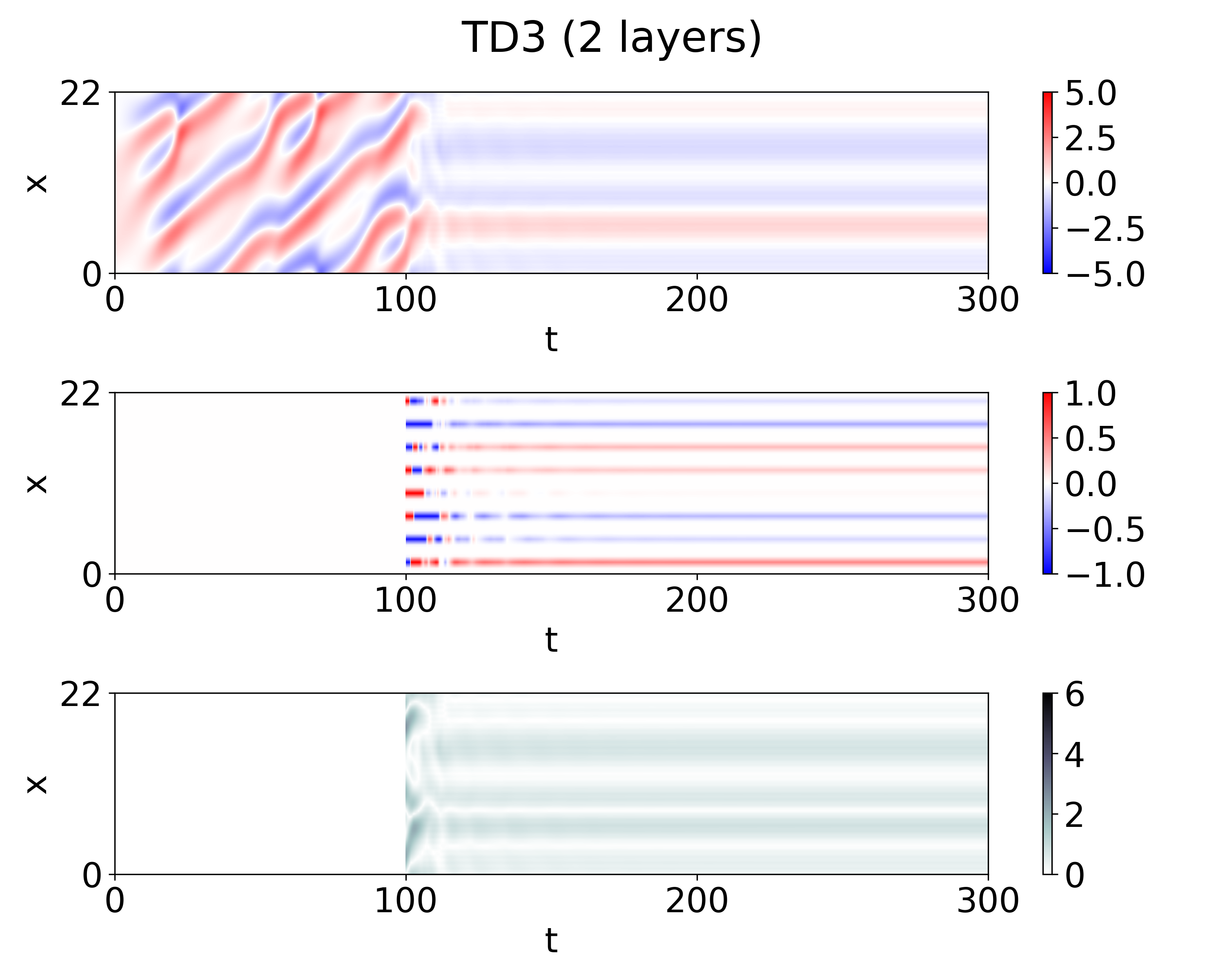}}
    \end{minipage}
    \centering
    \begin{minipage}{0.49\linewidth}
    \centering \subfloat{\includegraphics[height=0.85\textwidth]{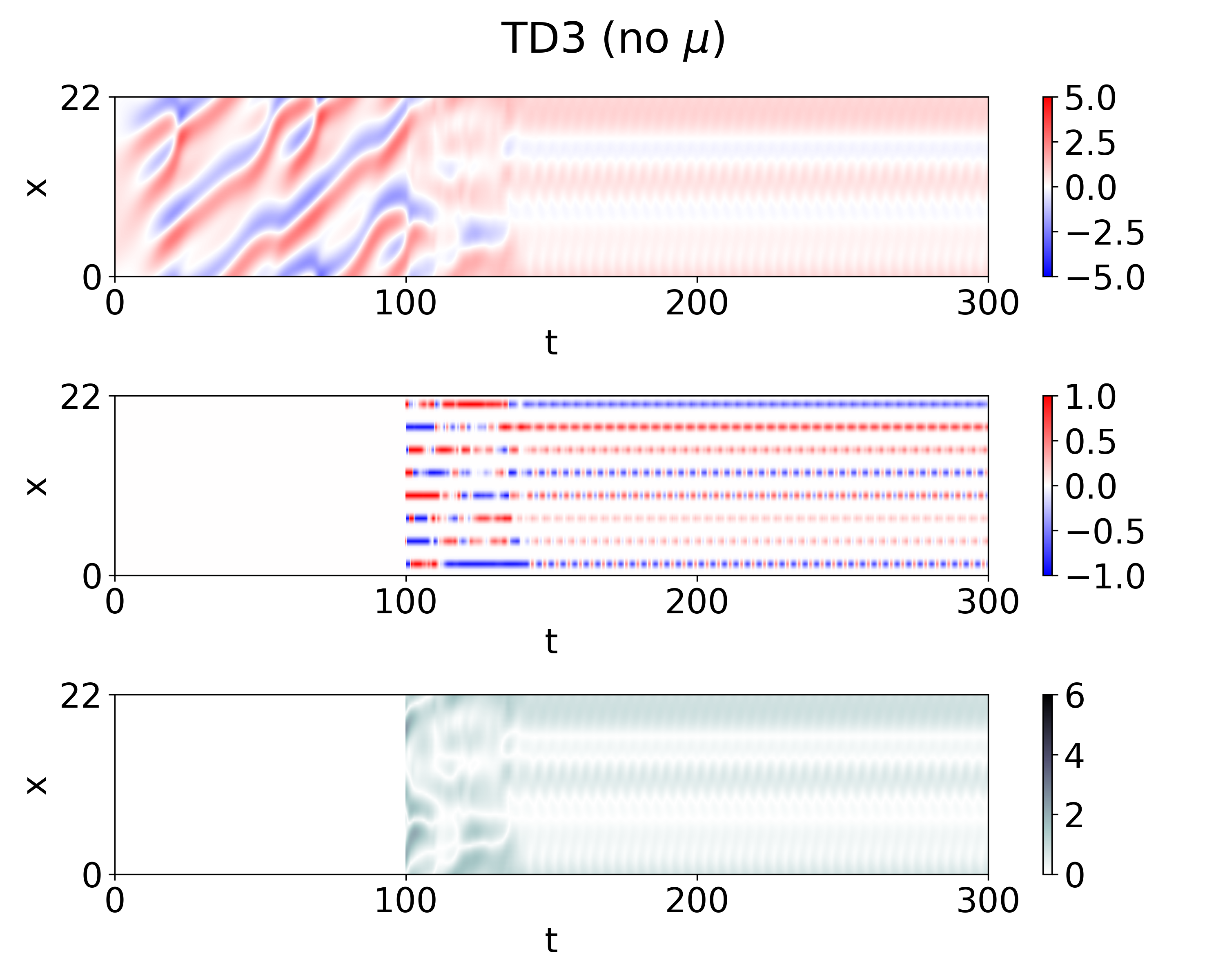}}
    \end{minipage}
    \caption{Controlled solutions of the KS equation when using the different agents for $\bm{y}_{\text{ref}}=\bm{0}$.}
    \label{fig:KS_controlled_extra1}
\end{figure*}

In Figure \ref{fig:KS_controlled_full} and \ref{fig:KS_controlled_extra1}, we show examples of controlled solution when tracking $\bm{y}_{\text{ref}}=-\bm{1}$ and $\bm{y}_{\text{ref}}=\bm{0}$.

\clearpage
\subsection{Stabilization of the State of a Parametric KS Equation to an Arbitrary Reference}\label{app:param_KS_extra_results}
\begin{figure}[h!]
     \centering
     \begin{subfigure}[b]{0.49\textwidth}
         \centering
         \includegraphics[width=0.85\textwidth]{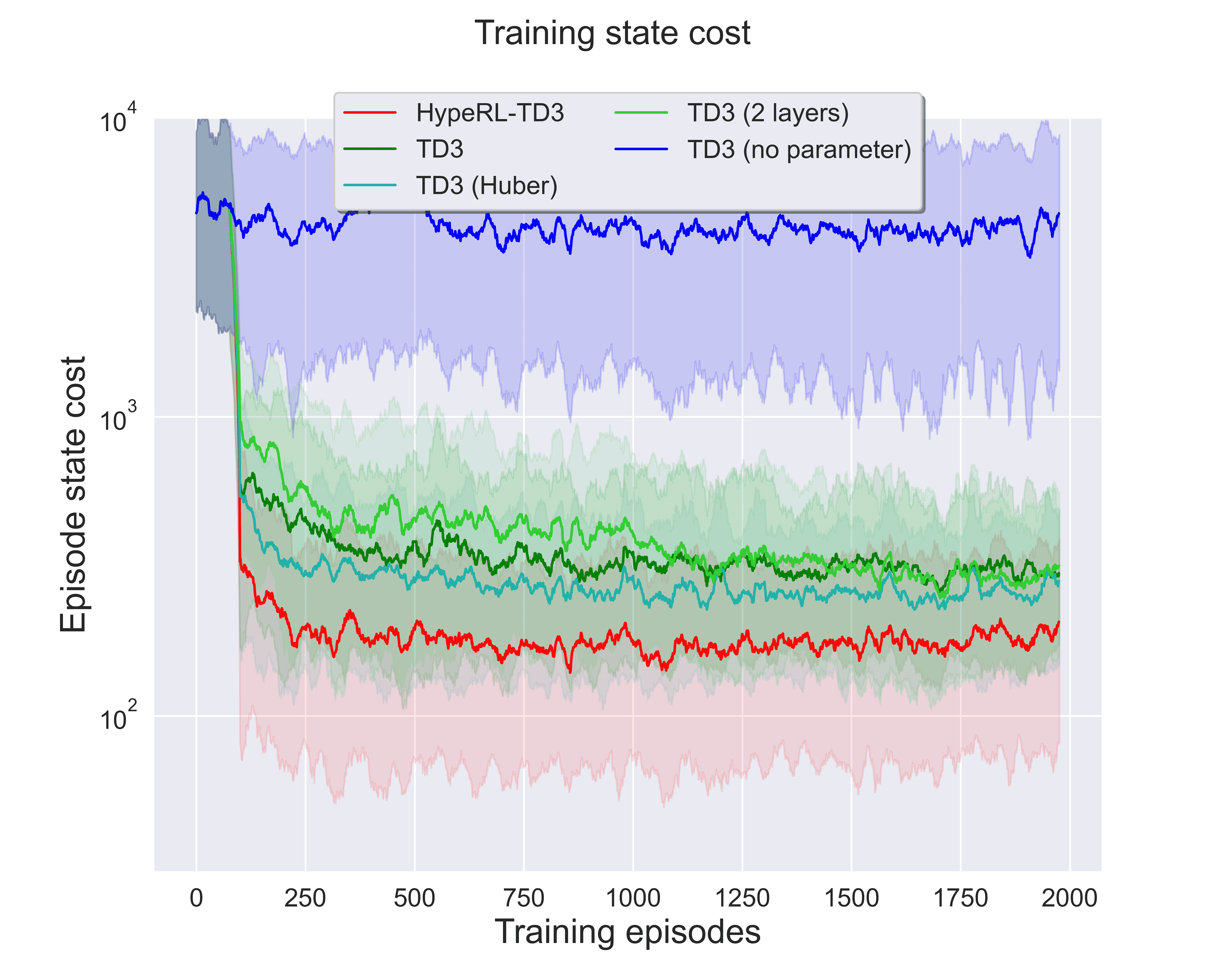}
         \caption{}
    \end{subfigure}
    \begin{subfigure}[b]{0.49\textwidth}
         \centering
         \includegraphics[width=0.85\textwidth]{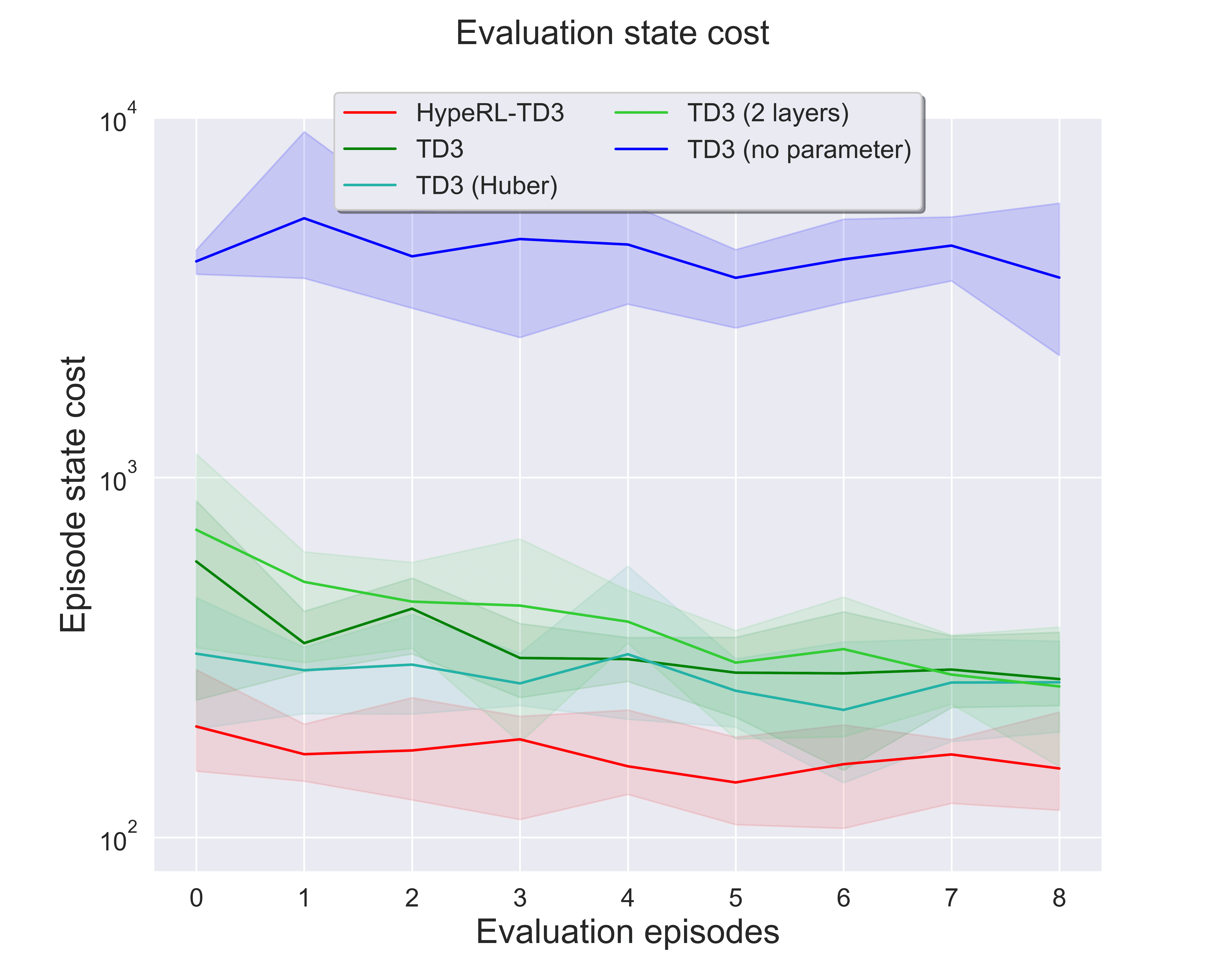}
         \caption{}
     \end{subfigure}
    \begin{subfigure}[b]{0.49\textwidth}
         \centering
         \includegraphics[width=0.85\textwidth]{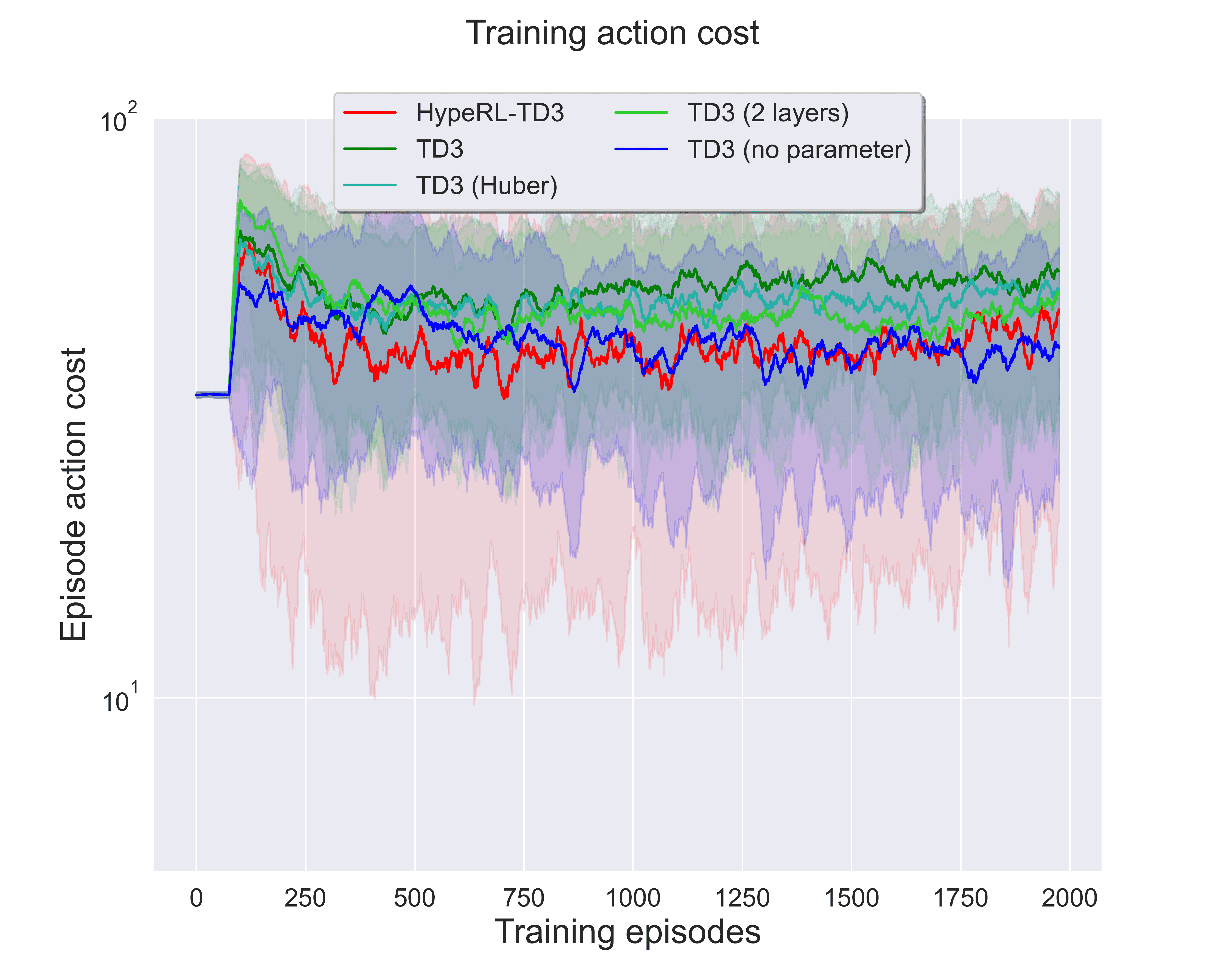}
         \caption{}
     \end{subfigure}
          \centering
    \begin{subfigure}[b]{0.49\textwidth}
         \centering
         \includegraphics[width=0.85\textwidth]{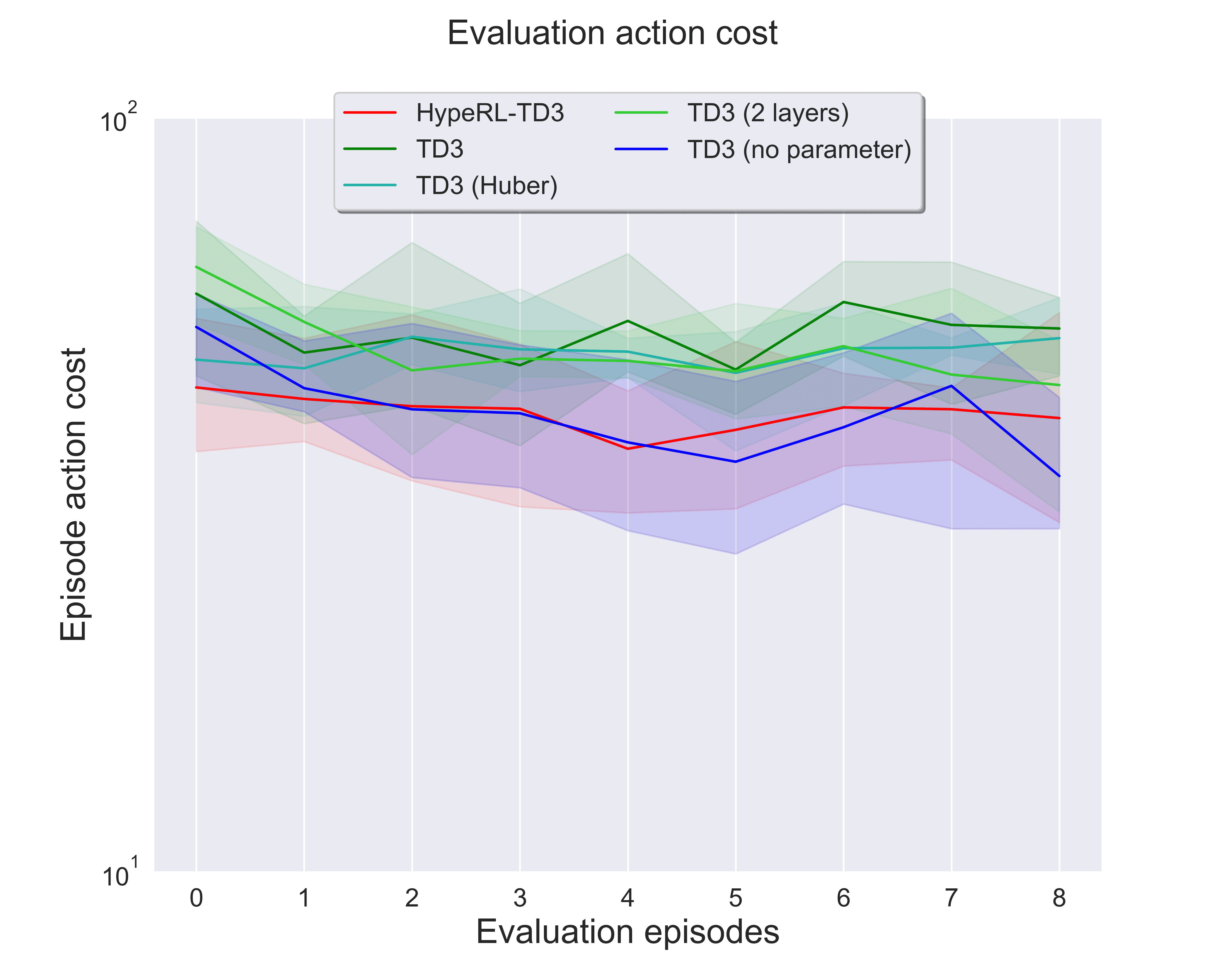}
         \caption{}
     \end{subfigure}
        \caption{Training and evaluation results. The solid line represents the mean and the shaded area the minimum and maximum values observed over 5 different random seeds.}
        \label{fig:app_KS_param_rewards_extra1}
\end{figure}

In Figure \ref{fig:app_KS_param_rewards_extra1}, we show the state and action costs over training and evaluation for the KS equation.

\begin{figure*}[h!]
    \begin{minipage}{0.49\linewidth}
    \centering \subfloat{\includegraphics[height=0.85\textwidth]{Pics/controlled_solutions_parametric_KS/hypeRL_td3_v2_1_1.png}}
    \vspace{-0.1cm}
    \subfloat{\includegraphics[height=0.85\textwidth]{Pics/controlled_solutions_parametric_KS/td3_huber_1_1.png}}
    \end{minipage}
    \begin{minipage}{0.49\linewidth}
    \centering \subfloat{\includegraphics[height=0.85\textwidth]{Pics/controlled_solutions_parametric_KS/td3_1_1.png}}
    \vspace{-0.1cm}
    \subfloat{\includegraphics[height=0.85\textwidth]{Pics/controlled_solutions_parametric_KS/td3_twolayers_1_1.png}}
    \end{minipage}
    \centering
    \begin{minipage}{0.49\linewidth}
    \centering \subfloat{\includegraphics[height=0.85\textwidth]{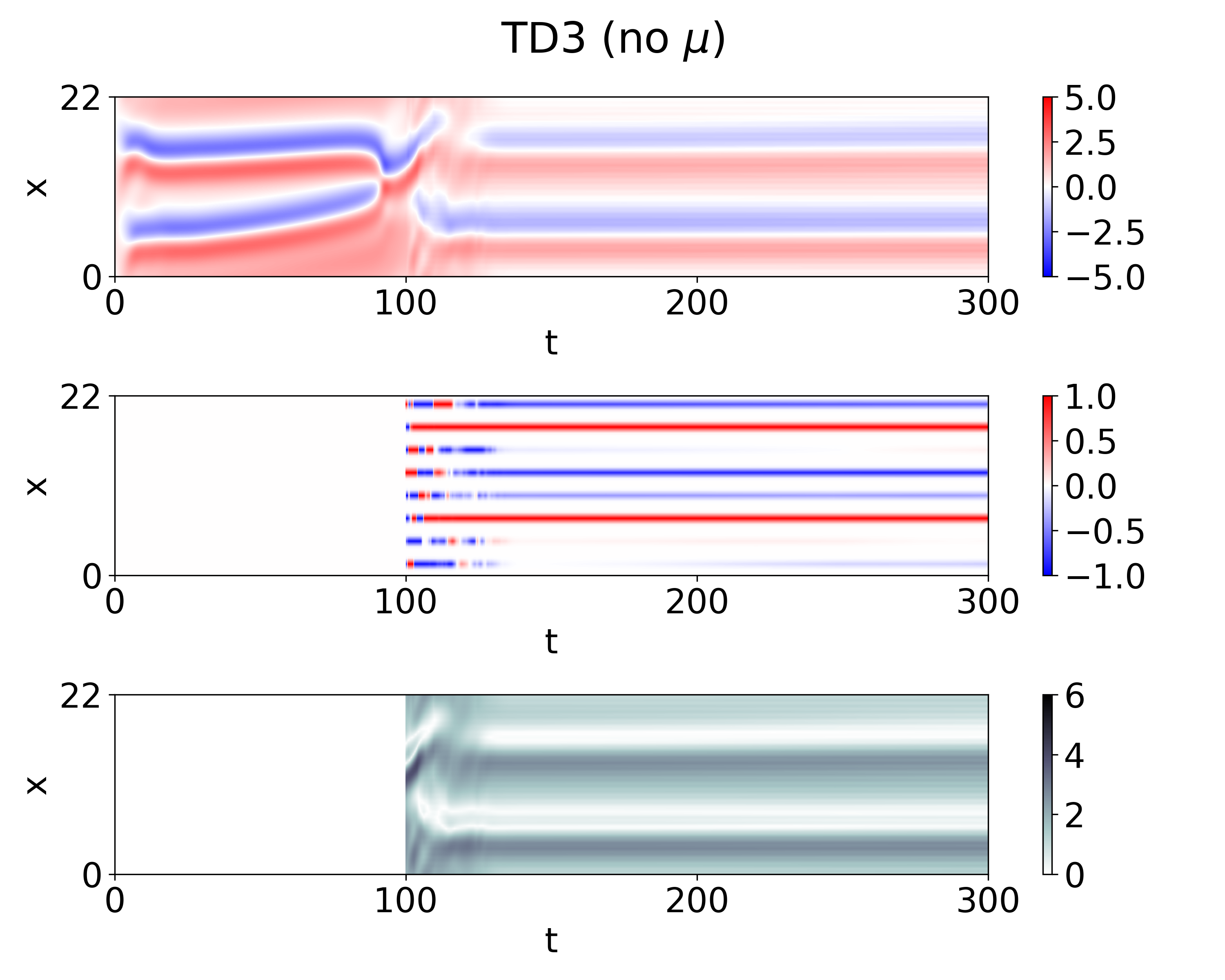}}
    \end{minipage}
    \caption{Controlled solutions of the parametric KS equation when using the different agents. The first row shows the state, the second the control, and the third one the error $|\bm{y}_k - \bm{y}_{\text{ref}}|$.}
    \label{fig:parametric_KS_controlled_full}
\end{figure*}

\begin{figure*}
    \begin{minipage}{0.49\linewidth}
    \centering \subfloat{\includegraphics[height=0.85\textwidth]{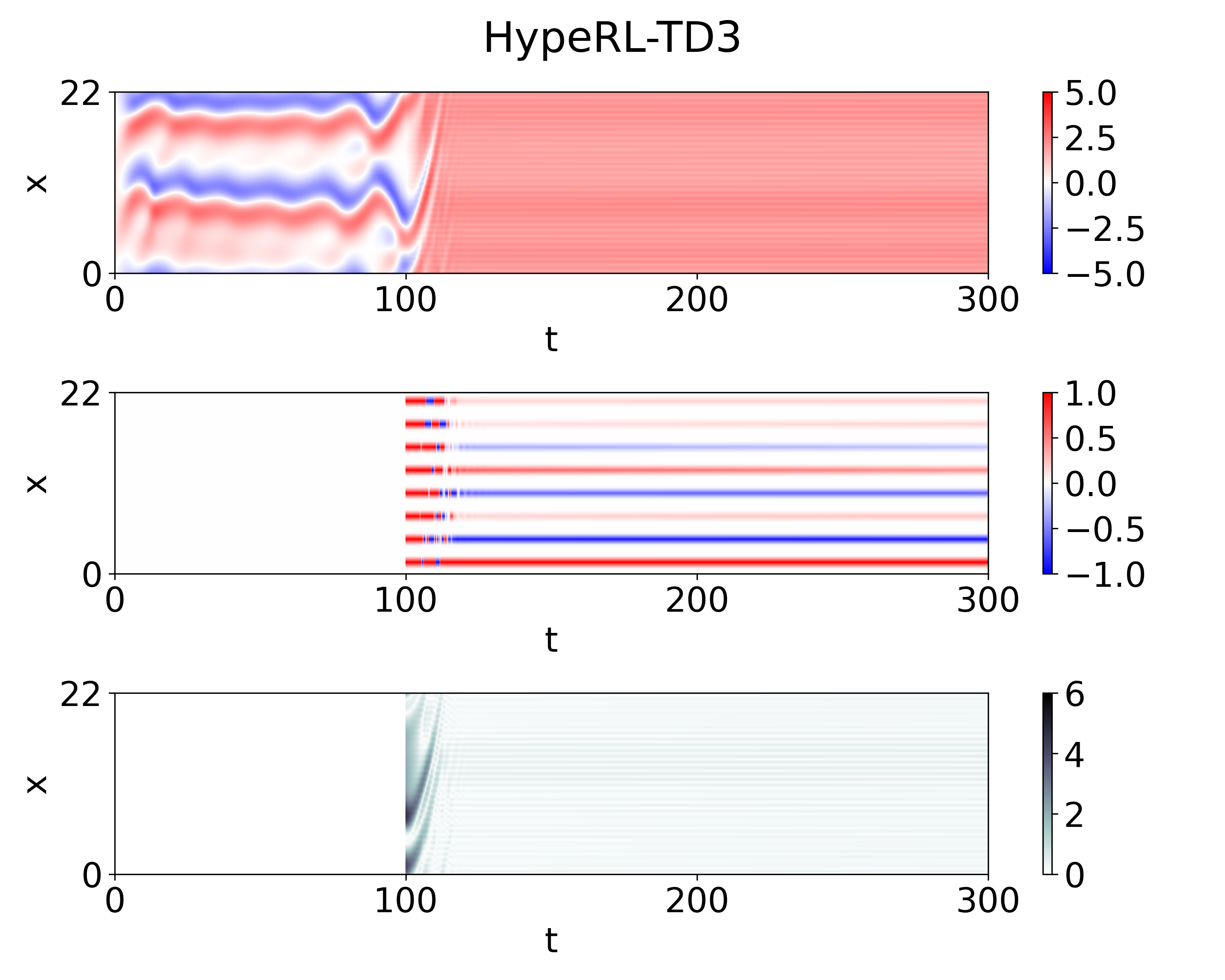}}
    \vspace{-0.1cm}
    \subfloat{\includegraphics[height=0.85\textwidth]{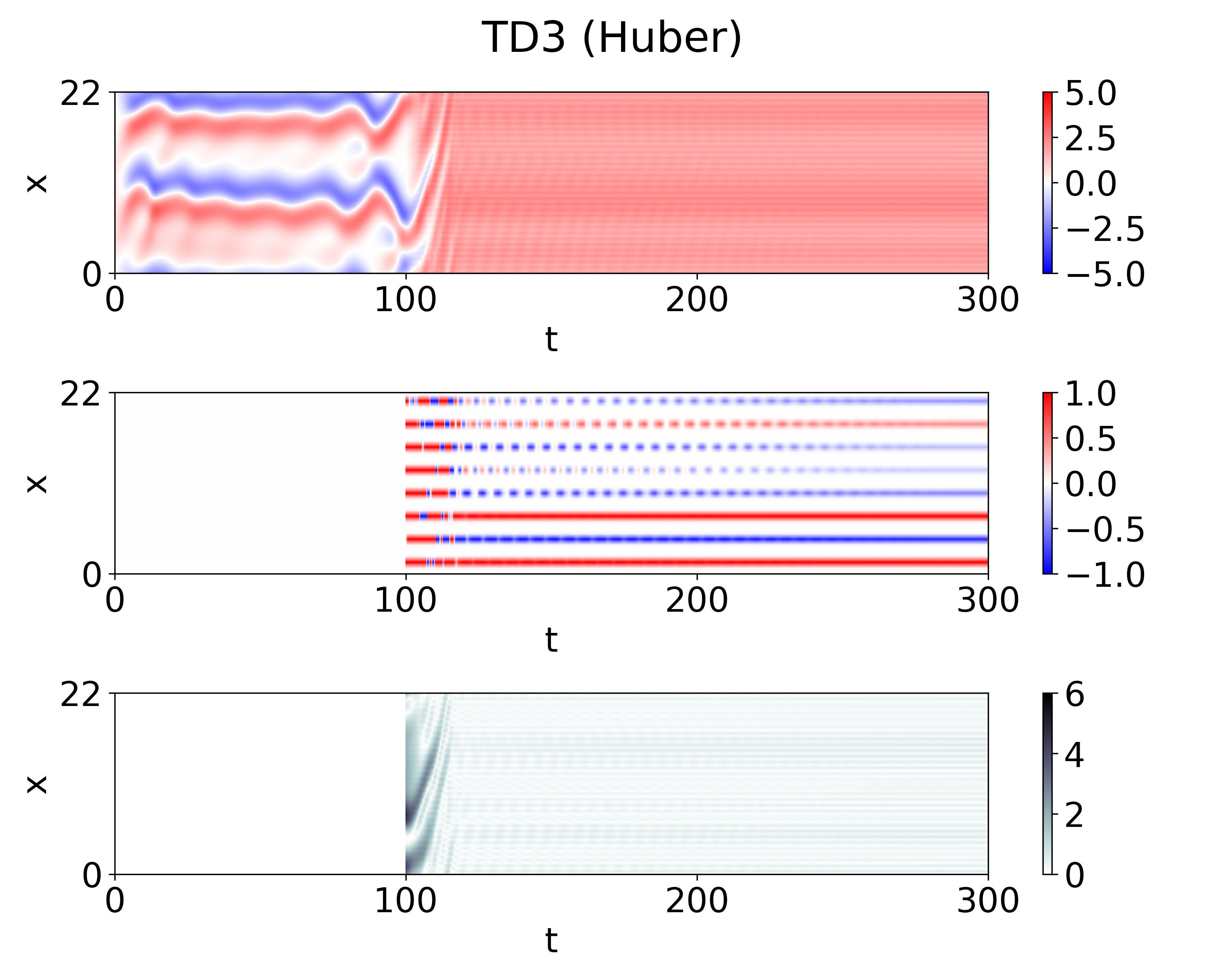}}
    \end{minipage}
    \begin{minipage}{0.49\linewidth}
    \centering \subfloat{\includegraphics[height=0.85\textwidth]{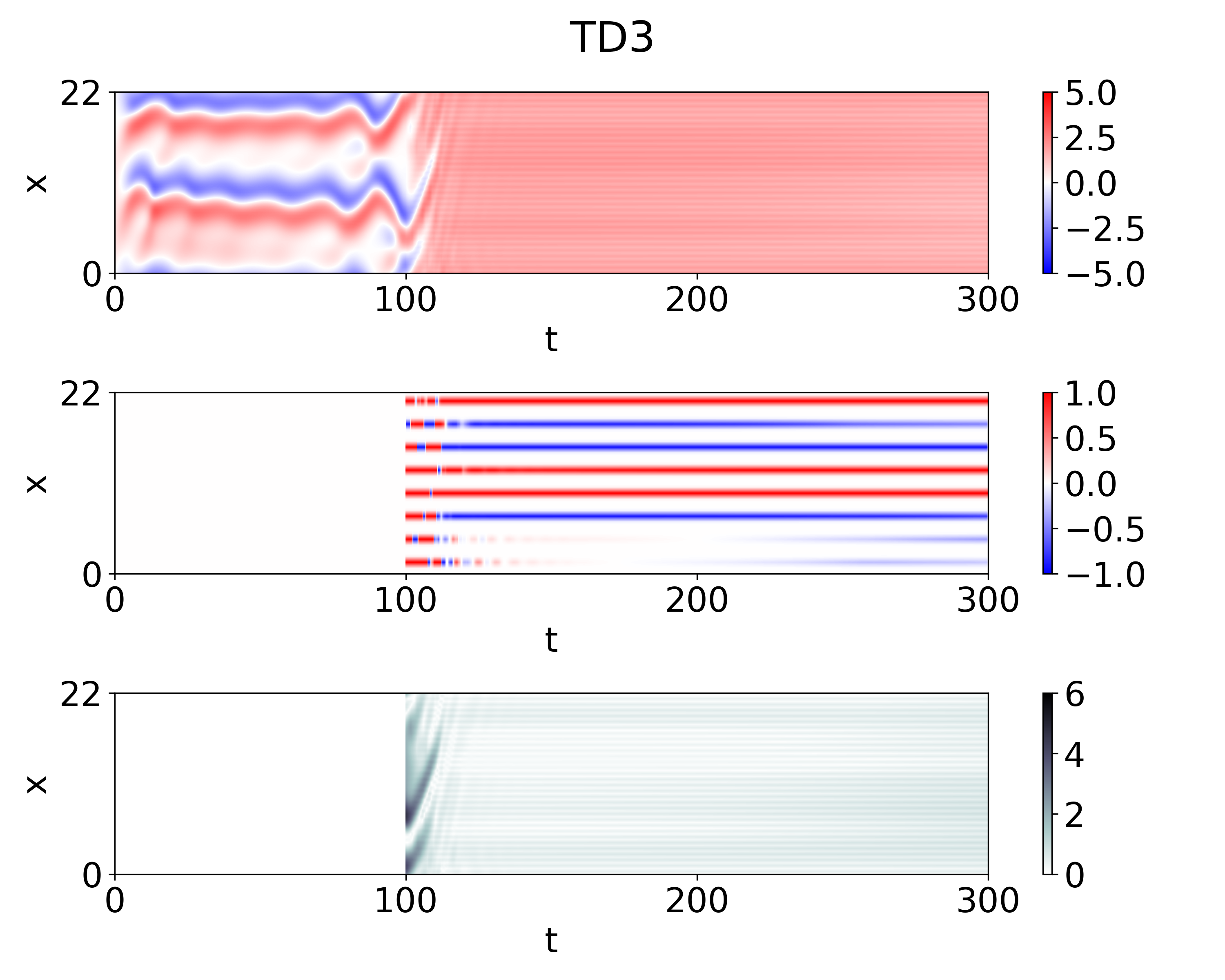}}
    \vspace{-0.1cm}
    \subfloat{\includegraphics[height=0.85\textwidth]{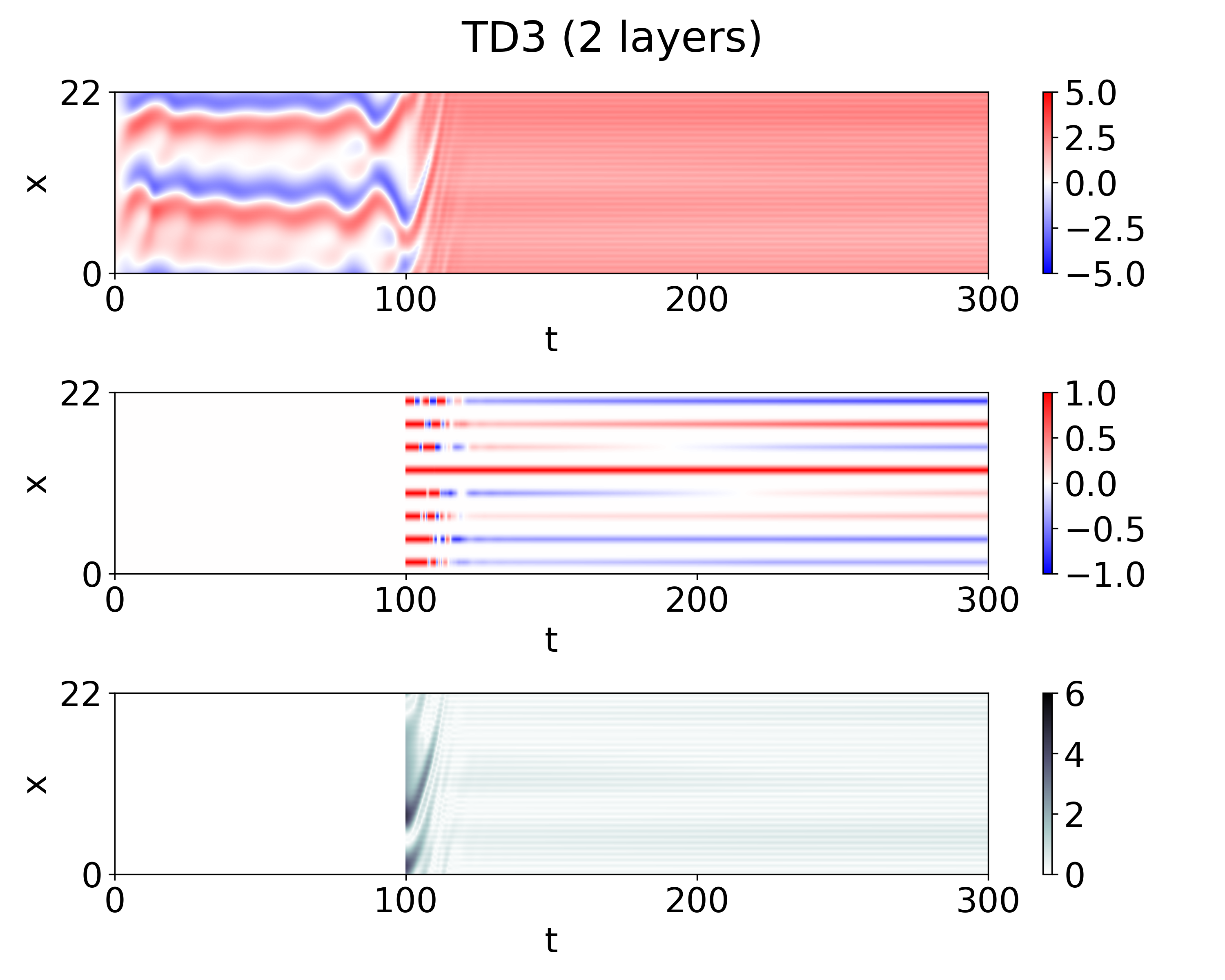}}
    \end{minipage}
    \centering
    \begin{minipage}{0.49\linewidth}
    \centering \subfloat{\includegraphics[height=0.85\textwidth]{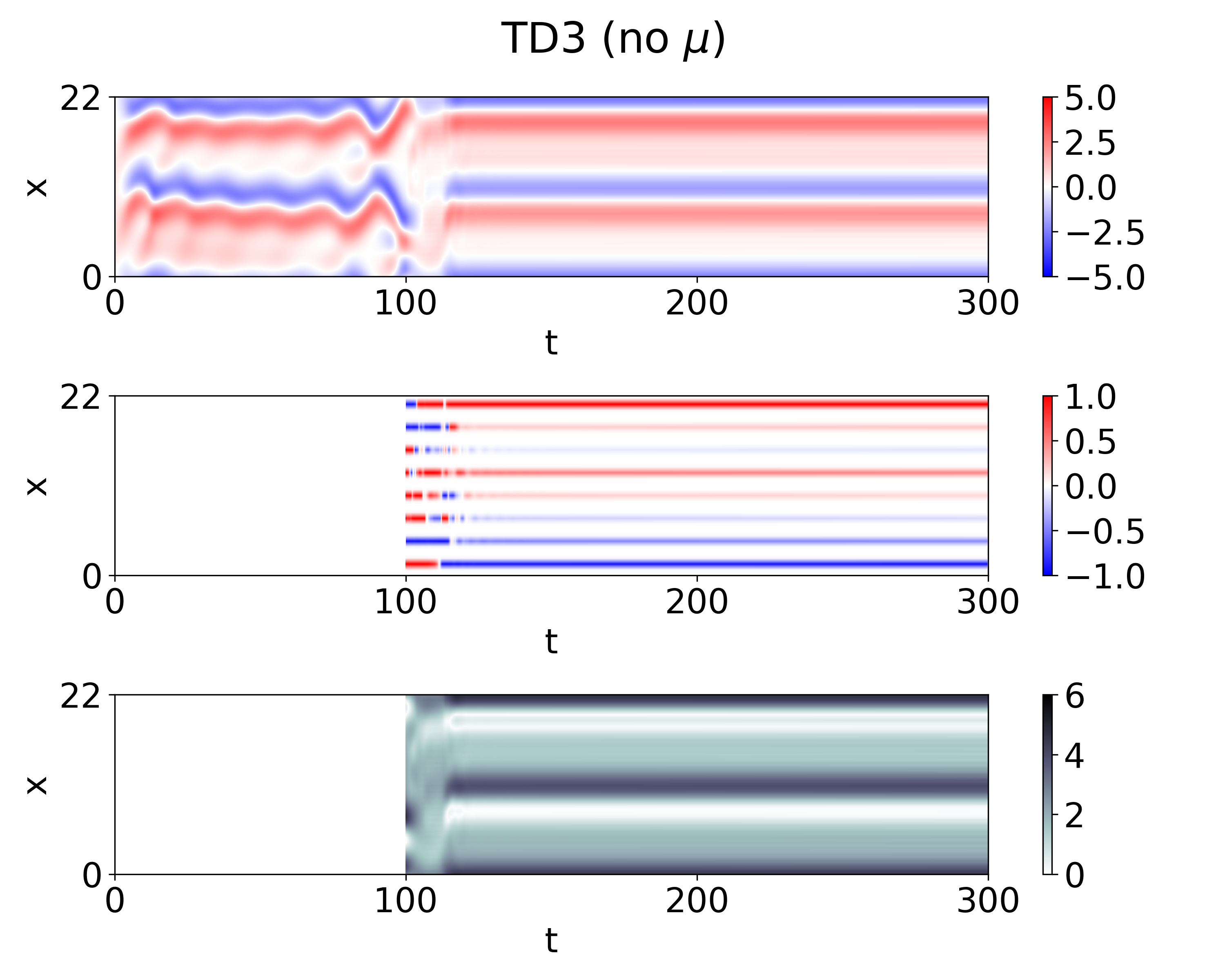}}
    \end{minipage}
    \caption{Controlled solutions of the KS equation when using the different agents. The first row shows the state, the second the control, and the third one the error $|\bm{y}_k - \bm{y}_{\text{ref}}|$.}
    \label{fig:parametric_KS_controlled_extra_1}
\end{figure*}

In Figure \ref{fig:parametric_KS_controlled_full} and \ref{fig:parametric_KS_controlled_extra_1}, we show examples of controlled solution when tracking $\bm{y}_{\text{ref}}=-\bm{1}$ and $\nu=0.25$ and $\bm{y}_{\text{ref}}=\bm{2}$ for $\nu=-0.2$.

\clearpage
\section{Additional Results: Particle Navigation in a Gyre Flow}
\subsection{Navigation of a Particle to Arbitrary Targets in a Gyre Flow}\label{app:gyro_extra}

\begin{figure}[h!]
     \centering
     \begin{subfigure}[b]{0.49\textwidth}
         \centering
         \includegraphics[width=0.85\textwidth]{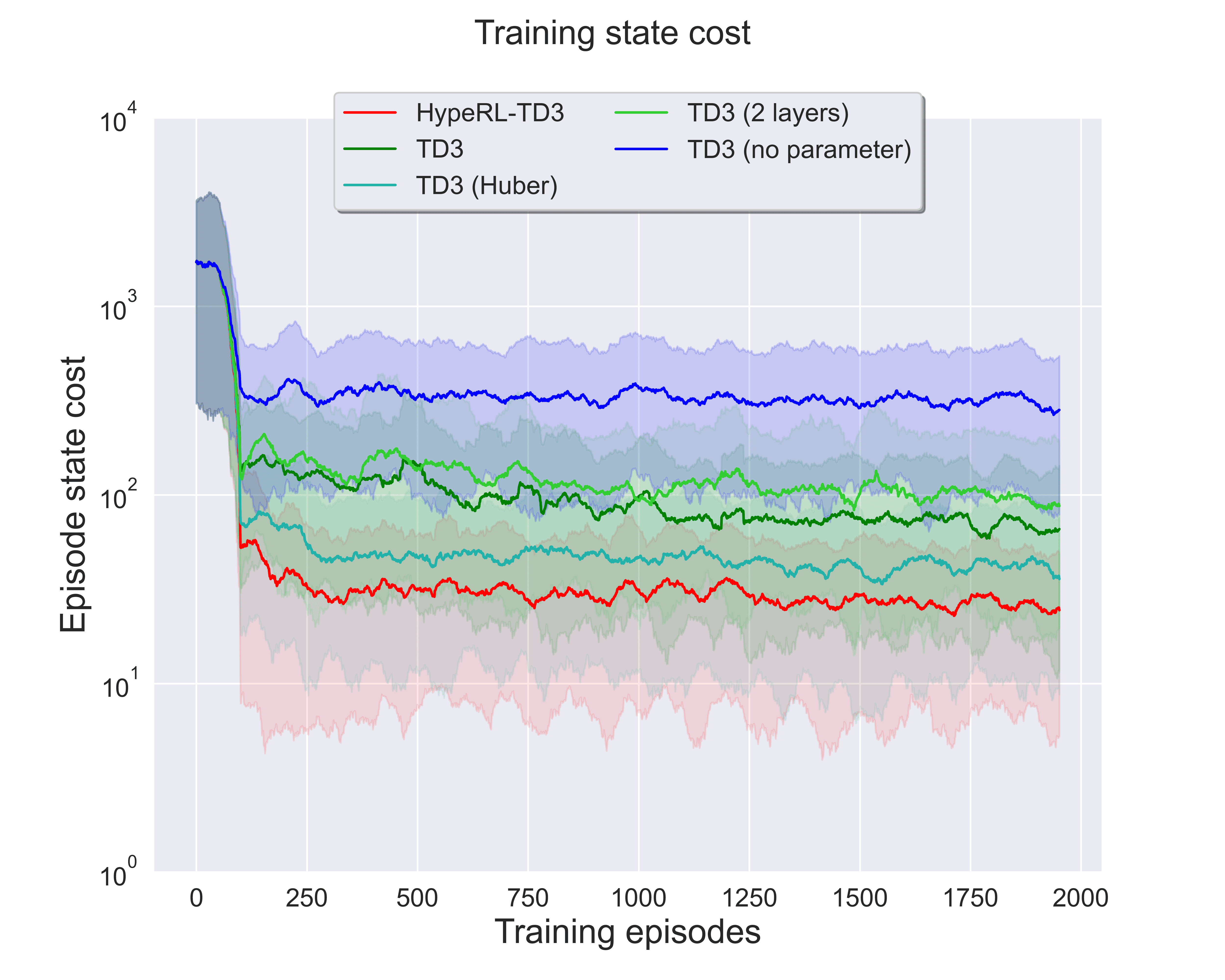}
         \caption{}
    \end{subfigure}
    \begin{subfigure}[b]{0.49\textwidth}
         \centering
         \includegraphics[width=0.85\textwidth]{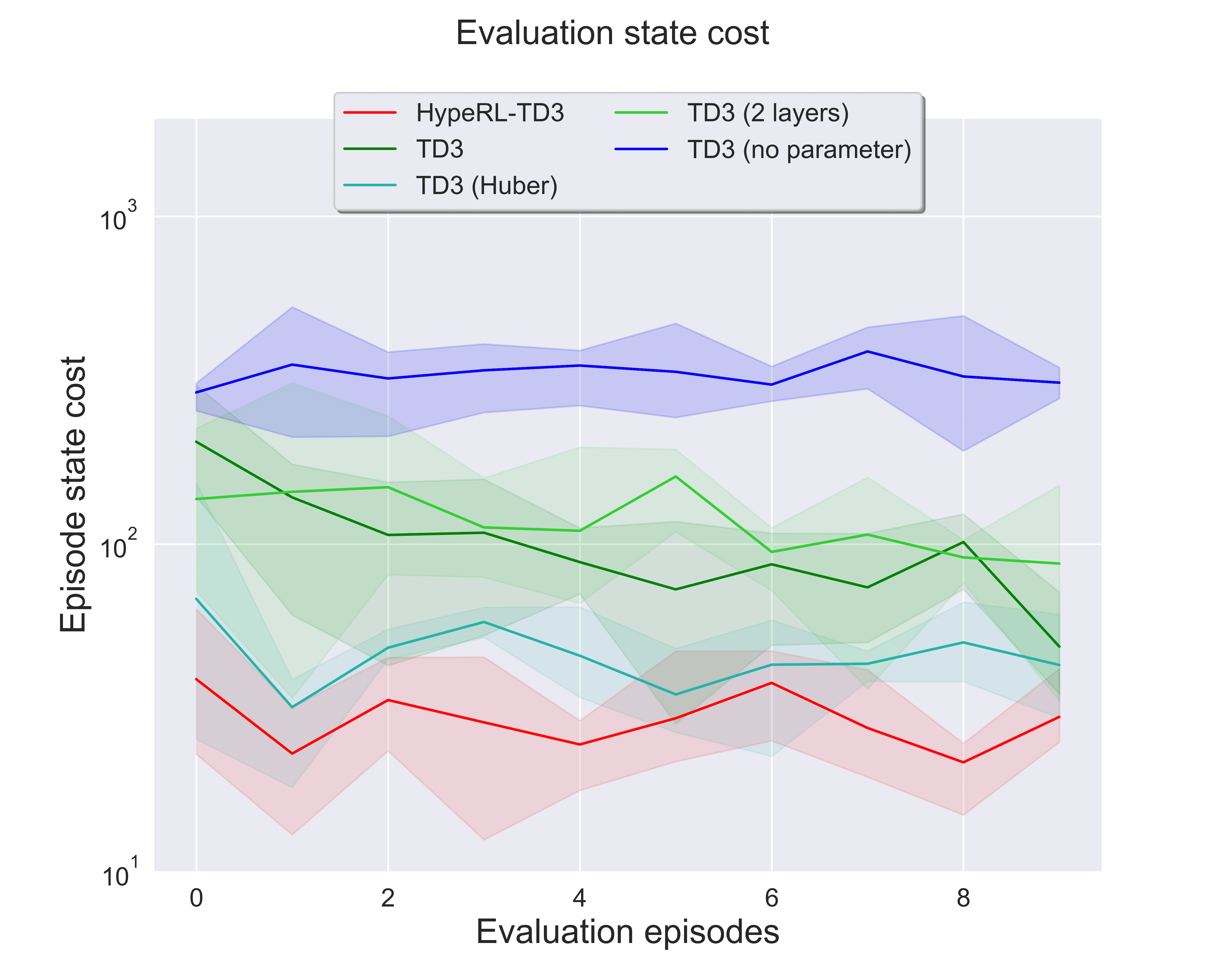}
         \caption{}
     \end{subfigure}
    \begin{subfigure}[b]{0.49\textwidth}
         \centering
         \includegraphics[width=0.85\textwidth]{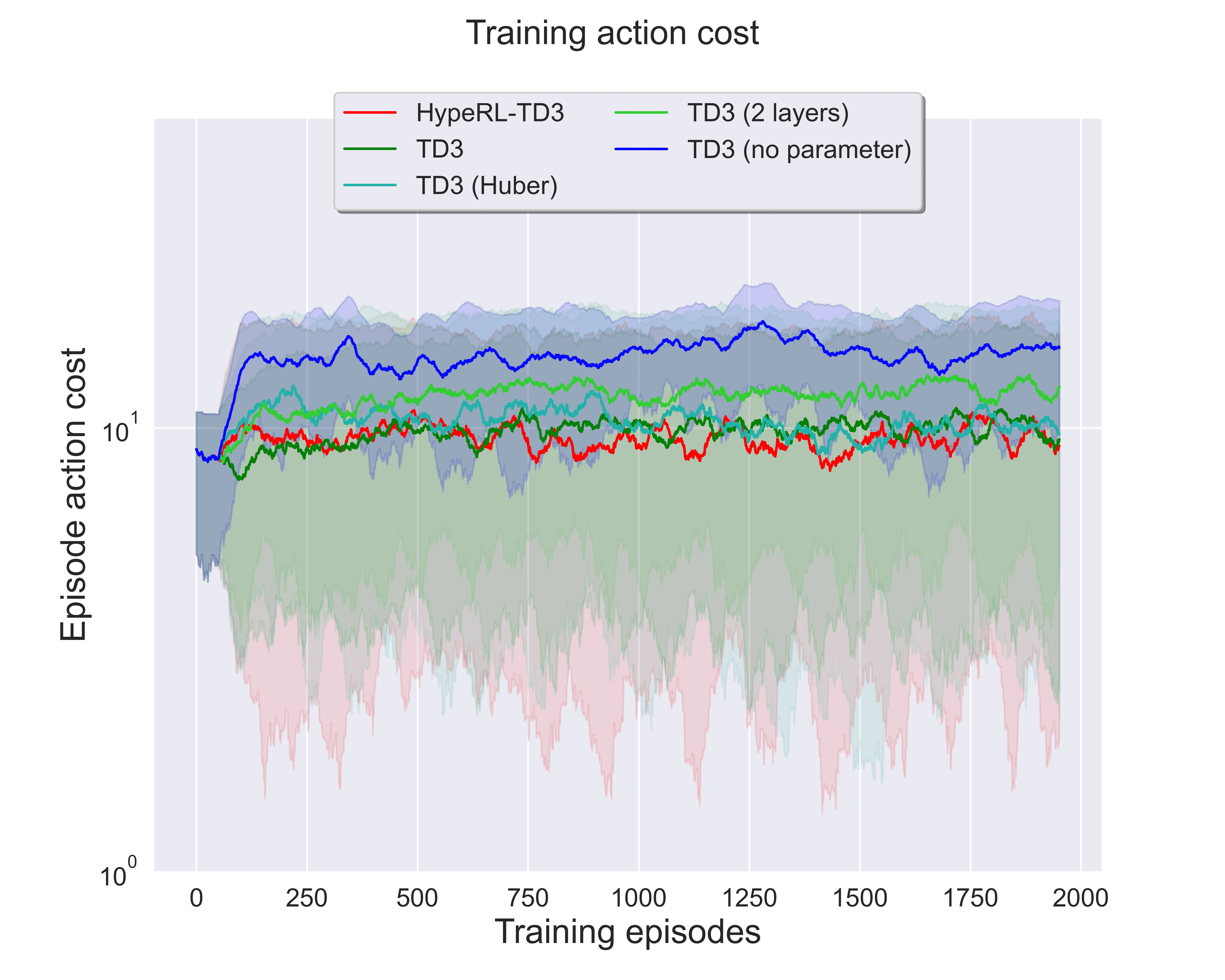}
         \caption{}
     \end{subfigure}
          \centering
    \begin{subfigure}[b]{0.49\textwidth}
         \centering
         \includegraphics[width=0.85\textwidth]{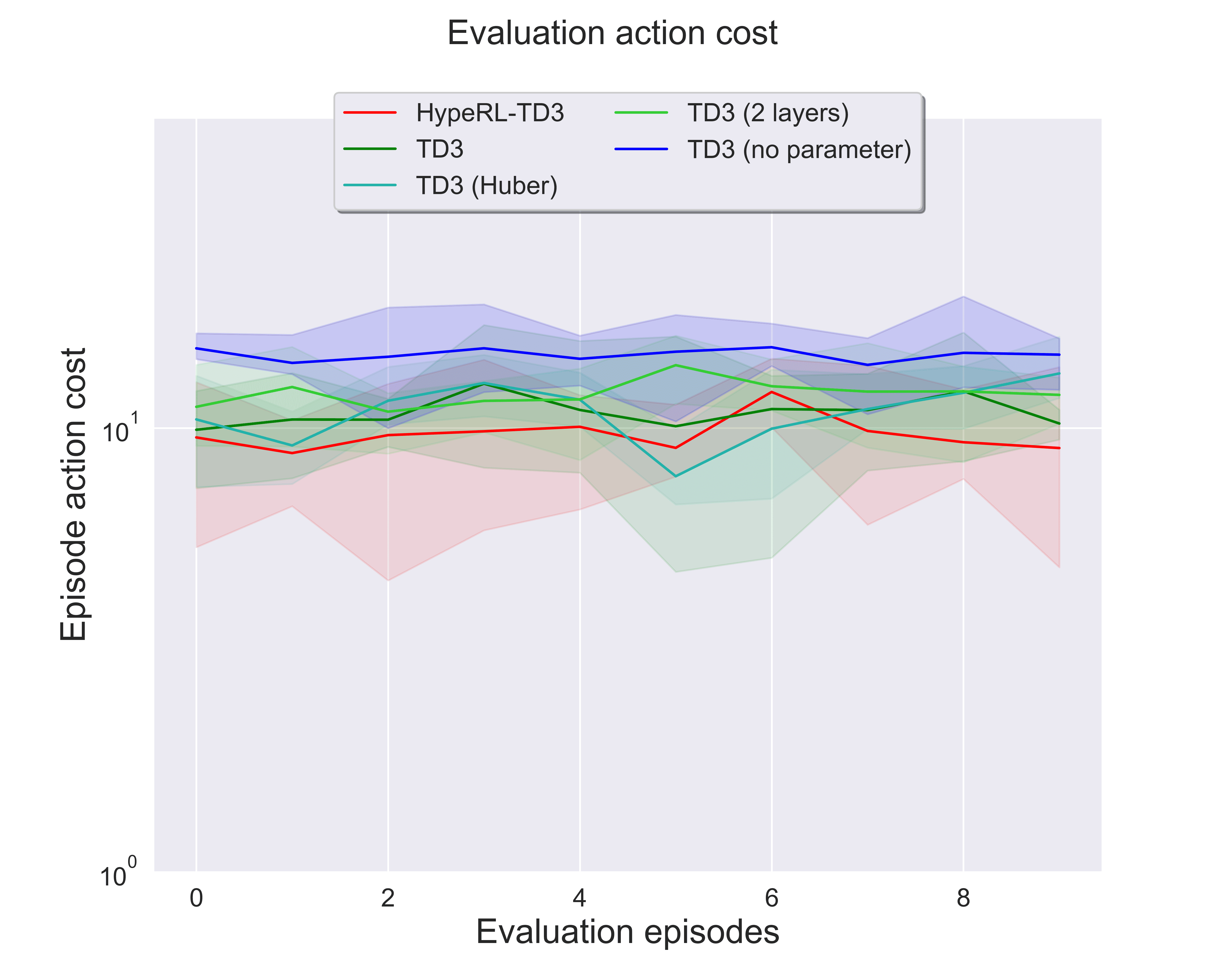}
         \caption{}
     \end{subfigure}
        \caption{Training and evaluation results. The solid line represents the mean and the shaded area the minimum and maximum values observed over 5 different random seeds.}
        \label{fig:10app}
\end{figure}

In Figure \ref{fig:10app}, we show the state and action costs over training and evaluation for the gyro flow test case.

\begin{figure*}[h!]
    \begin{minipage}{0.49\linewidth}
    \centering \subfloat{\includegraphics[height=0.7\textwidth]{Pics/controlled_solution_gyro/hypeRL_td3_v2_solution_1_0_timestep_130.png}}
    \vspace{-0.1cm}
    \subfloat{\includegraphics[height=0.7\textwidth]{Pics/controlled_solution_gyro/td3_huber_solution_1_0_timestep_184.png}}
    \end{minipage}
    \begin{minipage}{0.49\linewidth}
    \centering \subfloat{\includegraphics[height=0.7\textwidth]{Pics/controlled_solution_gyro/td3_solution_1_0_timestep_399.png}}
    \vspace{-0.1cm}
    \subfloat{\includegraphics[height=0.7\textwidth]{Pics/controlled_solution_gyro/td3_twolayers_solution_1_0_timestep_399.png}}
    \end{minipage}
    \centering
    \begin{minipage}{0.49\linewidth}
    \centering \subfloat{\includegraphics[height=0.7\textwidth]{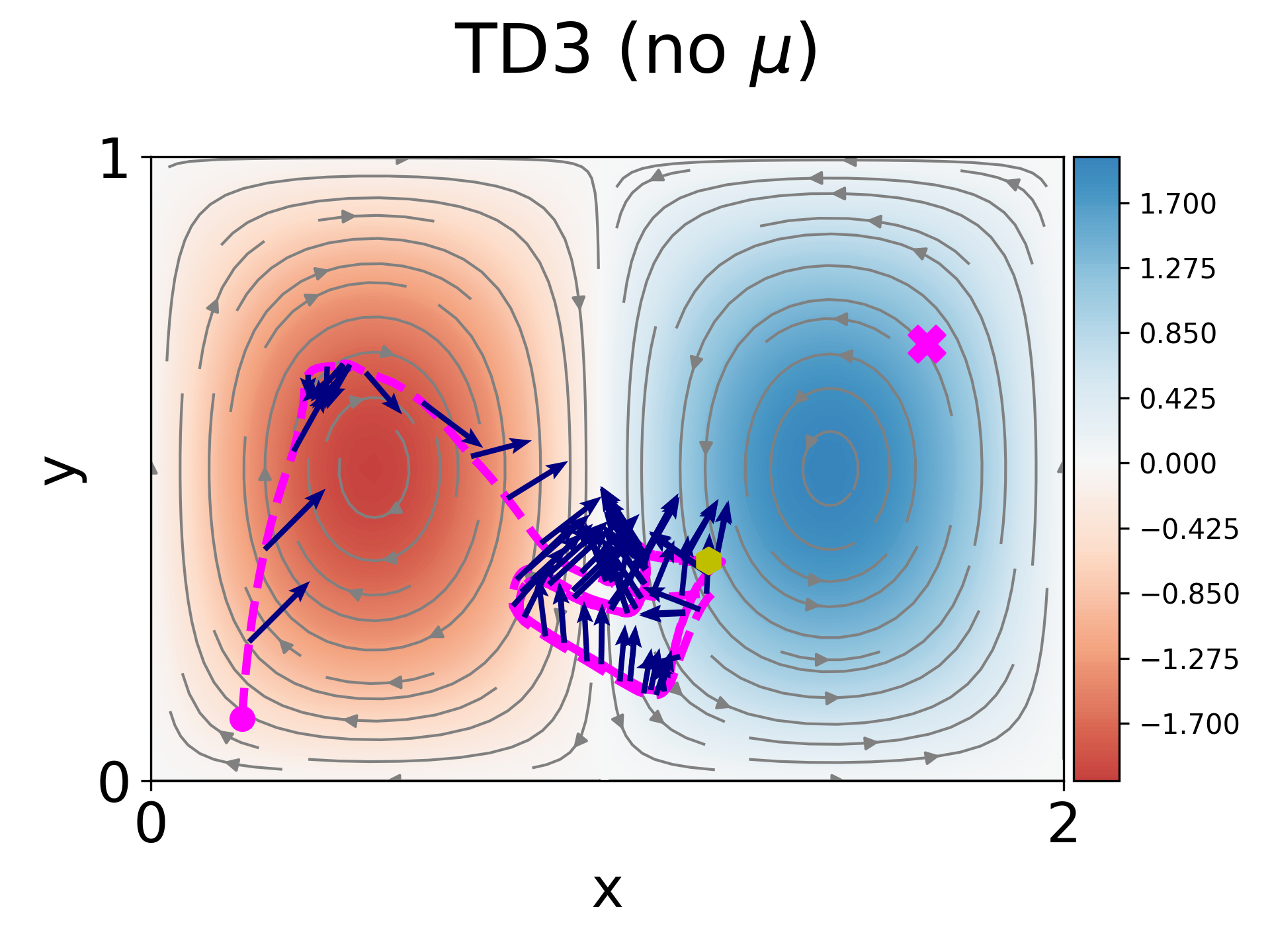}}
    \end{minipage}
    \caption{Trajectories (magenta dash line) of a controlled particle (yellow hexagon) in a gyre flow when using the different agents. The magenta circle indicates the starting location and the magenta cross the target one and the blue arrows represent the control inputs.}
    \label{fig:gyro_controlled_full}
\end{figure*}

\begin{figure*}[h!]
    \begin{minipage}{0.49\linewidth}
    \centering \subfloat{\includegraphics[height=0.7\textwidth]{Pics/controlled_solution_gyro/hypeRL_td3_v2_solution_FTLE_1_0_timestep_130.png}}
    \vspace{-0.1cm}
    \subfloat{\includegraphics[height=0.7\textwidth]{Pics/controlled_solution_gyro/td3_huber_solution_FTLE_1_0_timestep_184.png}}
    \end{minipage}
    \begin{minipage}{0.49\linewidth}
    \centering \subfloat{\includegraphics[height=0.7\textwidth]{Pics/controlled_solution_gyro/td3_solution_FTLE_1_0_timestep_399.png}}
    \vspace{-0.1cm}
    \subfloat{\includegraphics[height=0.7\textwidth]{Pics/controlled_solution_gyro/td3_twolayers_solution_FTLE_1_0_timestep_399.png}}
    \end{minipage}
    \centering
    \begin{minipage}{0.49\linewidth}
    \centering \subfloat{\includegraphics[height=0.7\textwidth]{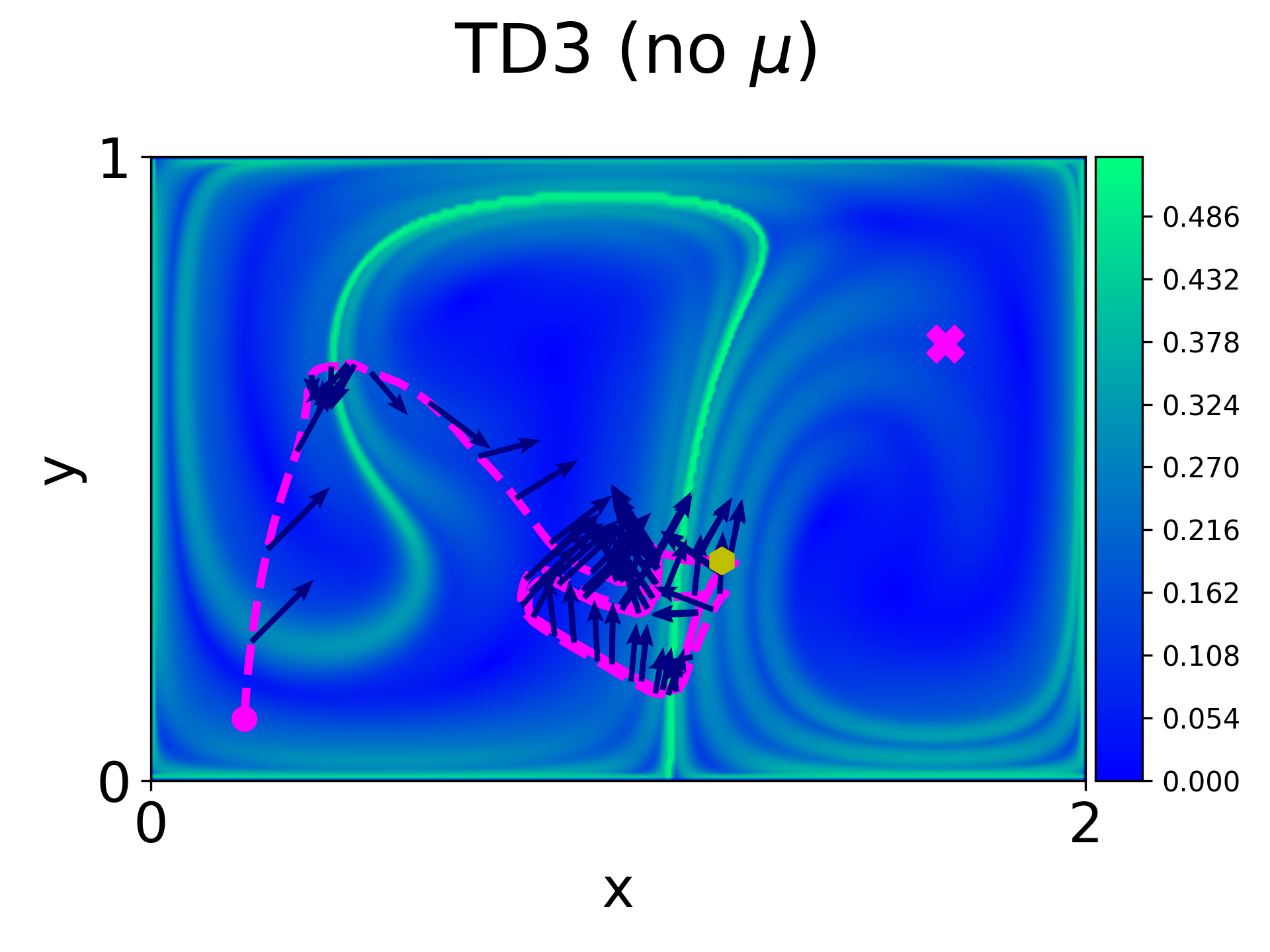}}
    \end{minipage}
    \caption{Trajectories (magenta dash line) of a controlled particle (yellow hexagon) in a gyre flow when using the different agents overlaid with the FTLE. The magenta circle indicates the starting location and the magenta cross the target one and the blue arrows represent the control inputs.}
    \label{fig:gyro_controlled_FTLE_full}
\end{figure*}

\begin{figure*}[h!]
    \begin{minipage}{0.49\linewidth}
    \centering \subfloat{\includegraphics[height=0.7\textwidth]{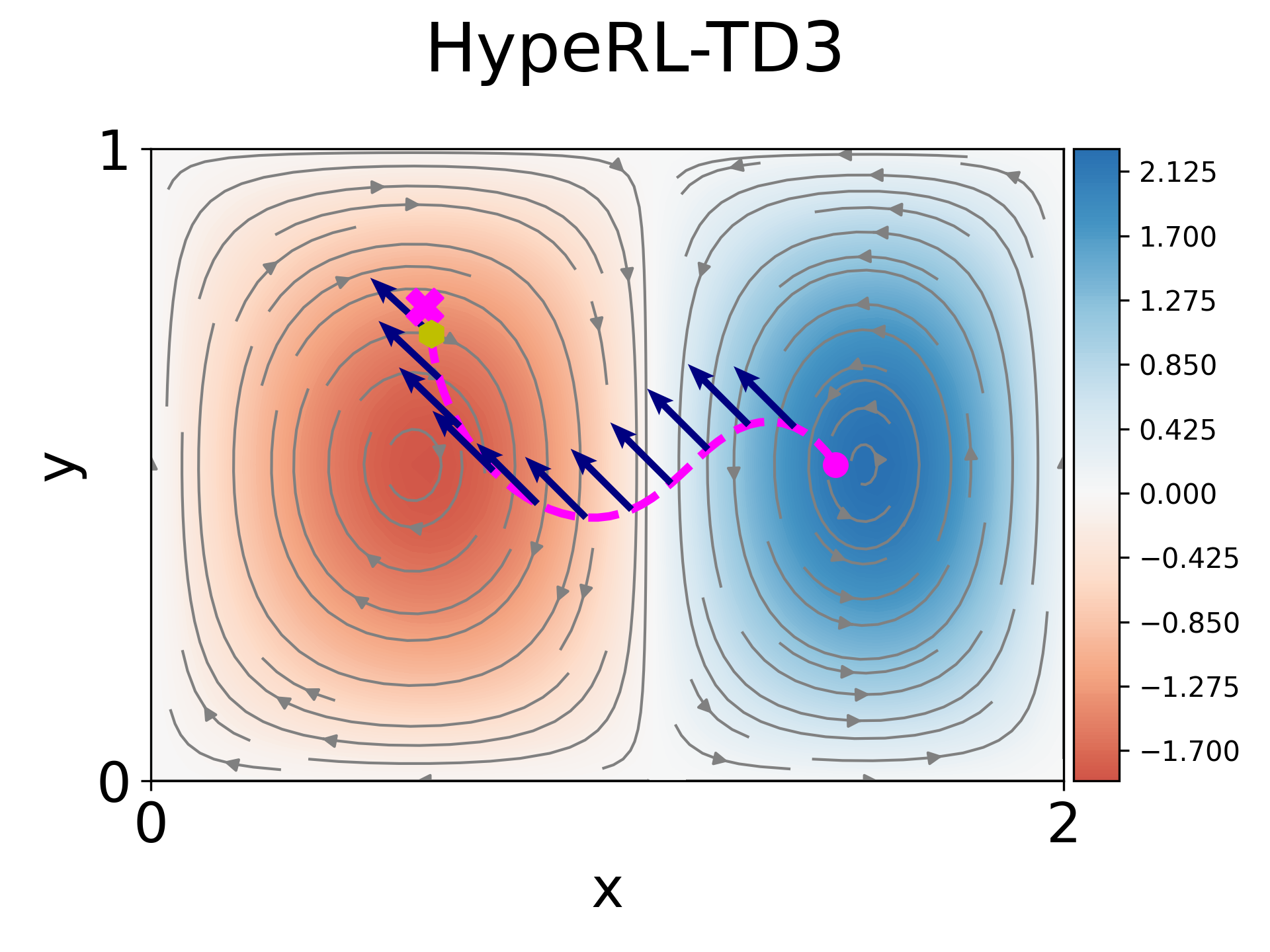}}
    \vspace{-0.1cm}
    \subfloat{\includegraphics[height=0.7\textwidth]{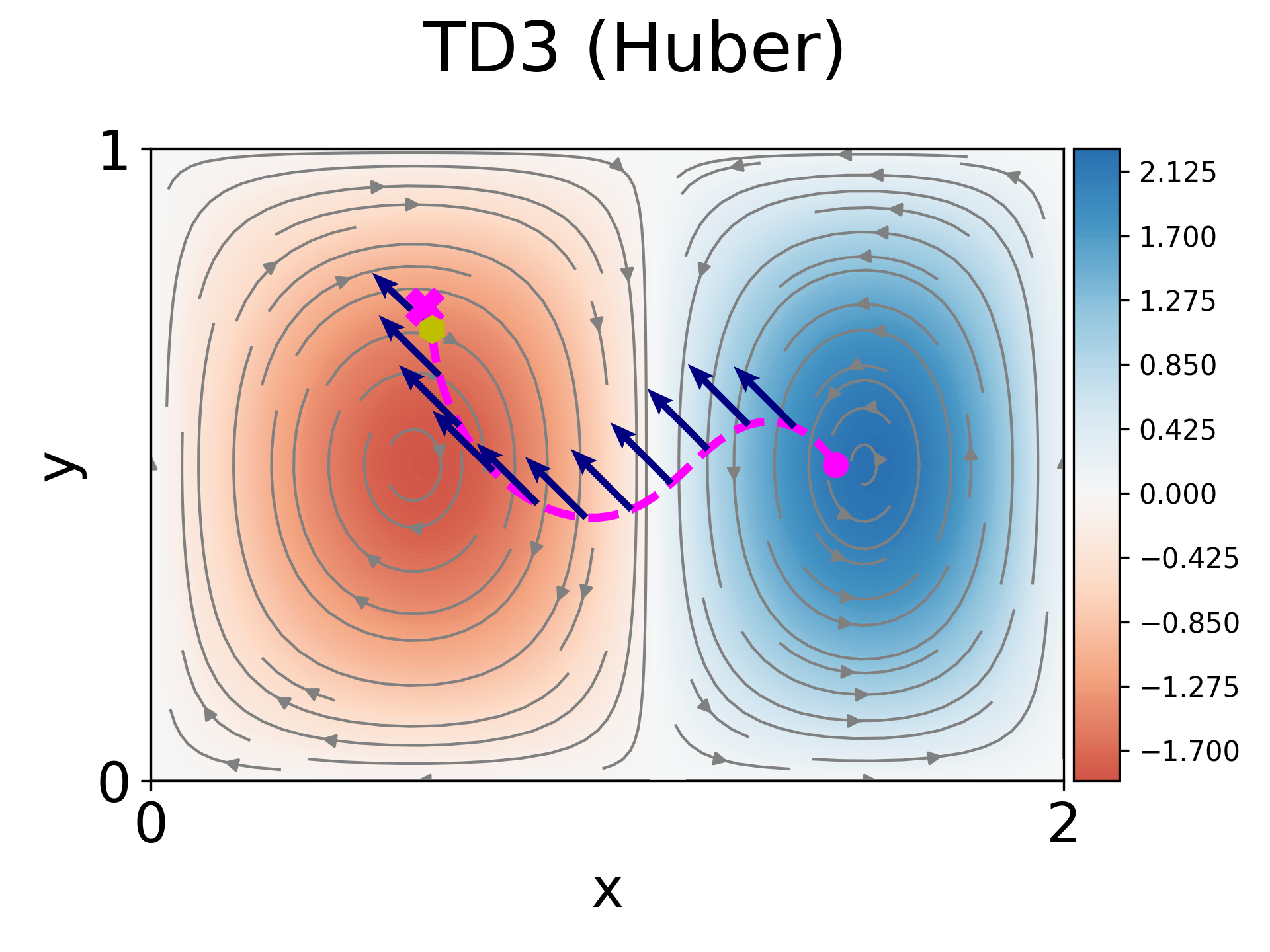}}
    \end{minipage}
    \begin{minipage}{0.49\linewidth}
    \centering \subfloat{\includegraphics[height=0.7\textwidth]{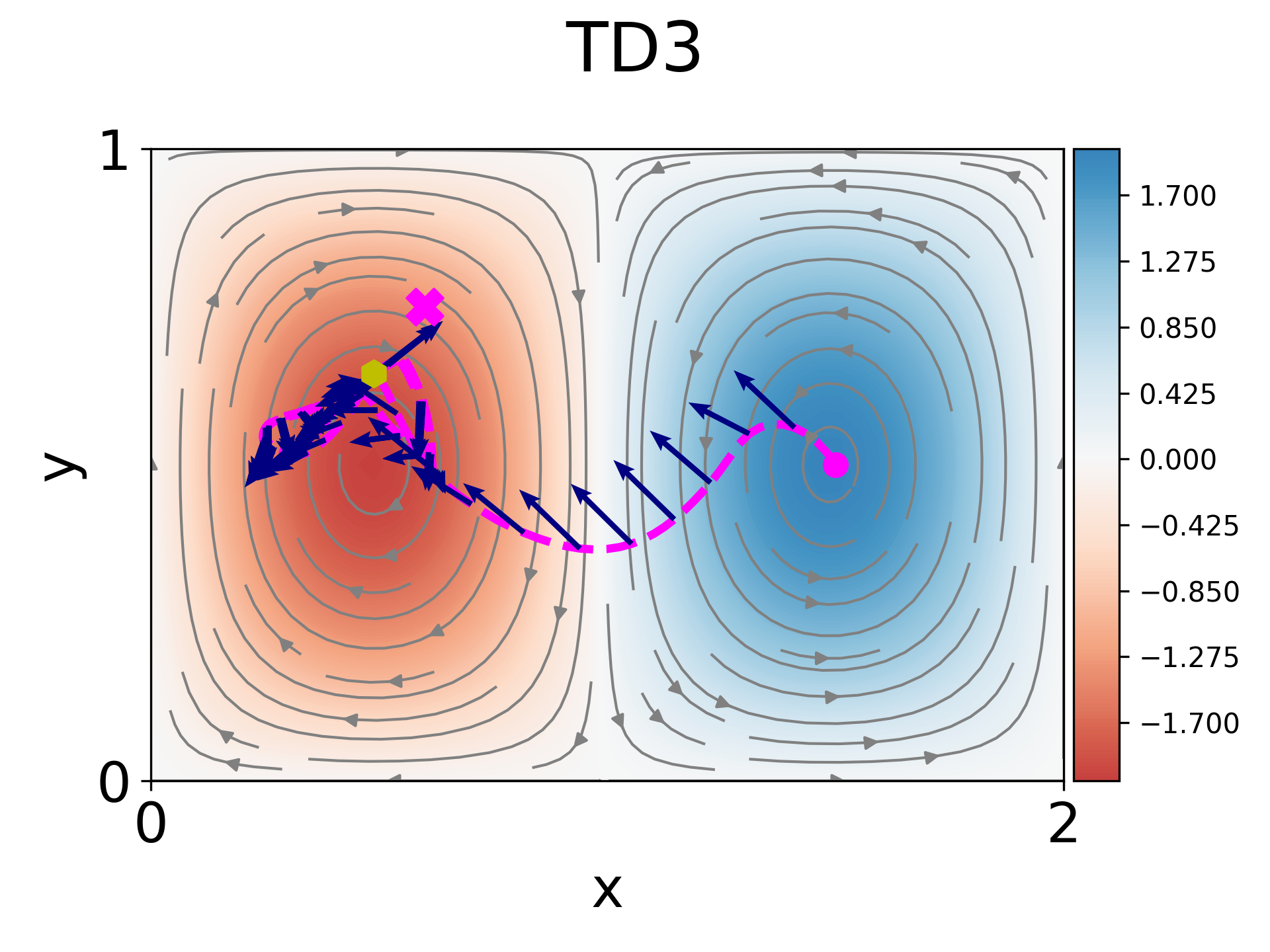}}
    \vspace{-0.1cm}
    \subfloat{\includegraphics[height=0.7\textwidth]{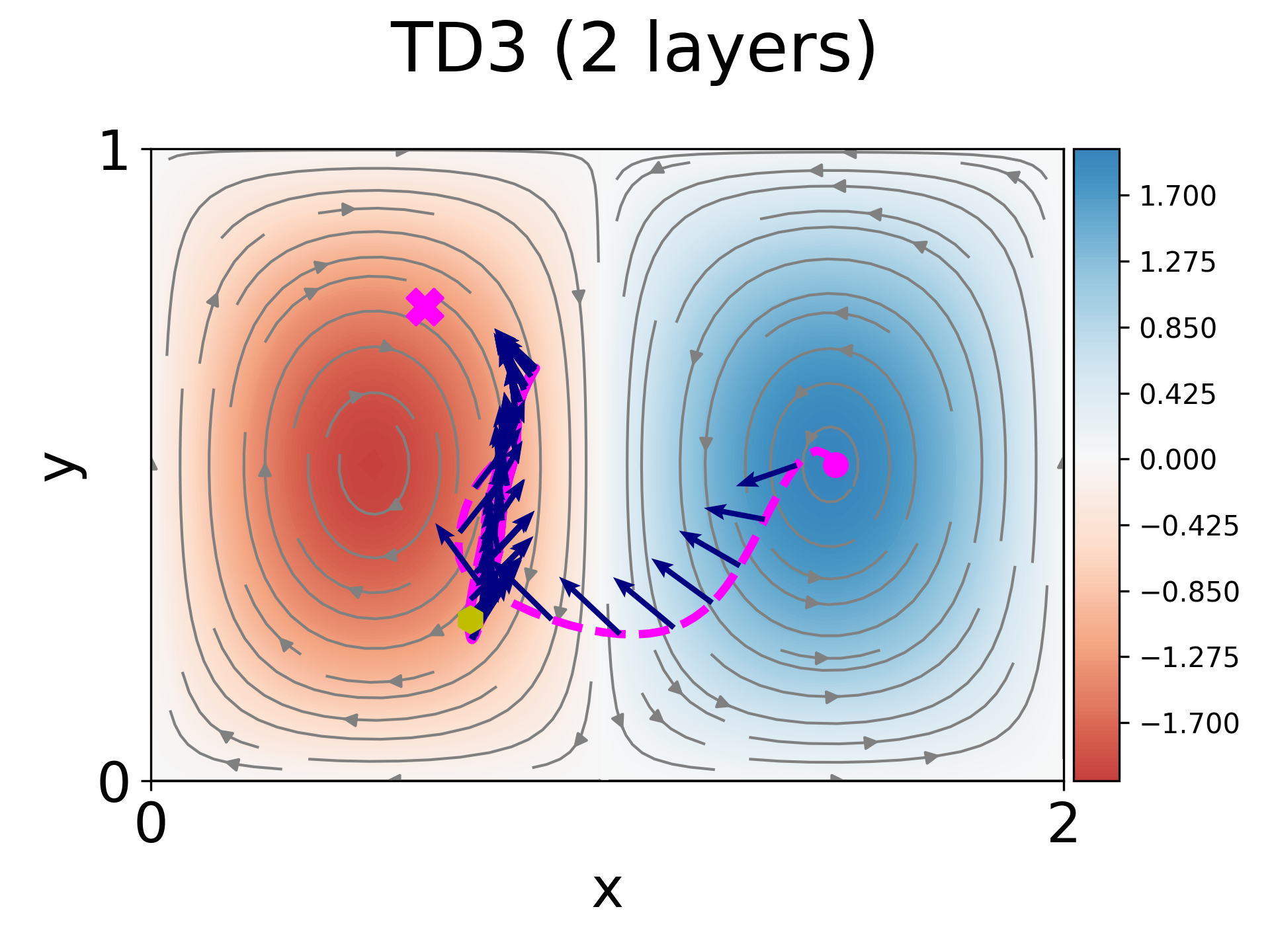}}
    \end{minipage}
    \centering
    \begin{minipage}{0.49\linewidth}
    \centering \subfloat{\includegraphics[height=0.7\textwidth]{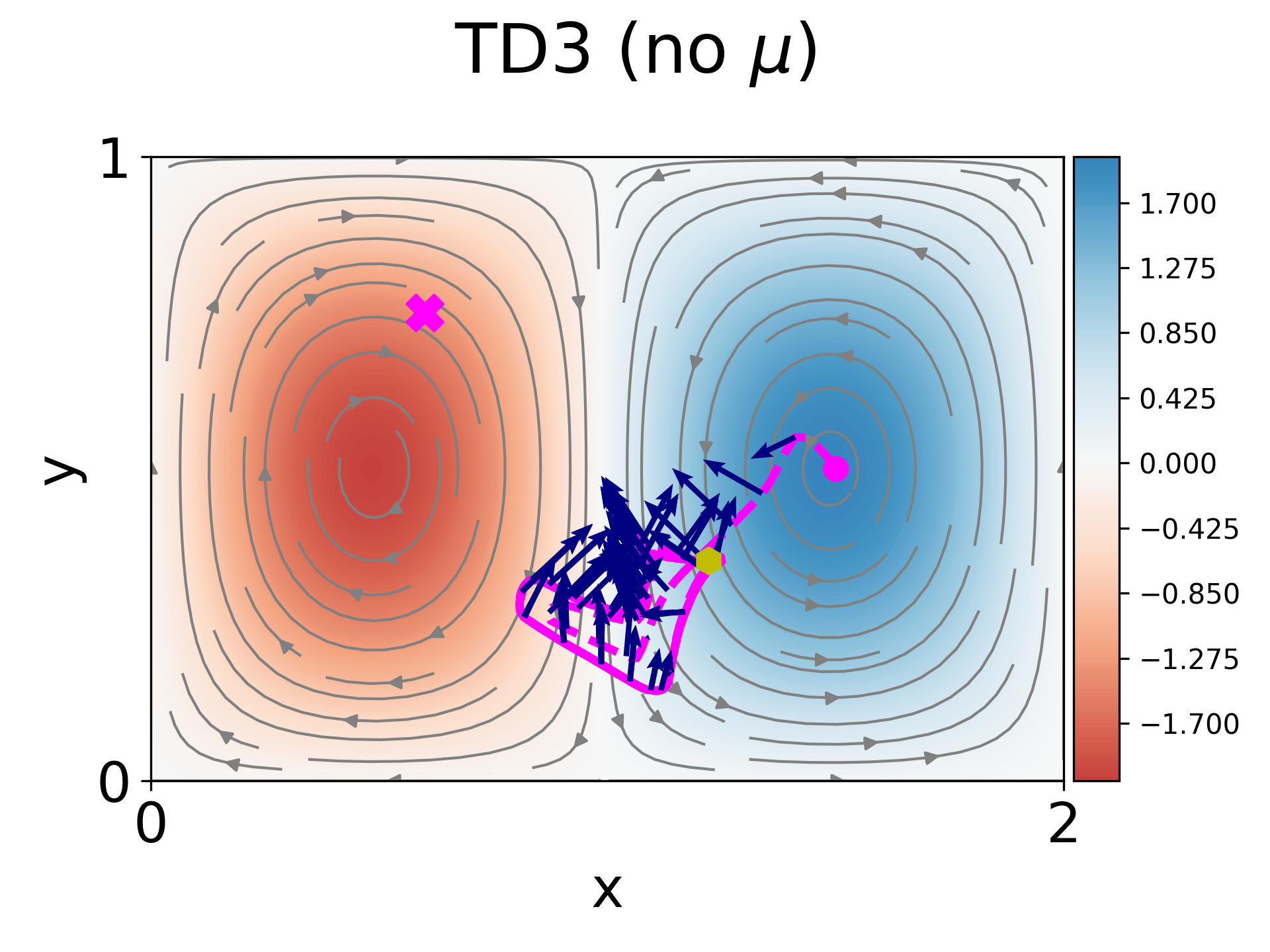}}
    \end{minipage}
    \caption{Trajectories of a controlled particle in a gyre flow when using the different agents.}
    \label{fig:gyro_controlled_full1}
\end{figure*}

\begin{figure*}[h!]
    \begin{minipage}{0.49\linewidth}
    \centering \subfloat{\includegraphics[height=0.7\textwidth]{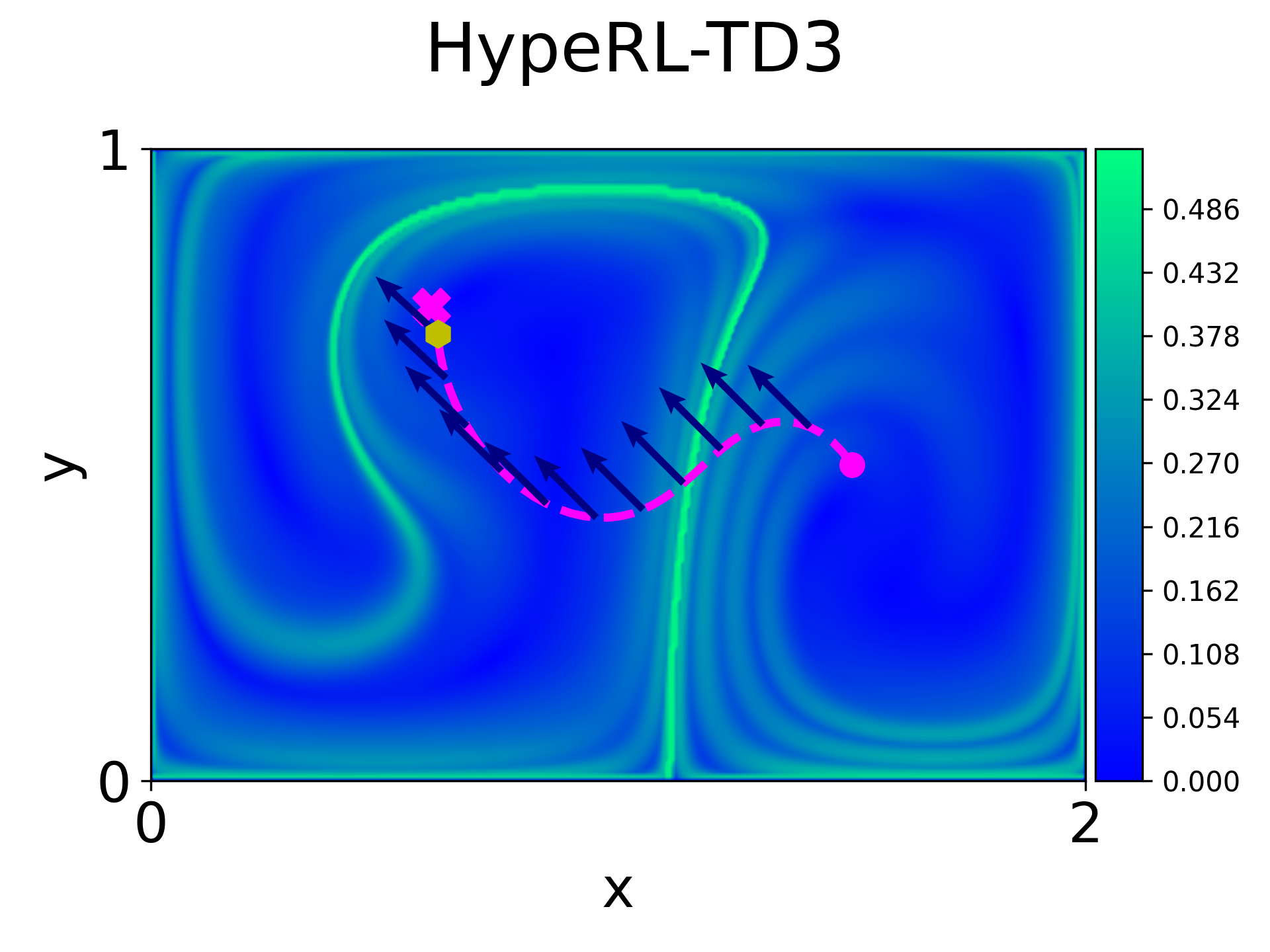}}
    \vspace{-0.1cm}
    \subfloat{\includegraphics[height=0.7\textwidth]{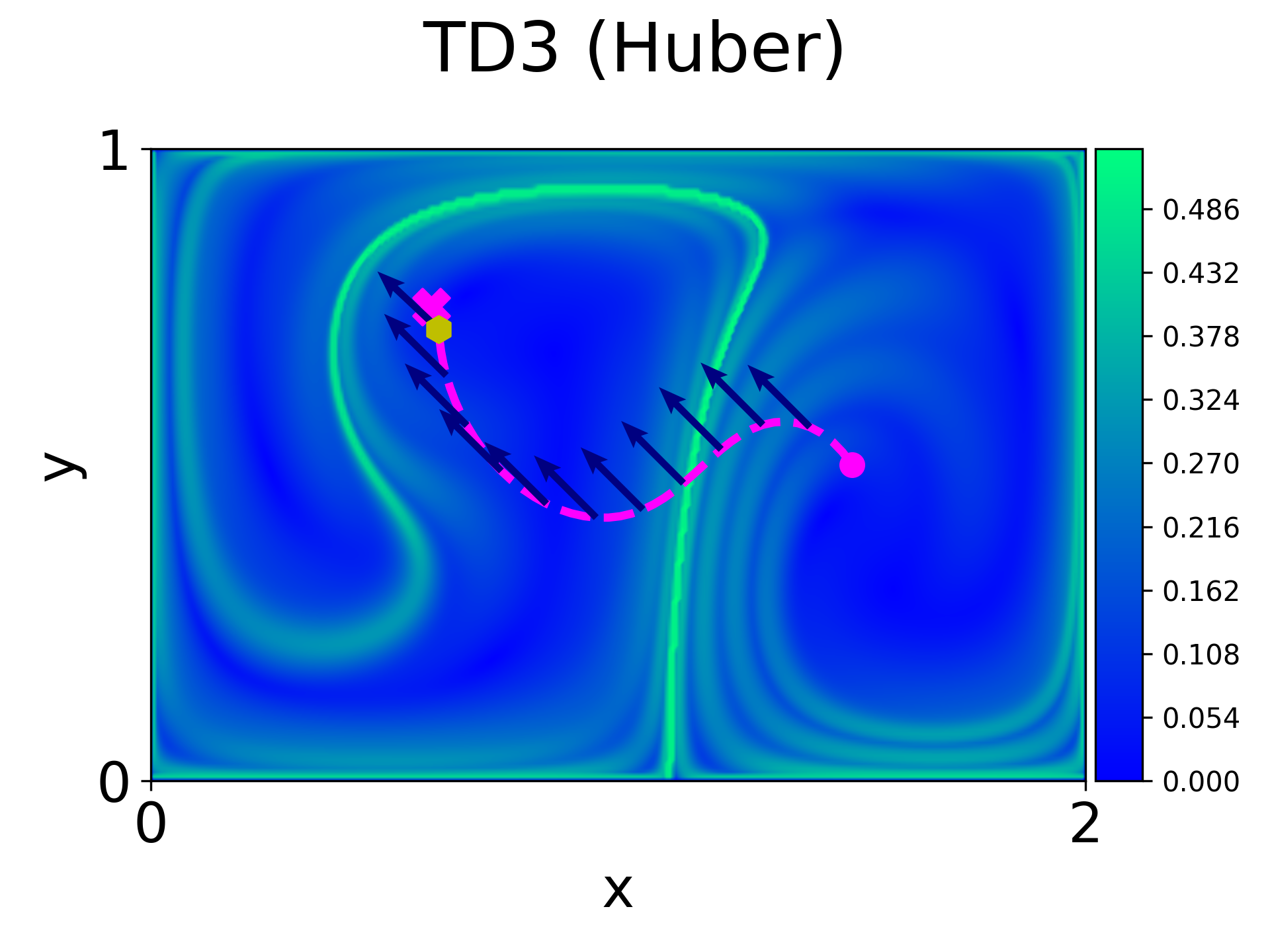}}
    \end{minipage}
    \begin{minipage}{0.49\linewidth}
    \centering \subfloat{\includegraphics[height=0.7\textwidth]{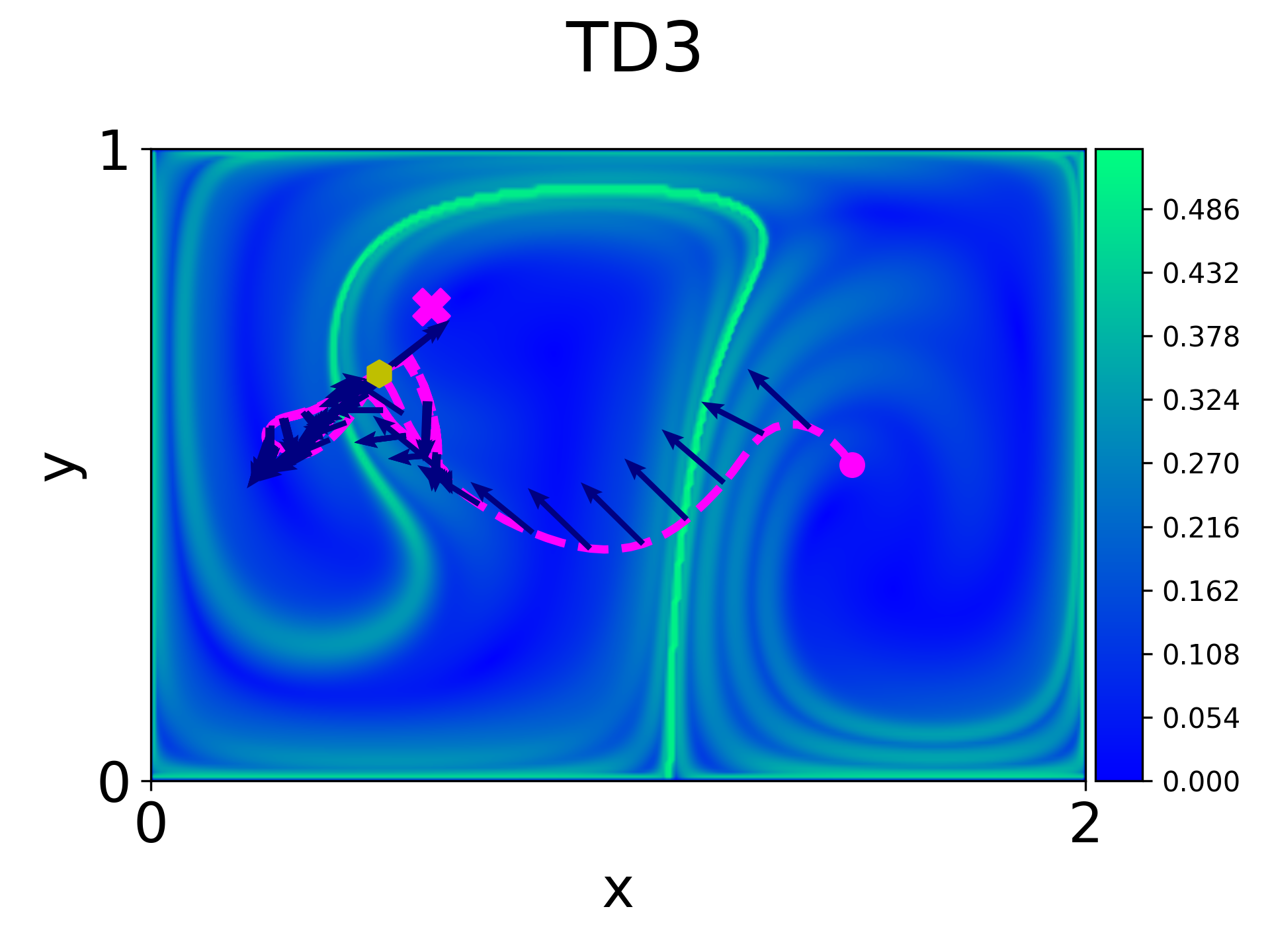}}
    \vspace{-0.1cm}
    \subfloat{\includegraphics[height=0.7\textwidth]{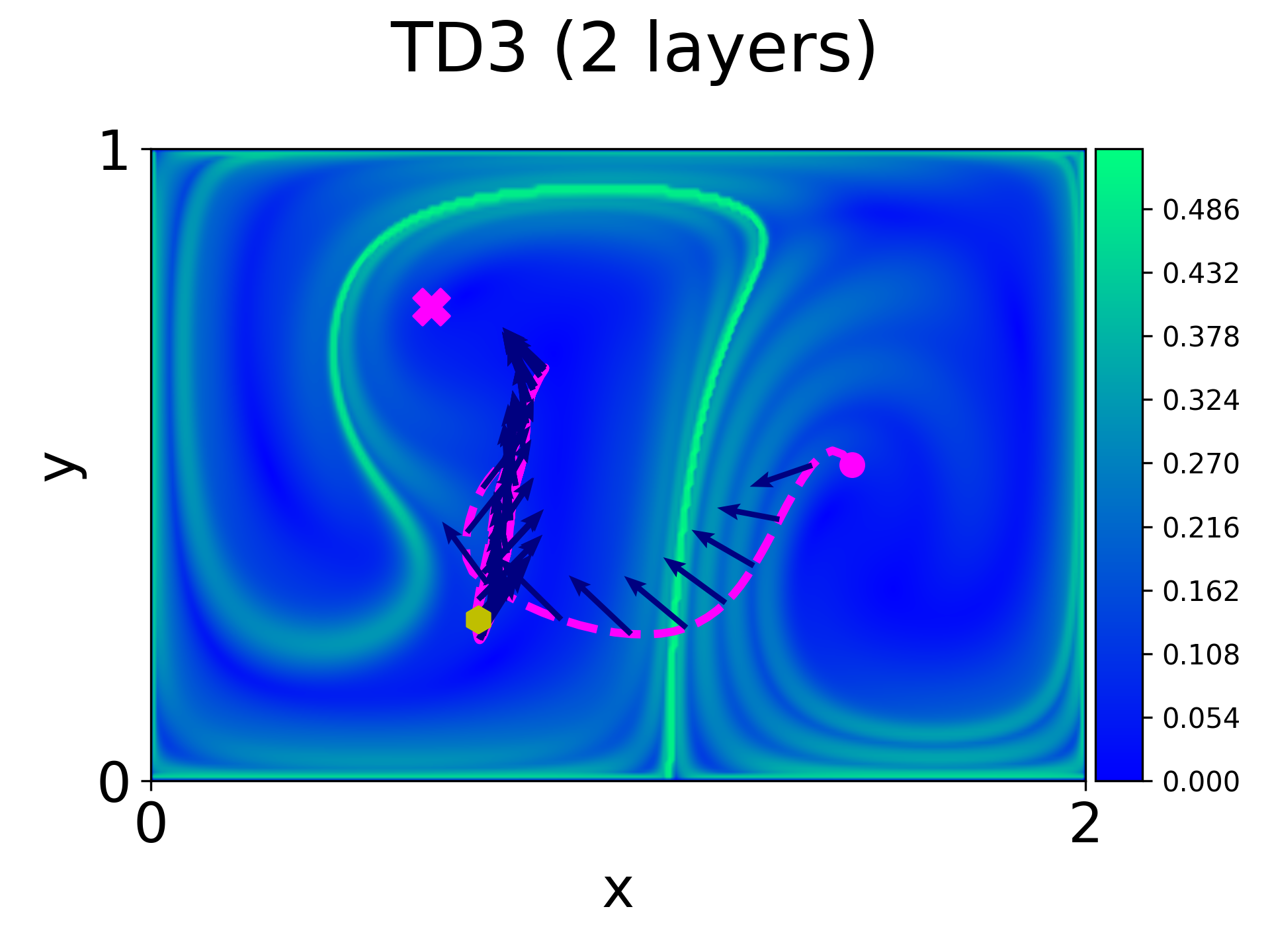}}
    \end{minipage}
    \centering
    \begin{minipage}{0.49\linewidth}
    \centering \subfloat{\includegraphics[height=0.7\textwidth]{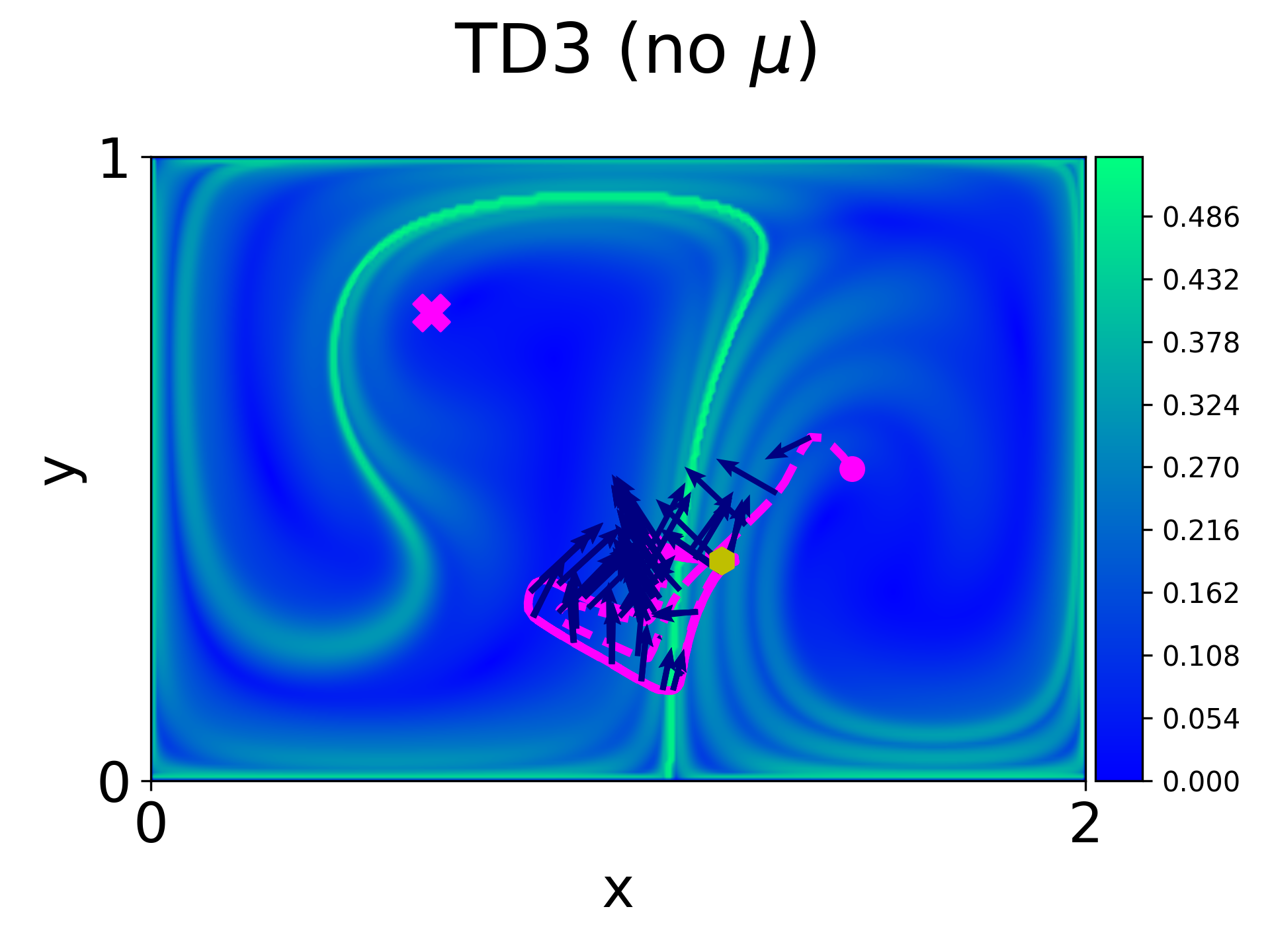}}
    \end{minipage}
    \centering
    \caption{Trajectories of a controlled particle in a gyre flow when using the different agents.}
    \label{fig:gyro_controlled_FTLE_full1}
\end{figure*}

In Figures \ref{fig:gyro_controlled_full} and \ref{fig:gyro_controlled_full1} we show examples of controlled trajectories overlaid with the gyro flow field, while in Figures \ref{fig:gyro_controlled_FTLE_full} and \ref{fig:gyro_controlled_FTLE_full1} we overlaid the trajectories with the FTLE.


\clearpage
\subsection{Navigation of a Particle to Arbitrary Targets in a Parametric Gyre Flow}

\begin{figure}[h!]
     \centering
     \begin{subfigure}[b]{0.49\textwidth}
         \centering
         \includegraphics[width=0.85\textwidth]{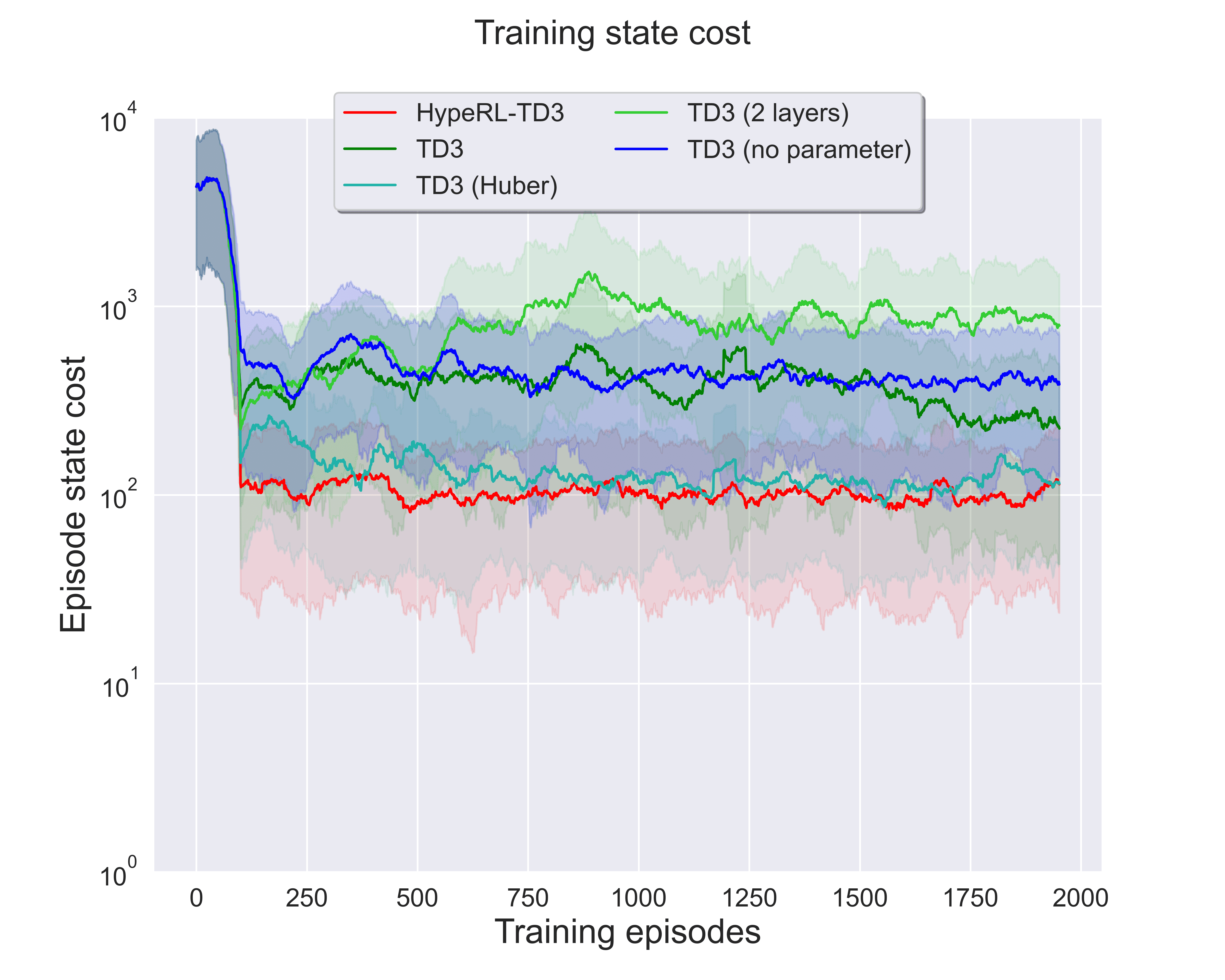}
         \caption{}
    \end{subfigure}
    \begin{subfigure}[b]{0.49\textwidth}
         \centering
         \includegraphics[width=0.85\textwidth]{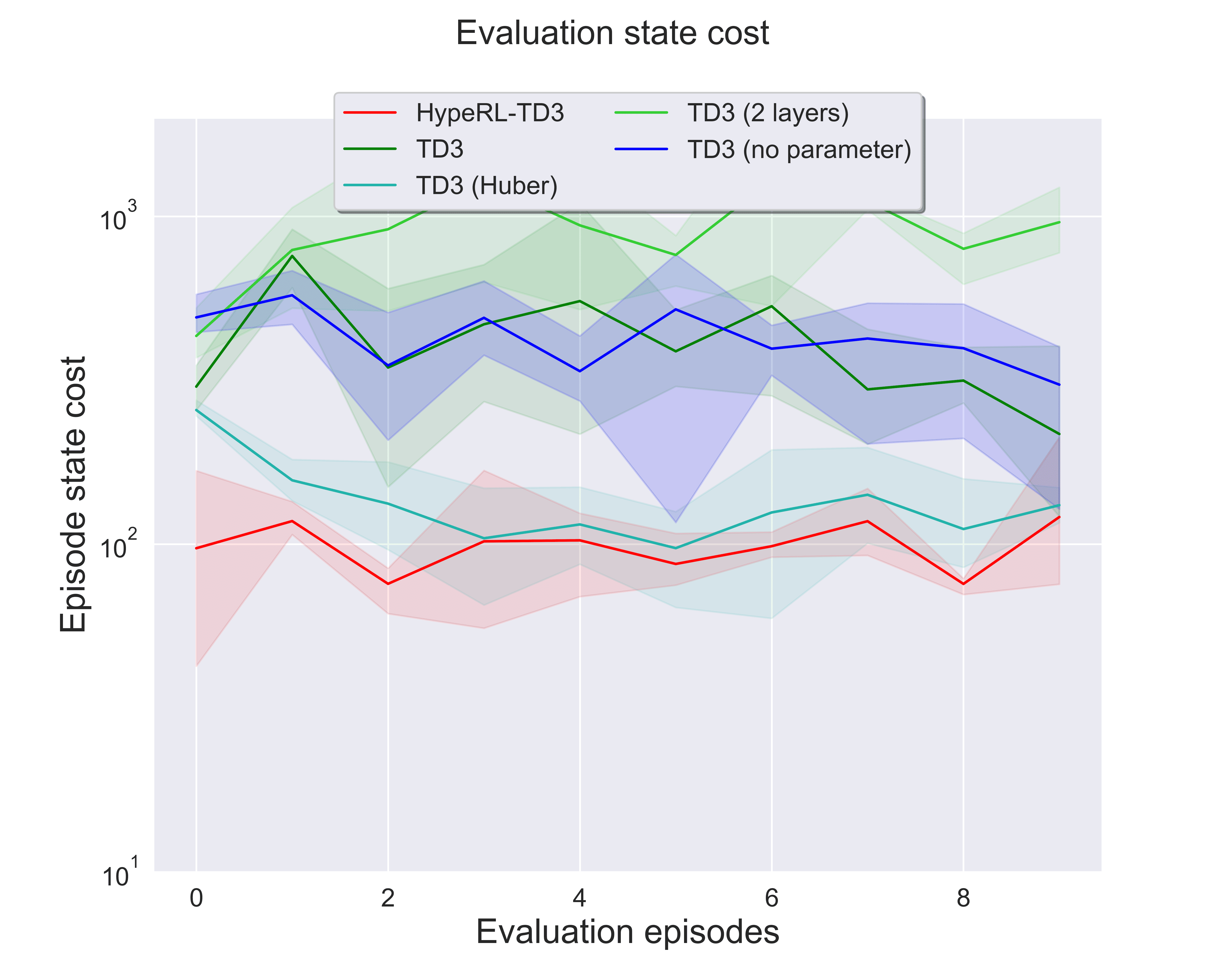}
         \caption{}
     \end{subfigure}
    \begin{subfigure}[b]{0.49\textwidth}
         \centering
         \includegraphics[width=0.85\textwidth]{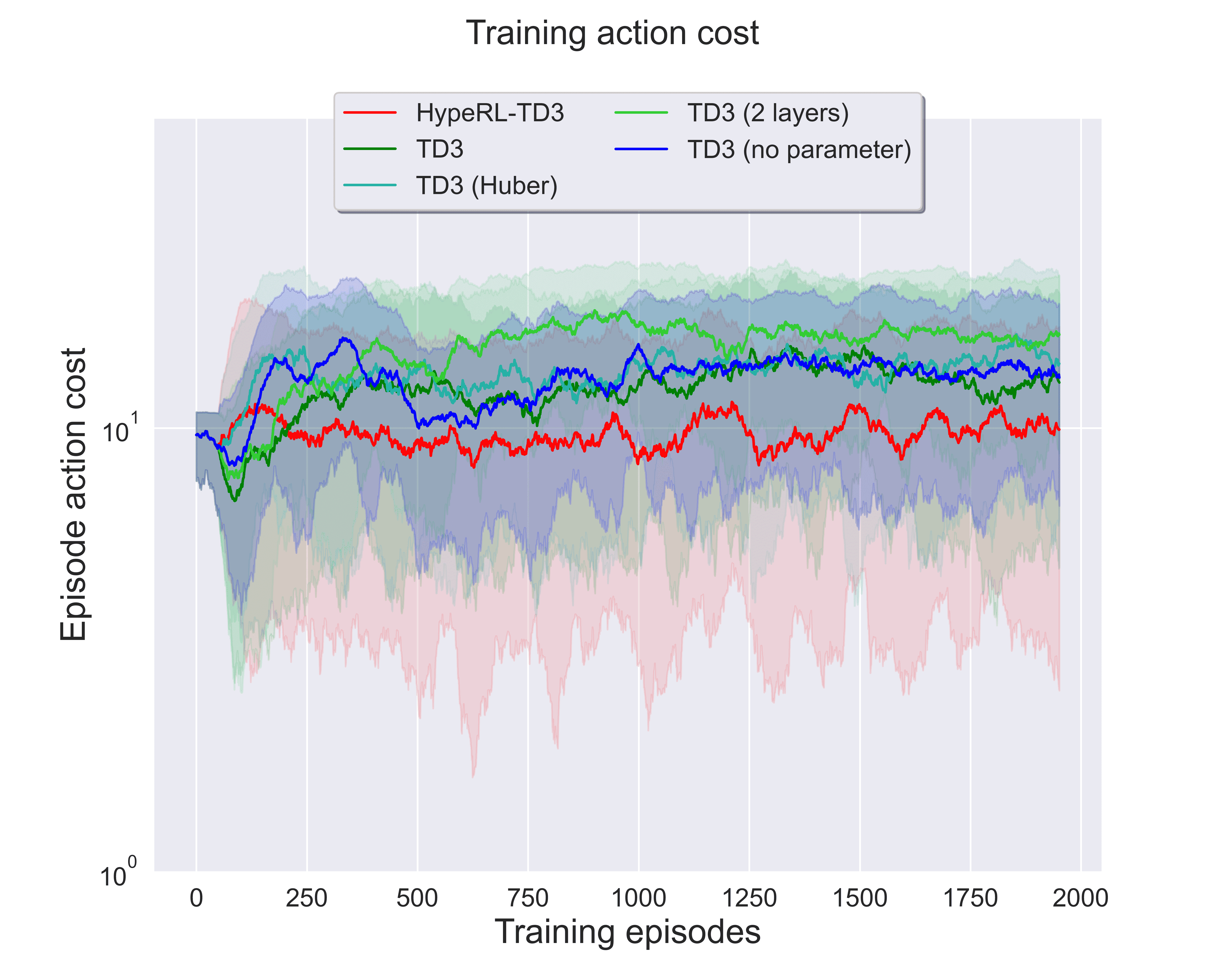}
         \caption{}
     \end{subfigure}
          \centering
    \begin{subfigure}[b]{0.49\textwidth}
         \centering
         \includegraphics[width=0.85\textwidth]{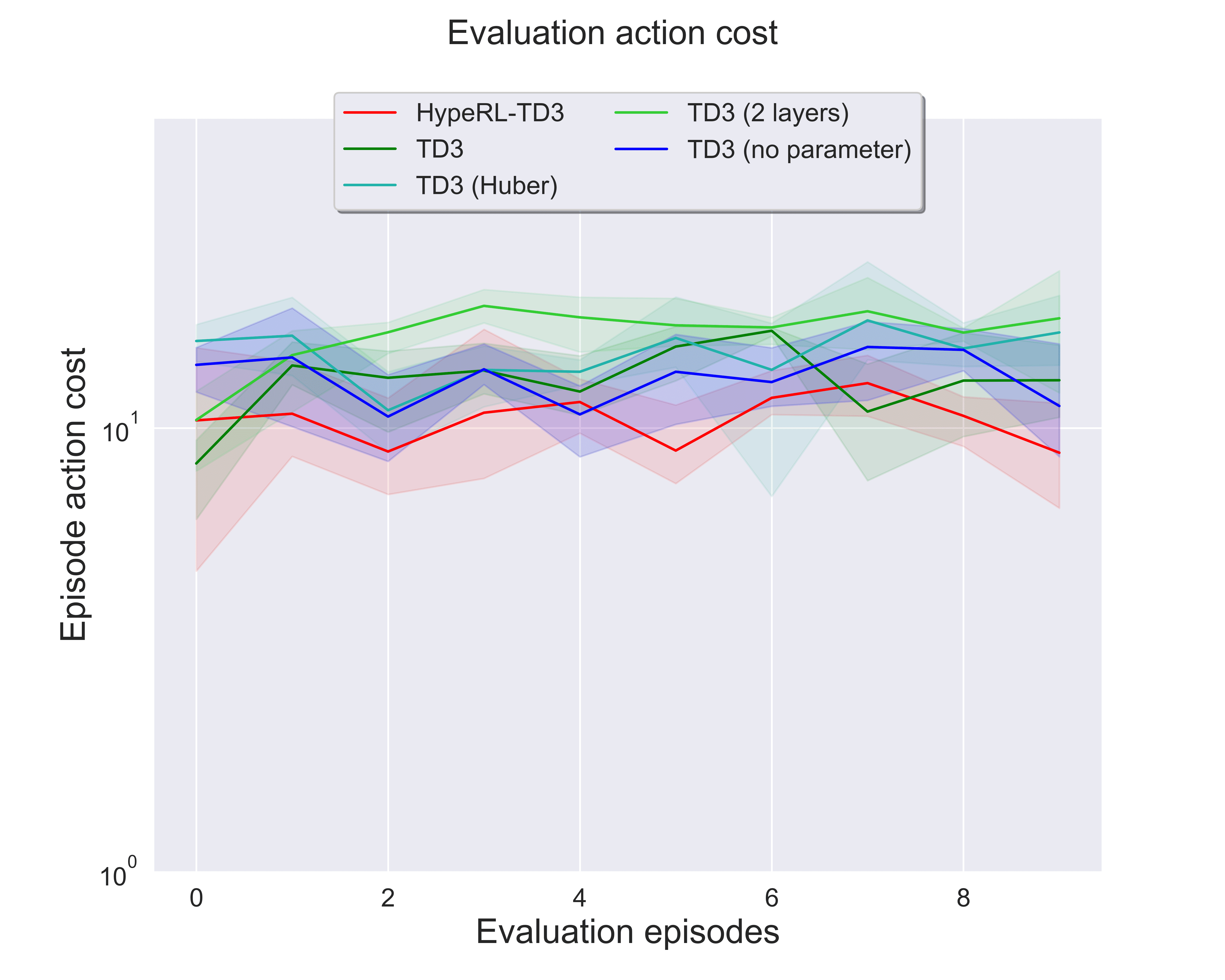}
         \caption{}
     \end{subfigure}
        \caption{Training and evaluation results. The solid line represents the mean and the shaded area the minimum and maximum values observed over 5 different random seeds.}
        \label{fig:param_gyro_costs_extra}
\end{figure}

In Figure \ref{fig:param_gyro_costs_extra}, we show the state and action costs over training and evaluation for the gyro flow test case.

\begin{figure*}[h!]
    \begin{minipage}{0.49\linewidth}
    \centering \subfloat{\includegraphics[height=0.7\textwidth]{Pics/controlled_solution_parametric_gyro/hypeRL_td3_v2_solution_1_1_timestep_67.png}}
    \vspace{-0.1cm}
    \subfloat{\includegraphics[height=0.7\textwidth]{Pics/controlled_solution_parametric_gyro/td3_huber_solution_1_1_timestep_105.png}}
    \end{minipage}
    \begin{minipage}{0.49\linewidth}
    \centering \subfloat{\includegraphics[height=0.7\textwidth]{Pics/controlled_solution_parametric_gyro/td3_solution_1_1_timestep_399.png}}
    \vspace{-0.1cm}
    \subfloat{\includegraphics[height=0.7\textwidth]{Pics/controlled_solution_parametric_gyro/td3_twolayers_solution_1_1_timestep_399.png}}
    \end{minipage}
    \centering
    \begin{minipage}{0.49\linewidth}
    \centering \subfloat{\includegraphics[height=0.7\textwidth]{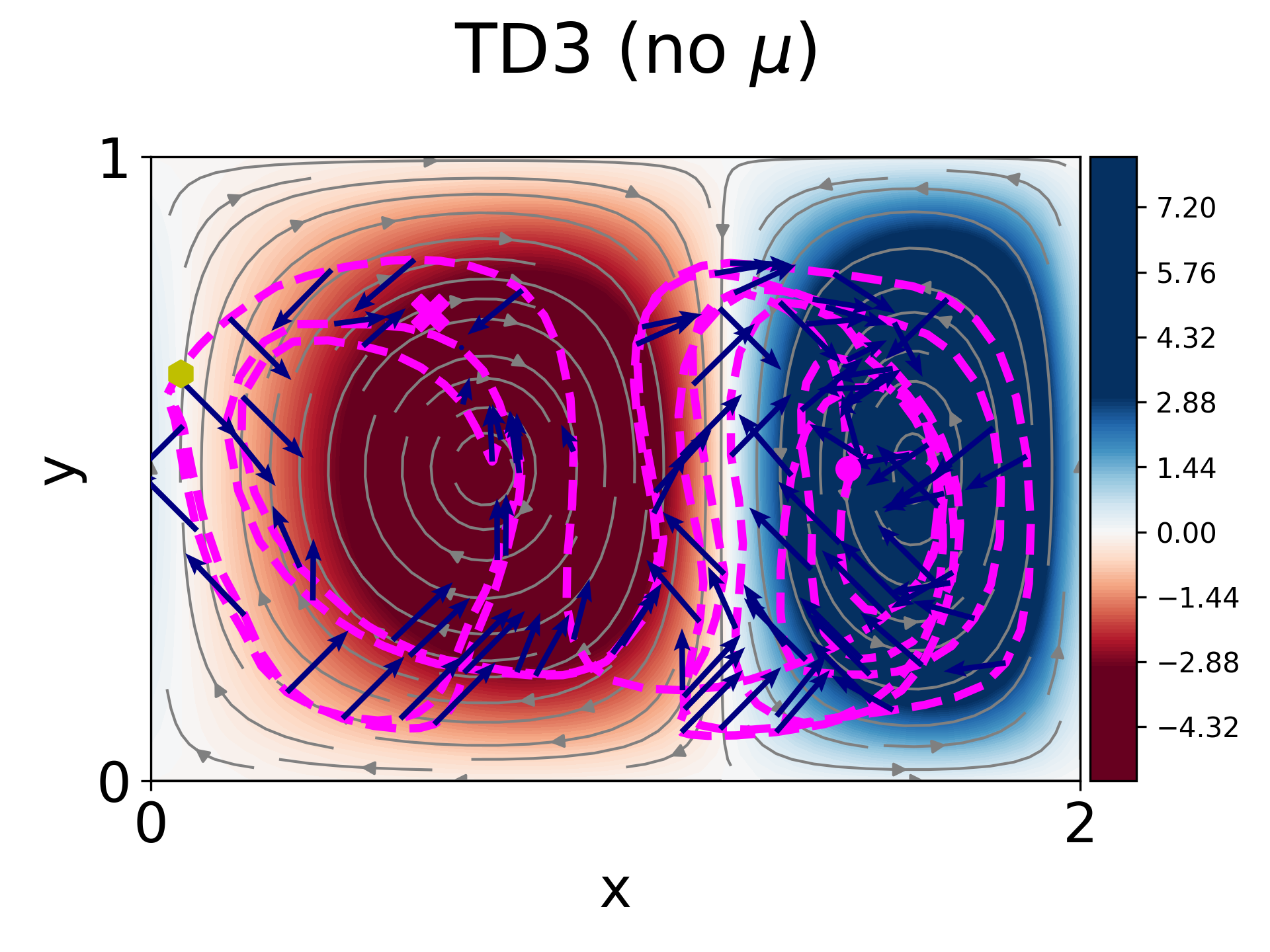}}
    \end{minipage}
    \caption{Trajectories (magenta dash line) of a controlled particle (yellow hexagon) in a gyre flow when using the different agents. The magenta circle indicates the starting location and the magenta cross the target one and the blue arrows represent the control inputs.}
    \label{fig:param_gyro_controlled_full}
\end{figure*}

\begin{figure*}[h!]
    \begin{minipage}{0.49\linewidth}
    \centering \subfloat{\includegraphics[height=0.7\textwidth]{Pics/controlled_solution_parametric_gyro/hypeRL_td3_v2_solution_FTLE_1_1_timestep_67.png}}
    \vspace{-0.1cm}
    \subfloat{\includegraphics[height=0.7\textwidth]{Pics/controlled_solution_parametric_gyro/td3_huber_solution_FTLE_1_1_timestep_105.png}}
    \end{minipage}
    \begin{minipage}{0.49\linewidth}
    \centering \subfloat{\includegraphics[height=0.7\textwidth]{Pics/controlled_solution_parametric_gyro/td3_solution_FTLE_1_1_timestep_399.png}}
    \vspace{-0.1cm}
    \subfloat{\includegraphics[height=0.7\textwidth]{Pics/controlled_solution_parametric_gyro/td3_twolayers_solution_FTLE_1_1_timestep_399.png}}
    \end{minipage}
    \centering
    \begin{minipage}{0.49\linewidth}
    \centering \subfloat{\includegraphics[height=0.7\textwidth]{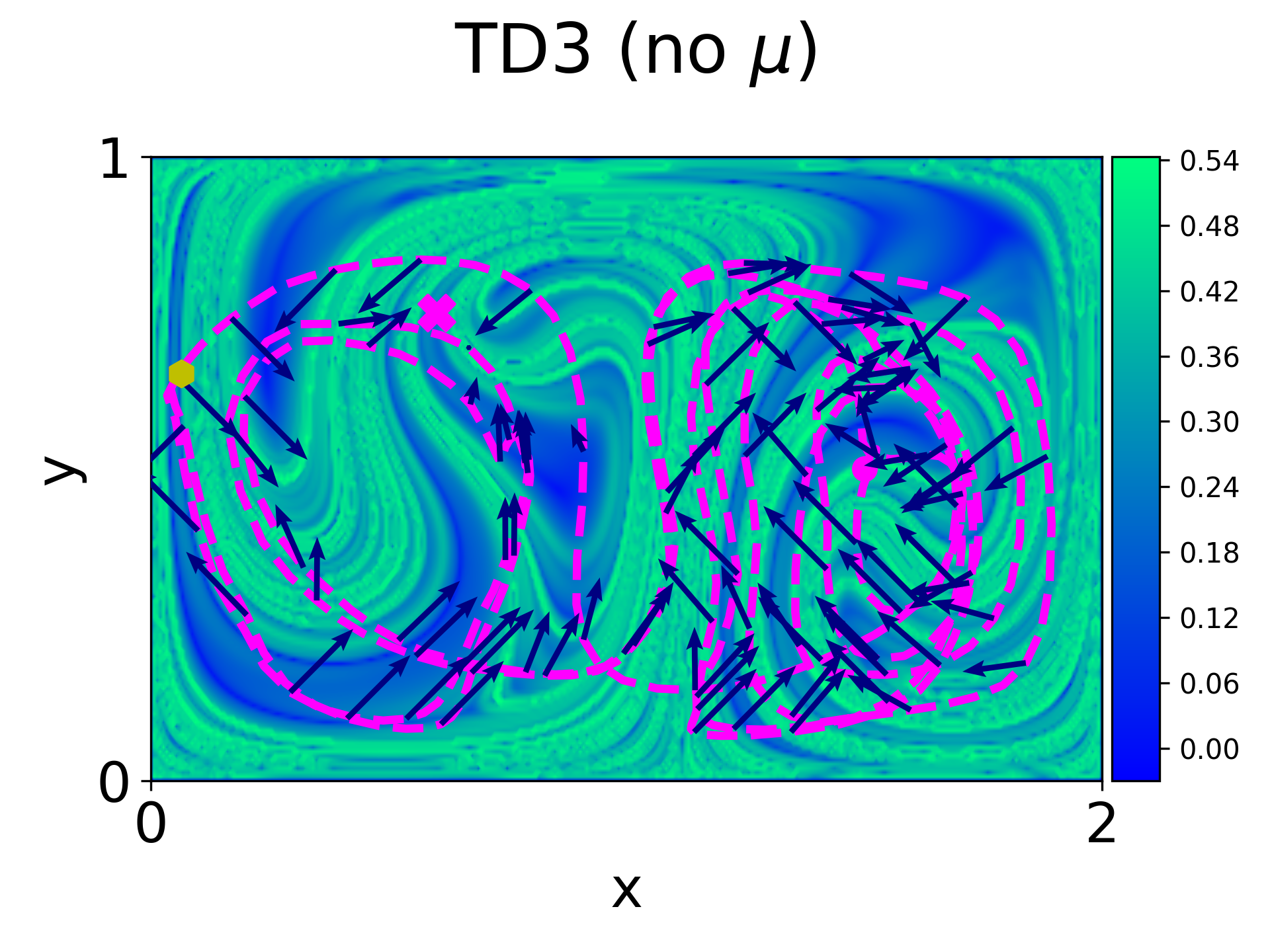}}
    \end{minipage}
    \caption{Trajectories (magenta dash line) of a controlled particle (yellow hexagon) in a gyre flow when using the different agents overlaid with the FTLE. The magenta circle indicates the starting location and the magenta cross the target one and the blue arrows represent the control inputs.}
    \label{fig:param_gyro_controlled_FTLE_full}
\end{figure*}

\begin{figure*}[h!]
    \begin{minipage}{0.49\linewidth}
    \centering \subfloat{\includegraphics[height=0.7\textwidth]{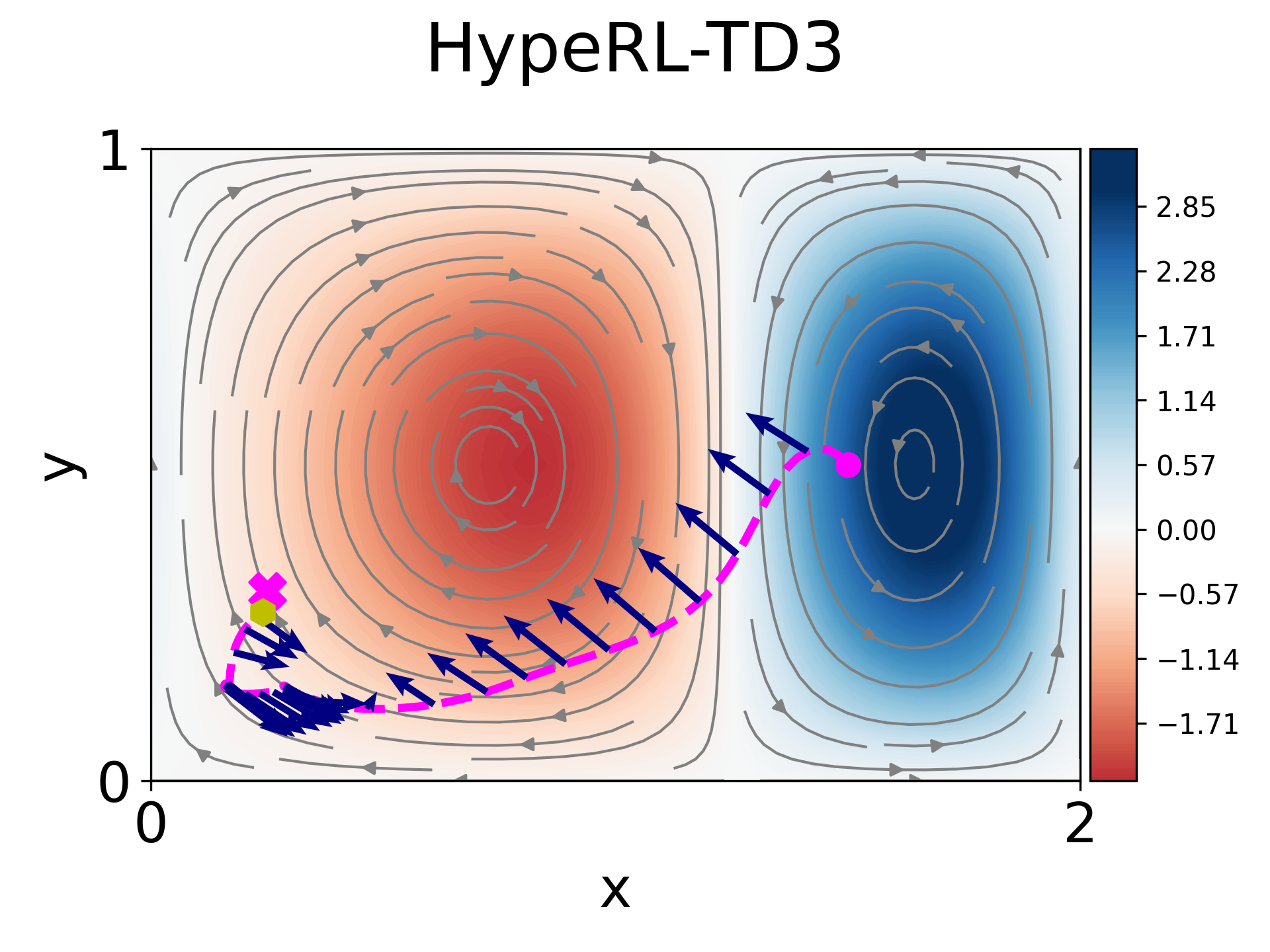}}
    \vspace{-0.1cm}
    \subfloat{\includegraphics[height=0.7\textwidth]{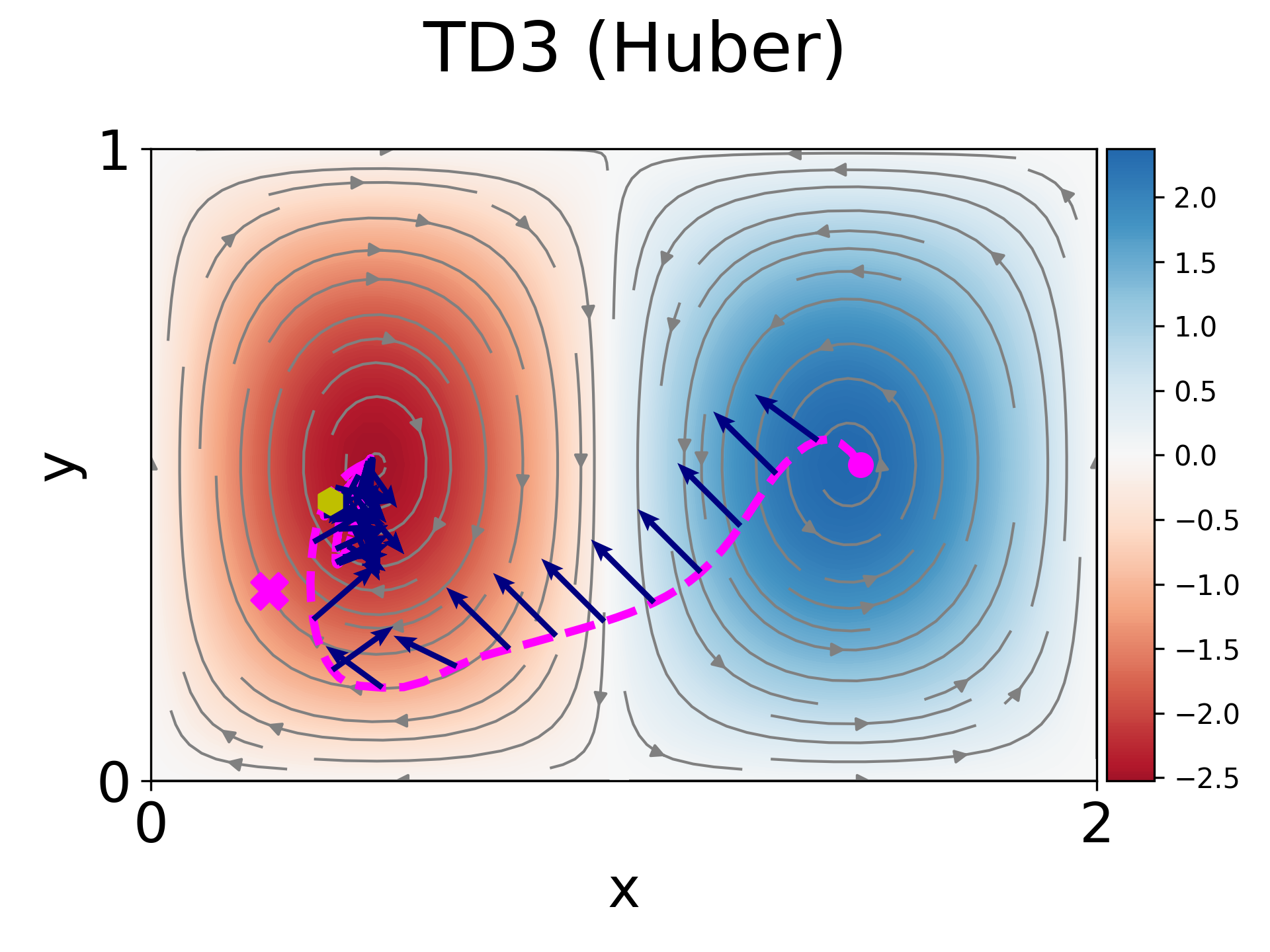}}
    \end{minipage}
    \begin{minipage}{0.49\linewidth}
    \centering \subfloat{\includegraphics[height=0.7\textwidth]{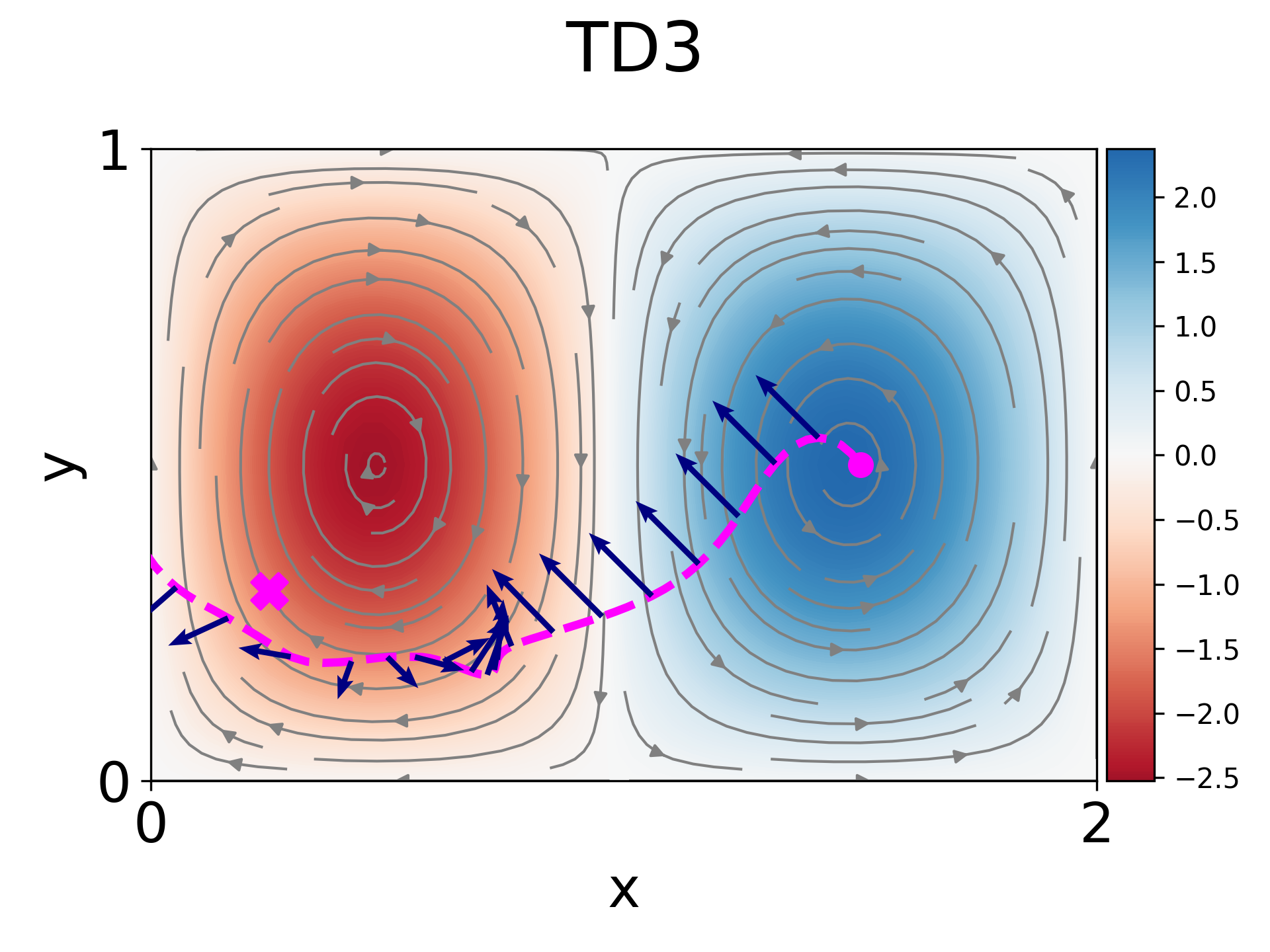}}
    \vspace{-0.1cm}
    \subfloat{\includegraphics[height=0.7\textwidth]{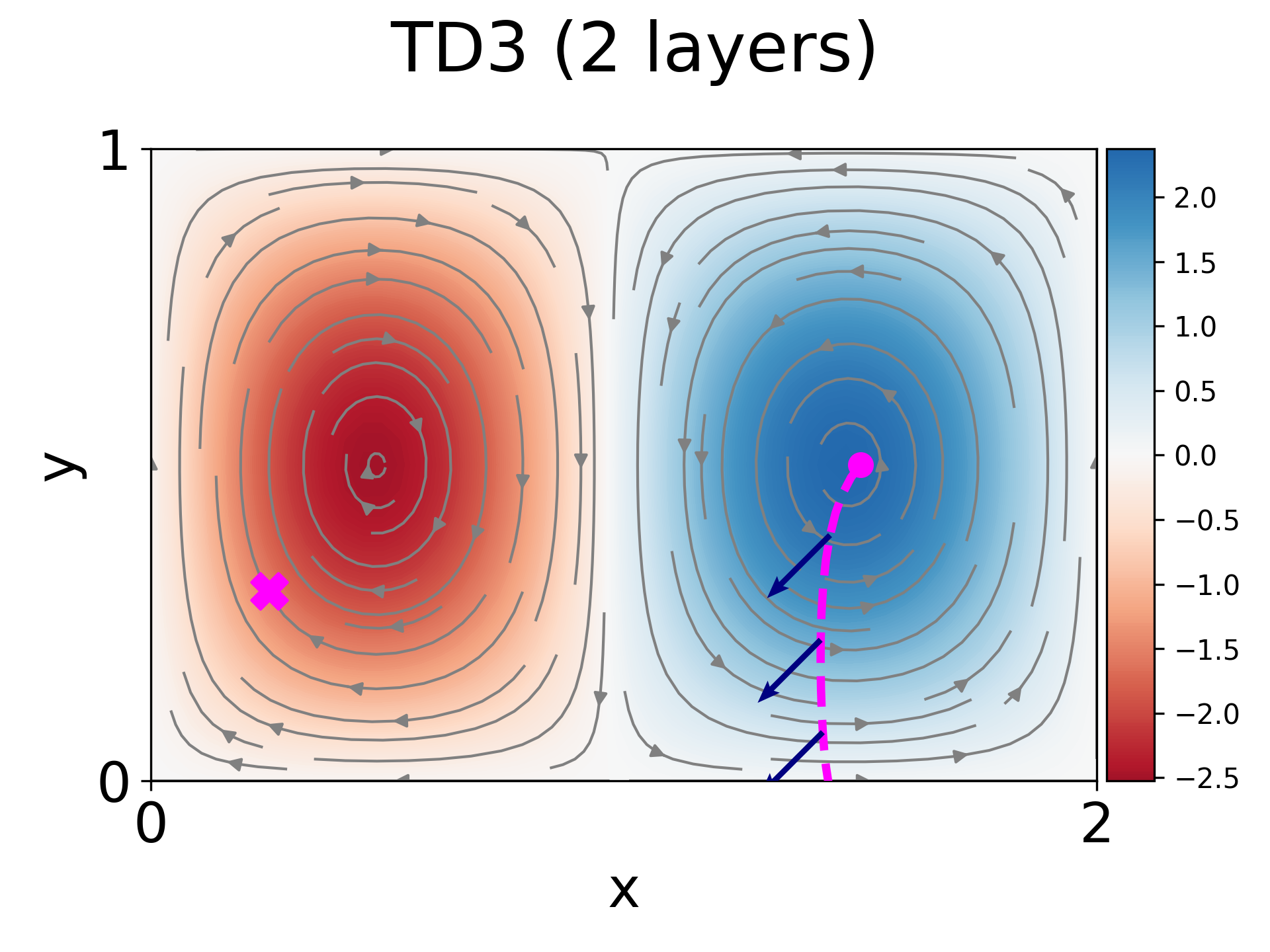}}
    \end{minipage}
    \centering
    \begin{minipage}{0.49\linewidth}
    \centering \subfloat{\includegraphics[height=0.7\textwidth]{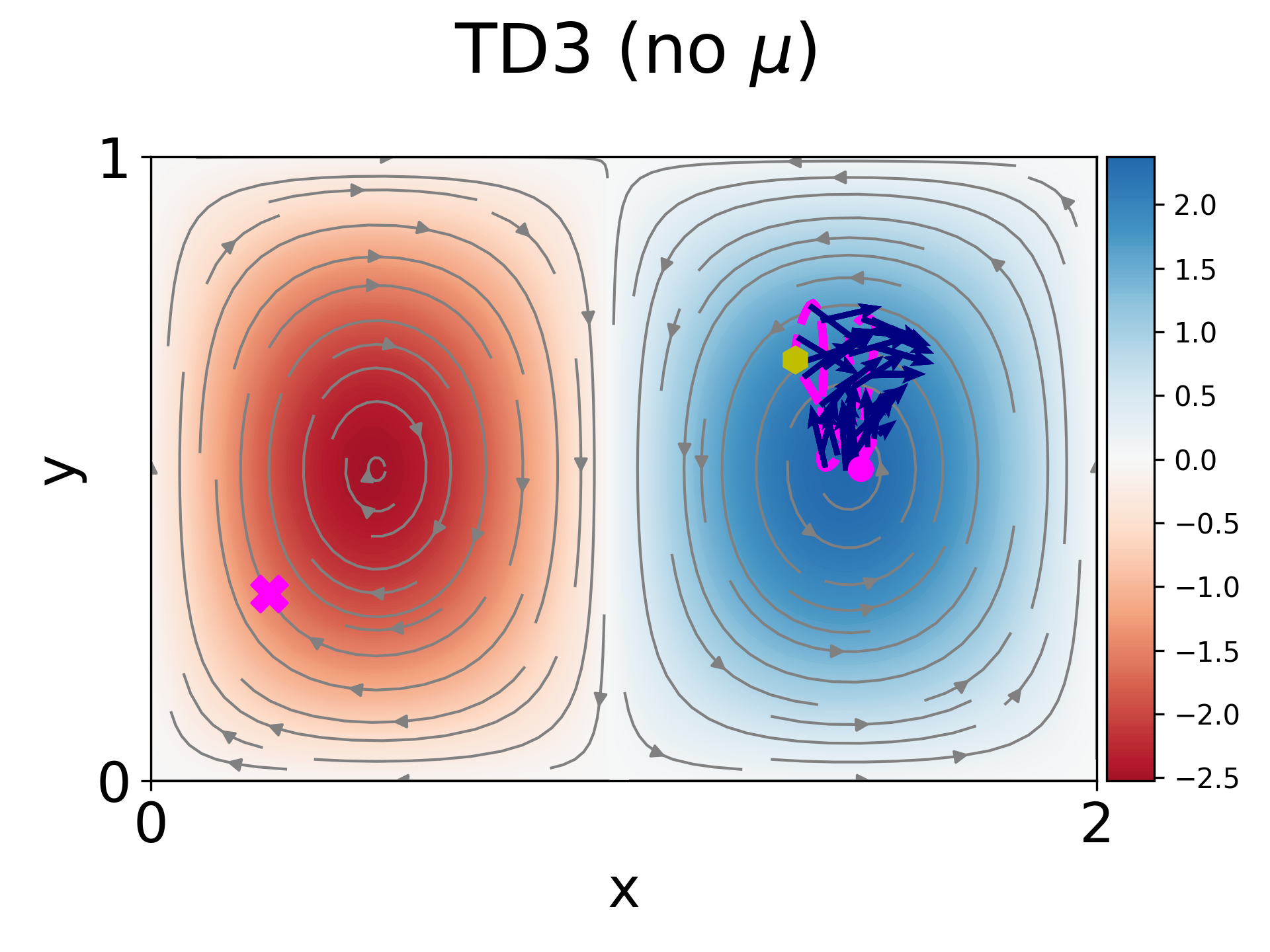}}
    \end{minipage}
    \caption{Trajectories (magenta dash line) of a controlled particle (yellow hexagon) in a gyre flow when using the different agents. The magenta circle indicates the starting location and the magenta cross the target one and the blue arrows represent the control inputs.}
    \label{fig:param_gyro_controlled_full1}
\end{figure*}

\begin{figure*}[h!]
    \begin{minipage}{0.49\linewidth}
    \centering \subfloat{\includegraphics[height=0.7\textwidth]{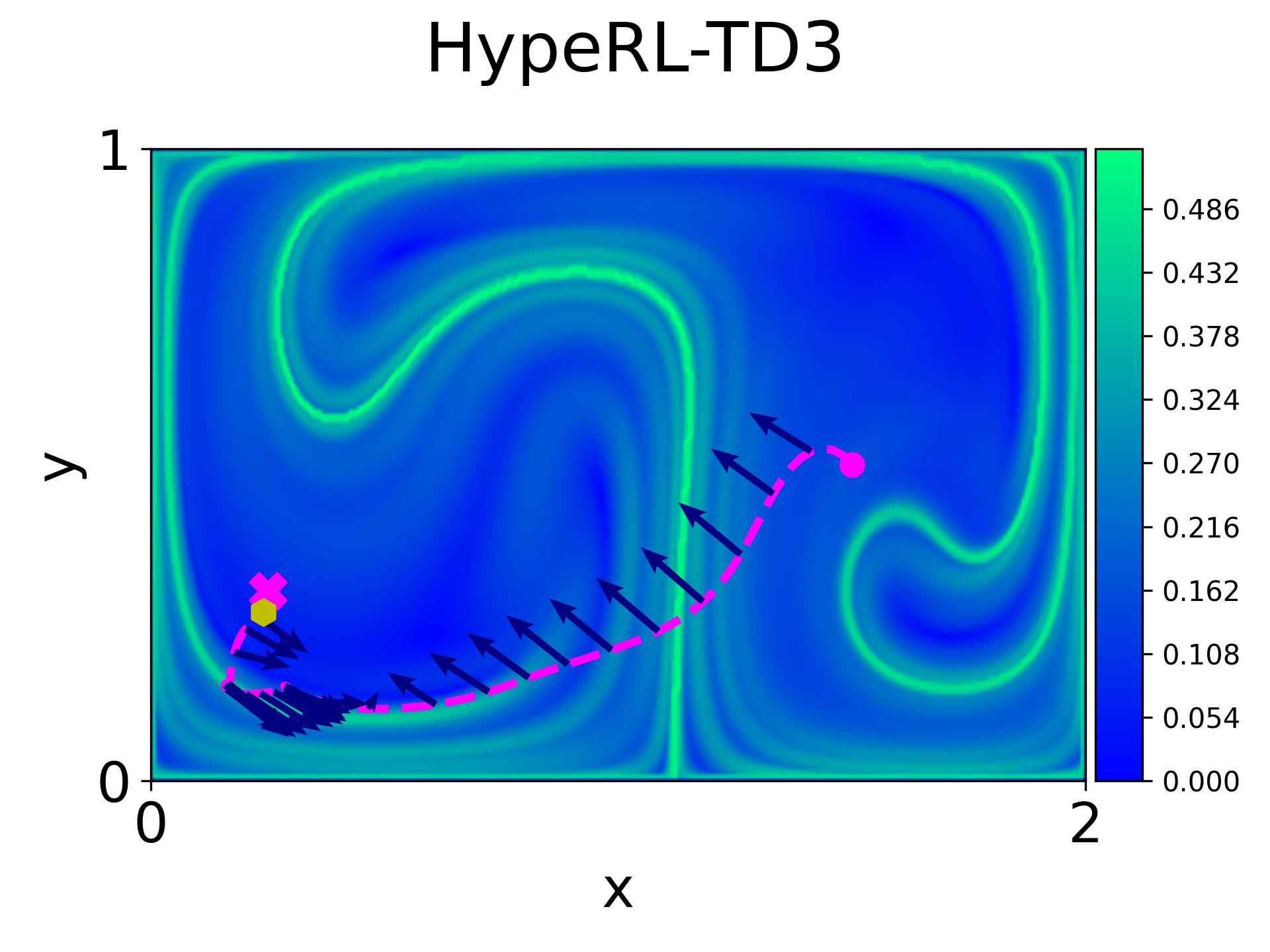}}
    \vspace{-0.1cm}
    \subfloat{\includegraphics[height=0.7\textwidth]{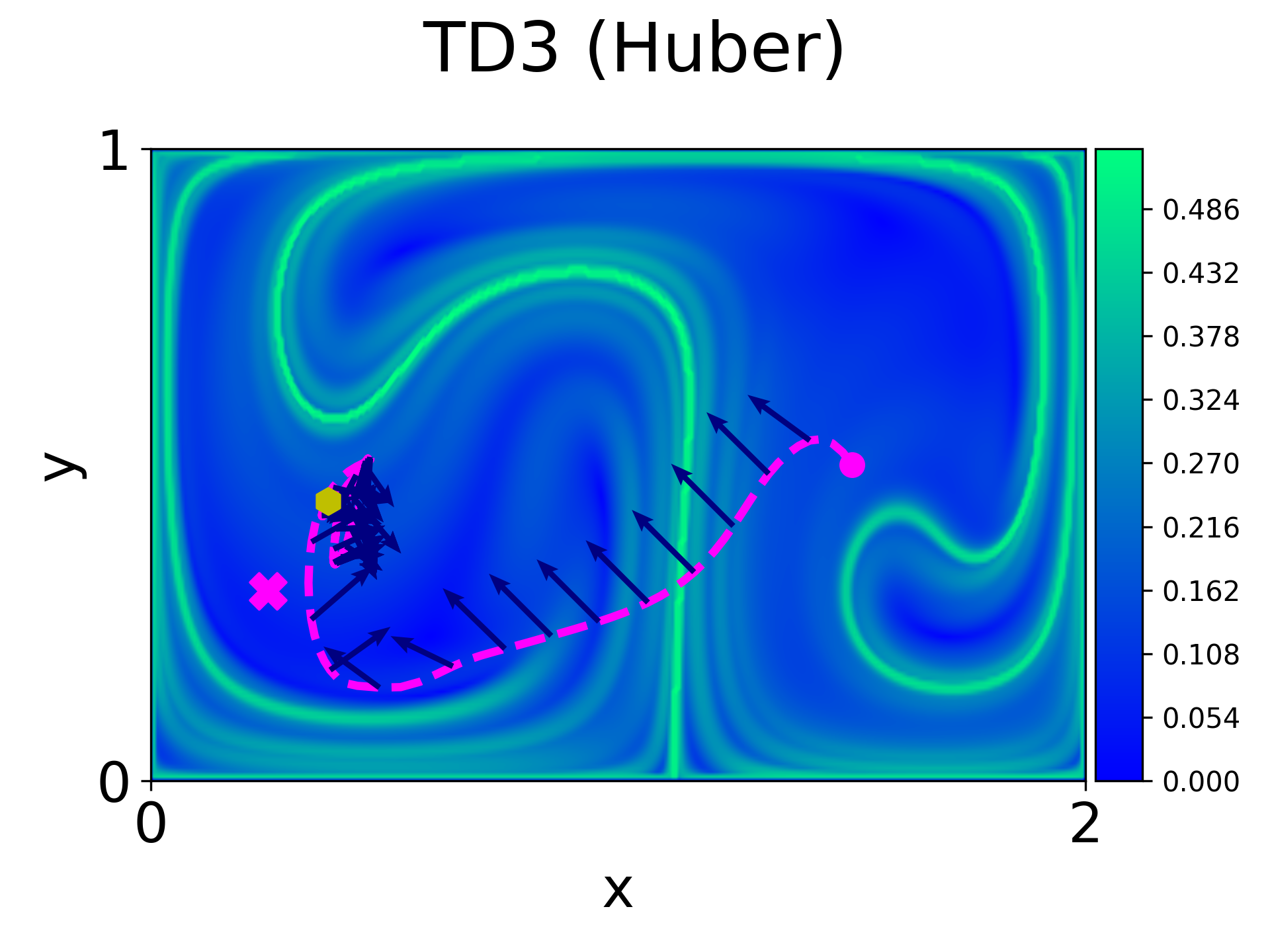}}
    \end{minipage}
    \begin{minipage}{0.49\linewidth}
    \centering \subfloat{\includegraphics[height=0.7\textwidth]{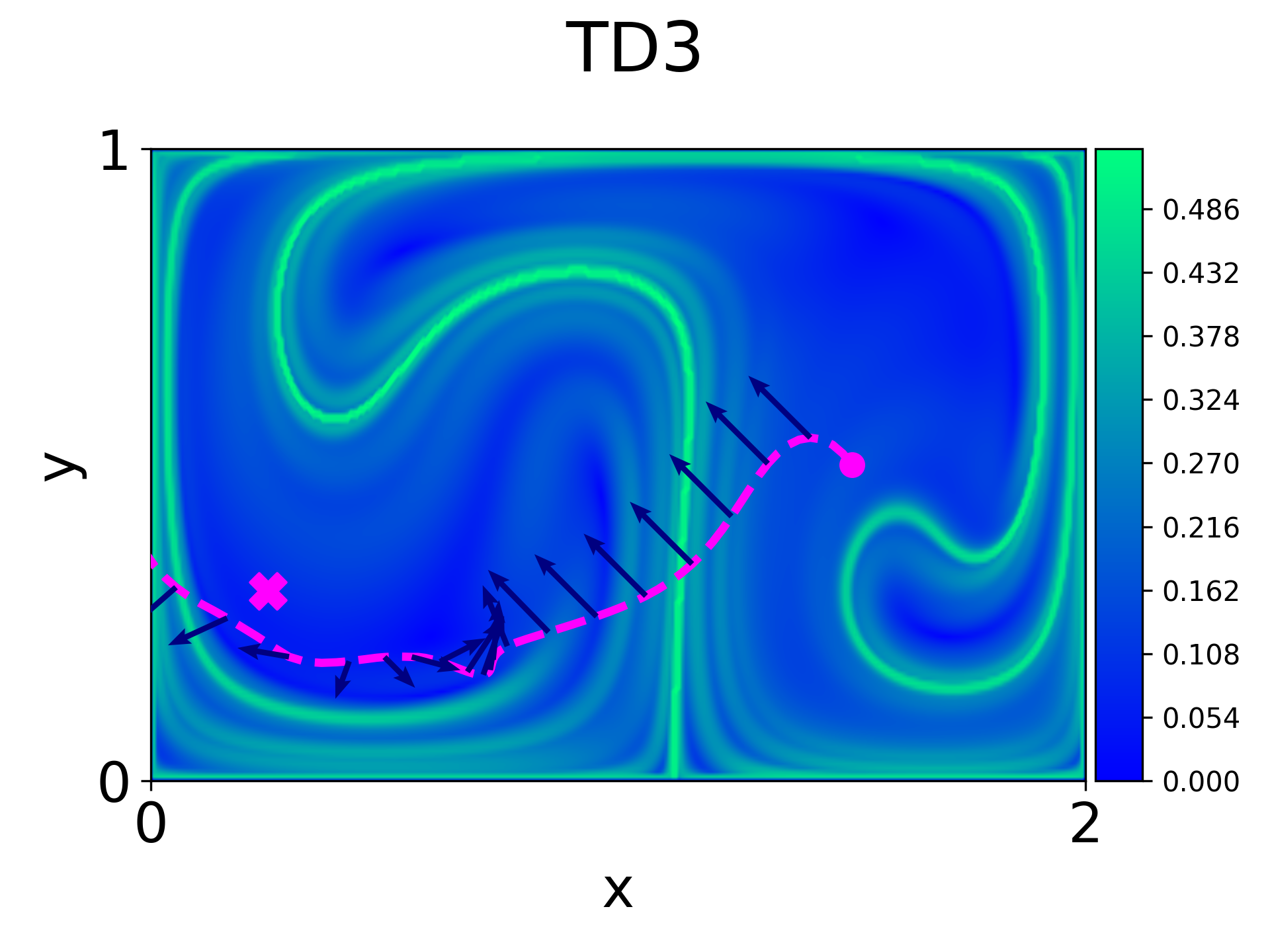}}
    \vspace{-0.1cm}
    \subfloat{\includegraphics[height=0.7\textwidth]{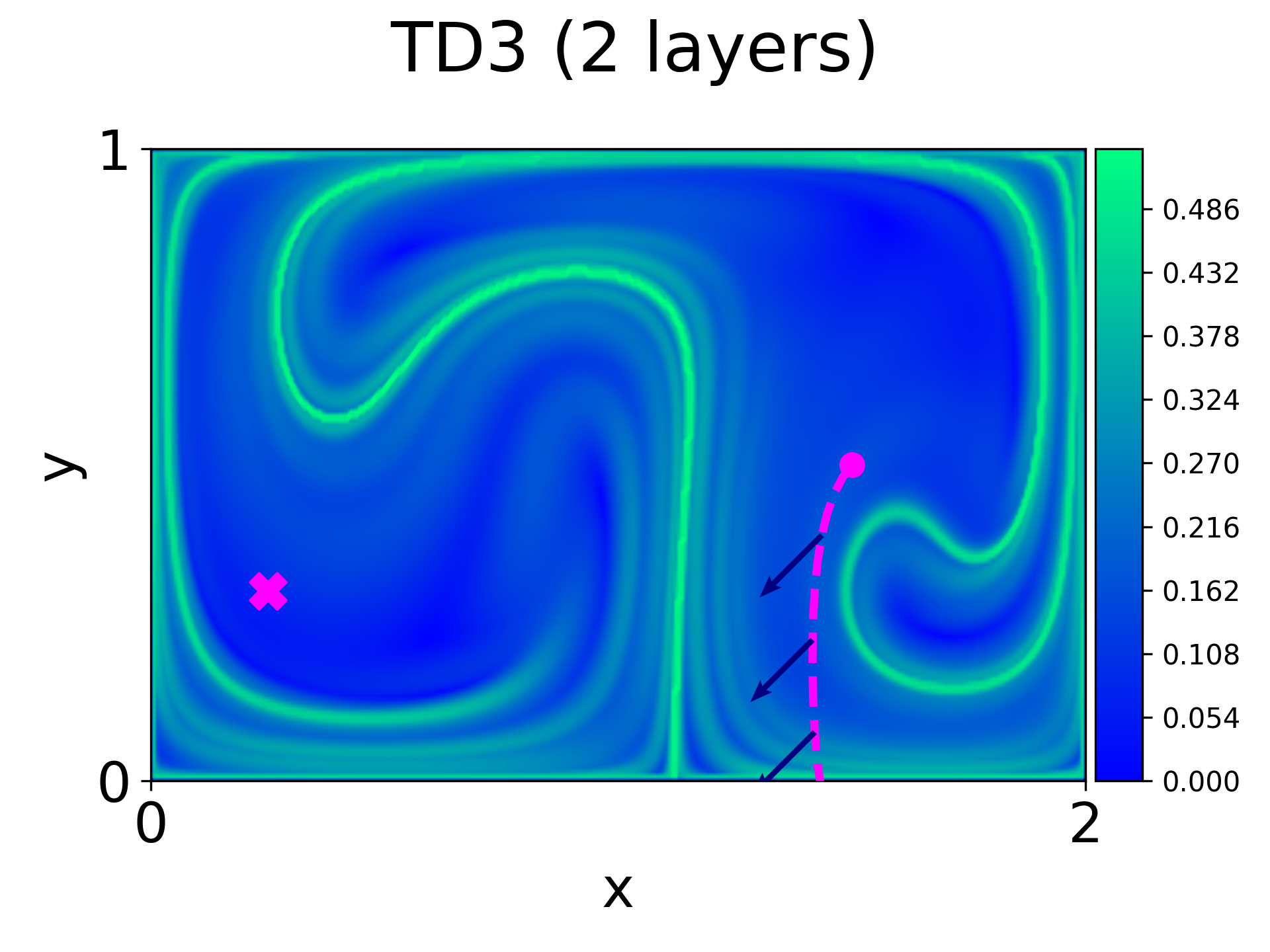}}
    \end{minipage}
    \centering
    \begin{minipage}{0.49\linewidth}
    \centering \subfloat{\includegraphics[height=0.7\textwidth]{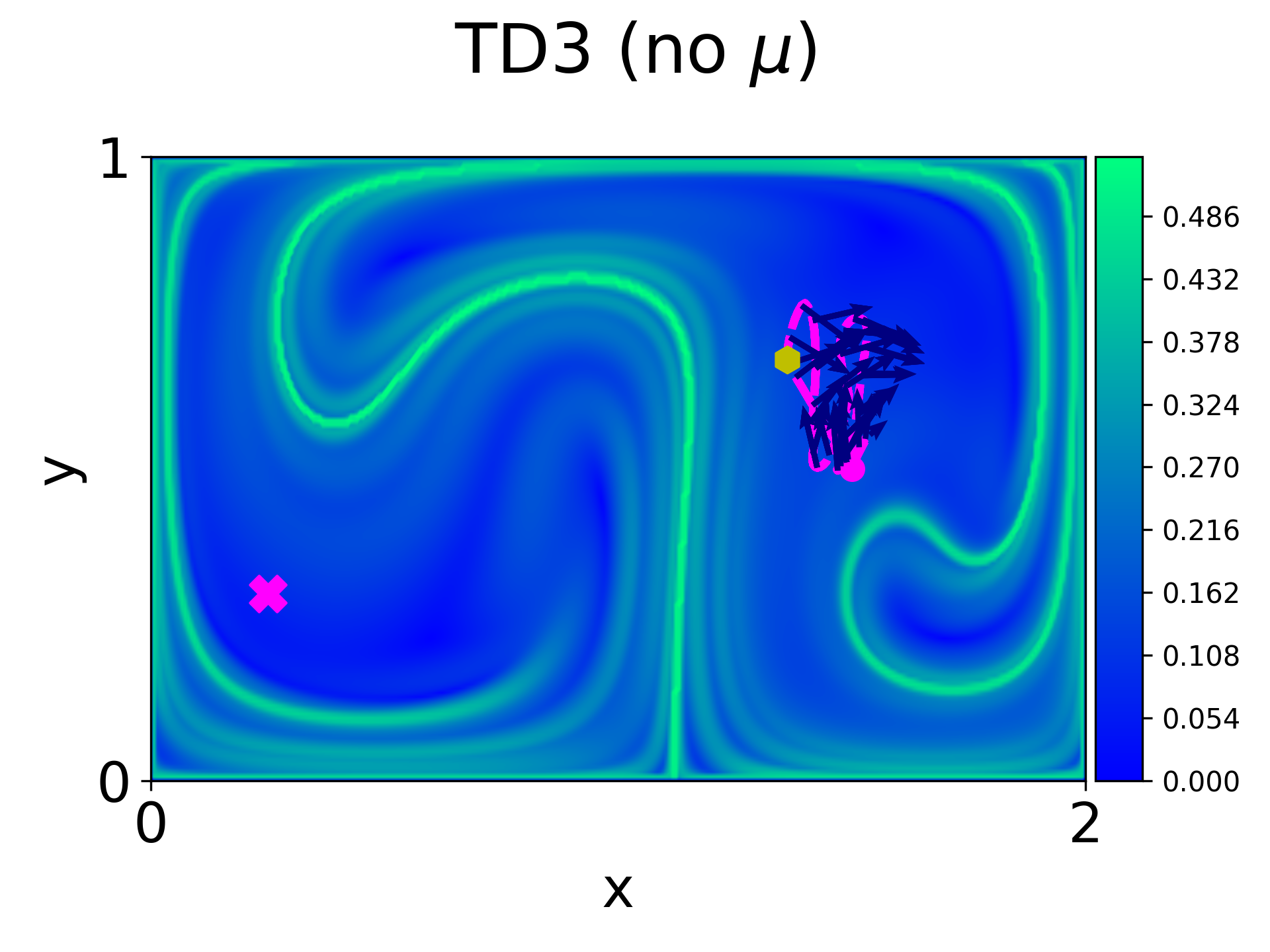}}
    \end{minipage}
    \caption{Trajectories (magenta dash line) of a controlled particle (yellow hexagon) in a gyre flow when using the different agents overlaid with the FTLE. The magenta circle indicates the starting location and the magenta cross the target one and the blue arrows represent the control inputs.}
    \label{fig:param_gyro_controlled_FTLE_full1}
\end{figure*}

In Figures \ref{fig:param_gyro_controlled_full} and \ref{fig:param_gyro_controlled_full1} we show examples of controlled trajectories overlaid with the gyro flow field, while in Figures \ref{fig:param_gyro_controlled_FTLE_full} and \ref{fig:param_gyro_controlled_FTLE_full1} we overlaid the trajectories with the FTLE.

